%% file: main.tex
\documentclass[conference, onecolumn]{IEEEtran}
\IEEEoverridecommandlockouts
\usepackage{cite}
\usepackage{amsmath,amssymb,amsfonts}
\usepackage{graphicx}
\usepackage{textcomp}
\usepackage{xcolor}
\def\BibTeX{{\rm B\kern-.05em{\sc i\kern-.025em b}\kern-.08em
    T\kern-.1667em\lower.7ex\hbox{E}\kern-.125emX}}

\usepackage{latexsym}
\usepackage{amsthm}
\usepackage{booktabs}
\usepackage{enumitem}
\usepackage{color}
\usepackage[hidelinks]{hyperref}
\usepackage{algorithm}
\usepackage{algpseudocode}
\usepackage{amssymb}
\usepackage{cases}
\usepackage[most]{tcolorbox}
\usepackage{bbm}
\usepackage{tabularx}
\usepackage{multirow}
\usepackage{tikz}
\usetikzlibrary{positioning, arrows.meta}

\usepackage{subcaption}
\usepackage{placeins}

\usepackage{eurosym}

\usepackage{siunitx}

\newcommand{\percent}[1]{\fpeval{round(100*(#1),1)}} 

\newcommand{\1}{\mathbbm{1}}

\usepackage{soul}

\newcommand{\twosideimages}[6]{
    \begin{figure}[htbp]
        \centering
        \begin{subfigure}{0.48\textwidth}
            \centering
            \includegraphics[width=\linewidth]{imgs/#1}
            \caption{#2}
        \end{subfigure}
        \hfill
        \begin{subfigure}{0.48\textwidth}
            \centering
            \includegraphics[width=\linewidth]{imgs/#3}
            \caption{#4}
        \end{subfigure}
        \caption{#5}        
        \label{fig:#6}
    \end{figure}
}

\newcommand{\threebiasimages}[2]{
    \begin{figure}[htbp]
        \centering
        \begin{subfigure}{0.3\textwidth}
            \centering
            \includegraphics[width=\linewidth]{imgs/#1_unbiased}
            \caption{Without Proposal}
        \end{subfigure}
        \hfill
        \begin{subfigure}{0.3\textwidth}
            \centering
            \includegraphics[width=\linewidth]{imgs/#1_biased}
            \caption{With Proposal}
        \end{subfigure}
        \hfill
        \begin{subfigure}{0.3\textwidth}
            \centering
            \includegraphics[width=\linewidth]{imgs/#1_post}
            \caption{Post-Processed}
        \end{subfigure}
        \caption{#2}        
        \label{fig:#1}
    \end{figure}
}

\renewcommand{\thetable}{\arabic{table}}

\begin{document}

\makeatletter

\newcommand{\linebreakand}{%
  \end{@IEEEauthorhalign}
  \hfill\mbox{}\par
  \mbox{}\hfill\begin{@IEEEauthorhalign}
}

\title{Object Detection Benchmarks are Incomplete:\\ The Role of Label Errors and Annotation Uncertainty}

\author{
  \IEEEauthorblockN{1\textsuperscript{st} Sarina Penquitt$^\ast$\thanks{$^\ast$ Equal contribution.} }
  \IEEEauthorblockA{\textit{Institute of Computer Science} \\
    \textit{Osnabrück University}\\
    Osnabrück, Germany \\
    sarina.penquitt@uni-osnabrueck.de}
  \and
  \IEEEauthorblockN{1\textsuperscript{st} Jonathan Klees$^\ast$}
  \IEEEauthorblockA{\textit{Institute of Computer Science} \\
    \textit{Osnabrück University}\\
    Osnabrück, Germany \\
    jonathan.klees@uni-osnabrueck.de}
  \linebreakand
  \IEEEauthorblockN{3\textsuperscript{rd} Antonia van Betteray}
  \IEEEauthorblockA{\textit{Institute of Computer Science} \\
    \textit{Osnabrück University}\\
    Osnabrück, Germany \\
    antonia.vanbetteray@uni-osnabrueck.de}
 \and
  \IEEEauthorblockN{4\textsuperscript{th} Parssa Jashnieh}
  \IEEEauthorblockA{\textit{Institute of Computer Science} \\
    \textit{Osnabrück University}\\
    Osnabrück, Germany \\
    parssa.jashnieh@uni-osnabrueck.de}
  \and
  \IEEEauthorblockN{5\textsuperscript{th} Peter Stehr}
  \IEEEauthorblockA{\textit{Department of Mathematics} \\
    \textit{University of Wuppertal}\\
    Wuppertal, Germany \\
    stehr@uni-wuppertal.de}
  \linebreakand
  \IEEEauthorblockN{6\textsuperscript{th} Matthias Rottmann$^\dagger$\thanks{$^\dagger$ Equal contribution.}}
  \IEEEauthorblockA{\textit{Institute of Computer Science} \\
    \textit{Osnabrück University}\\
    Osnabrück, Germany \\
    matthias.rottmann@uni-osnabrueck.de}
 \and
 \IEEEauthorblockN{7\textsuperscript{th} Lars Schmarje$^\dagger$}
 \IEEEauthorblockA{\textit{Wayve} \\
 London, United Kingdom \\
 lars.schmarje@wayve.ai}

}

\maketitle

\begin{abstract}
While object detection has advanced through improved architectures and open-vocabulary models, we provide strong evidence that benchmark quality is limited by annotation incompleteness. 
Across four widely used datasets (COCO, Pascal VOC, Cityscapes, KITTI), re-annotation reveals substantial increases in annotated objects (e.g., up to +60\% on KITTI and +40\% on COCO), driven primarily by previously unlabeled small, occluded, or densely packed instances. 
While some differences arise from dataset-specific annotation conventions, we consistently find that missing annotations are the main source of label errors across all datasets.
To achieve high data quality, we introduce a scalable annotation pipeline that emphasizes high recall and captures ambiguity through soft labels aggregated from at least 11 annotators per object. The resulting annotations improve coverage and align well with human calibration. We show that benchmark performance is highly sensitive to annotation quality, although model rankings remain largely stable.
We introduce two large-scale benchmarks: (i) an uncertainty-aware object detection benchmark, and (ii) a label error detection benchmark grounded in real label errors.
We show that current detectors are strongly depended on annotation quality and are misaligned with human perception.
Current label error detection methods, which have been shown to perform well on synthetic noise, struggle to achieve high recall and precision on real label errors.
Our results highlight the need for future object detection benchmarks to move beyond deterministic annotations toward high-recall, uncertainty-aware evaluation that maximizes valid instances and better reflects real-world ambiguity.
\end{abstract}


\section{Introduction}

In widely used object detection datasets, annotations are often incomplete, ambiguous, and overconfident~\cite{Wei2021cifar10n,schmarje2022benchmark,rechecked}, leading to systematic biases in both model training and evaluation~\cite{Ma2022Correction, northcutt2021confident}.
In particular, missing annotations are the main source of label errors, as many small, occluded, or densely packed objects are not labeled at all~\cite{rechecked}.
At the same time, existing datasets rely on deterministic hard labels, assigning a single class to each object and ignoring inherent ambiguity~\cite{peterson2019human,Basile2021ConsiderDisagree}.
This results in overconfident annotations and limits the ability to properly assess model uncertainty~\cite{schmarje2022benchmark}.

To address both annotation incompleteness and ambiguity, we re-annotate four widely used object detection benchmarks: COCO~\cite{COCO}, Pascal VOC~\cite{PascalVOC}, Cityscapes~\cite{Cityscapes}, and KITTI~\cite{kitti}. Using a multi-stage, crowdsourcing-based annotation pipeline, we construct annotations that emphasize high recall and capture inter-annotator disagreement through soft labels, i.e., probability distributions over classes derived from multiple independent annotations. Across datasets, this results in substantial increases in annotated objects (e.g., up to +60\% on KITTI and +40\% on COCO), primarily due to previously unlabeled small, occluded, or densely distributed instances. While some differences arise from dataset-specific annotation conventions (e.g., group annotations or “don’t care” regions), we consistently observe that missing annotations are the main source of label errors across all datasets.
Through comprehensive data quality assessment, we show that the rate of missing objects in re-annotated datasets is at least halved, compared to original data.
Additionally, the re-annotated data consists of soft labels which are well calibrated towards human perception. 
Overall, we not only find more objects, but annotations are of higher quality and capture human uncertainty.

Based on these validated annotations, we investigate how annotation quality affects model evaluation and find that benchmark performance differs between original and refined annotations, while model rankings remain largely stable. This suggests that absolute performance metrics depend on annotation quality and may be biased by incomplete ground truth (GT). Since current object detection evaluation protocols do not account for annotation uncertainty, we introduce a new evaluation protocol that generalizes average precision to soft labels and enables instance-wise comparison between predictions and soft labels.
We identify real label errors through direct comparison of refined annotations with the original GT. We categorize these errors into missing annotations, misaligned bounding boxes, and incorrect class labels, and provide a thorough analysis of their distribution across all four datasets. This way, we establish the first large-scale benchmark for label error detection in object detection on real data, evaluating state-of-the-art label error detection methods beyond synthetic settings~\cite{Yun2021Correction, Kim2024Correction}.
The code used in this work will be publicly available on \href{https://github.com/JonathanKlees/Uncertainty-aware-Object-Detection}{GitHub}, which also provides access to the validated annotations for all four datasets. 

\input{tables/introduction_tables}

\section{Related Work}

\paragraph*{Crowdsourcing and Re-annotation of Existing Datasets}
Several works have focused on improving existing datasets to study annotation quality and its effect on model performance. As summarized in \autoref{tab:related_work}, prior work typically addresses only subsets of this problem. For object detection, datasets such as LVIS~\cite{gupta2019lvis} and OpenImages~\cite{kuznetsova2020openimages} improve annotation coverage and increase label granularity, but do not explicitly model annotation uncertainty. Similarly, Kim et al.~\cite{Kim2024Correction} construct Pascal+ by refining annotations, while Ma et al.~\cite{Ma2022Correction} partially re-annotate COCO~\cite{COCO} and Open Images~\cite{kuznetsova2020openimages} to study the effect of corrected annotations on detector performance. Works such as CIFAR-10H~\cite{peterson2019human}, DCIC~\cite{schmarje2022benchmark}, and ReLabel~\cite{Yun2021Correction} focus on capturing human uncertainty or systematic re-annotation, but focus on image classification. Rechecked~\cite{rechecked} provides refined object detection annotations with soft labels for a subset of KITTI, but is limited to a single dataset and class.
In contrast, our work provides large-scale re-annotation for object detection across multiple datasets, enabling a systematic analysis of annotation completeness, human uncertainty and label errors.

\paragraph*{Soft Labels and Uncertainty Quantification in Object Detection}
Soft labels were originally introduced to represent confidence in binary classification~\cite{nguyen_learning_2011} and later extended to multiclass settings to capture aleatoric uncertainty~\cite{Peng2014LearningOP}. While prior work has explored training schemes that leverage soft labels~\cite{Peng2014LearningOP}, dataset construction typically focuses on deriving consensus or hard labels from multiple annotations~\cite{review_of_consensus_algorithms,learning_from_crowds}, rather than explicitly modeling annotation uncertainty.
Current research shows that a single annotation per instance is insufficient as it cannot reflect aleatoric uncertainty inherent in class affiliations~\cite{tailception,mammo-variability,schoening2020Megafauna,medicinecrowdsource,Vasudevan2022DoughBagel,planktonUncertain,Davani2022BeyondMajority,Basile2021ConsiderDisagree,schmarje2022dc3,Wei2021cifar10n}. Training models on such datasets may lead to overconfident predictions and harm model performance~\cite{Barragan2021QualityQuantity,deep_fish,noisy-labels-comparison,schmarje2022benchmark,Wei2021cifar10n,Tomasev2022Relicv2}. In contrast, using soft labels for training makes classification more robust~\cite{peterson2019human}. 
However, acquiring soft labels comes with substantial annotation effort as multiple annotators need to evaluate the same instance.
Consequently, benchmark datasets that include soft labels are rare and standard evaluation protocol assumes zero uncertainty in annotations~\cite{PascalVOC, COCO, ImageNet, SSD, focal_loss, fasterRCNN, Hall2020}.
We introduce soft labels for object detection at scale and a new evaluation metric that generalizes average precision~\cite{PascalVOC} to the domain of soft labels, enabling uncertainty-aware evaluation and direct comparison between model predictions and human perception.

\paragraph*{Label Error Detection for Object Detection Datasets}
Label errors in object detection datasets include overlooked objects, poor bounding box localization, and incorrect class affiliation~\cite{rechecked}.
They introduce bias in datasets, degrading model performance and generalization capability~\cite{Song2023LearningFromNoisyLabels, Nahum2024LLM, jakubik2024Improve, Barragan2021QualityQuantity, deep_fish, noisy-labels-comparison, Wei2021cifar10n, Tomasev2022Relicv2}, and affecting benchmark validity~\cite{northcutt2021confident, Song2023LearningFromNoisyLabels}.  
Prior work proposed automated label error detection methods for image classification~\cite{northcutt2021confident,northcutt2021pervasive,thyagarajan2022}, for object detection~\cite{loss-based-method,metadetect,objectlab} or semantic segmentation~\cite{Rottmann2023AutomatedDetection} datasets. However, their evaluation often relies on simulated label errors or small-scale manual inspection. 
In contrast, we leverage our refined annotations to construct a large-scale benchmark for label error detection based on real label errors across multiple object detection datasets, enabling realistic evaluation beyond synthetic noise settings.

\input{methodgraph/methodgraph}

\section{Dataset Construction}
\label{sec:dataset_construction}

The goal of the re-annotation is to achieve very high recall, i.e., to capture as many valid object instances as possible and minimize missed objects, by annotating every visible object in the images. For each object, we collect a bounding box together with a soft label, i.e., a probability distribution over all classes. To this end, we design a three-stage microtask (\autoref{fig:reannotation_pipeline}). A microtask is a simple, well-defined visual task with a short and easy-to-answer question. Each microtask is answered independently by multiple annotators, enabling parallel processing as well as the estimation of annotation uncertainty. 
We call the resulting refined annotations \emph{validated GT}.

\subsection*{Stage 1: Keypoint Collection}
In the first stage, we aim to localize all potential object instances. For each image and each class, annotators are asked to mark all instances that could belong to the respective class using keypoints. To maximize recall, annotators are instructed to also mark uncertain or partially visible objects. Across three repeats, annotators are shown previously placed keypoints and are asked to add any missing ones. This redundancy increases the likelihood that all relevant objects are captured. Objects not marked at this stage are not considered in subsequent stages.

\subsection*{Stage 2: Bounding Box Collection}
In the second stage, we obtain a tight bounding box for each keypoint. Annotators are shown one keypoint at a time and are asked to draw a bounding box tightly enclosing the corresponding object. This task is repeated with two annotators to ensure that each keypoint is processed and to improve the quality of the resulting bounding boxes by reducing missed instances and correcting inaccuracies. At this stage, all objects are marked with a bounding box.

\subsection*{Stage 3: Soft Label Generation}
In the final stage, we determine the class of each object by collecting class labels from multiple independent annotators. Each bounding box is presented to annotators, who are asked to assign it to one of the dataset classes or select \textit{can't solve} if the object is ambiguous or not recognizable. The class for which the keypoint was set serves as a proposal. By aggregating the response of at least 11 annotators, we obtain a soft label, which captures annotation uncertainty as a probability distribution over classes.

\subsection*{Stage 4: Post Processing}
After all stages, each image is associated with a set of validated object annotations, where each object is represented by a bounding box and a corresponding soft label derived from multiple annotators.
We further apply several post-processing steps, including the refinement of soft labels for ambiguous cases through 11 additional annotator responses, bias correction and the removal of duplicate annotations.
A potential bias is introduced due to the known effect of accepting a given proposal in uncertain situations~\cite{schmarje2022dc3,schmarje2023cleverlabel}.
Duplicate removal is conducted by experts, who review pairs of overlapping bounding boxes ranked by IoU until duplicate detection rate is sufficiently low. 
The resulting labels are our validated GT.
Details are provided in
appendix~\ref{sec: appendix - Dataset Details}.

\subsection*{Used Datasets}
We re-annotate four widely used object detection benchmarks: 
KITTI~\cite{kitti}, Cityscapes~\cite{Cityscapes}, COCO 2017~\cite{COCO} and Pascal VOC 2012~\cite{PascalVOC}, covering both street scenes and everyday environments (see overview in \autoref{tab:dataset_overview}).
For KITTI, we re-annotate all training images across seven classes. 
For Cityscapes, the validation set is fully re-annotated, while the training set is extended by adding previously missing objects to the original annotations.
For COCO 2017, we focus on the images of the validation split, covering all 80 classes across 12 supercategories. 
For Pascal VOC 2012, we re-annotate both training and validation splits across all 20 classes.

\input{tables/data_quality_objects_missing}
\section{Data Quality Assessment}
\label{sec:data_quality}

All evaluations and usage of our validated GT depend on the fact that its quality is actually higher than originally.
In this section, we compare the quality of the validated GT and original annotations. 

\subsection{Descriptive Statistics}
Through re-annotation, each bounding box is assigned a soft label, from which we derive a hard label by selecting the most probable class. The resulting validated GT consists of high-confidence annotations with $p>0.5$. \autoref{tab:gt_vs_vgt_all_main} shows that the number of annotated objects increases across all datasets, though to varying degrees. Cityscapes (+5.1\%) and Pascal VOC (+8.6\%) demonstrate only moderate gains, suggesting relatively complete original annotations. In contrast, KITTI (+60.0\%) and COCO (+40.0\%) show substantial increases, due to a large number of previously missing objects. Overall, the validated GT significantly improves coverage of annotations. 

\subsection{Annotation Completeness in Validated GT}
For the evaluation of missing objects (false negative), we randomly sample 600 images from each dataset, except for Cityscapes, for which we review the complete validation set of 500 images, since only the validation set was fully re-annotated. The sampled images are independently reviewed by three expert annotators, each assessing a disjoint subset of images. For each image, all visible objects are considered to determine the number of objects missing from the original and validated annotations. We report the resulting false negative rates (FNRs) together with 95\% Wilson confidence intervals.
As shown in \autoref{tab:fn_highlight}, the validated annotations consistently reduce false negatives across all datasets. For Pascal VOC, the FNR decreases from $18.6\%$ to $9.3\%$, while for COCO it decreases from $13.1\%$ to $6.1\%$, approximately halving the respective error rates. Substantial reductions are also observed for Cityscapes, from $11.1\%$ to $3.8\%$, and KITTI, from $41.5\%$ to $11.6\%$. The particularly high FNR of the original KITTI annotations reflects our evaluation of object completeness, as we count all visible objects across the entire image, including those located in \emph{don't care} regions of the original annotations.
Overall, these results demonstrate that our annotation pipeline substantially reduces the number of missed objects across all evaluated datasets.

\input{tables/label_error_table_main}

\subsection{Label Errors in Original GT}
To analyze annotation differences, we compare the validated GT with the original annotations. This comparison allows us to identify label errors and categorize discrepancies between these two annotation sets. Based on this, we distinguish three types of label errors: missing bounding box, misaligned bounding boxes, and classification errors. 

\textit{Missing bounding boxes} correspond to instances that are present in the validated annotations but not labeled in the original data. \textit{Misaligned bounding boxes} refer to cases where an object is annotated in both versions but the localization is inaccurate. \textit{Classification errors} occur when overlapping bounding boxes are assigned different class labels. Each label error is represented by a bounding box and an associated soft label probability $p$. We denote the set of all label errors as the label error ground truth (LEGT). We define two different configuration of LEGT: a less restrictive setting (all objects, confidence $p\geq 0.5$) and a stricter setting (objects $\geq 40$ pixels, confidence $p\geq 0.8$). These configurations capture both a comprehensive view of potential errors and a subset of  high-confidence cases. 
Further details on matching and configurations are provided in the appendix. 

\autoref{tab:label_error_table_main} summarizes the distribution of label errors. Across all datasets, missing bounding boxes clearly dominate, followed by misaligned boxes, while classification errors remain rare (typically below 1\%). Datasets such as COCO and KITTI show particularly high numbers of missing annotations. Applying the stricter setting substantially reduces the number of detected label errors, most strongly for missing boxes (e.g., on COCO from 16,219 to 2,685), while misaligned boxes decrease more moderately. This indicates that a large number of label errors is associated with small objects. Overall, this consistent pattern across datasets highlights missing annotations as the primary challenge for label quality.

\input{tables/data_quality_bias}

\subsection{Bias Correction} Annotator responses can be biased toward confirming a proposed class, leading to overconfident soft label probabilities for this class~\cite{schmarje2023annotating,schmarje2023cleverlabel}. To quantify this effect, we collect a small set of unbiased annotations without class proposals and compare them to the standard annotation setting (with proposals). \autoref{fig:introduced_bias} shows the differences of the probabilities for the proposed class with proposal (biased) and without proposal (unbiased).  
A confirmation bias is indicated by a positive probability difference.
We observe that the potential proposal bias is dataset- and class-dependent: while it is evident, e.g., for Pascal VOC and COCO kitchen superclass, it is weak or even absent in KITTI. This demonstrates that annotation bias is not uniform and depends on visual ambiguity and class distinctiveness.
To mitigate this effect, we apply a post-processing method from~\cite{schmarje2023cleverlabel}. \autoref{tab:post_processing_results} summarizes the results, showing a consistent reduction in the probability for the proposed class with relative improvements of up to $\sim$50\% in the proposed-class bias and corresponding reductions in Jensen–Shannon distance~\cite{Fuglede2004JensenShannonDA,liese1987convex} between annotations with and without proposal.
Jensen-Shannon distance is $\mathrm{\sqrt{JS}}(P \,\|\, Q) = \sqrt{\frac{1}{2} \mathrm{KL}(P \,\|\, M) + \frac{1}{2} \mathrm{KL}(Q \,\|\, M)}$ where $M = \frac{1}{2}(P + Q)$ and KL the Kullback-Leibler divergence~\cite{Kullback1951}. These improvements indicate that post-processing moves the observed distributions closer to unbiased human judgments. However, larger effects are only visible for datasets where we saw before in~\autoref{fig:introduced_bias} that an actual bias exists.
Full quantitative results across datasets and implementation details of the correction procedure are provided in the appendix.
Overall, this step ensures that the resulting soft labels more accurately reflect human uncertainty.

\input{tables/data_quality_calibration_wrongclass}

\subsection{Quality of Soft Labels}
We evaluate the quality of soft labels by comparing them to expert judgments.
For each dataset, three expert annotators assessed randomly sampled objects regarding the probability for a given class. Results were averaged and compared to the crowdsourced soft labels, see appendix for details.
As shown in \autoref{tab:human_perception}, soft label probabilities align well with expert perception across all datasets. Clear cases are typically rated around 0.75–0.85 (expected $\geq 0.75)$, ambiguous cases around 0.4–0.5 (expected $0.25-0.75)$, and incorrect cases around 0.15–0.3 (expected $< 0.25$). This consistently matches expected ranges reported in prior work~\cite{rechecked}, indicating that the constructed soft labels provide a realistic representation of human uncertainty.

\subsection{Classification Errors in Validated GT}
We estimate the frequency of incorrect class labels for original and validated annotations in \autoref{tab:wrong_label}. The full methodology, results and visual examples are provided in the appendix. Across datasets, severe classification errors remain low for both GT sets, typically in the range of $\sim$1–4\%. Slightly higher rates in the validated annotations likely reflect stricter alignment with expert judgments rather than reduced quality, and differences are not statistically significant.

\section{Object Detection Benchmark}
\label{sec:object detection benchmark}

\input{tables/map_recall_object_detectors}

\subsection{Performance of Object Detectors on Original and Validated Data}
We compare performance of state-of-the-art object detectors on original and validated GT to analyze, how GT quality affects benchmark results. 
We evaluate single-state detectors such as YOLOv8~\cite{yolov8_ultralytics} and RT-DETR~\cite{lv2023detrs}, Faster R-CNN~\cite{fasterRCNN} as a two-stage detector, as well as Grounding DINO~\cite{liu2023grounding}, YOLO-World~\cite{cheng2024yoloworld}, and Owl-ViT~\cite{minderer2022owlvit} as open-vocabulary object detectors (OVODs). 
Performance is measured in terms of mean average precision (AP), following the COCO evaluation protocol~\cite{COCO}.
To compute AP, we derive hard labels from soft labels as described in \ref{subsec:appendix-descriptive statistics} with threshold $0.8 \leq p$.

Across all datasets, we show in \autoref{tab:detection_org_validated} that object detection models trained on original GT generally exhibit lower performance when evaluated on validated GT, reflecting the increased difficulty due to more complete annotations. On COCO, AP decreases consistently across models (e.g., Grounding DINO from 56.9\% to 52.1\%), primarily due to reduced recall as additional small and previously unlabeled objects are introduced. Model rankings remain largely stable, indicating that annotation quality mainly affects absolute performance. Similar performance drops are observed on Pascal VOC. In contrast, KITTI shows a mixed pattern: OVODs improve, likely because newly annotated objects convert previous false positives into true positives, while fine-tuned models experience substantial degradation, suggesting a bias toward the original training distribution. On Cityscapes, where annotation differences are smaller, performance slightly improves. Notably, since evaluation is performed under the strict setting, these effects are not driven by ambiguous or low-confidence cases but reflect systematic differences in annotation completeness and object coverage. For a less strict evaluation ($p\geq 0.5)$, performance drops are slightly larger~(see \ref{sec: appendix - mAP results on other datasets}). Overall, the results highlight that benchmark outcomes strongly depend on annotation quality and validated GT exposes limitations in recall of current models, particularly for small objects which were previously missing in the data.

\begin{figure}[htbp]
\begin{minipage}[t]{0.6\linewidth}
\vspace{0pt}
\setlength{\tabcolsep}{4pt}
\centering
\captionof{table}{Soft label-based evaluation of object detector performance on COCO in terms of Average Similarity (AS), mean total variation distance (TVD) and Jensen-Shannon distance ($\sqrt{\mathrm{JS}}$) for all predictions and only those predictions that matched with a GT object (TP). Values are in \%.}
\label{tab:soft_label_based_benchmark}
\resizebox{\linewidth}{!}{
\begin{tabular}{cccccccc}
\toprule
{Model} 
& AS\textsubscript{50..95}\,$\uparrow$ & AS\textsubscript{S}\,$\uparrow$  & AS\textsubscript{L}\,$\uparrow$  & TVD\,$\downarrow$ & TVD\textsubscript{TP}\,$\downarrow$ & $\sqrt{\mathrm{JS}}\,\downarrow$ & $\sqrt{\mathrm{JS}}$\textsubscript{TP}\,$\downarrow$ \\
\midrule
Grounding DINO & \textbf{\percent{0.5508}} & \textbf{\percent{0.3842}} & \textbf{\percent{0.7875}} & \textbf{\percent{0.1198}} & \percent{0.3823} & \textbf{\percent{0.1636}} & \percent{0.3495} \\
YOLOv8-World v2 & \percent{0.4808} & \percent{0.3060} & \percent{0.7032} & \percent{0.1641} & \percent{0.3758} & \percent{0.2230} & \percent{0.3570} \\
OWL-ViT & \percent{0.3583} & \percent{0.1402} & \percent{0.6781} & \percent{0.1386} & \percent{0.7524} & \percent{0.1845} & \percent{0.6305} \\
RT-DETR-L & \percent{0.4910} & \percent{0.3226} & \percent{0.7454} & \percent{0.1406} & \percent{0.4688} & \percent{0.1934} & \percent{0.4397} \\
YOLOv8-L & \percent{0.4664} & \percent{0.3032} & \percent{0.6779} & \percent{0.1756} & \percent{0.3312} & \percent{0.2374} & \percent{0.3170} \\
Faster R-CNN (ResNet 50) & \percent{0.4395} & \percent{0.2965} & \percent{0.6129} & \percent{0.1815} & \textbf{\percent{0.2468}} & \percent{0.2190} & \textbf{\percent{0.2552}} \\
\bottomrule
\end{tabular}
}
\end{minipage}
\begin{minipage}[t]{0.37\linewidth}
\vspace{0pt}
\centering
\includegraphics[width=0.9\linewidth]{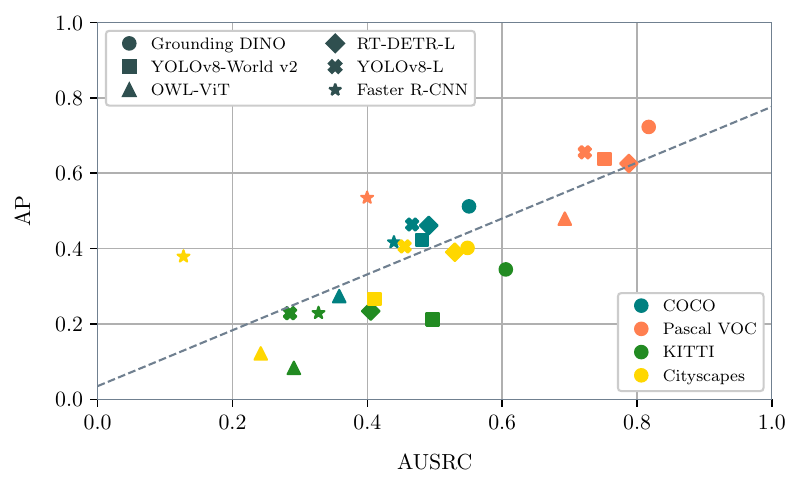}
    \captionof{figure}{Comparison of AP and AS across all datasets and models. Pearson's correlation coefficient is $0.78$.}
    \label{fig:correlation_AP_AUSRC}
\end{minipage}
\end{figure}

\subsection{Does Model Uncertainty Align with Human Perception?}

We evaluate how well predictions of current object detectors align with human perception by comparing predicted class probabilities with soft label annotations. 
The task therefore combines object detection with modeling human uncertainty.
Uncertainty-aware evaluation is enabled by the availability of soft labels in the validated ground truth.
For evaluation, we propose the Area under the Similarity-Recall Curve (AUSRC), or Average Similarity (AS). 
AS generalizes AP to soft labels by replacing binary precision with a continuous similarity measure to compare predicted and ground truth distributions. AS therefore captures both detection performance and alignment with human uncertainty.
We use $1-$ total variation distance (TVD) as a similarity measure, comparing unmatched predictions to background and sorting predictions by entropy, see~\ref{sec: appendix - AUSRC Definition} for more details on the metric definition. 
IoU-based evaluation follows standard AP procedure, averaging across thresholds and object sizes. 
We additionally report average TVD and Jensen-Shannon distance ($\sqrt{\mathrm{JS}}$) of predictions and GT.
For evaluation, we require a predicted probability distribution over all classes, including background. While some models provide this directly, others output independent class probabilities. In that case, we first calibrate predicted scores through isotonic regression and construct a background probability as the product of complementary class probabilities (see~\ref{sec: appendix - predictions processing}).

Resulting performance metrics on COCO are listed in Table~\ref{tab:soft_label_based_benchmark} and provided in~\ref{sec: appendix - additional soft label benchmark results} for other datasets.
Across all datasets and models, AS correlates with AP (cf.\ \autoref{fig:correlation_AP_AUSRC}), indicating that it meaningfully measures detection performance. Like AP, AS increases with object size and decreases at higher IoU thresholds.
However, AP shows no correlation with TVD ($r=-0.02$) or $\sqrt{\mathrm{JS}}$ ($r=-0.02$), whereas AS correlates with both ($r=-0.58$, $-0.60$), showing that AP fails to capture distributional alignment while AS does.
Distance measures indicate discrepancies between predictions and soft labels across models and datasets. In particular, correctly localized (TP) detections deviate from soft labels as shown by high TVD values of over 25\% on COCO and JS distance values well above the observed value of around 0.1 between human annotators in the bias correction analysis (cf.~\ref{sec: appendix - data quality - bias correction}).
These values provide a rough reference range for human-level agreement, highlighting the gap to model predictions.
Unmatched predictions are often of low confidence, resulting in small distances when compared to background.
$\sqrt{\mathrm{JS}}\textsubscript{TP}$, restricted to true positives and thus, directly comparable to the bias correction setting, remains even higher, confirming worse alignment for matched detections.
Notably, Faster R-CNN, the only model explicitly predicting background probabilities, achieves best alignment for matched predictions, while Grounding DINO performs best overall.
Preliminary experiments with soft label training slightly improve calibration and alignment, but gains remain limited due to current loss functions being optimized for hard labels (see~\ref{subsec: appendix - object detection benchmark - soft label training}).
Overall, beyond the proposed evaluation protocol, the availability of soft labels enables proper uncertainty quantification and future work towards aligning model uncertainty with human perception.

\input{labelerrors/main/plot_main_label_errors}
\input{tables/label_error_benchmark_table_main}

\section{Label Error Detection Benchmark}
\label{sec: label error detection benchmark}

In object detection, label error detection methods use object detector predictions in combination with the original annotations to generate label error proposals, identifying inconsistencies such as missing bounding boxes, misaligned bounding boxes, or incorrect class labels. We construct a large-scale label error detection benchmark based on real label errors using the validated GT. We evaluate loss-based instance-wise scoring~\cite{loss-based-method}, ObjectLab~\cite{objectlab}, and MetaDetect~\cite{metadetect}, and compare them to two baselines: (i) a naive baseline that reviews all false positive (FP) predictions with respect to the original GT, and (ii) a score baseline that ranks FPs by detector confidence. We use Grounding DINO as the underlying detector, while the loss-based method uses Cascade R-CNN due to architectural constraints. Each method produces label error proposals, which are matched to the less restrictive label error ground truth ($p\geq 0.5$) using an IoU threshold of 0.1 (examples of label errors in \autoref{fig:main_label_errors_plot}). This low threshold reflects a high-recall setting, where partial detections are sufficient to guide manual review. We report performance at the threshold that maximizes the F1-score, and additionally report average precision, computed as the area under the precision-recall curve (\autoref{tab:label_error_benchmark_main}). 
AP evaluates ranking quality across thresholds, while F1-score reflects performance at a fixed operating point. In practical applications, label error detection methods require selecting a threshold to determine which proposals are represented for review. We therefore additionally report the maximum F1-score to asses performance under such a threshold-based selection.

The naive baseline achieves near-perfect recall across datasets but suffers from low precision (around 0.4–2\%), making it impractical. Label error detection is therefore a prioritization problem: the goal is to detect as many label errors as possible while minimizing the number of samples that require manual review. The score baseline performs best across datasets, achieving the highest F1-scores (e.g., 59.3\% on KITTI and 43.5\% on COCO) and strong AP (up to 59.7\%), indicating that detector confidence provides a strong signal for ranking annotation inconsistencies. Loss inspection achieves moderate recall (e.g., up to 41.5\%) but low precision (often below 20\%), resulting in many false positives. ObjectLab provides a more balanced trade-off (e.g., F1-score up to 36.9\% on KITTI), but remains below the score baseline. MetaDetect also remains below the score baseline across datasets, with F1-scores ranging from 24.3\% to 36.1\%, suggesting that its IoU-based signal provides limited additional benefit in this setting. 
Compared to prior work that primarily evaluates on synthetic noise~\cite{loss-based-method,objectlab}, all methods seem to struggle to accurately detect real label errors, highlighting a significant gap between synthetic and real-world label errors.

\section{Limitations, Broader Impact \& Conclusion}
\label{sec: Limitations}

\paragraph*{Limitations} While we significantly improved label quality on all four datasets, there is no perfect data.
We ensure annotation quality with soft labels that are based on at least 11 annotator responses, however, these responses are subject to annotator reliability and confirmation bias exists.
This results in errors even in this data.
For most evaluation, we adapt current hard label based methods and benchmarks to incorporate the uncertainty of soft labels. 
E.g., we model a class distribution given the raw predictions and post-hoc calibration.
Our work argues that model architectures should incorporate soft labels as object detection datasets are full of ambiguous cases.
While our soft labels capture class-label uncertainty, they do not model spatial uncertainty in bounding box locations. Modeling such uncertainty, e.g., through spatial distributions, would require multiple independent bounding box annotations per object and was therefore beyond the scope and annotation budget of this work.

\paragraph*{Broader Impact} We take public datasets and provide high quality labels for these datasets. We see no direct broader negative impact on the society aside from general issues arising from the usage of object detection models, e.g., for the tracking and surveillance of humans.

\paragraph*{Conclusion} We show that object detection benchmarks are fundamentally limited by annotation incompleteness, with missing objects as the main source of label errors. By introducing high-recall, uncertainty-aware validated GT, we enable a systematic analysis of annotation quality, model evaluation, and uncertainty alignment. 
Our object detection benchmark based on this validated GT reveals that benchmark performance is highly dependent on annotation quality and exposes a clear mismatch between model confidence and human perception. 
Furthermore, our label error detection benchmark based on the validated GT shows that current models struggle to generalize from synthetic to real-world errors.
Collectively, these findings highlight the need to move beyond deterministic annotations toward high-recall, uncertainty-aware benchmarks for object detection.

\section*{Acknowledgment}
\noindent
S.P., J.K.\ and M.R.\ acknowledge support by the German Federal Ministry of Research, Technology and Space (BMFTR) within the project ``RELiABEL'' (grant no.\ 16IS24019B). L.S.\ also acknowledges support by the BMFTR within the same project ``RELiABEL'' (grant no.\ 16IS24019A).

\input{main.bbl}

\section*{Appendix Overview}

Our appendix complements the main paper with details on dataset construction, experimental setup, dataset statistics, and post-processing of annotations. It includes additional numerical results and visual examples. To facilitate the navigation of the appendix, we give a brief overview below:
\begin{itemize}[left=0pt]

    \item Appendix~\ref{sec: appendix - Dataset Details} describes the annotation process for dataset construction (see \autoref{sec:dataset_construction} in the main paper). It includes:
    \begin{itemize}[left=0pt]
        \item Additional details on the annotation pipeline and task design~(\ref{sec: appendix - dataset details - microtask design}).
        \item Annotation time breakdown and scalability analysis of the proposed annotation pipeline~(\ref{sec: appendix - dataset details - scalability analysis})
        \item Quality control measures to improve annotation quality~(\ref{sec: appendix - dataset details - quality measures}).
        \item The procedure for duplicate removal through manual inspection~(\ref{sec: appendix - dataset details - duplicate removal}).
        \item Details on the re-annotated datasets and label policies~(\ref{subsec:appendix-useddatasets}, \ref{subsec: appendix - dataset details - label policies}).
    \end{itemize}

    \item Appendix~\ref{sec: appendix - data quality assessment} reports additional results on data quality (see \autoref{sec:data_quality} in the main paper), including:
    \begin{itemize}[left=0pt]
        \item Comparison of original and validated annotations, including dataset statistics and a per-dataset analysis of label errors~(\ref{subsec:appendix-descriptive statistics}).
        \item Visual examples and full results for missing objects~(\ref{appendix:missing}).
        \item Refinement of soft labels for ambiguous cases~(\ref{appendix:refine_ambiguous}).
        \item Definition of label errors and their distribution across datasets, including special conditions such as \textit{don't care} regions~(\ref{sec:appendix - data quality - label error}).
        \item Analysis of original-only annotations and their impact~(\ref{sec: appendix- data quality - orgOnly}).
        \item Bias correction for soft labels, including methodology and results across datasets~(\ref{sec: appendix - data quality - bias correction}).
        \item Alignment of soft labels with expert judgments per dataset~(\ref{sec: appendix - data quality - calibration of soft labels}).
        \item Wrong label analysis and corresponding visual examples~(\ref{sec: appendix - data quality - wrong class}).
    \end{itemize}

    \item Appendix~\ref{sec: appendix - Additional Results} complements the main experiments of the object detection benchmark with additional results and experimental details (see \autoref{sec:object detection benchmark} in main paper), including:
    \begin{itemize}[left=0pt]
        \item Definition of the proposed Area under Similarity-Recall Curve (AUSRC)~(\ref{sec: appendix - AUSRC Definition}).
        \item Training details for fine-tuning object detectors~(\ref{sec: appendix - training details}).
        \item Performance in terms of AP on original vs.\ validated annotations across datasets~(\ref{sec: appendix - mAP results on other datasets}).
        \item Prediction processing, including post-hoc calibration and extraction of class probabilities~(\ref{sec: appendix - predictions processing}).
        \item Additional results for soft label-based evaluation~(\ref{sec: appendix - additional soft label benchmark results}).
        \item Comparison of soft label vs.\ hard label-based training for YOLOv8 and RT-DETR~(\ref{subsec: appendix - object detection benchmark - soft label training}).
        \item Cross-evaluation for training on original and refined annotations~(\ref{subsec: appendix - object detection benchmark - training cross evaluation})
        \item Visual examples for the divergence between human soft labels and model predictions~(\ref{subsec: appendix - object detection benchmark - visual examples})

    \end{itemize}

    \item Appendix~\ref{appendix: label error benchmark} provides a detailed analysis of the label error detection benchmark (see \autoref{sec: label error detection benchmark} in main paper), including:
\begin{itemize}[left=0pt]
    \item Description of the evaluated methods and how label error proposals are generated~(\ref{appendix: label error benchmark - methods}).
    \item Details on the prediction networks used to generate proposals and their characteristics~(\ref{appendix: label error benchmark - prediction networks}).
    \item Results on a stricter label error ground truth focusing on larger and high-confidence errors~(\ref{appendix: label error benchmark - stricter legt}).
    \item Precision-recall analysis showing how methods differ in ranking and recall behavior~(\ref{appendix: label error benchmark - pr curves}).
    \item Comparison using Cascade R-CNN predictions for a consistent evaluation across methods~(\ref{appendix: label error benchmark - cascade}).
    \item Analysis of how the choice of prediction network and practical implications~(\ref{appendix: label error benchmark - choice prediction}).
    \item Visual examples illustrating typical label errors across datasets~(\ref{appendix: label error benchmark - examples}).
\end{itemize}
\end{itemize}

\appendices
\renewcommand{\thesubsection}{\Alph{section}.\arabic{subsection}}
\renewcommand{\theparagraph}{\thesubsection.\alph{paragraph}}

\makeatletter
\renewcommand{\@seccntformat}[1]{%
  \csname the#1\endcsname\quad
}
\makeatother

\section{Dataset Construction}
\label{sec: appendix - Dataset Details}

\input{tables/microtask_parameters_table}

\subsection{Micro Task Design}
\label{sec: appendix - dataset details - microtask design}

\begin{figure}[htbp]
    \centering
    \begin{minipage}[t]{0.48\linewidth}
        \centering
        \includegraphics[width=\linewidth]{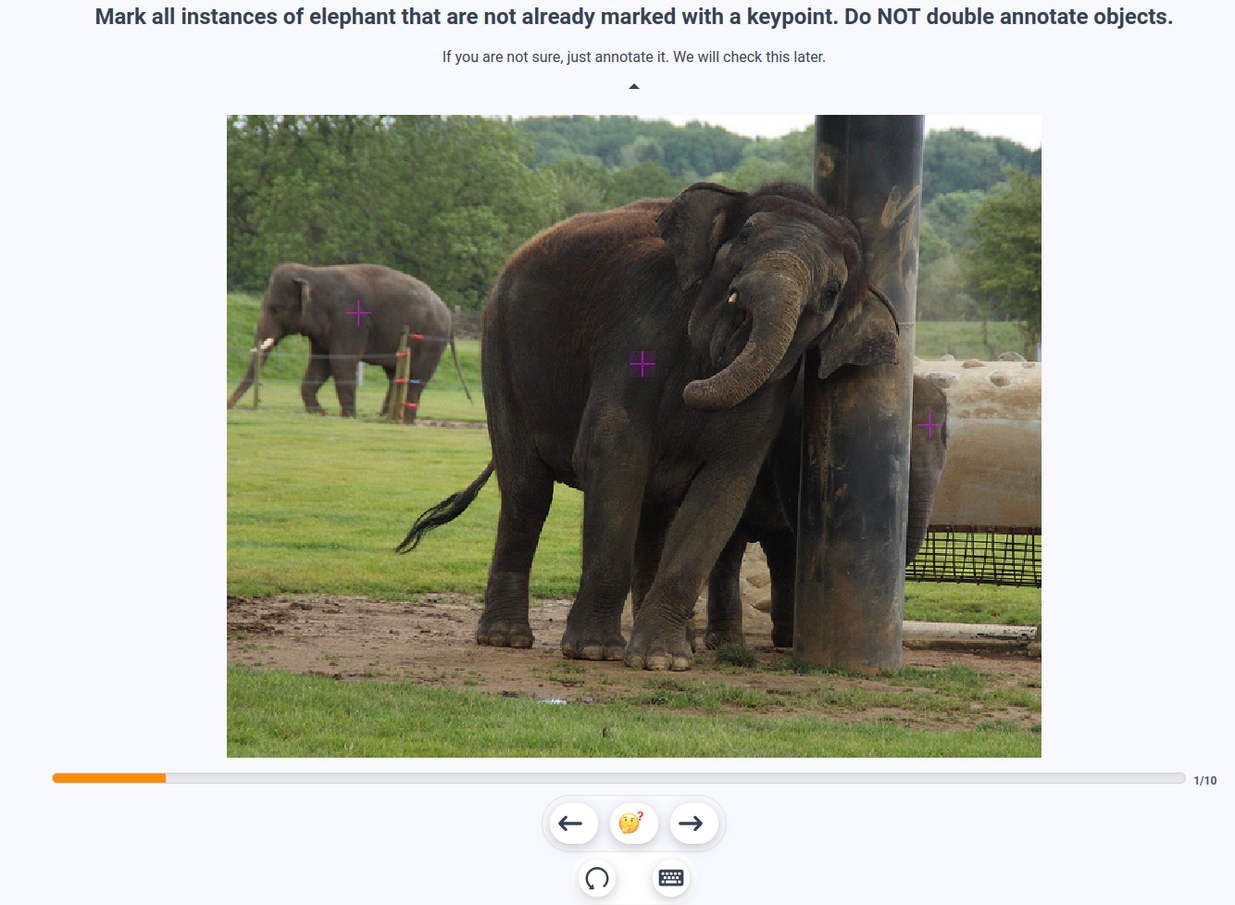}
        \caption{Annotator interface for the keypoint task on a COCO image, showing already placed keypoints.}
        \label{fig:keypoint_hari}
    \end{minipage}
    \hfill
    \begin{minipage}[t]{0.48\linewidth}
        \centering
        \includegraphics[width=\linewidth]{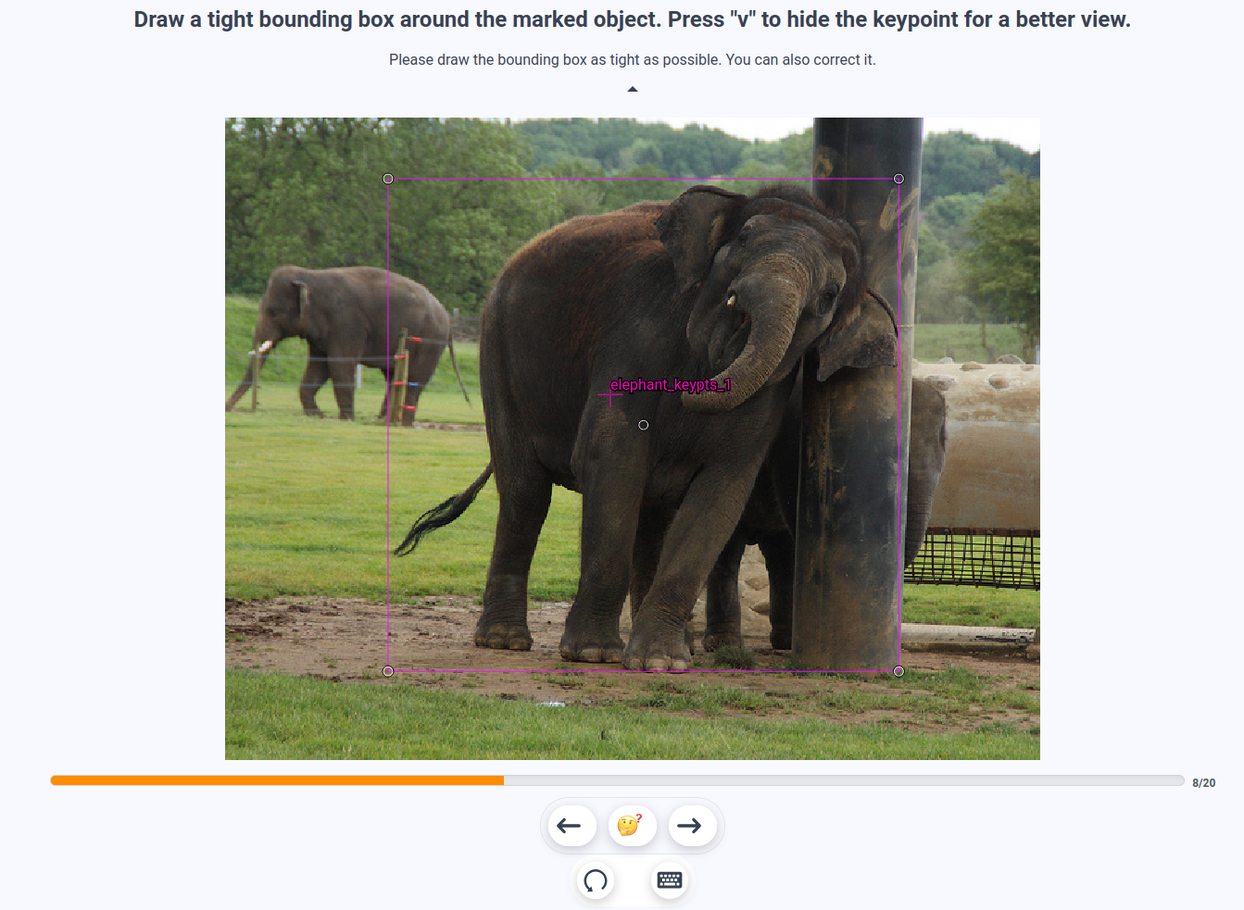}
        \caption{Annotator interface for the bounding box task on a COCO image, showing an already drawn bounding box.}
        \label{fig:bbox_hari}
    \end{minipage}
\end{figure}

\begin{figure}[htbp]
    \centering
    \includegraphics[width=0.5\linewidth]{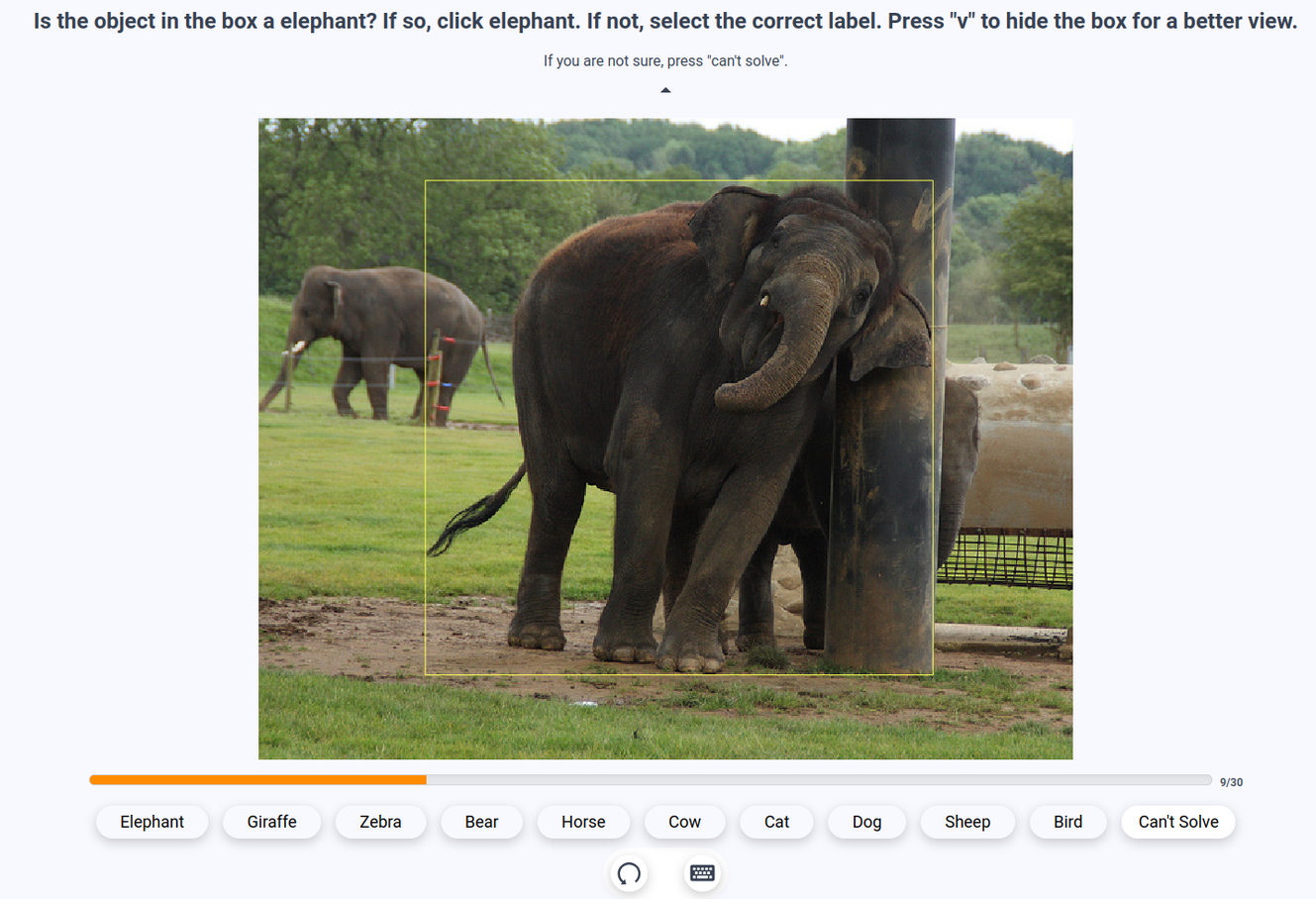}
    \caption{Annotator interface for soft label task on a COCO image.}
    \label{fig:softlabel_hari}
\end{figure}

Our annotation pipeline follows a multi-stage design that decomposes the complex annotation task into a sequence of simpler subtasks. This design is motivated by our experience with large-scale crowdsourcing projects and follows a divide-and-conquer strategy to reduce the complexity of individual annotation steps. In particular, identifying object locations via keypoints is considerably easier than directly drawing bounding boxes and helps annotators focus on finding all relevant object instances. The subsequent annotation stages then refine these object candidates by obtaining bounding boxes and collecting redundant class annotations. This decomposition simplifies the individual annotation tasks while promoting high annotation coverage and reliable soft labels.

To implement this multi-stage design, the annotation workflow is structured into a set of microtasks, each defined by a task-specific question and an additional description to guide annotators. In addition to the textual instructions, several parameters are configured to control the annotation process. An overview is provided in \autoref{tab:microtask_design}. The number of \textit{repeats} determines how many independent annotators perform the same task, enabling redundancy and improving annotation quality. The \textit{working package} specifies the number of items assigned per task instance. Depending on the task type, this corresponds to images (for keypoint annotation), individual keypoints (for bounding box annotation), or bounding boxes (for soft label classification). The time limit defines the maximum number of seconds available to complete one working package.

For the keypoint task (\autoref{fig:keypoint_hari}), we employ three repeats to ensure that each image is reviewed by several annotators, increasing the likelihood of detecting all object instances of a given class. Annotators can see previously placed keypoints and are explicitly asked to add any missing ones. The bounding box task (\autoref{fig:bbox_hari}) is comparatively simpler, as annotators are presented with a single keypoint at a time and are required to draw a corresponding bounding box. This task is repeated to ensure that no keypoints are missed and to improve the quality of the resulting bounding boxes. The class associated with the keypoint is displayed to provide additional context, which is particularly helpful in crowded scenes or for small objects where the correspondence between keypoint and object may be ambiguous. For the soft label task (\autoref{fig:softlabel_hari}), we use a higher number of repeats, with eleven annotators evaluating each instance. This redundancy enables the estimation of reliable soft labels by aggregating multiple independent class annotations, thereby capturing annotation uncertainty. In the interface, the class associated with the bounding box is presented first, as it is the most likely label. Annotators can select from all dataset classes as well as the additional option \textit{can't solve}, which is used when the object is ambiguous or cannot be reliably assigned to a class.

The Cityscapes training set was annotated using a slightly different procedure. The original ground truth bounding boxes were displayed during the keypoint annotation stage. Annotators were therefore only required to mark additional object instances that belong to the respective class but were not yet present in ground truth annotations. We generate soft labels for additional objects and the objects of the original annotations. 

\subsection{Scalability Analysis of the Annotation Pipeline}
\label{sec: appendix - dataset details - scalability analysis}

\input{tables/scalability_analysis.tex}

To provide an estimate of the human annotation effort required by our validation pipeline, \autoref{tab:annotation_times} reports the observed average annotation times for the individual annotation microtask. The reported times are derived from the recorded annotation times and normalized according to the respective annotation procedure.

For the keypoint task, each image is reviewed class-wise by three independent annotators. The reported keypoint time therefore represents the average time required by one annotator to review one image for one particular class and, if corresponding objects are present, annotate them with keypoints. Thus, this value captures both the inspection of images without objects of the respective class and the placement of keypoints when objects are present. 

For the subsequent bounding box task, the goal is to obtain a bounding box for every identified keypoint. For each keypoint, the annotator either draws a new bounding box or, if a box is already available from a previous annotation, verifies and corrects it if necessary. Since the recorded annotation times do not distinguish between drawing, correcting, and accepting a bounding box, we report the average time per bounding box annotation/review. Consequently, this value reflects the practical annotation effort of the bounding box microtask rather than the isolated time required to draw a new bounding box. 

For soft label annotation, the annotation time of each individual response is recorded directly. We therefore report the average time per response as well as the average cumulative annotation time for objects receiving 11, 22, 33, or 44 votes. The latter is obtained by summing the recorded response times for each object and averaging these cumulative times across objects with the corresponding number of votes. It therefore represents the average annotation effort required to create a soft label with a given number of responses. No annotation-time outliers are removed. All reported values reflect the observed annotation process and should be interpreted as approximate estimates of human annotation effort.

Across all datasets, the individual annotation steps can be completed within a few seconds on average. Image-class reviews during keypoint annotation take $1.29$--$3.47$\,s on average, while bounding box annotation/review requires $6.29$--$11.59$\,s. Individual soft label responses are similarly fast, requiring $1.73$--$2.28$\,s on average. Consequently, collecting standard 11 responses results in an average cumulative annotation time of $16.50$--$21.11$\,s per object across datasets. The average cumulative time increases to $46.76$--$53.58$\,s for 22 votes and reaches $100.74$--$114.04$\,s for 44 votes. The cumulative time does not scale linearly because additional evaluations were conducted only for ambiguous cases which take longer to evaluate. Thus, additional annotation effort is invested for ambiguous objects, while the standard 11-vote setting requires comparatively limited annotation time.

Overall, these measurements provide a practical breakdown of the human annotation effort associated with the individual stages of our pipeline and indicate how the soft label annotation effort scales with the number of collected responses.

\begin{figure}[htbp]
    \centering
    \includegraphics[width=0.5\linewidth]{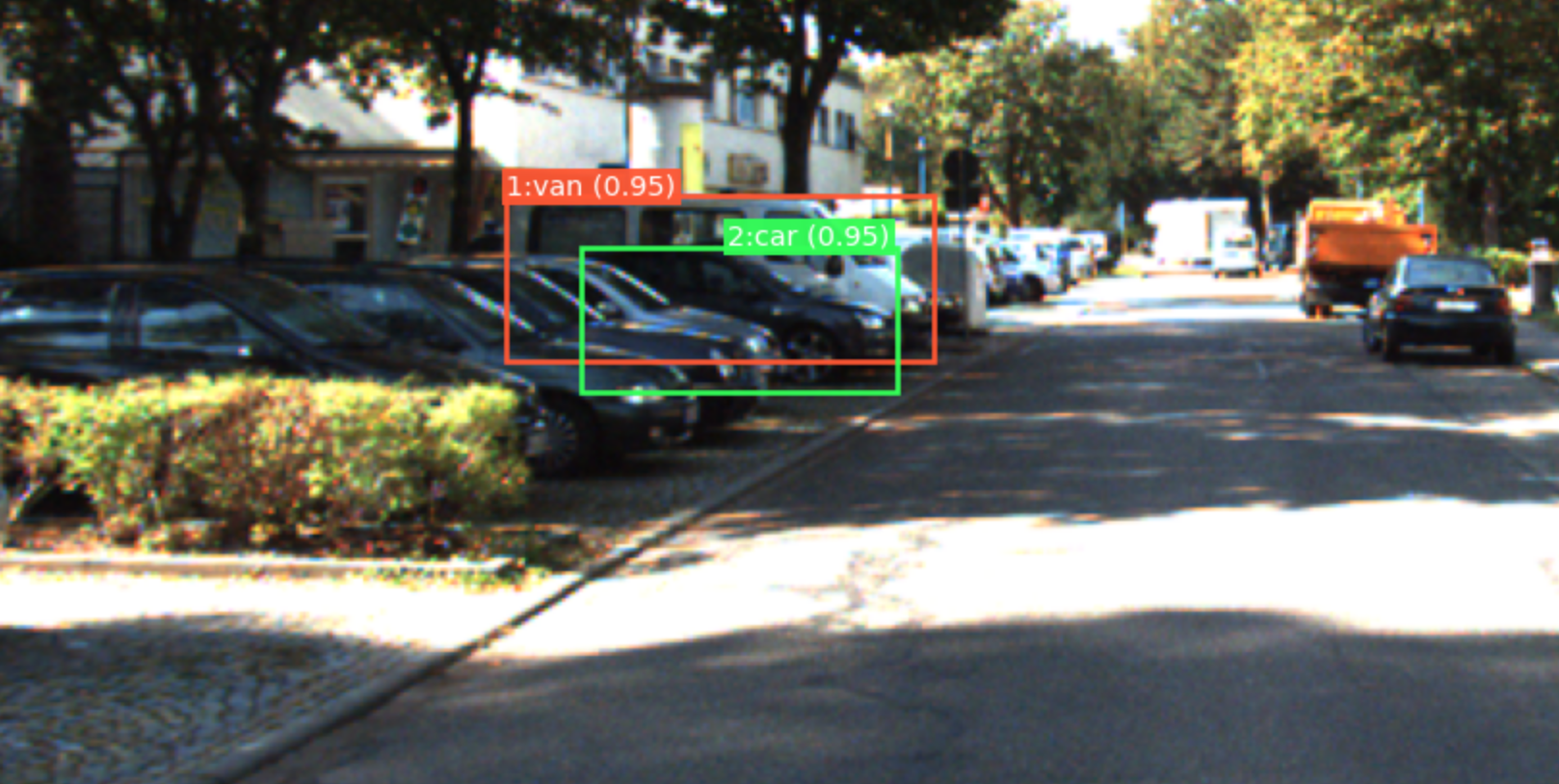}
    \caption{Interface for duplicate removal. Each pair of overlapping bounding boxes is displayed together with the information of the most voted class. The user can decide whether to keep both boxes or merge one of the boxes with the other if they form a duplicate. In this case, the van and the car are two separate objects so both bounding boxes are kept. Candidate pairs are displayed sorted by their IoU and reviewed until convergence.}
    \label{fig:duplicate_removal_interface}
\end{figure}

\subsection{Manual Quality Control and Completion}
\label{sec: appendix - dataset details - quality measures}

Throughout all stages of the re-annotation process, we continuously monitored and improved annotation quality through additional verification and refinement steps. Following the bounding box stage, we conducted systematic expert reviews to further improve annotation completeness and accuracy. Depending on dataset characteristics, different review strategies were applied. For datasets with a smaller number of classes, such as KITTI, images were reviewed jointly across classes to identify missing annotations. For more complex datasets with many classes or dense scenes, such as COCO and Cityscapes, the review was performed class-wise and on selected subsets of images. In these cases, we restricted the review to images that contained at least one ground truth annotation or at least one keypoint. For each class, the existing bounding boxes were visualized, and the images were inspected to identify any additional object instances of the respective class that had not yet been annotated. In addition to detecting missing instances, we also refined bounding box quality. In cases where bounding boxes were not sufficiently tight or multiple boxes were assigned to the same objects, annotations were corrected. Duplicate boxes resulting from multiple annotations were removed in a dedicated post-processing step. Overall, these additional quality control and refinement steps ensure a high level of annotation completeness and consistency, which is essential for constructing a reliable validated ground truth.

\begin{figure}[htbp]
    \centering
    \begin{subfigure}{0.46\textwidth}
        \centering
        \includegraphics[width=\linewidth]{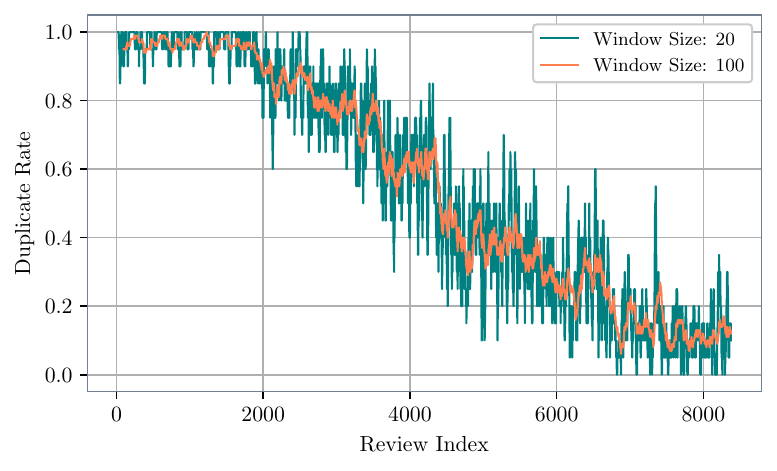}
        \caption{Duplicate rate as a moving average of previous reviewed cases with two different window sizes.}
    \end{subfigure}
    \hfill
    \begin{subfigure}{0.46\textwidth}
        \centering
        \includegraphics[width=\linewidth]{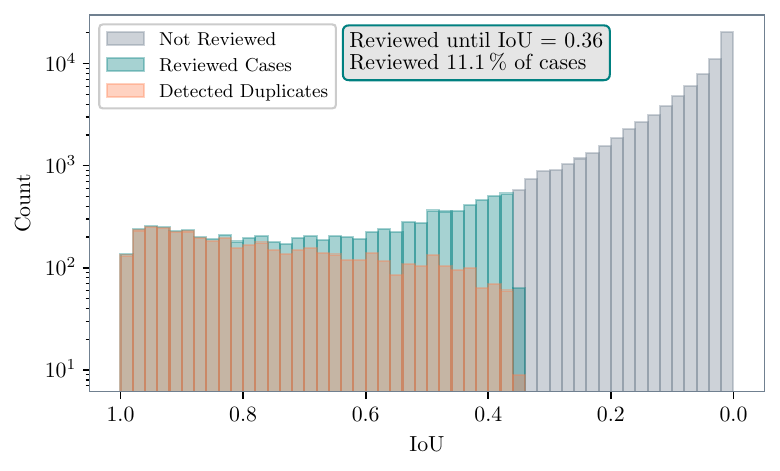}
        \caption{IoU distribution of overlapping bounding box pairs. Colors indicate reviewed cases and duplicates found.}
    \end{subfigure}
    \caption{Monitoring results for duplicate removal in COCO. In (a), duplicate rate drops as more cases are reviewed and resolved. In (b), the IoU distribution of duplicate candidates is shown.}
    \label{fig:duplicate_removal_COCO}
\end{figure}

\begin{figure}[htbp]
    \centering
    \begin{subfigure}{0.46\textwidth}
        \centering
        \includegraphics[width=\linewidth]{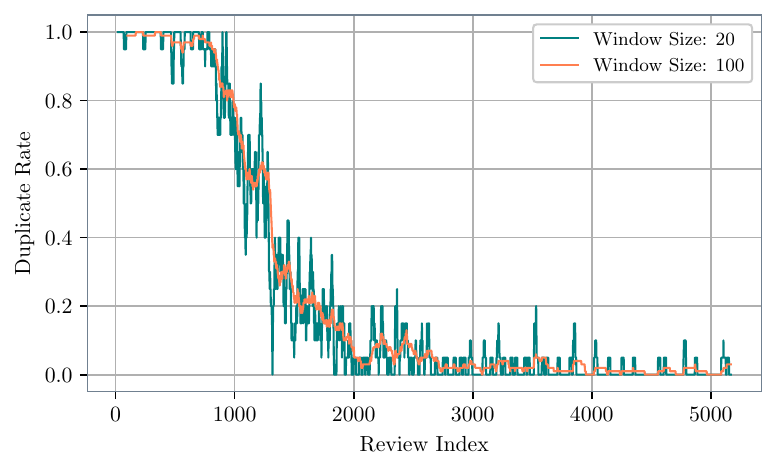}
        \caption{Duplicate rate as a moving average of previous reviewed cases with two different window sizes.}
    \end{subfigure}
    \hfill
    \begin{subfigure}{0.46\textwidth}
        \centering
        \includegraphics[width=\linewidth]{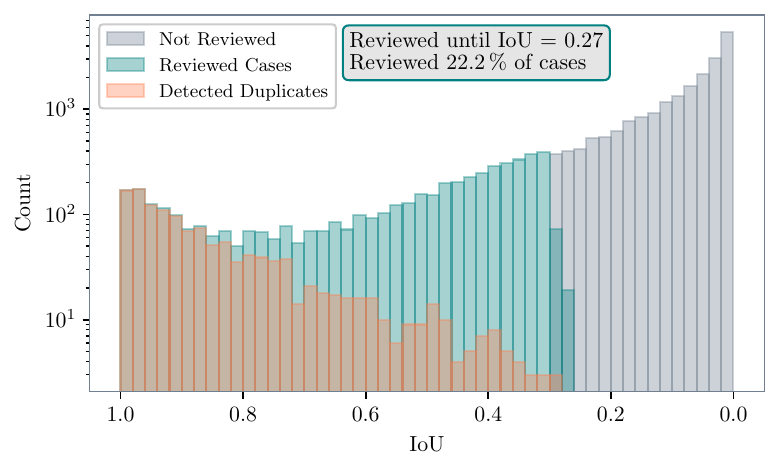}
        \caption{IoU distribution of overlapping bounding box pairs. Colors indicate reviewed cases and duplicates found.}
    \end{subfigure}
    \caption{Monitoring results for duplicate removal in Pascal VOC. In (a), duplicate rate drops as more cases are reviewed and resolved. In (b), the IoU distribution of duplicate candidates is shown.}
    \label{fig:duplicate_removal_PascalVOC}
\end{figure}

\begin{figure}[htbp]
    \centering
    \begin{subfigure}{0.46\textwidth}
        \centering
        \includegraphics[width=\linewidth]{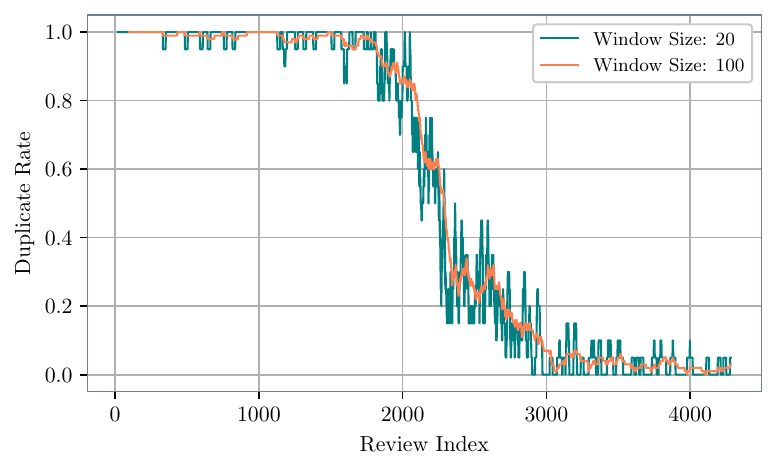}
        \caption{Duplicate rate as a moving average of previous reviewed cases with two different window sizes.}
    \end{subfigure}
    \hfill
    \begin{subfigure}{0.46\textwidth}
        \centering
        \includegraphics[width=\linewidth]{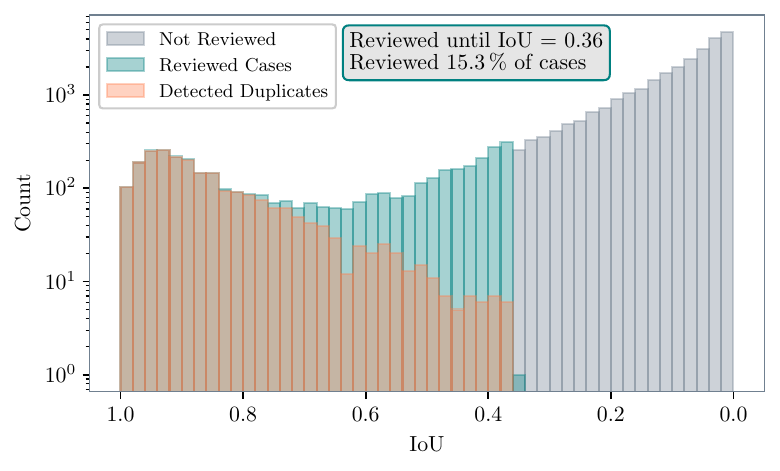}
        \caption{IoU distribution of overlapping bounding box pairs. Colors indicate reviewed cases and duplicates found.}
    \end{subfigure}
    \caption{Monitoring results for duplicate removal in KITTI. In (a), duplicate rate drops as more cases are reviewed and resolved. In (b), the IoU distribution of duplicate candidates is shown.}
    \label{fig:duplicate_removal_KITTI}
\end{figure}

\begin{figure}[htbp]
    \centering
    \begin{subfigure}{0.46\textwidth}
        \centering
        \includegraphics[width=\linewidth]{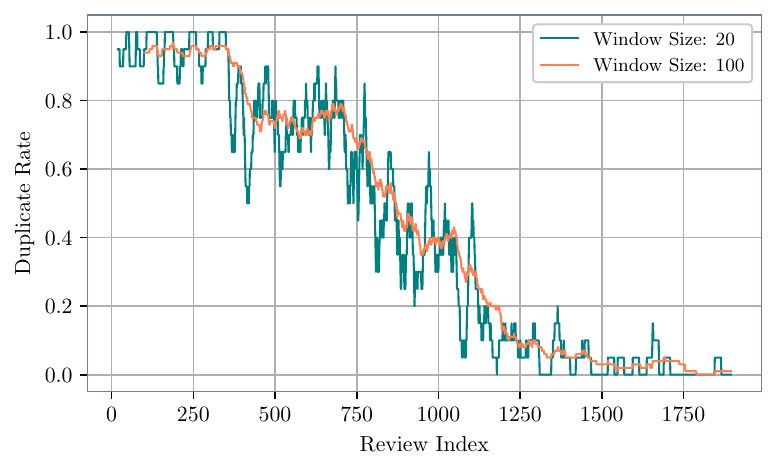}
        \caption{Duplicate rate as a moving average of previous reviewed cases with two different window sizes.}
    \end{subfigure}
    \hfill
    \begin{subfigure}{0.46\textwidth}
        \centering
        \includegraphics[width=\linewidth]{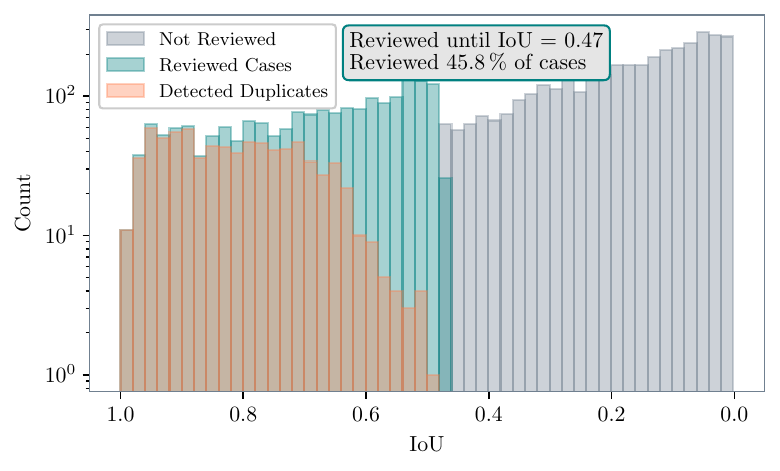}
        \caption{IoU distribution of overlapping bounding box pairs. Colors indicate reviewed cases and duplicates found.}
    \end{subfigure}
    \caption{Monitoring results for duplicate removal in Cityscapes. In (a), duplicate rate drops as more cases are reviewed and resolved. In (b), the IoU distribution of duplicate candidates is shown.}
    \label{fig:duplicate_removal_Cityscapes}
\end{figure}

\subsection{Duplicate Removal}
\label{sec: appendix - dataset details - duplicate removal}
During the annotation process, duplicates may occur for two reasons: For a given keypoint, the first annotator draws a bounding box and the second annotator adds a second one even though there exists one already. While this is not intended, such mistakes happen. The second reason is that during the keypoint stage, an object was marked for two different categories, e.g., once as a car and once as a van. This way, it receives two bounding boxes, which then form a duplicate. 
For duplicates, we manually select the better bounding box and merge the annotations of both objects to form a single soft label. Manual removal of duplicates is conducted by expert annotators who review potential duplicate cases based on IoU ranking.
More precisely, we compute IoU values for all bounding box pairs per-image and sort these duplicate candidate pairs by IoU from high to low. This way, similar bounding boxes are reviewed first and annotators decide for each pair, whether these bounding boxes indeed form a duplicate or not. If they form a duplicate, the better fitting bounding box is kept and annotations from the discarded bounding box are merged to the kept one. Throughout this process, we monitor duplicate detection rate as a moving average over reviewed cases and conduct reviewing until convergence is achieved, i.e., duplicate rate is sufficiently low. The review process is illustrated in \autoref{fig:duplicate_removal_interface} and results for the monitored duplicate detection rate are included in \autoref{fig:duplicate_removal_COCO}, \autoref{fig:duplicate_removal_PascalVOC}, \autoref{fig:duplicate_removal_KITTI}, and \autoref{fig:duplicate_removal_Cityscapes}.

\subsection{Re-Annotated Datasets}
\label{subsec:appendix-useddatasets}

For KITTI, we create an 80/20 train-validation split, as no test annotations are available.  We preserve \textit{don't care} regions in our evaluations for label error detection and exclude the \textit{misc} category due to inconsistent labeling guidelines.
For Cityscapes, bounding boxes are derived from segmentation masks to ensure comparability with the original ground truth.
For Pascal VOC 2012, we incorporate the \textit{difficult} flag to explicitly account for ambiguous or partially occluded objects.

While the annotation methodology is consistent across datasets, the selection of splits for re-annotation reflects practical annotation budget constraints and the availability of suitable annotated data. Our goal was to maximize the number of high-quality annotations within the available project budget of 5,000 annotation hours. Accordingly, we fully re-annotate the Cityscapes validation set and improve its training split by adding newly identified objects, whereas a complete re-annotation of the substantially larger COCO training set ($\sim$118,000 images) was beyond the scope of this project. For KITTI, the test annotations are internally withheld by the benchmark organizers, and we therefore follow the evaluation protocol described above.

\subsection{Label Policies and Label Provider}
\label{subsec: appendix - dataset details - label policies}

We provide the label policies used for the annotation task for all considered datasets:

\begin{itemize}
    \item Cityscapes: \url{https://drive.google.com/uc?export=download&id=134MEILCPpeqQQGcWIJ_cTwMi-DRSyd4k}
    \item COCO: \url{https://drive.google.com/uc?export=download&id=1lWBdhOV665ye4UBOMDaT2d0AUHiQy0l4}
    \item Kitti: \url{https://drive.google.com/uc?export=download&id=1BrvtNaXgjywWEDLXoM31Eni_by8eXdYd}
    \item Pascal VOC: \url{https://drive.google.com/uc?export=download&id=17za9WVbpNLjipyt4mNTFsSnidjKPrbCj}
\end{itemize}

They were defined to the best of our knowledge based on a careful review of the original ground truth annotations. For Cityscapes and Pascal VOC we closely followed the official annotation guidelines. In contrast, for COCO and KITTI, no explicit label policies were available, so we established our own guidelines based on a detailed inspection of the datasets and their annotations.

We employed a crowdsourced annotation provider based in India, with a compensation rate of €4.5 per hour per annotator. In total, we commissioned 5,000 hours of annotation work. We considered the economic context of the annotators and aimed to provide compensation that is appropriate with respect to local wage levels and cost of living. Based on a comparison of purchasing power, the selected rate corresponds to a competitive hourly wage in the local context.

\section{Data Quality Assessment}
\label{sec: appendix - data quality assessment}

\subsection{Descriptive Statistics}
\label{subsec:appendix-descriptive statistics}

This subsection describes in detail the descriptive comparison of the original vs. validated ground truth.
It start with the explanation of we converted soft to hard labels and then describes all ground truth differences per dataset.

\paragraph{Deriving Hard Labels from Soft Labels}

Through the re-annotation of all four datasets, each validated bounding box is associated with a soft label representing a probability distribution over classes. To enable direct comparisons with the original ground truth and to support label error detection, we derive a corresponding hard label for each annotation. Specifically, we assign each bounding box the class with the highest likelihood in its soft label distribution. To ensure reliable and unambiguous assignments, we consider only those annotations for which the maximum class probability is greater than 0.5. This criterion ensures that a single class is clearly preferred over all others, avoiding ambiguous cases where multiple classes receive equal support. We analyze such ambiguous cases separately in additional evaluations. Furthermore, we exclude annotations labeled as ``can't solve'', which are assigned when an object does not correspond to any class defined in the original ground truth and is therefore not comparable within our evaluation framework. The resulting set of high-confidence annotations, obtained through this hard label extraction process, forms the validated ground truth. Each validated bounding box is associated with a hard label and a corresponding soft label probability $p>0.5$.

\paragraph{Comparison of Original and Validated Ground Truth on Pascal VOC}

\input{tables/gt_vs_vgt_table_pascalvoc}

The comparison between the original and the validated ground truth annotations of the Pascal VOC dataset reveals a consistent increase in the number of annotated objects across both splits (\autoref{tab:gt_vs_vgt_table_pascalvoc}). Overall, the validated annotations contain $+8.0\%$ more objects on the training set and $+9.2\%$ on the validation set, indicating a noticeable level of underannotation in the original dataset, despite Pascal VOC already being considered a high-quality benchmark. This trend is observed for the majority of classes, with particularly strong increases for \textit{boat} ($+18.5\%$ train, $+17.1\%$ val), \textit{dining table} ($+11.8\%$ train, $+20.3\%$ val) and \textit{person} ($+12.8\%$ train, $+13.8\%$ val), suggesting that these categories are especially prone to missing annotations. At the same time, several classes such as \textit{cow}, \textit{sheep} and \textit{tv monitor} show notable decreases, while others like \textit{cat} and \textit{dog} remain largely unchanged. This indicates that the validation process does not only adds missing objects but also removes or corrects inaccurate annotations present in the original ground truth.

\paragraph{Comparison of Original and Validated Ground Truth on COCO}

\input{tables/gt_vs_vgt_table_coco}

The \textit{person} and \textit{animal} categories (\autoref{tab:coco_person_animal}) show a strong increase in annotated instances, particularly for \textit{person} ($+51.5\%$) and classes such as \textit{bird} ($+123.6\%$) and \textit{sheep} ($+73.1\%$). This indicates that many objects were missing in the original annotations. This behavior is closely related to the crowd-based annotation strategy of the COCO dataset, where groups of objects in dense scenes can be represented by a single crowd annotation rather than by annotations of all individual instances. For example, groups of people may be annotated as crowd regions, while individual persons within these regions are not necessarily annotated separately. In contrast, our validation aims to annotate every visible instance, resulting in significantly higher counts. A similar effect can be observed for animals in cluttered environments. The \textit{vehicle} and \textit{sports} categories (\autoref{tab:coco_vehicle_sports}) exhibit a more moderate increase ($+13.8\%$), with most classes showing relatively small changes. This suggests that these objects are generally well covered in the original annotations. Vehicles are typically large, making them less likely to be missed. However, some classes such as motorcycle ($+35.0\%$) still show notable increases, indicating that smaller or partially occluded instances can be overlooked. In the \textit{sports} category, the mixture of increases and decreases reflects annotation ambiguity for small objects and less consistent labeling. The \textit{outdoor}, \textit{accessory} and \textit{electronic} categories (\autoref{tab:coco_outdoor_accessory_electronic}) shows only minor overall changes ($+5.6\%$), but with noticeable variation across classes. While some categories such as \textit{traffic light} ($+30.5\%$) increase, others like \textit{stop sign} ($-30.7\%$) and \textit{tv} ($-40.3\%$) decrease. Since these objects rarely appear in dense groups, the influence of crowd-based labeling is less dominant. Instead, the differences are mainly due to inconsistent annotation criteria and ambiguity, leading to both the addition of previously missed objects and the removal of questionable annotations. The \textit{kitchen}, \textit{food} and \textit{appliance} categories (\autoref{tab:gt_vs_vgt_coco_kitchen_food_appliance}) show the largest increase, driven by extreme gains in classes such as \textit{carrot} ($+449.6\%$), \textit{apple} ($+396.2\%$), and \textit{donut} ($+120.7\%$). This clearly indicates systematic underannotation in the original ground truth. In scenes with many small and densely packed objects, such as fruits in a bowl, annotators often label only a subset or use coarse bounding boxes. In contrast, our validated annotations capture each individual object leading to a substantial increase in object counts. The \textit{furniture} and \textit{indoor} categories (\autoref{tab:coco_furniture_indoor}) also show a strong increase ($+63.1\%$), particularly for \textit{book} ($+177.2\%$) and \textit{chair} ($+61.5\%$). This again highlights missing annotations for dense and repetitive objects. A typical example is bookshelves, where the original annotations often include only a few instances or a single coarse box, while our validated annotations label each book individually. Similar effects occur for cluttered indoor scenes, where partially visible objects are frequently omitted in the original annotations.

\paragraph{Comparison of Original and Validated Ground Truth on Cityscapes}

\input{tables/gt_vs_vgt_table_cityscapes}

\autoref{tab:gt_vs_vgt_table_cityscapes} shows only moderate changes between the original and the validated ground truth, indicating that the Cityscapes annotations are already of high quality. Across all classes, we observe an increase of $+5.7\%$ on the training split, while the validation split remains more stable with only $+2.0\%$. It is important to note that, for the training split, the original ground truth annotations were treated as given and annotators were only asked to add additional objects. Therefore, the observed increases primarily reflect newly annotated instances. The most striking result is observed for the \textit{bicycle} class, with a strong increase of $+38.9\%$ in training and $+59.3\%$ in validation, indicating that individual bicycles, particularly in dense scenes, are frequently not annotated as separate instances in the original ground truth. Cityscapes group annotations (e.g., \emph{persongroup} and \emph{bicyclegroup}) are excluded from this comparison, as they represent groups of objects rather than individually annotated instances. Consequently, some of the additional individual instances in our validated ground truth may correspond to objects that were covered by a group annotation in the original ground truth. At the same time, some classes show decreases, particularly in the validation split, such as \textit{car} ($-12.7\%$) and \textit{truck} ($-9.7\%$). These reductions are not due to missing annotations, but rather stem from the filtering of uncertain labels. Since only objects with a confidence of at least $50\%$ are included in the validated ground truth, some original annotations are discarded if they are deemed ambiguous. In contrast to training split, the validation split was fully re-annotated from scratch, which explains why both additions and removals are more clearly reflected. Overall, this shows that the validation process primarily adds missing objects while also enforcing a consistent quality threshold on the annotations.

\paragraph{Comparison of Original and Validated Ground Truth on KITTI}

\input{tables/gt_vs_vgt_table_kitti}

\autoref{tab:gt_vs_vgt_table_kitti} shows a substantial increase in the number of annotated objects across almost all classes. In particular, classes such as \textit{car} and \textit{pedestrian} exhibit strong gains, while categories like \textit{person sitting} show a disproportionately large increase, indicating that smaller or more ambiguous instances are frequently missed in the original annotations. Overall, the total number of objects increases by $+56.9\%$ on the training set and more than $+72.3\%$ on the validation set, highlighting a significant level of underannotations in the original ground truth. At the same time, slight decrease (e.g., for \textit{van}) suggest that some original annotations are either ambiguous or re-assigned to more appropriate classes during the re-annotation process. It is also important to note that the original KITTI annotations contain \textit{don't care} regions, in which no objects were labeled. In contrast, our annotation process does not explicitly exclude these regions, which further contributes to the higher number of annotated instances. In the subsequent label error analysis, we therefore distinguish between label errors on the full image and those restricted to areas outside the \textit{don't care} regions. This allows for a more detailed and fair comparison.

\subsection{Missing Objects}
\label{appendix:missing}

The detailed results of the false negative analysis are reported in \autoref{tab:fn_full}. Visual examples of certain errors are given in \autoref{fig:missing_objects_examples}. For Pascal VOC, COCO, and KITTI, we randomly sample 600 images per dataset, while for Cityscapes, we evaluate the complete validation set of 500 images, since only the validation set was fully re-annotated. Across these images, the analysis covers between 2,072 objects for Pascal VOC and 11,594 objects for Cityscapes. The validated GT consistently exhibits substantially lower FNRs than the original annotations, with relative reductions ranging from $50.0\%$ for Pascal VOC to $72.0\%$ for KITTI. Overall, the detailed analysis confirms the consistent reduction in missing objects reported in the main paper.

\input{tables/false_negative_full}

\input{imgs/collages-false-negative}

\subsection{Refinement of Soft Labels \& Ambiguous Cases}
\label{appendix:refine_ambiguous}

\input{tables/further_analysis_vgt_table}
\input{ambiguous_cases/plot_ambiguous_cases}

To improve the reliability of the soft labels, we refine the annotations through multiple annotation stages. In the validated ground truth, each bounding box initially receives 11 label annotations (i.e., votes forming a soft label) during the soft label generation stage. A subset of bounding boxes is then re-annotated, resulting in 22 annotations per instance. In this second round, annotators assign labels again with randomized class order to avoid bias from the initial task. Ambiguous cases, identified by the presence of at least one \textit{can't solve} vote, are further re-evaluated, increasing the total to 33 annotations. Finally, during duplicate removal, soft labels of boxes that clearly correspond to the same object are merged, which can also increase the total number of annotations per instance. The left side of \autoref{tab:annotation_and_ambiguity} shows that most bounding boxes retain the initial set of 11 annotations, while only a subset accumulates additional annotations through subsequent annotation and post-processing stages.

As shown on the right side of \autoref{tab:annotation_and_ambiguity}, ambiguous instances, defined as cases where multiple classes share the highest probability in the soft label, are rare across all datasets, accounting for less than 1\% of all bounding boxes. Nevertheless, the qualitative examples in \autoref{fig:ambiguous_examples} illustrate that these cases correspond to genuinely challenging scenarios, such as strong visual similarity between classes (e.g., \textit{car} vs. \textit{van} or \textit{person} vs. \textit{rider}), small or partially occluded objects, and low image quality. In several classes, the ambiguity is inherent to the visual content rather than annotation noise, highlighting the benefit of soft labels in capturing such uncertainty instead of enforcing a potentially arbitrary hard decision. 

\subsection{Source of Label Errors}
\label{sec:appendix - data quality - label error}

This section defines how we calculated the label errors between original and validated ground truth in detail. 
Then the label errors are described per dataset.

\paragraph{Definition of Label Error Ground Truth}

To construct the label error ground truth (LEGT), we compare the validated ground truth with the original annotations and define the LEGT as the set of identified label errors. We perform greedy one-to-one matching between validated and original bounding boxes based on descending IoU, using a matching threshold of $\mathit{IoU}\geq 0.5$. Each bounding box can be matched at most once. Based on this comparison, we identify three types of label errors: missing bounding boxes, misaligned bounding boxes, and classification errors.

A \textit{missing bounding box} refers to an object that is present in the validated ground truth but has no corresponding match in the original annotations and demonstrates only negligible overlap with any original bounding box ($\mathit{IoU}\leq 0.1$). A \textit{misaligned bounding box} describe a case where no match is found, but a partial overlap exists with an original bounding box ($0.1<\mathit{IoU}\leq0.5$), indicating inaccurate localization. A \textit{classification error} occurs when two bounding boxes are matched with high spatial agreement ($\mathit{IoU}>0.7$) but are assigned different class labels. To further refine the analysis, we introduce additional criteria based on object size and annotation confidence. Specifically, we distinguish between small and large objects using bounding box height thresholds of 25 and 40 pixels. In addition, we consider confidence thresholds of $p\geq0.5$ and $p\geq 0.8$, derived from the soft label probabilities of the validated annotations. These parameters allow us to define different LEGT configurations with varying levels of strictness. In our experiments, we report two default configurations. The \textit{less restrictive} LEGT includes all detected label errors with no restriction on objects size and a confidence threshold of $p\ge 0.5$, providing a comprehensive view of all potential label errors. The \textit{stricter} LEGT is more restrictive and considers only objects with a height of at least 40 pixels and a confidence threshold of $p\geq 0.8$, thereby focusing on the most reliable and clearly defined label errors. For the KITTI dataset, we evaluate label errors across the entire image, independent of the annotated \textit{don't care} regions.

Finally, it is important to note that there is no single definitive definition of a label error ground truth. The resulting LEGT depends on the chosen thresholds and, more generally, on the underlying soft label distributions. Different configurations emphasize different aspects of label quality. Lower thresholds include a broader set of potential label errors, including ambiguous and low-confidence case, while stricter thresholds focus on fewer but more reliable and clearly defined label errors.
Note that the categorization of label error types depends on the chosen thresholds as well. In particular, the IoU threshold strongly influences the distinction between missing and misaligned annotations. Since these error types can overlap, it is not possible to define a single threshold that perfectly separates them.

\paragraph{Identified Label Errors - KITTI}

\input{tables/Kitti_label_error_table_train}
\input{tables/Kitti_label_error_table_val}

We first analyze the distribution of label errors across the KITTI dataset. As shown in \autoref{tab:train_label_errors_kitti} and \autoref{tab:val_label_errors_kitti}, missing bounding boxes clearly dominate, with 17,371 instances on the training split at a probability threshold of $p \ge 0.5$, compared to only 5,533 misaligned and 829 classification label errors. A key finding is that the majority of label errors occur for small objects. Restricting the analysis to objects with a height of at least 25 pixels reduces missing bounding boxes from 17,371 to 3,579, and further to 944 for objects larger than 40 pixels. A large fraction of label errors occurs within ``don't care'' regions. Excluding these regions reduces missing bounding boxes from 17,371 to 4,933, highlighting the impact of ambiguous or difficult areas. Increasing the probability threshold from $p \ge 0.5$ to $p \ge 0.8$ substantially reduces missing bounding boxes and classification label errors, while misaligned bounding boxes remain comparatively stable. Similar trends are observed on the validation split, indicating consistent behavior across the dataset. Overall, missing bounding boxes are the dominant source of label errors, far outweighing misaligned bounding boxes and classification label errors. Most label errors occur for small objects and within ``don't care'' regions, and high-confidence filtering primarily reduces missing bounding boxes and classification label errors.

\paragraph{Identified Label Errors - Cityscapes}

\input{tables/Cityscapes_label_error_table_train}
\input{tables/Cityscapes_label_error_table_val}

In contrast to KITTI, the number of label errors in Cityscapes is considerably lower (\autoref{tab:train_label_errors_city}, \autoref{tab:val_label_errors_city}). This is mainly due to the annotation procedure of the training set, where annotators were instructed to only add missing objects while treating the original ground truth as given. As a result, missing bounding boxes clearly dominate, while classification label errors remain very limited. Similar to our observations on KITTI, the majority of label errors occur for small objects. Increasing the minimum object height leads to a substantial reduction in all label error types, highlighting the difficulty of consistently annotating small instances. Furthermore, the relative distribution of label error types remains stable across different probability thresholds, indicating that the threshold primarily affects the absolute number of detected label errors rather than their overall structure. The validation split, which was fully re-annotated, shows a more balanced distribution of label errors compared to the training set, further confirming the bias introduced by the partial re-annotation strategy.

\paragraph{Identified Label Errors - COCO}

\input{tables/COCO_label_error_table_val}

\autoref{tab:val_label_errors_coco} presents the number of identified label errors on the validation split of the COCO dataset under different probability and object size thresholds. Overall, missing bounding boxes clearly dominate, exceeding both misaligned bounding boxes and classification label errors across all settings. This indicates that annotation incompleteness is the primary source of label errors in COCO. A substantial proportion of these missing bounding boxes can be attributed to COCO crowd annotations, where groups of densely packed objects are often annotated using a single large bounding box. In contrast, our method frequently identifies multiple individual objects within these crowd scenes, which are consequently counted as missing bounding box. Increasing the probability threshold from $p\ge 0.5$ to $p\ge 0.8$ consistently reduces the number of detected label errors for all label errors types. This reduction is most noticeable for classification label errors, which decrease most strongly, although they remain minor in absolute terms. Furthermore, the number of detected label errors decreases significantly with increasing object height. For example, the number of missing bounding boxes drops from 16,219 for all objects to 3,182 for objects with a height of at least 40 pixels. This trend highlights that small objects are particularly prone to annotation issues, whereas larger objects are annotated more reliably. Misaligned bounding boxes represent a secondary source of label errors, with consistently lower counts than missing bounding boxes but still substantially higher than classification label errors. In contrast, classification label errors occur only rarely, indicating that class labels in COCO are largely accurate.

\paragraph{Identified Label Errors - Pascal VOC}

\input{tables/PascalVOC_label_error_table_train}
\input{tables/PascalVOC_label_error_table_val}

\input{labelerrors/original_only/plot_original_only}

The results on the Pascal VOC dataset show a consistent pattern (\autoref{tab:train_label_errors_voc}, \autoref{tab:val_label_errors_voc}). Across both training and validation splits, missing bounding boxes are by far the dominant label error type. For example, on the training split at $p\ge 0.5$, 1,835 missing bounding boxes are detected, compared to 1,120 misaligned bounding boxes and only 29 classification label errors. A similar pattern is observed on the validation split (1,819 missing vs. 1,090 misaligned and 23 classification label errors). This indicates that most label errors stem from objects that are not annotated at all, while classification label errors remain minor. Increasing the probability threshold from $p\ge 0.5$ to $p\ge 0.8$ consistently reduces the number of detected label errors across all label error types, indicating that higher confidence requirements result in fewer identified label errors. Object size has a similarity strong impact. When considering only larger objects (e.g., $\ge$ 40 pixels), the number of label errors decreases significantly. For example, missing bounding boxes on the training split decrease from 1,835 to 597 at $p\ge 0.5$. This indicates that a substantial portion of the label errors originates from small objects, which are more difficult to annotate consistently. Misaligned bounding boxes represent the second most frequent label error type, but remain clearly below the number of missing bounding boxes. In contrast, classification label errors are consistently low across all settings. Overall, the results indicate that missing bounding boxes are the main source of label errors in Pascal VOC, particularly for small objects, while classification label errors play only a minor role.

\input{tables/original_only_table}

\begin{figure}[htbp]
    \centering
    \includegraphics[width=0.6\linewidth]{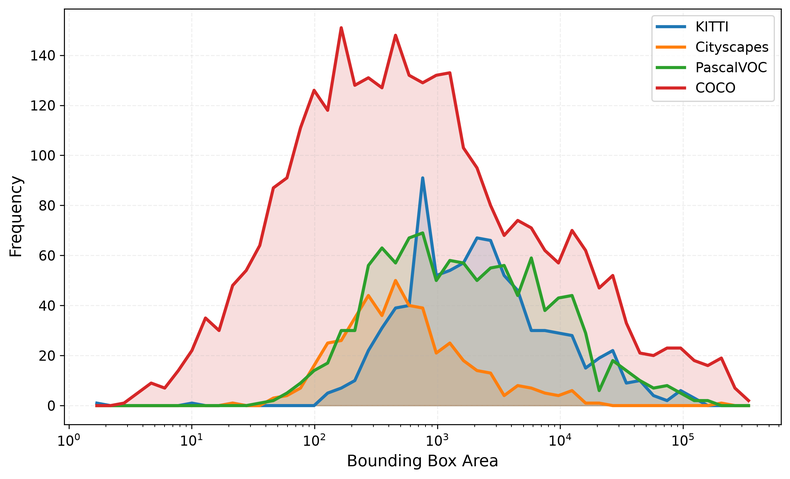}
    \caption{
    Distribution of bounding box areas for original-only annotations across all datasets. 
    Each curve shows the frequency of bounding box sizes on a logarithmic scale.
    }
    \label{fig:original_only_area_distribution}
\end{figure}

\input{tables/reviewed_original_only_table}
\input{tables/fnr_original_only}

\subsection{Original-Only Bounding Boxes}
\label{sec: appendix- data quality - orgOnly}

In addition to missing bounding boxes, misaligned bounding boxes, and classification label errors, we analyze \textit{original-only} annotations, i.e., bounding boxes that are present in the original ground truth but not in our validated ground truth. Such annotations are inherently ambiguous: they may either correspond to incorrect annotations in the original ground truth or to valid objects that were missed during re-annotation (\autoref{fig:originalonly_examples}). For the Cityscapes training split, no original-only annotations exist because we kept the original annotations and only added newly identified objects. Therefore, we do not consider the Cityscapes training split in the following original-only analysis.

To determine whether original-only annotations correspond to valid objects, we manually review all original-only annotations and decide for each bounding box whether it should be included in the dataset. \autoref{tab:original_only_review} summarizes the results of this review. Across all dataset, a substantial fraction of the original-only annotation is accepted, although the acceptance rate varies considerably between datasets. For Pascal VOC, more than $93\%$ of the original-only annotations are accepted, whereas the acceptance rates are lower for KITTI ($67.39$--$70.46$\%), COCO ($66.63\%$), and Cityscapes ($53.30\%$). Similar to the observed missing bounding box errors, original-only annotations are dominated by small objects (\autoref{fig:original_only_area_distribution}), indicating that discrepancies between the original and validated annotations frequently occur for small and visually challenging instances.

After the manual review, we further validate the class labels of the accepted original-only annotations by collecting five annotations per box from five independent annotators. The resulting votes form a soft label, from which we derive a hard label if a single class receives a strict majority ($p>0.5$) and the majority class is not \textit{can't solve}. Original-only annotations satisfying these criteria are subsequently added to the VGT. \autoref{tab:vgt_original_only_hardlabels} compares the number of hard-labeled annotations in the original GT and the VGT with the resulting VGT after adding these original-only annotations.

Adding the reviewed original-only annotations further increases the number of hard-labeled annotations beyond the VGT across all considered splits. For example, the increase relative to the original GT changes from $+39.98\%$ to $+44.20\%$ for COCO and from $+7.96\%$ to $+11.28\%$ for the Pascal VOC training split. For KITTI, where the VGT already contains substantially more annotations than the original GT, the additional increase is smaller, from $+56.86\%$ to $+58.05\%$ for training and from $72.29\%$ to $+73.19\%$ for validation. These results show that the original-only review provides an additional refinement step that recovers valid annotations missed during re-annotation while ensuring that potentially incorrect original annotations are not automatically added to the VGT.

To further assess the effect of adding the reviewed original-only annotations to the validated GT, we repeat the false negative analysis using the resulting validated GT + original-only annotations. We evaluate 600 randomly samples images for Pascal VOC, COCO, and KITTI and the complete Cityscapes validation set of 500 images. For each image, all visible objects are considered when determining the number of missing objects. As shown in \autoref{tab:fnr_original_only}, the accepted original-only annotations further reduces the FNR across all datasets. The largest relative reduction is observed for COCO, where the FNR decreases from $6.1\%$ to $3.6\%$, while smaller but consistent improvements are observed for Pascal VOC, Cityscapes, and KITTI. These results indicate that the original-only review recovers additional valid annotations that were missed during re-annotation.

\subsection{Bias Correction}
\label{sec: appendix - data quality - bias correction}

Annotators tend to be biased towards confirming the proposed class~\cite{schmarje2023cleverlabel}. This bias is due to the fact that clear cases are more frequent than ambiguous ones and annotators may develop a certain pace in their annotations or lose focus over time.
Additionally, it is common for humans to accept a proposal if it sounds reasonable.
We configured a different task for bounding box verification that reduces this bias by shuffling objects of all classes. In this task, no class is proposed and annotators have to select a different answer option every time so that they cannot develop a confirmation bias. This task takes significantly longer to answer. We find about factor 2 as shown in \autoref{fig:times-0} and \autoref{fig:times-1} which renders this procedure very costly at scale. $11$ different annotators conducted this task for $1,000$ randomly selected objects of each dataset and for COCO, for each one of the $11$ super categories. Through comparison of the potentially biased annotations with the ones obtained using the unbiased approach, we can measure confirmation bias in the data.

\twosideimages
{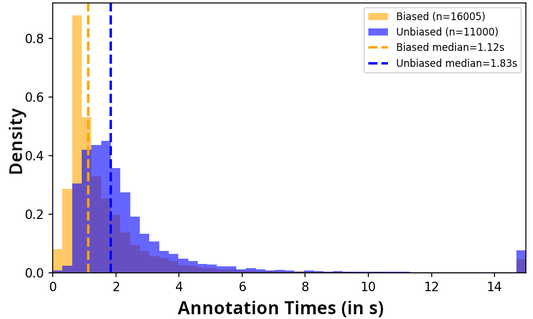}          
{KITTI}        
{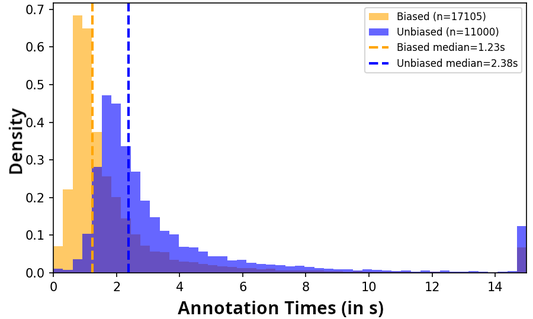}          
{Coco Kitchen Superclass}       
{Annotation Times with proposal (biased, yellow) and without proposal (unbiased, blue). The graphic shows a histogram plot of the raw annotation times and its median.}     
{times-0}

\twosideimages
{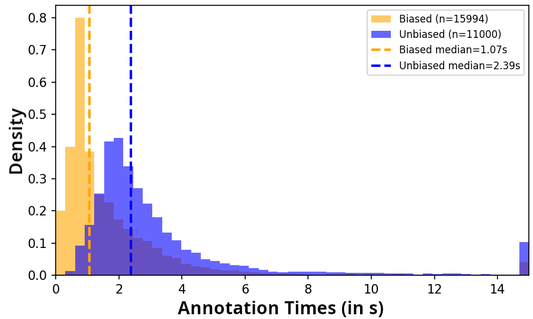}          
{Cityscapes}        
{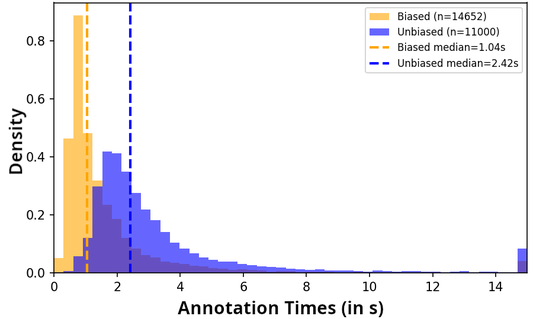}          
{Pascal VOC}       
{Annotation Times with proposal (biased, yellow) and without proposal (unbiased, blue). The graphic shows a histogram plot of the raw annotation times and its median.}     
{times-1}

\paragraph{Used Methodology}

We use CleverLabel~\cite{schmarje2023cleverlabel} for the bias correction of proposal-based annotation
workflows.
In the original proposal-based setup,
annotators are shown a ``proposed class'' and tend to accept it too readily, inflating its
probability. CleverLabel's \textbf{bias correction (BC)} stage models this: it estimates the
acceptance rate~$A$ as the fraction of votes for the proposed class, then computes a corrected
belief $B = \frac{A - \delta}{1^* - \delta}$ that separates the true class probability from
annotation bias, redistributing the rejected-case mass over the remaining classes. An optional
\textbf{class blending (CB)} stage then mixes the corrected distribution with a row from a
precomputed \emph{transition matrix} (which encodes empirical confusion patterns between
classes, e.g.\ truck$\leftrightarrow$car) weighted by parameter~$\mu$.

We extend this methodology by creating \textbf{CleverLabel Lite} and adding \textbf{overreaction avoidance}.
CleverLabel Lite does something inherently very similar to CleverLabel but far more simply: it multiplies the proposed class's probability by a \texttt{downweight\_factor}
(e.g.\ $0.5$) and renormalises. 
Thus, if the class is very likely it will be still very likely after correction, since not much mass can be transferred to the other clases in the rebalancing step.
Both methods reduce the inflated proposed-class probability and
redistribute mass to other classes; CleverLabel does so via a generative model of the acceptance
bias with the transition matrix providing class-specific redistribution, while downweight
achieves this with a single scalar reduction and uniform renormalisation.

The Overreaction avoidance relies on the  \textbf{\texttt{avoid\_overcorrection\_threshold}}
parameter which guards against over-correction.
If the proposed class already received less than this fraction of votes (meaning annotators already disagree significantly), bias correction is skipped entirely, since the distribution already reflects
genuine ambiguity.

Finally, we apply smoothing to the corrected distributions to ensure non zero values. 
Additionally, it is more realistic that a small percentage of annotations can be an arbitary class e.g. due to a misclicked annotation.
\textbf{Epsilon smoothing} ($\epsilon = 10^{-10}$) is applied after
post-processing to ensure no class probability is exactly zero, each probability receives an
additive~$\epsilon$ and the result is renormalised, which prevents numerical issues in
downstream log-likelihood or calibration computations without meaningfully altering the
distribution.
We also experimented with Dirichlet smoothing~\cite{dirchilet} using $\alpha = 0.1$ and $\alpha = 0.5$, but found its effect to be overly strong. Despite having at least 11 annotations per instance (and up to 44 in some cases), the smoothing significantly distorts the resulting distributions, especially when annotations are aggregated. Even for relatively small values such as $\alpha = 0.1$, the impact is substantial. This is consistent with prior work, which typically recommends $\alpha \in {0.5, 1.0}$. However, in our setting with 9 classes and 11 annotations, smoothing reduces the maximum achievable class probability from 1.0 to below 0.6. As a result, even unanimous agreement among annotators would lead to a softened label with a maximum probability of only about 0.6, which we consider undesirable for our use case.
\[
p_{\max} = \frac{1 + \alpha}{1 + K \alpha}
= \frac{1.1}{1 + 0.9}
= \frac{1.1}{1.9}
\approx 0.5789.
\]

\twosideimages
    {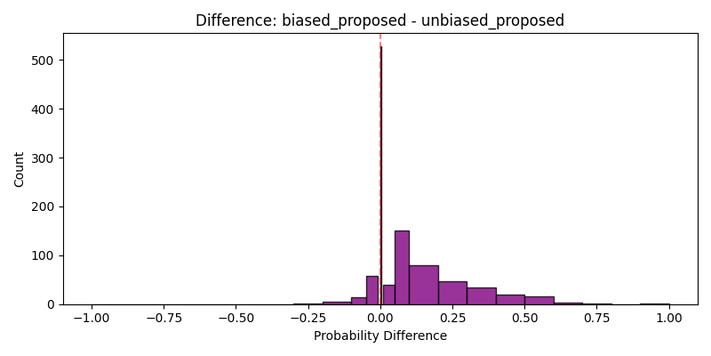}          
    {Pascal VOC}        
    {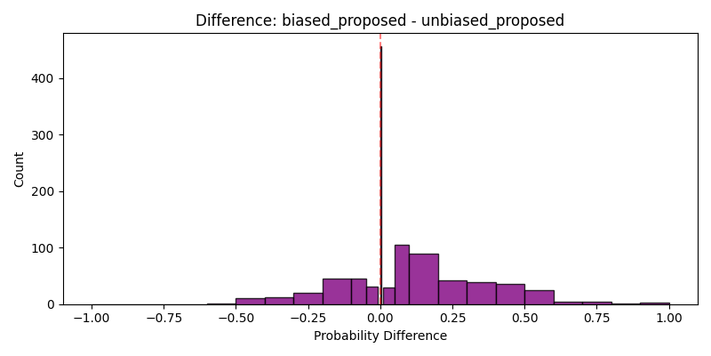}          
    {Cityscapes}       
    {Biased minus unbiased probability for proposed label, expect values above 0 which indicate an introduced bias}     
    {diff_biased_0}

\twosideimages
    {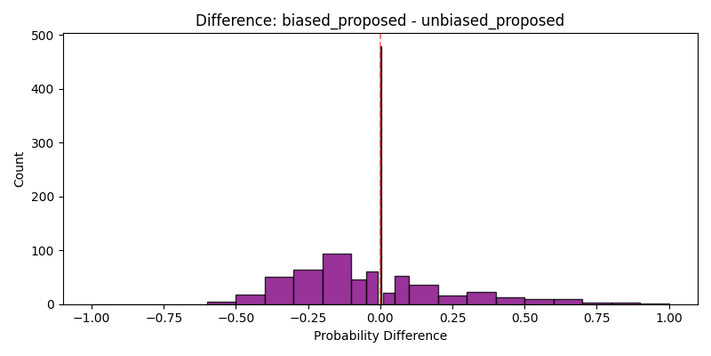}          
    {KITTI}        
    {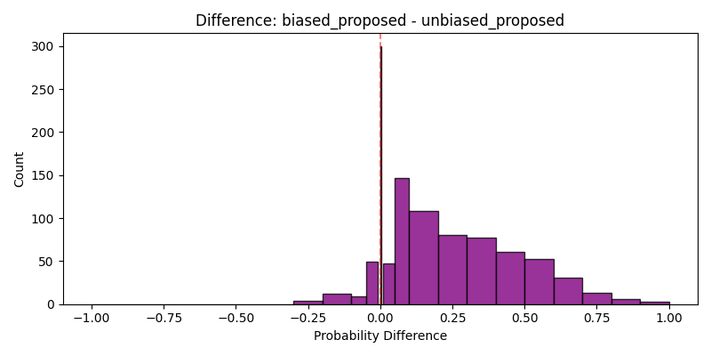}          
    {Coco Kitchen Superclass}       
    {Biased minus unbiased probability for proposed label, expect values above 0 which indicate an introduced bias}     
    {diff_biased_1}

\input{tables/post_processing_results_full}

\paragraph{Parameter Tuning}
The post-processing parameters are selected via exhaustive grid search over a
discrete parameter space. For the CleverLabel method, the search covers the Cartesian product
of $\delta \in \{0.0, 0.05, 0.1, \ldots, 0.6\}$ (9~values),
$\mu \in \{0.0, 0.25, 0.5, 0.75, 1.0\}$, and the overcorrection threshold
$\in \{0.0, 0.1, \ldots, 0.5, 0.8, 0.9\}$ (8~values), yielding 360~configurations.
CleverLabel Lite adds 10~configurations
(\texttt{downweight\_factor} $\in \{0.1, 0.2, \ldots, 0.95\}$), plus a single identity
baseline, for a total of 371~candidates. Smoothing is either epsilon or dirichlet smoothing.

We use a subset of 1000 annotations per dataset or for COCO per class superset for calibration and validation.
Each dataset's annotated objects carry both \emph{biased} frequency counts (collected with a
proposed class shown to annotators) and \emph{unbiased} frequency counts (collected without
any suggestion), the latter serving as ground truth. The data is split into a training and
validation set (50/50 by default, deterministic seed). A transition matrix~$C$, a
$K \times K$ class-confusion matrix recording how often class~$j$ is annotated when the true
majority class is~$i$, is estimated exclusively from the training split's unbiased counts.

For every parameter combination~$\theta$, the post-processing transform
$\mathrm{PP}_\theta$ is applied to the biased distributions in the training set. The
optimization target is the mean per-object delta divergence:
\begin{equation}
    \theta^* \;=\; \underset{\theta \in \Theta}{\arg\min} \;\;
    \frac{1}{N}\sum_{i=1}^{N} \Bigl[
        \sqrt{\mathrm{JS}\!\bigl(\mathrm{PP}_\theta(x_i),\, g_i\bigr)}
        \;-\;
        \sqrt{\mathrm{JS}\!\bigl(x_i,\, g_i\bigr)}
    \Bigr],
\end{equation}
where $x_i$ is the biased distribution, $g_i$ the unbiased ground truth, and
$\mathrm{JS}$ denotes Jensen--Shannon divergence\cite{Fuglede2004JensenShannonDA}. A negative delta indicates that
post-processing moved the distribution closer to ground truth. The configuration with the
most negative mean delta on the training set is selected.
The held-out
validation split is used solely for reporting generalization performance and does not influence parameter selection.

For COCO, the procedure is applied independently per super-category (e.g.\ \emph{vehicle},
\emph{animal}, \emph{food}), and the per-category results are merged into a single
multi-configuration file with a shared default fallback.

We also investigated the option of a small Multi Layer Peceptron (MLP) which can be
trained to predict the unbiased distribution from the concatenation of the smoothed biased
probabilities and the corresponding transition
matrix row. It is trained with KL-divergence loss and early stopping, and its validation
performance is compared against the grid-search winner to determine which approach is more
effective for a given dataset. This method yielded worse performance an all datasets in comparison to CleverLabel.

\paragraph{Experimental Results}

From the literature, we can expect that a proposal acceptance bias would lead to higher class labels proposed class if the proposal is shown~\cite{schmarje2023annotating}
However, we also know this bias is not general but depends on the dataset~\cite{schmarje2023cleverlabel}, thus it can also be the case that no bias is introduced when a proposal is shown  during annotation.

In \autoref{fig:diff_biased_0} and \autoref{fig:diff_biased_1} we see the difference of the proposed classed for a with proposal and without proposal shown.
We see that a clear bias is present for Pascal VOC and the Kitchen Superclass of COCO. 
For Cityscape a small bias towards (positive) is visible but there are also multiple cases where the opposite effect (negative values) is visible.
For KITTI, we can not that a majority of difference are negative (thus no bias is determinable).
Thus we can measure a bias in some datasets while not in other datasets.

This results are in alignment with the improvements we measured after the application of the best post processing method based on the training data.
An overview is provided in \autoref{tbl:post_process_results_full}.
We see for Pascal VOC and COCO Kitchen Superclass a clear reduction (over 5\%) in the  $\sqrt{\text{JS}}$.
The introduced biased probability for the proposed can reduced by more than 50\% towards the unbiased expected value.
For Cityscapes and KITTI the gains are smaller, but a reduction is still measurable with 10-20\% relative improvement in the proposed class probability and the averaged reduction difference of $\sqrt{\text{JS}}$.

Results of \autoref{tbl:post_process_results_full} are calculated on the complete calibration data to allow broader understanding of the performance on the full dataset.  For tuning only the training subset was used.

\subsection{Comparison with Human Perception}
\label{sec: appendix - data quality - calibration of soft labels}

We evaluate the calibration of soft labels before and after bias correction by comparing them to the perception of expert annotators. For this, three expert annotators evaluated randomly sampled objects regarding the proposed class by specifying a probability of the proposed class being present. 

For each dataset, we randomly sampled 100 objects with soft label probability below $0.5$ and 100 objects with probability larger than $0.5$. Each expert annotator evaluated an object on a scale from 0 to 1 in steps of 0.1 to give a probability estimate. The results of three expert annotators are averaged and then compared to the results obtained by crowdsourcing.

\threebiasimages{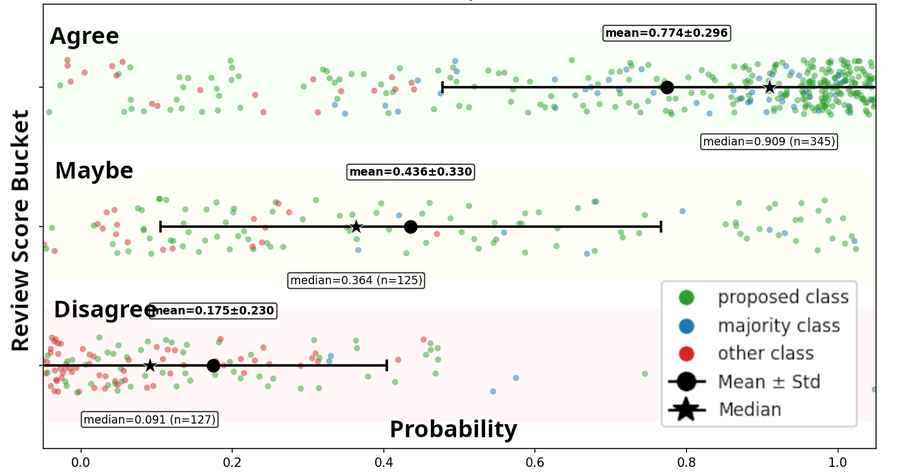}{Alignment of Human Perception with and without proposal and our postprocessing on the dataset COCO - 
The figure shows the ratings of a expert human labeler if the label is clearly ($\geq 0.75$), maybe ($0.25 - 0.75$) or not ($< 0.25$) correct. 
The x-axis are the corresponding soft probability for the evaluated class. 
Median is given as star, mean and std given as line plot. }
\threebiasimages{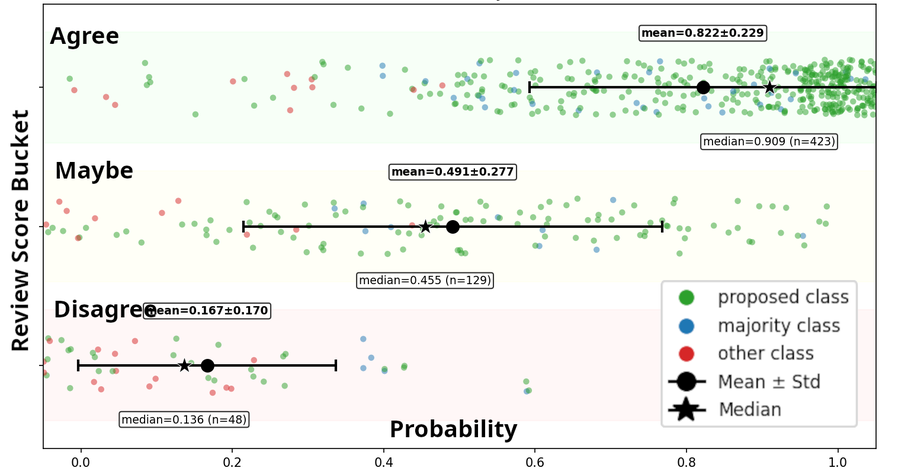}{Alignment of Human Perception with and without proposal and our postprocessing on the dataset Cityscapes - 
The figure shows the ratings of a expert human labeler if the label is clearly ($\geq 0.75$), maybe ($0.25 - 0.75$) or not ($< 0.25$) correct. 
The x-axis are the corresponding soft probability for the evaluated class. 
Median is given as star, mean and std given as line plot.}
\threebiasimages{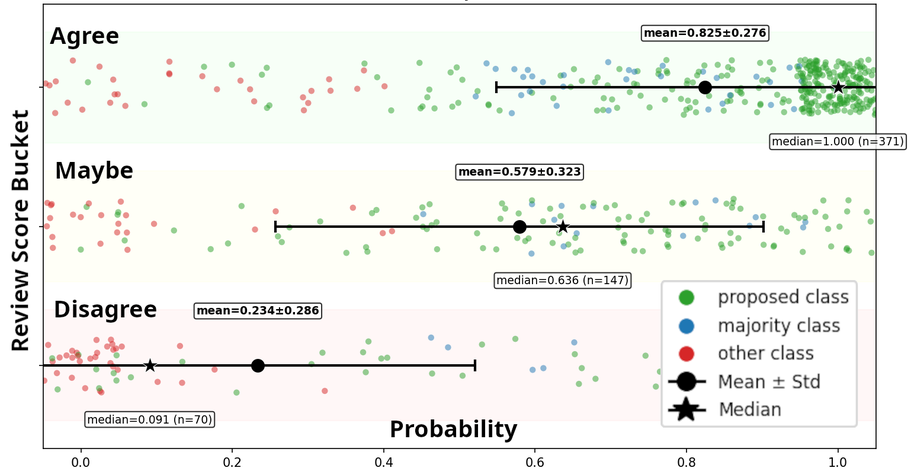}{Alignment of Human Perception with and without proposal and our postprocessing on the dataset KITTI - 
The figure shows the ratings of a expert human labeler if the label is clearly ($\geq 0.75$), maybe ($0.25 - 0.75$) or not ($< 0.25$) correct. 
The x-axis are the corresponding soft probability for the evaluated class. 
Median is given as star, mean and std given as line plot. }
\threebiasimages{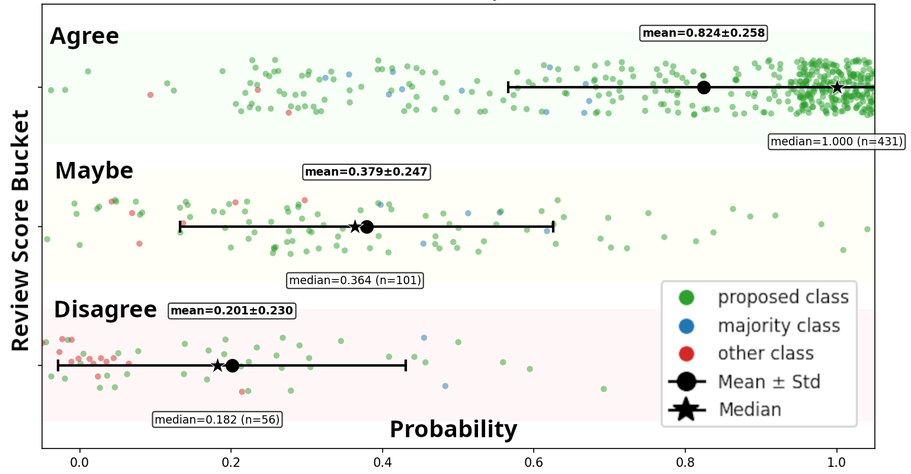}{Alignment of Human Perception with and without proposal and our postprocessing on the dataset Pascal VOC - 
The figure shows the ratings of a expert human labeler if the label is clearly ($\geq 0.75$), maybe ($0.25 - 0.75$) or not ($< 0.25$) correct. 
The x-axis are the corresponding soft probability for the evaluated class. 
Median is given as star, mean and std given as line plot. 
}

The results for all datasets are in \autoref{fig:COCO_human_perception}, \autoref{fig:CITYSCAPES_human_perception}, \autoref{fig:KITTI_human_perception}, and \autoref{fig:PASCALVOC_human_perception}.
We see that especially for Pascal VOC the post processing significantly move the the human perception back to the original estimates. 
This means post processing helps to realign potentially biased distributions more into the unbiased version.

\input{tables/wrong_labels_full}

\input{imgs/collage_wrong_class}

\subsection{Wrong Class}
\label{sec: appendix - data quality - wrong class}

For the estimate how many classes are wrong we used two approaches one suited for soft label like our validated GT and one suited for hard labels from the original GT.
For the original data we picked 200 random samples from the original GT and a human expert check if certainly wrong labels were provided.
For the validated GT we used the average scores across the 3 human calibration checks (see above) as baseline with values above 0.75 and below 0.25. 
In both cases, we did not evaluate uncertain wrong classes, since this highly depend on the used annotators and inlcude uncertainty.
Nevertheless, we provide the full results for the soft labels in \autoref{fig:bucket_coco}, \autoref{fig:bucket_cityscapes}, \autoref{fig:bucket_kitti}, and \autoref{fig:bucket_pascal}.
For all datasets, we provide the accuracy across buckets, Error1+ number of samples at least one bucket off (1-accuracy) and Error2+ which means at least two buckets off. This is the reported certain error from the paper.
These tables also include comparison of human experts with each other. 
We see no severe errors 2+ in any datasets.
In \autoref{tab:original_gt_certain_errors}, we provide the raw numbers for the evaluation on the original ground truth.
Additionally, example images are given in \autoref{fig:wrong_label_examples}.

\section{Object Detection Benchmark}
\label{sec: appendix - Additional Results}

\subsection{Definition Area Under Similarity-Recall Curve}
\label{sec: appendix - AUSRC Definition}
With its introduction in the Pascal VOC challenge in 2007~\cite{PascalVOC}, average precision (AP) has become the standard evaluation measure in object detection~\cite{COCO, ImageNet, SSD, focal_loss, fasterRCNN}. Object detectors are evaluated in terms of mean AP (mAP or just AP) over all classes, which is typically also averaged over multiple IoU thresholds to reward well-localized predictions~\cite{COCO}.
However, in this evaluation procedure, GT annotations are assumed to be of zero uncertainty and confidence levels are only used to sort predictions but not evaluated directly. 
To estimate the calibration of model confidence, precision is computed for batches of predictions with a certain confidence range to generate a so-called calibration plot~\cite{Chuan2017}. If confidence is well-calibrated, the resulting precision should align with the average confidence of this batch of predictions. However, this approach does not allow for instance-wise evaluation of model uncertainty. In the domain of soft labels, AP can be generalized by replacing precision as an evaluation metric with a normalized similarity measure that captures how well predicted and GT distribution align.

For evaluation of object detectors on object detection datasets with soft labels, we define the Area Under Similarity-Recall Curve (AUSRC) as a suitable performance metric.
Let a model produce a set of detections ${d_i}, i=1,\ldots, N$, each associated with a predicted class probability vector $\mathbf{p}_i \in \Delta^{C+1}$ and a matched GT soft label $\mathbf{q}_i \in \Delta^{C+1}$, where $\Delta^{C}$ denotes the probability simplex over $C$ classes and background. Since soft labels are only available for existing GT objects, soft labels of predictions that do not match with any GT object are compared to a one-hot vector for background. 
We rely on a similarity function
$S(\mathbf{p}, \mathbf{q}) \in [0,1]$,
such as cosine similarity, $1 -$ total variation distance (TVD), or Jensen-Shannon distance (JSD) to quantify agreement between $p_i$ and $q_i$. Per default, we use TVD due to its simple interpretation as the largest absolute deviation in the entries of two probability vectors. To account for the full distribution of predictions, we sort them by entropy instead of confidence.
For the top-$k$ detections, we compute geometric recall and the average similarity score:
\begin{align}
R(k) &:= \frac{1}{|\mathcal{G}|} \sum_{i=1}^{k} \1_{\{d_i \text{ has a match in }\mathcal{G}\}}, \quad
\bar{S}(k) := \frac{1}{k} \sum_{i=1}^{k} S(\mathbf{p}_i, \mathbf{q}_i),
\end{align}
where $\mathcal{G}$ is the set of GT objects and matching follows greedy IoU-based assignment.

This yields a Similarity-Recall curve $\bar{S}(R)$, analogous to the classical precision-recall curve.
AUSRC is then defined as:
\begin{align}
    \mathrm{AUSRC} := \int_0^1 \bar{S}(R), dR,
\end{align}
which we approximate via interpolation over discrete recall levels, similar to standard AP computation.

While AP measures how well a model ranks true positives above false positives, AUSRC additionally captures how well the predicted class distributions align with soft labels. High AUSRC therefore requires accurate localization and ranking, as well as high semantic agreement between predicted and target distributions.
In the case of hard labels, AUSRC reduces to standard average precision if predictions are sorted by confidence and checked for correct class assignment.

\subsection{Details on Training Schemes}
\label{sec: appendix - training details}

To fine-tune COCO-pretrained closed set object detectors on other datasets, we apply the following training procedures. For YOLOv8 and RT-DETR, we mainly rely on the training routine and the default parameters of the Ultralytics library~\cite{ultralytics}. These default parameters specify an image size of $640\times 640$, a batch size of 16, an initial learning rate is determined automatically and reduced to $1\%$ of its initial value over time, as well as, that the model is trained for 100 epochs. A comprehensive list of training parameters is provided by the Ultralytics library. Since models are COCO-pretrained, we choose an initial learning rate of $10^{-5}$ and optimize using stochastic gradient descent when fine-tuning on our COCO annotations. For Pascal VOC, choosing an initial learning rate of $10^{-4}$ led to better results than the automatically determined learning rate, so we chose this setup for fine-tuning. We fine-tune V2 of Faster R-CNN, available in Pytorch for 50 epochs with an initial learning rate of $10^{-3}$, which is then halved every 10 epochs. For optimization, we apply stochastic gradient descent with a momentum of $0.9$ and a weight decay of $5\cdot 10^{-4}$. We fine-tune models on a single NVIDIA A100 GPU with 80 GB of RAM, which takes about 1 - 2 hours depending on model and dataset, so overall around 2 days for all 29 trainings on this hardware.

\input{tables/map_comparison_coco}
\input{tables/map_comparison_PascalVOC}
\input{tables/map_comparison_KITTI}
\input{tables/map_comparison_Cityscapes}

\subsection{Comparison of mAP on Other Datasets}
\label{sec: appendix - mAP results on other datasets}
The performance results of the evaluated object detectors on COCO, Pascal VOC, KITTI, and Cityscapes are listed in
\autoref{tab:mAP_comparison_COCO}, \autoref{tab:mAP_comparison_VOC}, \autoref{tab:mAP_comparison_KITTI}, and  \autoref{tab:mAP_comparison_Cityscapes}. Similar to the COCO evaluation results, model performance on Pascal VOC generally drops on validated annotations due to the increased difficulty of the detection task, which leads to lower recalls. On the KITTI dataset, we note an increase in mAP of OVODs such as Grounding DINO and YOLO World, presumably due to the fact that missing annotations were added to the dataset such that accurate model predictions are now evaluated as true positives. For fine-tuned models, we observe an extreme performance drop from original annotations to our validated annotations, indicating that these models overfit to the training data and do not generalize well for the actual task they were trained on. On the Cityscapes dataset, we note a slight increase in performance on validated annotations. While the original and validated annotations are most similar for this dataset, performance gains may be explained on the one hand, by the addition of missing annotations and on the other hand, by the filtering of uncertain annotations. 

\input{tables/ECE_before_and_after_calibration}

\begin{figure}[htbp]
    \centering
    \begin{subfigure}{0.48\textwidth}
        \centering
        \includegraphics[width=\linewidth]{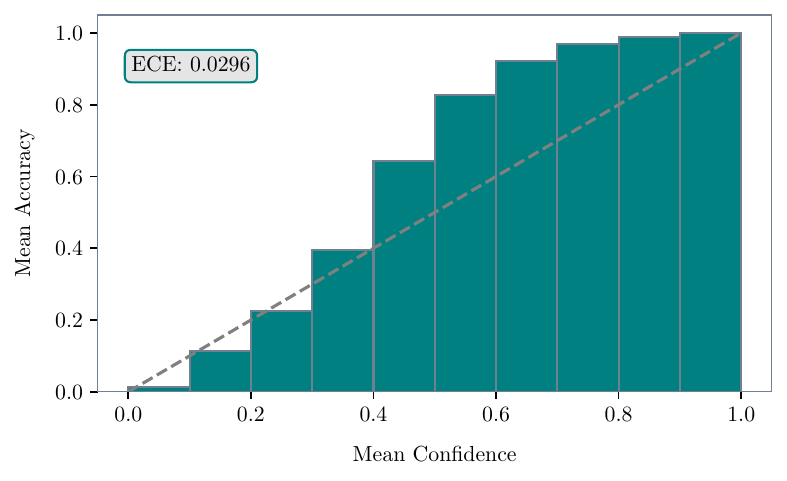}
        \caption{Calibration plot for raw predictions.}
    \end{subfigure}
    \hfill
    \begin{subfigure}{0.48\textwidth}
        \centering
        \includegraphics[width=\linewidth]{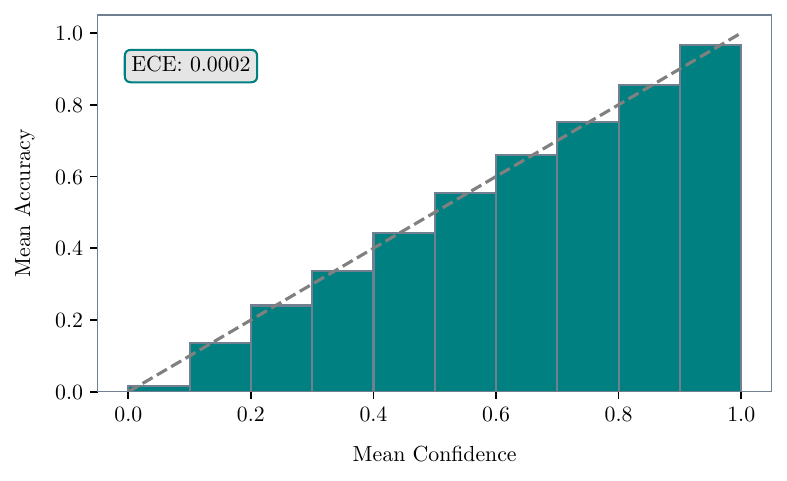}
        \caption{Calibration plot after post-hoc calibration.}
    \end{subfigure}
    \caption{Calibration plots for Grounding DINO on COCO before and after post-hoc calibration through isotonic regression.}
\label{fig:gdino_calibration_plot_before_and_after}
\end{figure}

\subsection{Processing of Raw Predictions}
\label{sec: appendix - predictions processing}

By default, detection models only output a confidence or score, which reflects the probability for the most likely class. We adjusted post-processing of predictions to extract probabilities for all classes. This information is reflected by the logit scores, which however are usually discarded for classes other than the most likely one. We recover this information by according adjustment of post-processing of raw predictions without affecting original predictions. 

Most models predict probabilities independently for each class. Hence, these probabilities do not sum up to 1. Further, a background probability is not explicitly predicted. Nonetheless, a probability distribution can be obtained the following way: First, we introduce a background probability as the product of individual counter probabilities. More precisely, let $p_i, i\in\{1,\ldots,C\}$ be the predicted probability for class $i$. Then, we define the background probability $p_{bg}:=\prod_{i=1}^C (1-p_i)$, which reflects the product model assuming that class predictions are independent. Second, after including the background probability, we normalize all class probabilities to obtain a predicted class distribution that also includes background.

We follow this procedure for all models except for RT-DETR and Faster R-CNN. The RT-DETR model outputs a softmax distribution for each predicted bounding box already, which we extend with a zero probability for the background class, as this aligns with the model design. Due to its two-staged design, Faster R-CNN outputs a probability distribution directly, which also includes a background probability. Still, post-processing had to be modified slightly: Originally, Faster R-CNN adjusts initial region proposals depending on class predictions. In particular, each region proposal is refined with respect to each class independently and thresholding is then conducted over the resulting set of $n_p C$ bounding boxes. Instead, we keep only the refined region proposal for the most likely class and associate the class probabilities with this bounding box.

To model background probabilities, we first have to ensure that the predicted probabilities are calibrated such that model confidence and probability of correct prediction are better aligned. Post-hoc calibration of neural networks and calibration plots were introduced by~\cite{Chuan2017}. Adaptations of the expected calibration error (ECE) to the domain of object detection, incorporating geometric correctness have been proposed~\cite{Kuzucu2024_calibration_object_detection}. However, our work focuses on semantic alignment of predictions and labels, so we analyze calibration of predictions for a fixed IoU threshold of $0.5$. Consequently, we evaluate correctness of predictions on  validation data following COCO evaluation protocol. Through $k$-fold cross validation ($k=5$), we train an isotonic regression model using sklearn's implementation~\cite{scikit-learn}. We also apply a logistic regression model for reference, which is referred to as Platt scaling. We use original annotations for post-hoc calibration. The regression models learn to adjust confidence scores based on binary targets corresponding to correctness of predictions. For simplicity, we do not learn separate models for each class but only one model per dataset. Overall, ECE is drastically reduced through post-hoc calibration for all models and datasets as can be seen in \autoref{tab:calibration_results}. Since isotonic regression outperforms Platt scaling, we chose this method for post-hoc calibration. A comparison of calibration plots for raw predictions and calibrated predictions is show in \autoref{fig:gdino_calibration_plot_before_and_after}. It can be concluded that post-hoc calibration significantly improves the alignment of model confidence with the probability of that prediction being indeed correct. After applying the calibration models to all class probabilities, we can compute a background probability and obtain a full probability distribution that we can compare to soft labels.

\input{tables/soft_label_evaluation_coco}
\input{tables/soft_label_evaluation_VOC}
\input{tables/soft_label_evaluation_KITTI}
\input{tables/soft_label_evaluation_Cityscapes}

\subsection{Additional Results of Soft Label-based Evaluation}
\label{sec: appendix - additional soft label benchmark results}
Here, we list additional evaluation results for soft label-based evaluation of current object detectors. We discuss performance metrics and evaluation protocol in the main paper where we also listed results for the COCO dataset. \autoref{tab:soft_label_based_benchmark_coco}, \autoref{tab:soft_label_based_benchmark_VOC}, \autoref{tab:soft_label_based_benchmark_KITTI}, and \autoref{tab:soft_label_based_benchmark_Cityscapes} list the results for COCO 2017, Pascal VOC 2012, KITTI, and the Cityscapes validation data. Across datasets, Grounding DINO achieves best performance in terms of Average Similarity but all models exhibit a significant deviation from soft label probabilities represented by high TVD values particularly for geometrically matched predictions.

\input{tables/soft_label_training_results}

\subsection{Improving Model Confidence Calibration with Soft Label Information}
\label{subsec: appendix - object detection benchmark - soft label training}

Object detectors are typically trained with a combination of spatial and semantic loss so that they learn both location and class affiliation of objects. 
Here, we modify the semantic loss by incorporating soft label information and analyze how training on soft labels affects model calibration and performance.
We fine-tune COCO-pretrained RT-DETR and YOLOv8 on our annotations using soft labels and compare the results to fine-tuning on the corresponding hard labels with the same training scheme. To train on soft labels, we only modify the targets for the Varifocal Loss used in RT-DETR and Binary Cross Entropy for YOLOv8.
We first construct two matching sets of annotations. On the one hand, the soft labels and on the other hand, the hard labels that result from thresholding at $p\geq 0.5$. We include only soft labels that meet the criterion that the maximal probability exceeds $0.5$ so that the annotation sets match exactly. We apply the same training setup as described in section~\ref{sec: appendix - training details} and fine-tune COCO-pretrained models over 100 epochs with the only difference being the targets that enter the classification loss. As model architectures expect hard labels, we inject the corresponding soft labels right before loss computation, which naturally works with soft labels, i.e., binary cross entropy in YOLOv8 and varifocal loss in RT-DETR are directly compatible with soft labels. As the range of the classification loss remains of the same order, we keep the weights for individual loss terms at default values.

\autoref{tab:training_comparison} compares model performance in terms of AP, AS as well as model calibration in terms of ECE and distance measures for the two training strategies. 
For RT-DETR, training with soft labels consistently improves calibration in terms of ECE, TVD and $\sqrt{\mathrm{JS}}$, often also improving detection performance in terms of AP and AS. Results are mixed for YOLOv8-L with improved ECE on two out of four datasets.
However, the effect of soft labels may not be fully unlocked as model architectures and hyperparameters are tailored to the use of hard labels, which limits uncertainty-aware object detection. 
Overall, these findings indicate that training on soft labels can improve model calibration and uncertainty quantification. However, calibration gains remain limited which may be due to the current design of loss functions. For instance, the focal loss introduces a focusing factor that scales the cross entropy loss according to predicted confidence and additionally uses pre-computed class weights for refactoring. While these mechanisms proved effective in terms of detection performance, they may hinder the calibration of models towards soft label probabilities.

\subsection{Cross-Evaluation of Downstream Detection Performance}
\label{subsec: appendix - object detection benchmark - training cross evaluation}

\begin{table}[t]
    \centering
    \caption{Change in AP, when training on our validated annotations, compared to training on original annotations. We evaluate on both sets of annotations and report changes in AP in percentage points. Positive numbers indicate that training on our validated annotations improved AP.}
    \label{tab:cross_training_results}
    \begin{subtable}[t]{0.32\linewidth}
        \centering
        \caption{Cityscapes}
        \begin{tabular}{lrr}
            \toprule
            Model & Eval=Orig & Eval=Ours \\
            \midrule
            Faster R-CNN & $+0.7$ & $+1.5$ \\
            RT-DETR-L    & $-0.1$ & $+1.1$ \\
            YOLOv8-L     & $-0.5$ & $+0.4$ \\
            \bottomrule
        \end{tabular}
    \end{subtable}
    \hfill
    \begin{subtable}[t]{0.32\linewidth}
        \centering
        \caption{KITTI}
        \begin{tabular}{lrr}
            \toprule
            Model & Eval=Orig & Eval=Ours \\
            \midrule
            Faster R-CNN & $-36.3$ & $+28.6$ \\
            RT-DETR-L    & $-44.0$ & $+34.1$ \\
            YOLOv8-L     & $-43.4$ & $+27.5$ \\
            \bottomrule
        \end{tabular}
    \end{subtable}
    \hfill
    \begin{subtable}[t]{0.32\linewidth}
        \centering
        \caption{Pascal VOC}
        \begin{tabular}{lrr}
            \toprule
            Model & Eval=Orig & Eval=Ours \\
            \midrule
            Faster R-CNN & $-2.7$ & $+0.3$ \\
            RT-DETR-L    & $+0.1$ & $+1.8$ \\
            YOLOv8-L     & $-2.2$ & $+0.9$ \\
            \bottomrule
        \end{tabular}
    \end{subtable}
\end{table}

Here, we investigate downstream performance of models when training on different sets of annotations. To this end, we compare performance in terms of AP of models when trained on original data, compared to training on our refined annotations.
We evaluate the absolute changes in AP (in percentage points) on both sets of annotations and list the results in the tables below. For the difference in AP, we use $\Delta AP = AP\textsubscript{trained on ours} - AP\textsubscript{trained on orig.}$.
We use the same training parameters as listed in~\ref{sec: appendix - training details} and list the results in~\autoref{tab:cross_training_results}.
As can be expected, a model generally performs best when trained and evaluated on the same distribution of annotations. However, we also observe that for some models (such as RT-DETR) and datasets (such as Cityscapes), performance increases on both sets of annotations when training on our refined data. Again, we observe a significant discrepancy of original and validated annotations for KITTI. To assess performance of trained models independently, a third set of annotations or another task would be required.
From the main experiments in section \ref{sec:object detection benchmark}, when looking at the performance of open-vocabulary models that are not biased toward any of the GT distributions, we see that high-capacity models such as Grounding DINO score better on our annotations (e.g., on KITTI and Cityscapes). This is presumably due to the addition of previously missing objects and we interpret performance increase of open-vocabulary models as a signal for previously insufficient data quality.

\subsection{Visual Examples}
\label{subsec: appendix - object detection benchmark - visual examples}

\paragraph{Overconfident Ground Truth Annotations}
To illustrate the importance of soft labels in the domain of object detection, we provide a sample of original annotations that are also present in our validated GT but the corresponding soft label indicates a certain degree of ambiguity. To sample these cases, we consider objects present in both sets of annotations (IoU $\geq$ 0.9) where the highest soft label probability is in the range of 0.5 to 0.8, so ambiguity is indicated. Samples for each dataset are shown in \autoref{fig:visual-overconfident-gt} where we see hard to identify background objects or objects that exhibit class ambiguity due to related classes such as car and van in KITTI or unusual representations, for instance a different type of cow that may not be identified as such by every annotator.
\input{imgs/visual_examples/overconfident_gt}

\paragraph{Divergence of Model Predictions and GT Soft Labels}
To give an intuition for the discrepancy of model predictions and ground truth soft labels, we illustrate examples of disagreement for the YOLOv8 model on each dataset in \autoref{fig:visual-model-gt-discrepancy}. We excluded the trivial case where the model simply did not detect the object at all to sample meaningful cases here. A pattern across examples is that the model is mislead by visual similarity of similar classes (e.g., a bowl that looks like a cup when viewed from above or a table that looks similar to a bench or distinguishing vans and regular cars). There are also edge cases where a cyclist has a foot on the ground or just unusual representations of a class that are not frequent in the training data distribution.
\input{imgs/visual_examples/model_confidence_discrepancy}

\FloatBarrier

\section{Label Error Detection Benchmark}
\label{appendix: label error benchmark}

\subsection{Label Error Detection Methods}
\label{appendix: label error benchmark - methods}

For this benchmark, we consider five different label error detection methods.

In the \textit{naive baseline}, we analyze the predictions of an object detector and treat false positives with respect to the ground truth as label error proposals, as they indicate discrepancies between predictions and annotations. A prediction is considered a false positive (FP) if its IoU with any ground truth object of the same class is below 0.5. The \textit{score baseline} extends this approach by additionally incorporating the object detector's confidence score, which is used for thresholding during evaluation.

The \textit{loss inspection} method by~\cite{loss-based-method} is based on instance-wise object detection loss. As the method depends on the underlying detectors architecture, we apply it only to predictions from Cascade R-CNN, following the original work. 

\textit{ObjectLab}~\cite{objectlab} assigns label quality score at the image level, where lower scores indicate potentially erroneous annotations. Since our benchmark requires bounding box-level label error proposals, we first compute label quality score for all images and then apply the naive baseline to extract false positive predictions. Each FP is assigned a score by combining the image-level label quality $s_\textit{img}$ with the detector's objectness score $s$, i.e., $(1-s_\textit{img})\cdot s$. This prioritizes high-confidence predictions from images with low label quality sores, which are more likely to contain label errors. Compared to the score baseline, which ranks FPs solely by objectness, this approach incorporates a global image-level signal and thus yields a different ranking, emphasizing label errors in systematically problematic images rather than only highly confident detections.

\textit{MetaDetect}~\cite{metadetect} is a meta classification approach based on gradient boosting. We use the default hyperparameters and perform predictions using 5-fold cross-validation on the respective hold-out set. To generate training labels for the meta classifier, we apply an IoU threshold of 0.5 to distinguish between correct predictions and false positives, thereby converting IoU values into binary labels. For each predicted bounding box, the model outputs a score indicating the likelihood of the prediction being correct. We use this score to identify label errors by selecting false positive predictions and ranking them in descending order according to the MetaDetect score. Thus, high-scoring false positives represent predictions that MetaDetect considers likely to be correct despite their mismatch with the ground truth, indicating potential label errors.  

\input{tables/preds_perf_gt_table}
\input{tables/cascade_rcnn_training_parameters}

\subsection{Prediction Networks}
\label{appendix: label error benchmark - prediction networks}

Label error detection methods for object detection datasets operate on both ground truth annotations and predictions from an object detector, analyzing their agreement and discrepancies to identify potential label errors. 

\paragraph{Grounding DINO}

We first employ Grounding DINO to generate these predictions, using a box threshold of 0.01 and a text threshold of 0.2. Subsequently, we apply non-maximum suppression (NMS) with an IoU threshold of 0.6 across all datasets to suppress duplicate detections. Since the predictions serve as input for label error detection, we choose comparatively relaxed confidence thresholds to increase the number of candidate detections and improve coverage of potential discrepancies with the ground truth annotations. 

The resulting detection performance with respect to the ground truth annotations is reported in \autoref{tab:gd_detection_original}. These results are not obtained through dataset-specific optimization, but reflect the behavior of the model under settings chosen for label error detection. The observed performance differences across datasets are consistent with their characteristics: performance is highest on Pascal VOC due to simpler scenes and larger objects, while COCO, Cityscapes, and KITTI present increasing challenges in terms of scene complexity, object density, and the presence of small or distant objects. In particular, KITTI shows the lowest performance, which can be attributed to a strong domain shift as well as the presence of small and distant objects such as pedestrians and cyclists.

Since Grounding DINO does not directly provide class-wise probabilities, we first obtain prediction bounding boxes by applying low box and text thresholds. For each predicted bounding box, we aggregate the token-level scores into class-level scores by grouping tokens associated with each class and taking the maximum score within each group. The predicted class is then determined as the class with the highest aggregated score, and the corresponding class score is used as the confidence score of the prediction. To obtain a distribution-like representation, we normalize the resulting class scores across all classes. Finally, we apply class-wise non-maximum suppression to remove redundant predictions.

\begin{figure}[htbp]
    \centering

    \resizebox{0.85\linewidth}{!}{%
    \begin{minipage}{\linewidth}
        \centering

        \begin{subfigure}{0.44\linewidth}
            \centering
            \includegraphics[width=\linewidth]{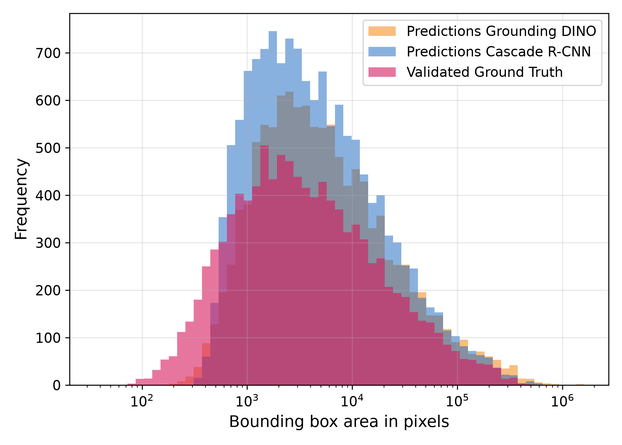}
            \caption{Cityscapes}
        \end{subfigure}
        \hfill
        \begin{subfigure}{0.44\linewidth}
            \centering
            \includegraphics[width=\linewidth]{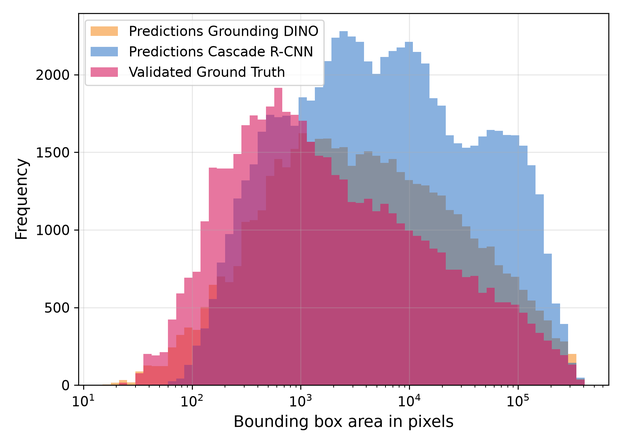}
            \caption{COCO}
        \end{subfigure}

        \vspace{0.1cm}

        \begin{subfigure}{0.44\linewidth}
            \centering
            \includegraphics[width=\linewidth]{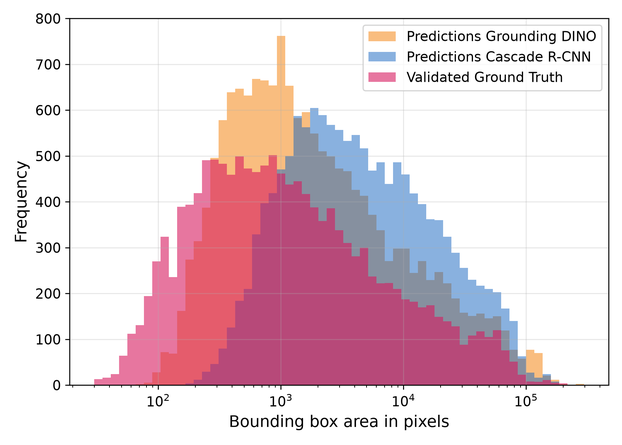}
            \caption{KITTI}
        \end{subfigure}
        \hfill
        \begin{subfigure}{0.44\linewidth}
            \centering
            \includegraphics[width=\linewidth]{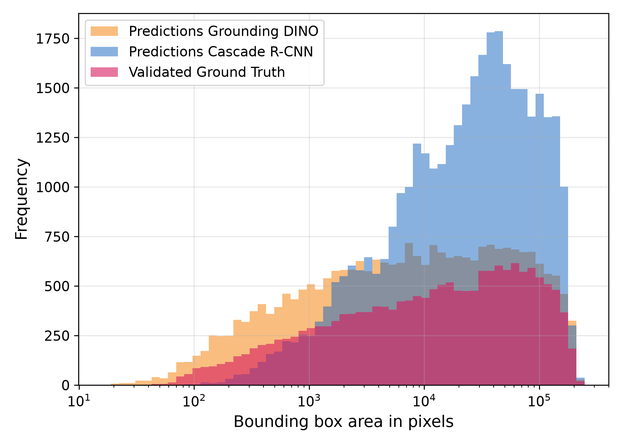}
            \caption{Pascal VOC}
        \end{subfigure}

    \end{minipage}%
    }

    \caption{Distribution of bounding box areas for validated ground truth annotations compared to Grounding DINO and Cascade R-CNN predictions across the evaluated datasets. For visualization purposes, predictions are filtered using a confidence threshold of 0.2, while the label error detection benchmark itself uses a lower threshold of 0.01 to maximize object coverage.}
    \label{fig:bbox_area_distribution}
\end{figure}

\paragraph{Cascade R-CNN}

As a second prediction network for the label error detection benchmark, we employ Cascade R-CNN~\cite{cascadercnn}. The training parameters are summarized in \autoref{tab:cascade_rcnn_training}. We follow the configuration proposed by~\cite{loss-based-method} and use it as a guideline, while selecting suitable parameters for the Cityscapes dataset, which is not covered in their work. We train the model on a single NVIDIA A100 GPU with 80 GB of RAM, which takes about 12--15 h per dataset, so overall not more than 3 days for all datasets on this hardware.

To obtain proposals suitable for label error detection, we apply a very low score threshold of 0.01 in order to retain a high number of candidate detections. Additionally, non-maximum suppression with an IoU threshold of 0.6 is used to remove duplicate detections while still preserving closely overlapping predictions. This setup prioritizes high recall over precision, which is essential for identifying potential label errors.

The resulting detection performance with respect to the ground truth annotation is reported in \autoref{tab:cascade_detection_original}. It is important to note that these results are not obtained through dataset-specific optimization, but rather through a unified configuration designed to ensure broad object coverage across all datasets. The confidence scores correspond directly to the softmax probabilities produced by the network.

\paragraph{Distribution of Bounding Box Areas}

\autoref{fig:bbox_area_distribution} compares the bounding box area distribution of the validated ground truth annotation with the predictions of Grounding DINO and Cascade R-CNN. Across all datasets, both models broadly follow the distribution of the validated annotation, indicating that they capture the overall objects scale characteristics of the data. However, systematic differences can be observed, particularly for small objects. The validated ground truth  contains a larger fraction of small bounding boxes, whereas the predictions of both models are shifted towards medium and larger object sizes. This effect is more pronounced for Grounding DINO, while Cascade R-CNN generally demonstrates a slightly better coverage of smaller objects, although it still underrepresents them compared to the validated annotations. This behavior is expected, as small objects are inherently more difficult to detect due to their limited visual detail and higher sensitivity to localization label errors. In addition, they are often underrepresented in training data, which further complicates learning robust representations. The discrepancy is especially evident in datasets with many small or distant objects, such as KITTI and Cityscapes. In these cases, both models produce fewer predictions in the lower area ranges, consistent with the reduced detection performance observed on these datasets.

\input{tables/label_error_benchmark_table_appendix}

\subsection{Label Error Detection Benchmark - Stricter LEGT}
\label{appendix: label error benchmark - stricter legt}

As in the main paper, label error detection focuses on prioritizing label error proposals such that true annotation errors are ranked highly while minimizing manual review effort. We follow the same evaluation protocol as in the main paper and use predictions generated by Grounding DINO. All methods are evaluated at the method-specific threshold that maximizes the F1-score. An IoU threshold of 0.1 is used to emphasize high recall. In contrast to the less restrictive setting, we now evaluate on a stricter label error ground truth that includes only larger label errors (with a minimum height of 40 pixels) and high-confidence soft labels ($p\geq 0.8$), thereby focusing on more obvious label errors (\autoref{tab:label_error_benchmark_appendix}).

Compared to the results on the less restrictive label error ground truth (\autoref{tab:label_error_benchmark_main}), we observe a consistent increase in recall across most methods, particularly for the naive baseline, which achieves near perfect recall across all datasets. This indicates that larger and more confident label errors are generally easier to detect. ObjectLab also benefits from this setting, showing improved recall on most datasets. However, this increase in recall does not translate into a corresponding improvement in overall performance. Precision remains low for most methods, indicating that a large fraction of the proposed label error candidates are still false positives. The loss inspection method continues to generate many proposals, achieving moderate recall but at the cost of very low precision. In contrast, the score baseline remains more conservative, with moderate recall but improved precision compared to the naive baseline. MetaDetect becomes more competitive in the stricter setting, achieving the highest F1-score on COCO, Kitti, and Pascal VOC, while remaining slightly below the score baseline on Cityscapes.

\begin{figure*}[htbp]
    \centering

    \resizebox{0.85\linewidth}{!}{%
    \begin{minipage}{\linewidth}
        \centering
        
        \begin{subfigure}[t]{0.42\textwidth}
            \centering
            \includegraphics[width=\linewidth]{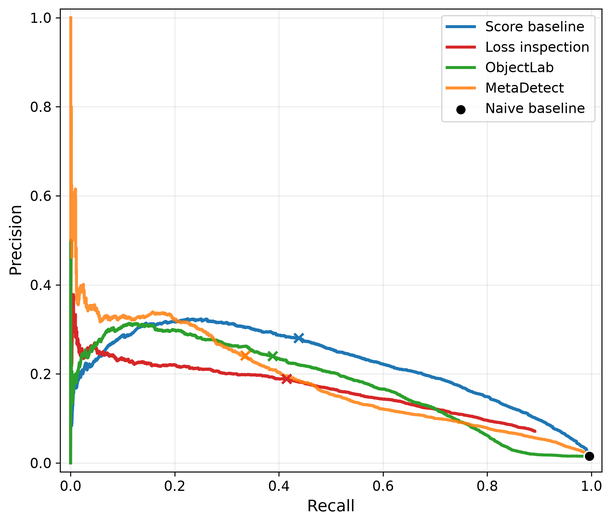}
        \end{subfigure}
        \hfill
        \begin{subfigure}[t]{0.42\textwidth}
            \centering
            \includegraphics[width=\linewidth]{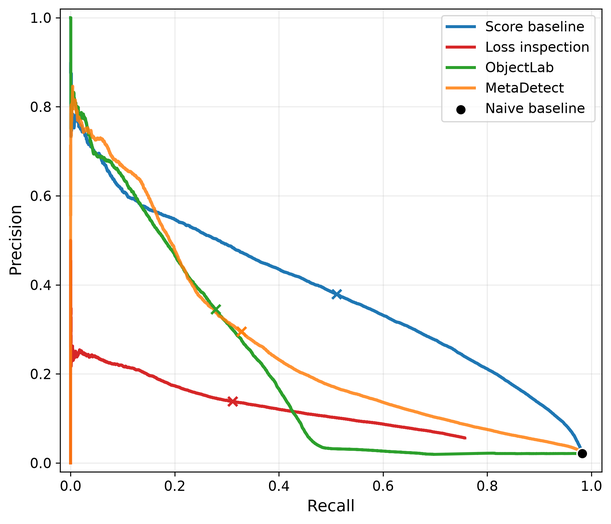}
        \end{subfigure}

        \vspace{0.3cm}

        \begin{subfigure}[t]{0.42\textwidth}
            \centering
            \includegraphics[width=\linewidth]{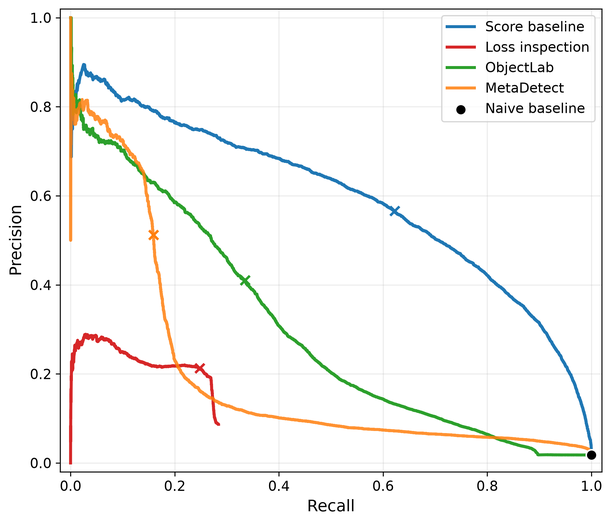}
        \end{subfigure}
        \hfill
        \begin{subfigure}[t]{0.42\textwidth}
            \centering
            \includegraphics[width=\linewidth]{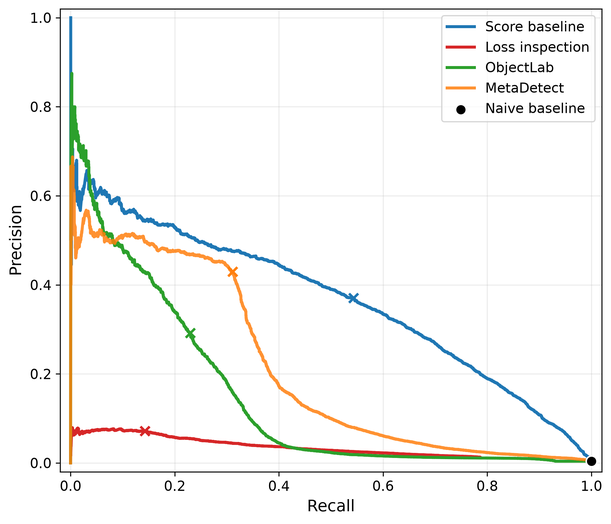}
        \end{subfigure}

    \end{minipage}%
    }

    \caption{
    Precision-recall curves of label error detection methods across all datasets, evaluated on the less restrictive label error ground truth. All methods operate on predictions from Grounding DINO, except for the loss-based method, which uses Cascade R-CNN. The cross on each curve indicates the operating point with the highest F1-score, at which all methods are evaluated. The naive baseline is shown for reference but is not directly comparable, as it has no ranking score (and thus no AP) and is evaluated at a single operating point.
    }
    
    \label{fig:pr_curves_all_datasets}
\end{figure*}

\subsection{Precision-Recall Analysis}
\label{appendix: label error benchmark - pr curves}

The precision-recall curves in \autoref{fig:pr_curves_all_datasets} illustrate clear differences in how the evaluated methods rank label error proposals. All methods operate on predictions from Grounding DINO, except for the loss-based method, which relies on Cascade R-CNN due to architectural constraints. All results are reported on the less restrictive label error ground truth. The naive baseline is included for reference but is not directly comparable, as it has no ranking score and is evaluated at a single operating point.

The score baseline consistently achieves the strongest performance across datasets and closely approaches the naive baseline at high recall. This behavior is expected, as both methods rely on the same set of false positive predictions. However, the score baseline benefits from a meaningful ranking based on the detector's confidence, which prioritizes more informative proposals earlier in the ranking. As a result, it recovers most of the label errors identified by the naive baseline while maintaining substantially higher precision.
ObjectLab follows a similar trend to the score baseline but demonstrates different ranking behavior. By combining detector confidence with an image-level label quality score, it reorders the label error proposals compared to the score baseline. Since image-level scores are available for all images, ObjectLab considers the same set of proposals as the baseline and achieves full coverage, comparable to the naive and score baseline. However, despite reaching the same maximum recall, the different ranking of proposals leads to a changed precision-recall trade-off and thus different overall performance. In particular, ObjectLab prioritizes proposals from images with lower estimated label quality, which can improve precision in certain regions of the curve but may also lead to suboptimal ordering compared to the score baseline.
In contrast, loss inspection and MetaDetect demonstrate fundamentally different behavior. The loss inspection method tends to assign high scores to a large number of proposals, resulting in relatively high recall but low precision. This indicates that the loss score is not sufficiently discriminative for identifying real label errors, leading to many false positives. MetaDetect, on the other hand, produces a different ranking of proposals and achieves competitive precision in some regions of the curves, but remains below the score baseline in overall performance across datasets. This suggests that the MetaDetect score provides useful information for prioritizing label errors, but does not consistently match the ranking quality of the detector confidences scores.

Importantly, the differences in behavior are largely driven by how each method defines and ranks label error proposals. While all methods operate on predictions from an object detector, they not only assign scores based on fundamentally different signals (confidence, loss, image-level quality, or estimated prediction correctness), but may also differ in how label error proposals are generated, leading to variations in the number of proposals considered. These differences introduce two key factors: the quality of the ranking signal and the coverage of the considered proposals. As a result, methods may produce comparable numbers of label error proposals but differ significantly in how effectively they prioritize true label errors, or they may operate on different subsets of proposals altogether, further affecting recall and overall performance.

\input{tables/cascade_label_error_benchmark_less_restrictive}
\input{tables/cascade_label_error_benchmark_stricter}

\begin{figure*}[htbp]
    \centering

    \resizebox{0.85\linewidth}{!}{%
    \begin{minipage}{\linewidth}
        \centering
        
        \begin{subfigure}[t]{0.42\textwidth}
            \centering
            \includegraphics[width=\linewidth]{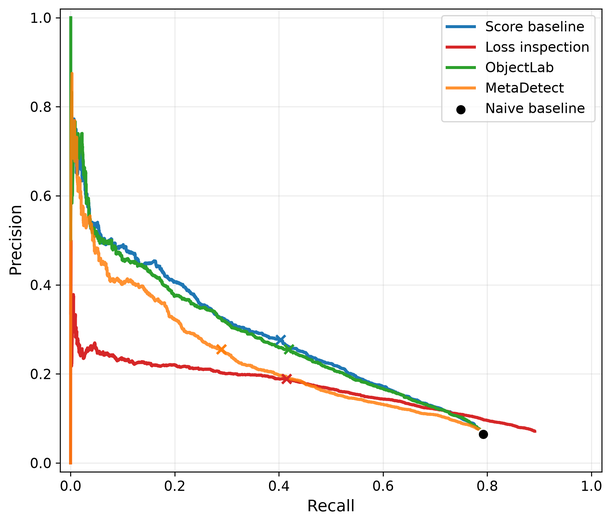}
        \end{subfigure}
        \hfill
        \begin{subfigure}[t]{0.42\textwidth}
            \centering
            \includegraphics[width=\linewidth]{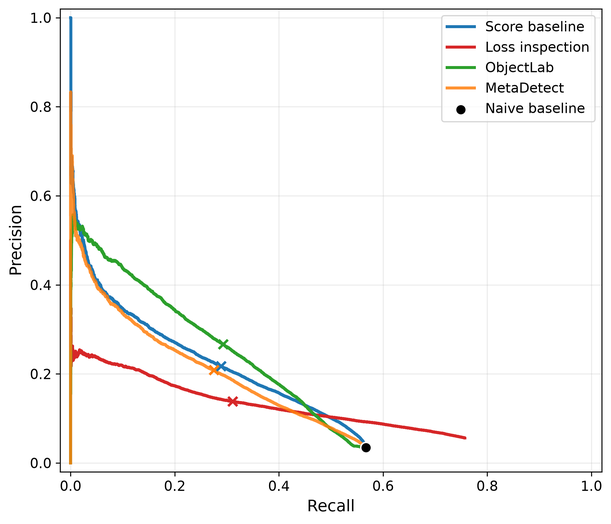}
        \end{subfigure}

        \vspace{0.3cm}

        \begin{subfigure}[t]{0.42\textwidth}
            \centering
            \includegraphics[width=\linewidth]{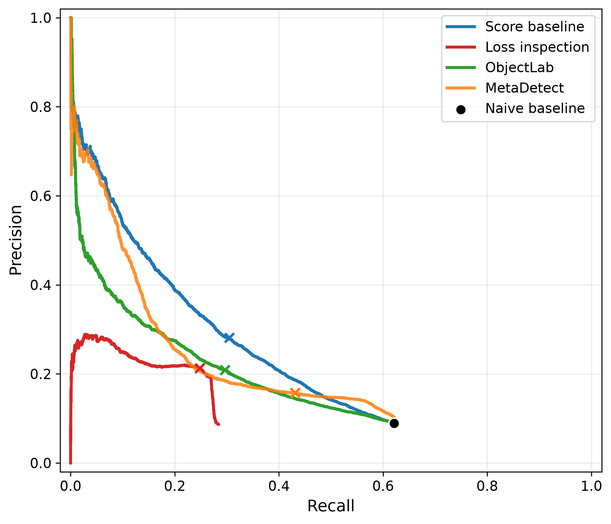}
        \end{subfigure}
        \hfill
        \begin{subfigure}[t]{0.42\textwidth}
            \centering
            \includegraphics[width=\linewidth]{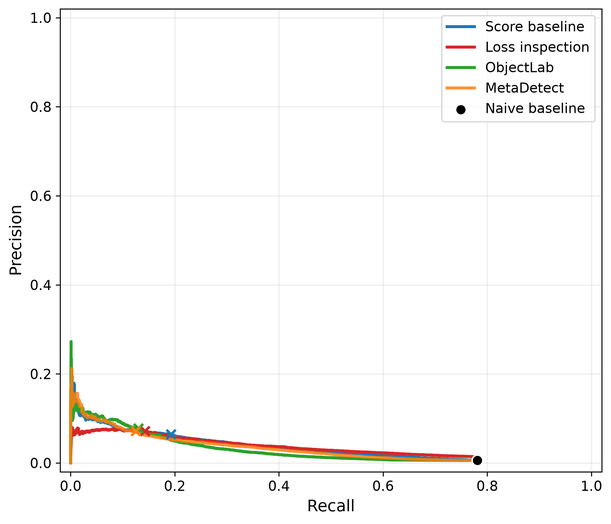}
        \end{subfigure}

    \end{minipage}%
    }

    \caption{
    Precision-recall curves of label error detection methods across all datasets, evaluated on the less restrictive label error ground truth. All methods operate on predictions from Cascade R-CNN. The cross on each curve indicates the operating point with the highest F1-score, at which all methods are evaluated. The naive baseline is shown for reference but is not directly comparable, as it has no ranking score (and thus no AP) and is evaluated at a single operating point.
    }
    
    \label{fig:cascade_pr_curves_all_datasets}
\end{figure*}

\subsection{Label Error Detection Benchmark Cascade-RCNN}
\label{appendix: label error benchmark - cascade}

Using Cascade R-CNN as unified prediction network enables a direct and fair comparison between all evaluated label error detection methods. The results are shown in \autoref{tab:cascade_label_error_benchmark_less_Restrictive} and \autoref{tab:cascade_label_error_benchmark_stricter} as well as in \autoref{fig:cascade_pr_curves_all_datasets}. Across all dataset, the score baseline achieves strong overall performance, frequently yielding the highest average precision and a favorable precision-recall trade-off. This indicates that the objectness score already provides a strong signal for identifying label errors when evaluated on Cascade R-CNN predictions. The naive baseline, which considers all false positive predictions equally, achieves the maximum possible recall but suffers from very low precision, highlighting the importance of effective ranking.

ObjectLab reaches the same maximum recall as the naive baseline, confirming that it operates on the full set of proposals. However, its different ranking strategy leads to a weaker precision-recall trade-off compared to the score baseline on most datasets. While ObjectLab prioritizes predictions from images with lower estimated label quality, this reordering does not consistently improve the ranking of true label errors, although it achieves the strongest performance on COCO and competitive performance on Pascal VOC. The loss inspection shows moderate performance but remains below the strongest methods across most datasets. Its behavior suggests that loss-based signals contain useful information for detecting label errors, but are less reliable than direct confidence-based ranking in most setting. MetaDetect also demonstrates competitive performance in some cases, particularly in terms of recall on KITTI, but does not consistently outperform the score baseline across datasets and label error definitions.

The observed trends are consistent across both the less restrictive and stricter label error settings. While the stricter setting generally leads to higher recall values, due to focusing on larger and more confident label errors, the relative ranking of methods varies across datasets. This suggests that both the choice of label error detection method and the exact definition of label errors influence performance. These results demonstrate that, when evaluated on Cascade R-CNN predictions, the score baseline remains a strong and robust approach, while more complex methods can provide better rankings in individual settings but do not consistently outperform it across datasets.

\subsection{Prediction Network and Practical Implications}
\label{appendix: label error benchmark - choice prediction}

Compared to the results obtained using Cascade R-CNN predictions, several notable differences can be observed when evaluating on Grounding DINO predictions. Most prominently, the overall performance of all methods is generally higher, particularly in terms of average precision. This indicates that Grounding DINO provides prediction scores that are more informative for ranking label error proposals, leading to a clearer separation between true and false positives. In contrast to the Cascade R-CNN setting, the gap between the score baseline and the other methods is more notable. The score baseline consistently achieves the strongest performance across all datasets, suggesting that the confidence estimates of Grounding DINO are particularly well calibrated for this task. ObjectLab and loss inspection benefit from this as well, but still fail to match the ranking quality of the raw confidence scores. Another key difference lies in the behavior of ObjectLab. While it achieves full coverage similar to the naive baseline in both settings, its relative performance is slightly stronger when using Grounding DINO predictions. This suggests that the image-level label quality scores align better with the prediction characteristics of Grounding DINO than with those of Cascade R-CNN. Finally, MetaDetect shows a stronger dependence on the prediction network and label error definition. While it remains below the score baseline in the less restrictive setting, it becomes more competitive under the stricter label error definition, particularly when using Grounding DINO predictions. This highlights that its meta-classification score is sensitive to the underlying detector and does not generalize equally well across different prediction models and label error definitions. These observations demonstrate that the effectiveness of label error detection methods is strongly influenced by the properties of the underlying prediction network, especially the quality of its predictions.

These findings also provide practical insights for dataset cleaning. In particular, the strong performance of the score baseline indicates that detector confidence can serve as a simple and effective signal for prioritizing candidate label errors for manual review. This makes confidence-based ranking a practical starting point for semi-automated dataset cleaning. More sophisticated methods such as MetaDetect and ObjectLab can provide improved rankings in individual settings, but their effectiveness depends more strongly on the underlying dataset, prediction network, and label error definition. Overall, our benchmark provides a foundation for systematically evaluating and developing future label error detection methods for real-world dataset cleaning.

\subsection{Visual Examples of Label Errors}
\label{appendix: label error benchmark - examples}

In the following, we present examples of missing bounding boxes (\autoref{fig:missing_label_errors}), misaligned bounding boxes (\autoref{fig:misaligned_label_errors}), and classification label errors (\autoref{fig:classification_label_errors}). Ground truth bounding boxes are shown in green, while label errors are highlighted in red, together with their associated labels and soft label probabilities $p$. The shown examples are randomly sampled, and each image highlights a single label error. Additional label errors may still be present but are not explicitly marked. Due to the definition of label errors, some misaligned bounding boxes may effectively correspond to missing bounding boxes, as their IoU with neighboring boxes exceeds 0.1. 

\input{labelerrors/missing/plot_missing_label_errors}
\input{labelerrors/misaligned/plot_misaligned_label_errors}
\input{labelerrors/classification/plot_classification_label_errors}

\end{document}

%% file: tables/introduction_tables.tex
\begin{table}[t]
    \centering
    \small
        \caption{
    Summary of related annotation-focused datasets.
    In Table~\subref{tab:related_work}, \textbf{Detection (Det.)} refers to object detection datasets, \textbf{Re-annotation (Re-ann.)} denotes efforts to improve annotation completeness through systematic relabeling, \textbf{Uncertainty (Uncert.)} denotes annotation uncertainty captured via multiple annotators and soft labels, and \textbf{Multi-dataset (Multi-DS)} indicates application of a consistent methodology across multiple datasets.
    Prior work typically addresses either annotation completeness or uncertainty, while our work combines both across multiple object detection benchmarks.
    Table~\subref{tab:dataset_overview} summarizes the number of images and classes that were re-annotated in this paper for each dataset.
    }
    \label{tab:datasets_combined}

    \begin{subtable}[t]{0.44\linewidth}
\input{tables/introduction_related_work}
 \end{subtable}
    \hfill
    \begin{subtable}[t]{0.55\linewidth}
\input{tables/introduction_numbers}
    \end{subtable}

\end{table}

%% file: tables/introduction_related_work.tex
    \centering

\caption{Related datasets
}
\label{tab:related_work}

\resizebox{0.9\linewidth}{!}{
\begin{tabular}{lcccc}
\toprule
\textbf{Work} & \textbf{Det.} & \textbf{Re-ann.} & \textbf{Uncert.} & \textbf{Multi-DS} \\
\midrule
LVIS~\cite{gupta2019lvis}         & \checkmark & \checkmark &  &  \\
OpenImages~\cite{kuznetsova2020openimages} & \checkmark & \checkmark &  &  \\
ReLabel~\cite{Yun2021Correction}     &  & \checkmark & \checkmark &  \\
Pascal+~\cite{Kim2024Correction}  & \checkmark & \checkmark &  &  \\
CIFAR-10H~\cite{peterson2019human} &  &  & \checkmark &  \\
DCIC~\cite{schmarje2022benchmark} &  &   & \checkmark &  \checkmark \\
REC\checkmark D~\cite{rechecked} & \checkmark & \checkmark & \checkmark &  \\
\midrule
\textbf{Ours}                    & \checkmark & \checkmark & \checkmark & \checkmark \\
\bottomrule
\end{tabular}
}
    

%% file: tables/introduction_numbers.tex
\centering
\caption{Overview of re-annotated datasets}
\label{tab:dataset_overview}
\small
\resizebox{0.7\linewidth}{!}{
\begin{tabular}{lccc}
\toprule
Dataset & \#Train & \#Val & \#Cls. \\
\midrule
KITTI \cite{kitti} & 7,481 & -- & 7 \\
Cityscapes \cite{Cityscapes} & 2,975 & 500 & 8 \\
COCO \cite{COCO} & -- & 5,000 & 80 \\
Pascal VOC \cite{PascalVOC} & 5,717 & 5,823 & 20 \\
\bottomrule
\end{tabular}
}
    

%% file: methodgraph/methodgraph.tex
\begin{figure*}[t]
\centering
\resizebox{\textwidth}{!}{
\begin{tikzpicture}[
    node distance=0.1cm,
    stage/.style={
        rectangle,
        rounded corners,
        draw=black,
        line width=0.6pt,
        align=center,
        inner sep=4pt,
        minimum width=3.0cm
    },
    imagebox/.style={
        rectangle,
        draw=black,
        line width=0.5pt,
        inner sep=0pt
    },
    arrow/.style={
        -{Latex[length=2.2mm]},
        line width=0.7pt
    }
]


\node[imagebox] (img0) {
    \includegraphics[width=0.25\linewidth]{methodgraph/sofa.jpg}
};

\node[imagebox, right=0.1 of img0] (img1) {
    \includegraphics[width=0.25\linewidth]{methodgraph/sofa_keypoints.jpg}
};

\node[imagebox, right=0.1cm of img1] (img2) {
    \includegraphics[width=0.25\linewidth]{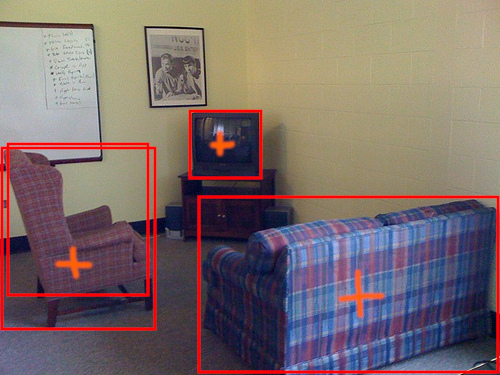}
};

\node[imagebox, right=0.1cm of img2] (img3) {
    \includegraphics[width=0.25\linewidth]{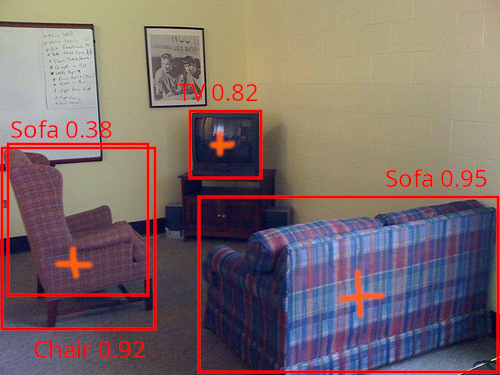}
};

\node[imagebox, right=0.1cm of img3] (img4) {
    \includegraphics[width=0.25\linewidth]{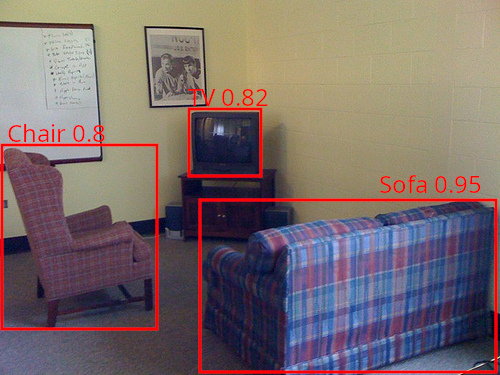}
};


\node[stage, above=0.35cm of img0] (s0) {
    Input Image \\
    No GT
};

\node[stage, above=0.35cm of img1] (s1) {
    Stage 1\\
    Add Keypoints
};

\node[stage, above=0.35cm of img2] (s2) {
    Stage 2\\
    Add Bounding Boxes
};

\node[stage, above=0.35cm of img3] (s3) {
    Stage 3\\
    Add Soft Labels
};

\node[stage, above=0.35cm of img4] (s4) {
    Post-Processing\\
    Validated GT
};


\end{tikzpicture}
}
\caption{Overview of our annotation pipeline. In stage 1, all potential objects of a given class are marked with keypoints. For each keypoint, a bounding box is then drawn and verified in stage 2. In the third stage, 11 annotators evaluate the proposed object, which results in a soft label. Ambiguous cases receive additional annotations and finally, post-processing removes duplicates and corrects bias. The resulting annotations are our validated ground truth.
}
\label{fig:reannotation_pipeline}
\end{figure*}

%% file: tables/data_quality_objects_missing.tex
\begin{table}[t]
\centering
\begin{minipage}{0.48\linewidth}
   \input{tables/gt_vs_vgt_main}
\end{minipage}
\hfill
\begin{minipage}{0.48\linewidth}
   \input{tables/false_negative_highlight}
\end{minipage}
\end{table}

%% file: tables/gt_vs_vgt_main.tex
\centering
\caption{Number of annotated objects in original vs.\ validated GT across datasets. We consider all objects in an image and also label previously ignored regions or groups like in COCO or KITTI.
}
\label{tab:gt_vs_vgt_all_main}
\resizebox{\linewidth}{!}{
\begin{tabular}{l|ccc}
\toprule
Dataset & Original & Validated (Ours) & $\Delta[\%]\uparrow $ \\
\midrule
Cityscapes & 62{,}649 & 65{,}856 & +5.1 \\
Pascal VOC  & 31{,}561 & 34{,}262 & +8.6 \\
KITTI      & 39{,}597 & 63{,}348 & +60.0 \\
COCO       & 36{,}781 & 51{,}487 &  +40.0 \\
\hline
\end{tabular}
}

%% file: tables/false_negative_highlight.tex
\centering
\caption{Comparison of false negative rates for original and validated GT (lower is better). Values are reported in \% with Wilson~\cite{wilson1927probable} 95\% confidence intervals.}
\label{tab:fn_highlight}
\small
\resizebox{0.8\linewidth}{!}{
\begin{tabular}{l cc}
\toprule
Dataset & Original & Validated (Ours) \\
\midrule
Pascal VOC  & $18.6 \pm 1.7$ & $9.3 \pm 1.3$ \\
COCO        & $13.1 \pm 1.0$ & $6.1 \pm 0.7$ \\
KITTI       & $41.5 \pm 1.3$ & $11.6 \pm 0.8$ \\
Cityscapes  & $11.1 \pm 0.6$ & $3.8 \pm 0.3$ \\
\bottomrule
\end{tabular}
}

%% file: tables/label_error_table_main.tex
\begin{table}[t]
\centering
\small
\caption{Comparison of identified label errors across datasets for two default label error configurations: a less restrictive setting ($p \geq 0.5$ considering all objects), and a stricter setting ($p \geq 0.8$ considering objects with height $\geq 40$ pixels). Percentages are given relative to the original ground truth. Counts are aggregated over training and validation data, except for COCO (validation only).}
\label{tab:label_error_table_main}
\setlength{\tabcolsep}{5pt}
\scalebox{1.0}{
\begin{tabular}{lccc|ccc}
\toprule
 & \multicolumn{3}{c|}{\textit{Less Restrictive}} 
 & \multicolumn{3}{c}{\textit{Stricter}} \\
Dataset & Missing & Misaligned & Classification & Missing & Misaligned & Classification \\
\midrule
Cityscapes 
& 5,249 (8.4\%) 
& 2,385 (3.8\%) 
& 85 (0.1\%) 
& 2,229 (3.6\%) 
& 1,318 (2.1\%) 
& 56 (0.1\%) \\

COCO       
& 16,219 (44.1\%) 
& 7,736 (21.0\%) 
& 254 (0.7\%) 
& 2,685 (7.3\%) 
& 3,497 (9.5\%) 
& 98 (0.3\%) \\

KITTI      
& 22,710 (57.4\%) 
& 7,167 (18.1\%) 
& 1,026 (2.6\%) 
& 961 (2.4\%) 
& 2,174 (5.5\%) 
& 479 (1.2\%) \\

Pascal VOC 
& 3,654 (11.6\%) 
& 2,210 (7.0\%) 
& 52 (0.2\%) 
& 761 (2.4\%) 
& 1,290 (4.1\%) 
& 25 (0.1\%) \\
\bottomrule
\end{tabular}}
\end{table}

%% file: tables/data_quality_bias.tex
\begin{figure}[tb]
\centering
\begin{minipage}{0.45\linewidth}
    \centering
    \includegraphics[width=\linewidth]{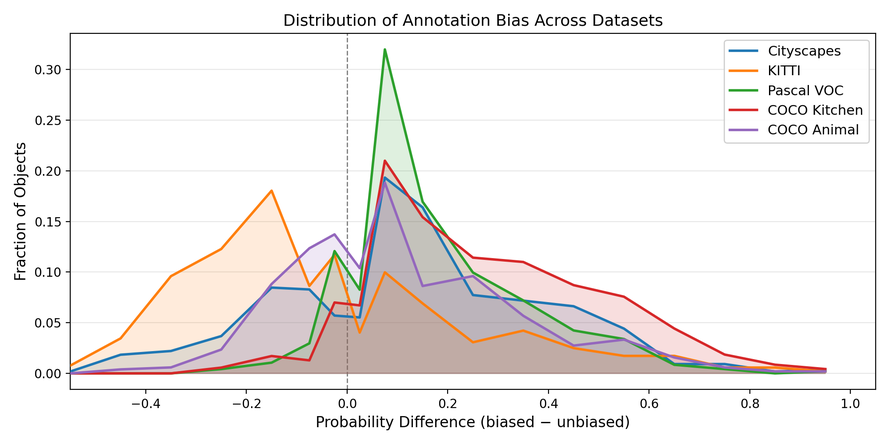}
    \captionof{figure}{Distribution of per-object annotation bias across selected datasets. Objects with near-zero difference ($|\Delta| < 0.01$) are excluded to better visualize distribution tails. Individual per-dataset histograms are provided in the appendix without near-zero filtering.}
    \label{fig:introduced_bias}
\end{minipage}
\hfill
\begin{minipage}{0.52\linewidth}
  \input{tables/post_processing_results}
\end{minipage}
\end{figure}

%% file: tables/post_processing_results.tex
\centering
\captionsetup{type=table}
\caption{Post-processing results per dataset.
We compare potentially biased annotations before and after post-processing to the unbiased annotations for the same objects. We report the relative reduction of soft label probability of the proposed class and Jensen-Shannon distance of soft labels~($\sqrt{\mathrm{JS}}$) to the unbiased reference.
Gray values show the corresponding absolute changes. See appendix for detailed table.}
\label{tab:post_processing_results}
\small
\resizebox{\linewidth}{!}{
\begin{tabular}{l c c}
\toprule
Dataset
  & Reduction [\%] $\downarrow$
  & $\Delta\!\sqrt{\mathrm{JS}}$ [\%] $\downarrow$ \\
\midrule
Cityscapes     & $-9.4$  {\color{gray}(+6.1 $\rightarrow$ +5.5 pp)}   & $-0.06$ {\color{gray}($-0.0001$)} \\
KITTI          & $-17.5$ {\color{gray}($-1.8 \rightarrow -1.5$ pp)}   & $-1.44$ {\color{gray}($-0.0019$)} \\
Pascal VOC      & $-54.5$ {\color{gray}(+7.0 $\rightarrow$ +3.2 pp)}   & $-7.16$ {\color{gray}($-0.0065$)} \\
\midrule
COCO/animal    & $-20.7$ {\color{gray}(+4.5 $\rightarrow$ +3.5 pp)}   & $-0.95$ {\color{gray}($-0.0009$)} \\
COCO/kitchen   & $-57.2$ {\color{gray}(+16.8 $\rightarrow$ +7.2 pp)}  & $-9.20$ {\color{gray}($-0.0173$)} \\
\bottomrule
\end{tabular}
}

%% file: tables/data_quality_calibration_wrongclass.tex
\begin{table}[tb]
\centering
\begin{minipage}{0.55\linewidth}
   \input{tables/human_perception}
\end{minipage}
\hfill
\begin{minipage}{0.42\linewidth}
   \input{tables/wrong_labels}
\end{minipage}
\end{table}

%% file: tables/human_perception.tex
\centering
\caption{Mean $\pm$ std of proposed-class probabilities grouped by expert review buckets: clearly ($\geq 0.75$), maybe ($0.25$--$0.75$), and not ($< 0.25$) correct, reflecting alignment of soft labels with expert perception.}
\label{tab:human_perception}
\resizebox{0.8\linewidth}{!}{
\begin{tabular}{l ccc}
\toprule
\textbf{Dataset} 
 & \textbf{Clearly} & \textbf{Maybe} & \textbf{Not}\\
\midrule
Cityscapes
 & .806 $\pm$ .262 & .448 $\pm$ .261 & .285 $\pm$ .209 \\

KITTI
 & .741 $\pm$ .306 & .424 $\pm$ .233 & .250 $\pm$ .197 \\

Pascal VOC
 & .843 $\pm$ .256 & .386 $\pm$ .242 & .258 $\pm$ .185 \\

COCO
 & .737 $\pm$ .321 & .419 $\pm$ .304 & .169 $\pm$ .188 \\
\bottomrule
\end{tabular}%
}

%% file: tables/wrong_labels.tex
\centering
\caption{Wrong class error rates (in \%) for original and validated GT, reported as Wilson confidence interval. Differences are small and not statistically significant.}
\label{tab:wrong_label}
\small
\resizebox{0.9\linewidth}{!}{
\begin{tabular}{l cc}
\toprule
Dataset & Original & Validated (Ours) \\
\midrule
COCO       & $3.9 \pm 2.5$ & $4.4 \pm 2.7$ \\
Cityscapes & $2.4 \pm 1.9$ & $3.9 \pm 2.5$ \\
KITTI      & $3.4 \pm 2.3$ & $3.5 \pm 2.4$ \\
Pascal VOC  & $1.4 \pm 1.3$ & $3.0 \pm 2.2$ \\
\bottomrule
\end{tabular}
}

%% file: tables/map_recall_object_detectors.tex
\begin{table}[tb]
\centering
\caption{Performance of object detectors trained on original GT and evaluated on original vs.\ validated GT. Underlined values indicate improvements $>10$ points, bold decreases $>10$ points.}
\label{tab:detection_org_validated}
\begin{subtable}[t]{0.49\linewidth}
\centering
\caption{AP$_{50..95}$ ($\uparrow$)}
\resizebox{\linewidth}{!}{
\begin{tabular}{lcc|cc|cc|cc}
\toprule
& \multicolumn{2}{c}{COCO}
& \multicolumn{2}{c}{Pascal VOC}
& \multicolumn{2}{c}{KITTI}
& \multicolumn{2}{c}{Cityscapes} \\
\cmidrule(lr){2-3} \cmidrule(lr){4-5} \cmidrule(lr){6-7} \cmidrule(lr){8-9}
{Model} & Orig. & Val. & Orig. & Val. & Orig. & Val. & Orig. & Val. \\
\midrule
Grounding DINO & \percent{0.5690} & \percent{0.5210} & \percent{0.8005} & \percent{0.7378} & \percent{0.2918} & \underline{\percent{0.3942}} & \percent{0.3419} & \percent{0.4382} \\
YOLOv8-World v2 & \percent{0.4453} & \percent{0.4377} & \percent{0.6965} & \percent{0.6544} & \percent{0.1986} & \percent{0.2444} & \percent{0.2411} & \percent{0.3035} \\
OWL-ViT & \percent{0.2868} & \percent{0.2855} & \percent{0.5275} & \percent{0.4921} & \percent{0.0928} & \percent{0.0955} & \percent{0.1160} & \percent{0.1378} \\
RT-DETR-L & \percent{0.5239} & \percent{0.4720} & \percent{0.7028} & \percent{0.6418} & \percent{0.7846} & \textbf{\percent{0.2733}} & \percent{0.3822} & \percent{0.4248} \\
YOLOv8-L & \percent{0.5168} & \percent{0.4779} & \percent{0.7442} & \percent{0.6739} & \percent{0.7468} & \textbf{\percent{0.2666}} & \percent{0.3814} & \percent{0.4435} \\
Faster R-CNN (ResNet 50) & \percent{0.4709} & \percent{0.4312} & \percent{0.6125} & \percent{0.5557} & \percent{0.6529} & \textbf{\percent{0.2672}} & \percent{0.3591} & \percent{0.4102} \\
\bottomrule
\end{tabular}
}
\end{subtable}%
\hfill%
\begin{subtable}[t]{0.49\linewidth}
\centering
\caption{AR$_{10}$ ($\uparrow$)}
\resizebox{\linewidth}{!}{
\begin{tabular}{lcc|cc|cc|cc}
\toprule
& \multicolumn{2}{c}{COCO} 
& \multicolumn{2}{c}{Pascal VOC}
& \multicolumn{2}{c}{KITTI}
& \multicolumn{2}{c}{Cityscapes} \\
\cmidrule(lr){2-3} \cmidrule(lr){4-5} \cmidrule(lr){6-7} \cmidrule(lr){8-9}
{Model} & Orig. & Val. & Orig. & Val. & Orig. & Val. & Orig. & Val. \\
\midrule
Grounding DINO & \percent{0.7053} & \percent{0.6373} & \percent{0.8824} & \percent{0.8255} & \percent{0.5067} & \underline{\percent{0.6168}} & \percent{0.5164} & \underline{\percent{0.6312}} \\
YOLOv8-World v2 & \percent{0.5847} & \percent{0.5664} & \percent{0.8190} & \percent{0.7661} & \percent{0.3171} & \percent{0.3768} & \percent{0.3055} & \percent{0.3870} \\
OWL-ViT & \percent{0.4430} & \percent{0.4344} & \percent{0.7427} & \percent{0.6939} & \percent{0.2226} & \percent{0.2196} & \percent{0.1735} & \percent{0.2122} \\
RT-DETR-L & \percent{0.6359} & \percent{0.5764} & \percent{0.8234} & \percent{0.7603} & \percent{0.8216} & \textbf{\percent{0.3691}} & \percent{0.4230} & \percent{0.5007} \\
YOLOv8-L & \percent{0.6243} & \percent{0.5799} & \percent{0.8269} & \percent{0.7640} & \percent{0.7834} & \textbf{\percent{0.3538}} & \percent{0.4369} & \percent{0.5151} \\
Faster R-CNN (ResNet 50) & \percent{0.6063} & \percent{0.5484} & \percent{0.7287} & \percent{0.6747} & \percent{0.7191} & \textbf{\percent{0.4040}} & \percent{0.4489} & \percent{0.5136} \\
\bottomrule
\end{tabular}
}
\end{subtable}
\end{table}

%% file: labelerrors/main/plot_main_label_errors.tex
\begin{figure}[htbp]
    \centering

    \begin{subfigure}[t]{0.48\textwidth}
        \centering
        \includegraphics[width=0.48\linewidth,trim={0 3.5cm 0 3cm}, clip]{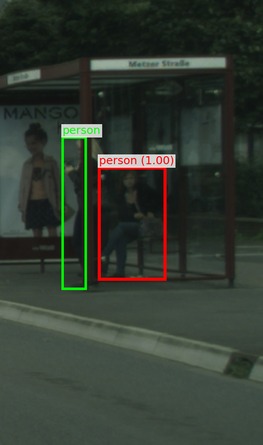}
        \hfill
        \includegraphics[width=0.48\linewidth]{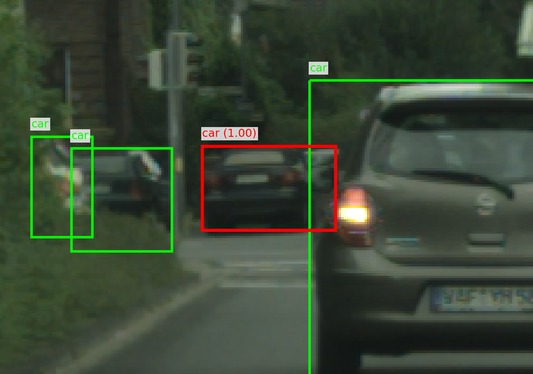}
        \caption{Cityscapes}
        \label{fig:main_label_errors_cityscapes}
    \end{subfigure}
    \hfill
    \begin{subfigure}[t]{0.48\textwidth}
        \centering
        \includegraphics[width=0.48\linewidth,trim={0 3cm 0 3cm}, clip]{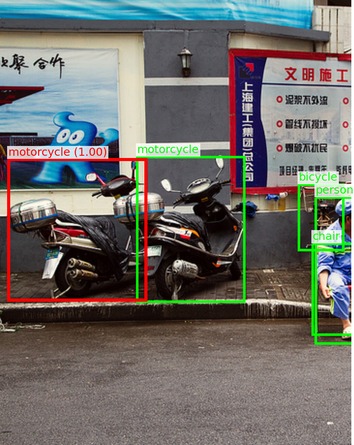}
        \hfill
        \includegraphics[width=0.48\linewidth,trim={0 3cm 0 3cm}, clip]{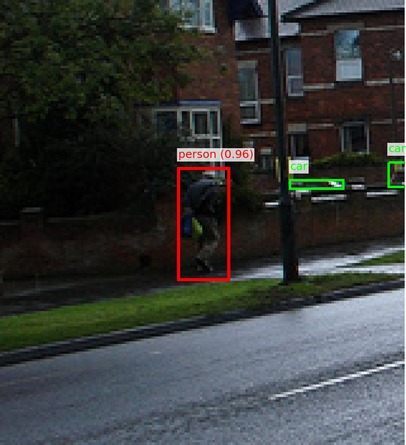}
        \caption{COCO}
        \label{fig:main_label_errors_coco}
    \end{subfigure}

    \vspace{0.5em}

    \begin{subfigure}[t]{0.48\textwidth}
        \centering
        \includegraphics[width=0.48\linewidth,trim={0 3cm 0 3cm}, clip]{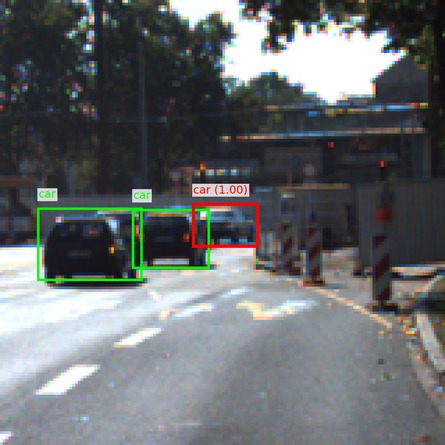}
        \hfill
        \includegraphics[width=0.48\linewidth]{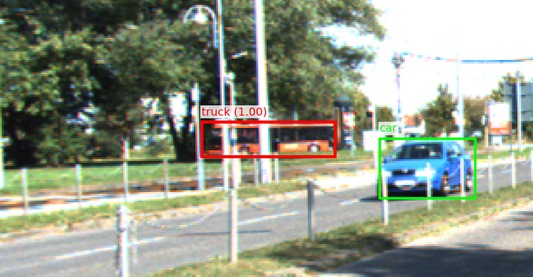}
        \caption{KITTI}
        \label{fig:main_label_errors_kitti}
    \end{subfigure}
    \hfill
    \begin{subfigure}[t]{0.48\textwidth}
        \centering
        \includegraphics[width=0.48\linewidth,trim={0 4cm 0 0}, clip]{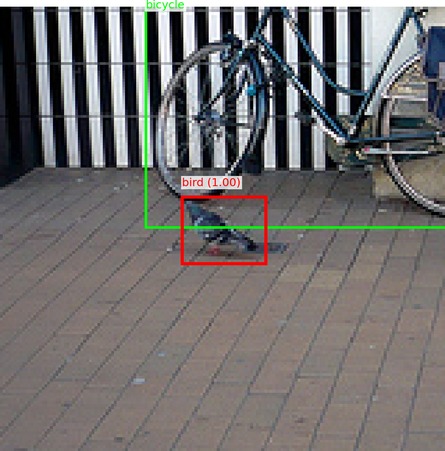}
        \hfill
        \includegraphics[width=0.48\linewidth,trim={0 0 0 3cm}, clip]{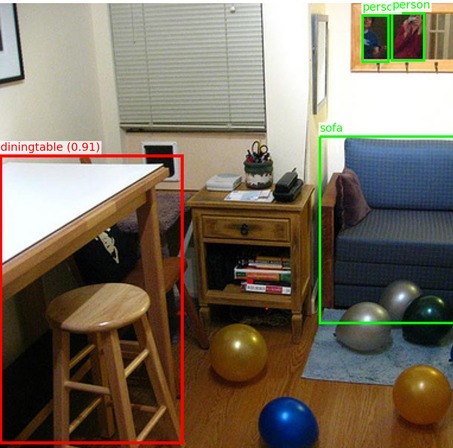}
        \caption{Pascal VOC}
        \label{fig:main_label_errors_voc}
    \end{subfigure}

    \caption{Examples of missing label errors across all datasets. Green boxes denote original annotations, red boxes highlight label errors with soft label probability $p$ found based on our validated GT.}
    \label{fig:main_label_errors_plot}
\end{figure}

%% file: tables/label_error_benchmark_table_main.tex
\begin{table*}[htbp]
\centering
\scalebox{1.0}{
\begin{minipage}{\textwidth}

\caption{
Label error detection performance across datasets.
Evaluation is performed on the less restrictive label error ground truth (soft label probability $\geq 0.5$). Methods use Grounding DINO predictions, except loss inspection, which uses Cascade R-CNN due to architectural constraints. We report AP for ranking quality and Precision, Recall, and F1 at the F1-maximizing threshold. The naive baseline has no ranking score and is reported as a single operating point without AP.
}

\label{tab:label_error_benchmark_main}

\begin{minipage}{0.49\textwidth}
\centering
\small
\begin{tabular}{lcccc}
\toprule
\textbf{Cityscapes} & \textbf{AP} & \textbf{Prec} & \textbf{Rec} & \textbf{F1}\\
\midrule
Naive Baseline & - & 1.50 & 99.61 & 2.96\\
Score Baseline & 23.44 & 28.07 & 43.77 & 34.21 \\
Loss Inspection & 15.53 & 18.89 & 41.47 & 25.96 \\
ObjectLab & 18.44 & 24.02 & 38.81 & 29.68 \\
MetaDetect & 18.70 & 24.14 & 33.54 & 28.07 \\
\bottomrule
\end{tabular}
\end{minipage}
\hfill
\begin{minipage}{0.49\textwidth}
\centering
\small
\begin{tabular}{lcccc}
\toprule
\textbf{COCO} & \textbf{AP} & \textbf{Prec} & \textbf{Rec} & \textbf{F1}\\
\midrule
Naive Baseline & - & 2.10 & 98.25 & 4.12 \\
Score Baseline & 38.21 & 37.92  & 51.09  & 43.53   \\
Loss Inspection & 10.42 & 13.85 & 31.12 & 19.17 \\
ObjectLab & 21.01 & 34.51 & 27.88 & 30.84  \\
MetaDetect & 26.37 & 29.51 & 32.82 & 31.08 \\
\bottomrule
\end{tabular}
\end{minipage}

\vspace{0.25cm}

\begin{minipage}{0.49\textwidth}
\centering
\small
\begin{tabular}{lcccc}
\toprule
\textbf{KITTI} & \textbf{AP} & \textbf{Prec} & \textbf{Rec} & \textbf{F1}\\
\midrule
Naive Baseline & - & 1.78 & 99.99 & 3.50 \\
Score Baseline & 59.65 & 56.60  & 62.25 & 59.29  \\
Loss Inspection & 6.72 & 21.28 & 24.81 & 22.91 \\
ObjectLab & 29.99 & 41.02 & 33.51 & 36.89 \\
MetaDetect & 19.87 & 51.23 & 15.93 & 24.30 \\
\bottomrule
\end{tabular}
\end{minipage}
\hfill
\begin{minipage}{0.49\textwidth}
\centering
\small
\begin{tabular}{lcccc}
\toprule
\textbf{Pascal VOC} & \textbf{AP} & \textbf{Prec} & \textbf{Rec} & \textbf{F1}\\
\midrule
Naive Baseline & - & 0.39 & 100.00 & 0.78 \\
Score Baseline & 36.68 & 37.08 & 54.30 & 44.06  \\
Loss Inspection & 3.27 & 7.19 & 14.29 & 9.56 \\
ObjectLab & 14.70 & 29.18 & 22.99 & 25.72  \\
MetaDetect & 21.20 & 43.00 & 31.11 & 36.10 \\
\bottomrule
\end{tabular}
\end{minipage}

\end{minipage}
}
\end{table*}

%% file: main.bbl

%% file: tables/microtask_parameters_table.tex
\begin{table}[htbp]
\centering
\caption{Overview of the microtask parameters used during the annotation process. Each task consists of a question and an additional description to guide annotators. The parameter \textit{repeats} defines how many independent annotators complete the same task, while the \textit{working package} (WP) specifies how many items (images, keypoints, or bounding boxes depending on the task) are processed per assignment. The time limit indicates the maximum duration available per working package. The placeholder \textit{class} is dynamically replaced during task execution by the respective dataset class currently under review.}
\label{tab:microtask_design}
\small
\setlength{\tabcolsep}{4pt}
\begin{tabular}{p{2cm} p{3.5cm} p{3.5cm} c c c}
\hline
Task & Question & Description & Repeats & WP & Time [s]/WP \\
\hline
Keypoint 
& Mark all instances of \textit{class} that are not already marked with a keypoint. Do not double annotate objects. 
& If you are not sure, just annotate it. We will check this later. 
& 3 & 10 & 1000 \\
\hline
Bounding Box 
& Draw a tight bounding box around the marked object. 
& Please draw the bounding box as tight as possible. You can also correct it. 
& 2 & 20 & 1800 \\
\hline
Soft Label 
& Is the object in the box a \textit{class}? If so, click \textit{class}. If not, select the correct label. 
& If you are not sure, press ``can't solve''. 
& 11 & 30 & 600 \\
\hline
\end{tabular}
\end{table}

%% file: tables/scalability_analysis.tex
\begin{table*}[htbp]
    \centering
    \caption{
        Average annotation times for the individual annotation microtasks. Keypoint time denotes the average time for a single image-class review by one annotator, and bounding box time denotes the average annotation/review time for a single keypoint by one annotator. Per-answer time denotes the average time for an individual soft label response, while the times for 11, 22, 33, and 44 votes denote the average cumulative annotation time required to obtain the corresponding soft label. All times are reported in seconds.
    }
    \label{tab:annotation_times}
    \resizebox{\textwidth}{!}{%
    \begin{tabular}{lccc|cccc}
        \hline
        Dataset
        & Keypoint Time [s]
        & Bounding Box Time [s]
        & Per Answer Time [s]
        & 11 Votes [s]
        & 22 Votes [s]
        & 33 Votes [s]
        & 44 Votes [s] \\
        \hline
        Cityscapes
        & 2.74 & 11.59 & 1.73
        & 17.90 & 51.43 & 72.06 & 110.95 \\

        Pascal VOC
        & 1.29 & 6.29 & 1.84
        & 16.50 & 49.18 & 70.07 & 101.14 \\

        KITTI
        & 3.47 & 7.34 & 1.99
        & 18.46 & 46.76 & 72.11 & 100.74 \\

        COCO
        & 2.50 & 6.94 & 2.28
        & 21.11 & 53.58 & 79.19 & 114.04 \\
        \hline
    \end{tabular}%
    }
\end{table*}

%% file: tables/gt_vs_vgt_table_pascalvoc.tex
\begin{table}[htbp]
\centering
\caption{Class-wise comparison of annotation counts between the original and validated Pascal VOC ground truth for the training and validation splits.}
\label{tab:gt_vs_vgt_table_pascalvoc}
\begin{tabular}{l|ccc|ccc}
\hline
\multirow{2}{*}{Class} & \multicolumn{3}{c|}{Train (5717 images)} & \multicolumn{3}{c}{Val (5823 images)} \\
 & Original GT & Validated GT & $\Delta$[\%] & Original GT & Validated GT & $\Delta$[\%] \\
\hline
aeroplane     & 470  & 506  & +7.7  & 484  & 510  & +5.4  \\
bicycle       & 410  & 436  & +6.3  & 380  & 416  & +9.5  \\
bird          & 592  & 673  & +13.7 & 629  & 642  & +2.1  \\
boat          & 508  & 602  & +18.5 & 491  & 575  & +17.1 \\
bottle        & 749  & 805  & +7.5  & 733  & 806  & +10.0 \\
bus           & 317  & 313  & -1.3  & 320  & 324  & +1.3  \\
car           & 1,191 & 1,281 & +7.6  & 1,173 & 1,308 & +11.5 \\
cat           & 609  & 601  & -1.3  & 618  & 615  & -0.5  \\
chair         & 1,457 & 1,654 & +13.5 & 1,449 & 1,679 & +15.9 \\
cow           & 355  & 323  & -9.0  & 347  & 331  & -4.6  \\
dining table   & 373  & 417  & +11.8 & 374  & 450  & +20.3 \\
dog           & 768  & 756  & -1.6  & 773  & 766  & -0.9  \\
horse         & 377  & 384  & +1.9  & 373  & 369  & -1.1  \\
motorbike     & 375  & 397  & +5.9  & 376  & 398  & +5.9  \\
person        & 5,019 & 5,660 & +12.8 & 5,110 & 5,813 & +13.8 \\
potted plant   & 557  & 600  & +7.7  & 542  & 606  & +11.8 \\
sheep         & 509  & 464  & -8.8  & 485  & 465  & -4.1  \\
sofa          & 399  & 410  & +2.8  & 387  & 419  & +8.3  \\
train         & 327  & 347  & +6.1  & 329  & 343  & +4.3  \\
tv monitor     & 412  & 400  & -2.9  & 414  & 398  & -3.9  \\
\hline
\textbf{Total} & 15,774 & 17,029 & +8.0 & 15,787 & 17,233 & +9.2 \\
\hline
\end{tabular}
\end{table}

%% file: tables/gt_vs_vgt_table_coco.tex
\begin{table}[htpb]
\centering

\begin{minipage}{0.48\textwidth}
\centering
\captionof{table}{Comparison of annotation counts between the original and validated COCO ground truth for the validation split across the \textit{person} and \textit{animal} categories.}
\label{tab:coco_person_animal}
\scriptsize
\setlength{\tabcolsep}{4pt}
\begin{tabular}{l|cc|c}
\hline
Class & Original GT & Validated GT & $\Delta[\%]$ \\
\hline
\multicolumn{4}{c}{\textit{Person}} \\
person & 11,004 & 16,667 & +51.5 \\
\hline
\multicolumn{4}{c}{\textit{Animal}} \\
bird & 440 & 984 & +123.6 \\
cat & 202 & 200 & -1.0 \\
dog & 218 & 208 & -4.6 \\
horse & 273 & 290 & +6.2 \\
sheep & 361 & 625 & +73.1 \\
cow & 380 & 346 & -8.9 \\
elephant & 255 & 305 & +19.6 \\
bear & 71 & 72 & +1.4 \\
zebra & 268 & 300 & +11.9 \\
giraffe & 232 & 241 & +3.9 \\
\hline
\textbf{Total} & 13,704 & 20,238 & +47.7 \\
\hline
\end{tabular}
\end{minipage}
\hfill
\begin{minipage}{0.48\textwidth}
\centering
\captionof{table}{Comparison of annotation counts between the original and validated COCO ground truth for the validation split across the \textit{vehicle} and \textit{sports} categories.}
\label{tab:coco_vehicle_sports}
\scriptsize
\setlength{\tabcolsep}{4pt}
\begin{tabular}{l|cc|c}
\hline
Class & Original GT & Validated GT & $\Delta[\%]$ \\
\hline
\multicolumn{4}{c}{\textit{Vehicle}} \\
bicycle & 316 & 381 & +20.6 \\
car & 1,932 & 2,404 & +24.4 \\
motorcycle & 371 & 501 & +35.0 \\
airplane & 143 & 151 & +5.6 \\
bus & 285 & 294 & +3.2 \\
train & 190 & 203 & +6.8 \\
truck & 415 & 330 & -20.5 \\
boat & 430 & 486 & +13.0 \\
\hline
\multicolumn{4}{c}{\textit{Sports}} \\
frisbee & 115 & 117 & +1.7 \\
skis & 241 & 256 & +6.2 \\
snowboard & 69 & 86 & +24.6 \\
sports ball & 263 & 289 & +9.9 \\
kite & 336 & 430 & +28.0 \\
baseball bat & 146 & 182 & +24.7 \\
baseball glove & 148 & 142 & -4.1 \\
skateboard & 179 & 183 & +2.2 \\
surfboard & 269 & 257 & -4.5 \\
tennis racket & 225 & 220 & -2.2 \\
\hline
\textbf{Total} & 6,073 & 6,912 & +13.8 \\
\hline
\end{tabular}
\end{minipage}

\end{table}

\begin{table}[htbp]
\centering

\begin{minipage}{0.48\textwidth}
\centering
\captionof{table}{Comparison of annotation counts between the original and validated COCO ground truth for the validation split across the \textit{outdoor}, \textit{accessory} and \textit{electronic} catgories.}
\label{tab:coco_outdoor_accessory_electronic}
\scriptsize
\setlength{\tabcolsep}{4pt}
\begin{tabular}{l|cc|c}
\hline
Class & Original GT & Validated GT & $\Delta[\%]$ \\
\hline
\multicolumn{4}{c}{\textit{Outdoor}} \\
traffic light & 637 & 831 & +30.5 \\
fire hydrant & 101 & 99 & -2.0 \\
stop sign & 75 & 52 & -30.7 \\
parking meter & 60 & 52 & -13.3 \\
bench & 413 & 321 & -22.3 \\
\hline
\multicolumn{4}{c}{\textit{Accessory}} \\
backpack & 371 & 364 & -1.9 \\
umbrella & 413 & 531 & +28.6 \\
handbag & 540 & 630 & +16.7 \\
tie & 254 & 322 & +26.8 \\
suitcase & 303 & 341 & +12.5 \\
\hline
\multicolumn{4}{c}{\textit{Electronic}} \\
tv & 288 & 172 & -40.3 \\
laptop & 231 & 232 & +0.4 \\
mouse & 106 & 107 & +0.9 \\
remote & 283 & 315 & +11.3 \\
keyboard & 153 & 108 & -29.4 \\
cell phone & 262 & 263 & +0.4 \\
\hline
\textbf{Total} & 4,490 & 4,740 & +5.6 \\
\hline
\end{tabular}
\end{minipage}
\hfill
\begin{minipage}{0.48\textwidth}
\centering
\captionof{table}{Comparison of annotation counts between the original and validated COCO ground truth for the validation split across the \textit{furniture} and \textit{indoor} categories.}
\label{tab:coco_furniture_indoor}
\scriptsize
\setlength{\tabcolsep}{4pt}
\begin{tabular}{l|cc|c}
\hline
Class & Original GT & Validated GT & $\Delta[\%]$ \\
\hline
\multicolumn{4}{c}{\textit{Furniture}} \\
chair & 1,791 & 2,892 & +61.5 \\
couch & 261 & 207 & -20.7 \\
potted plant & 343 & 299 & -12.8 \\
bed & 163 & 142 & -12.9 \\
dining table & 697 & 789 & +13.2 \\
toilet & 179 & 163 & -8.9 \\
\hline
\multicolumn{4}{c}{\textit{Indoor}} \\
book & 1,161 & 3,218 & +177.2 \\
clock & 267 & 382 & +43.1 \\
vase & 277 & 374 & +35.0 \\
scissors & 36 & 37 & +2.8 \\
teddy bear & 191 & 301 & +57.6 \\
hair drier & 11 & 7 & -36.4 \\
toothbrush & 57 & 51 & -10.5 \\
\hline
\textbf{Total} & 5,434 & 8,862 & +63.1 \\
\hline
\end{tabular}
\end{minipage}

\end{table}

\begin{table}[htbp]
\centering
\captionof{table}{Comparison of annotation counts between the original and validated COCO ground truth for the validation split across \textit{kitchen}, \textit{food} and \textit{appliance} categories.}
\label{tab:gt_vs_vgt_coco_kitchen_food_appliance}
\scriptsize
\setlength{\tabcolsep}{4pt}
\begin{tabular}{l|cc|c}
\hline
Class & Original GT & Validated GT & $\Delta[\%]$ \\
\hline
\multicolumn{4}{c}{\textit{Kitchen}} \\
bottle & 1,025 & 1,123 & +9.6 \\
wine glass & 343 & 384 & +12.0 \\
cup & 899 & 736 & -18.1 \\
fork & 215 & 205 & -4.7 \\
knife & 326 & 239 & -26.7 \\
spoon & 253 & 214 & -15.4 \\
bowl & 626 & 459 & -26.7 \\
\hline
\multicolumn{4}{c}{\textit{Appliance}} \\
microwave & 55 & 52 & -5.5 \\
oven & 143 & 132 & -7.7 \\
toaster & 9 & 13 & +44.4 \\
sink & 225 & 226 & -0.4 \\
refrigerator & 126 & 107 & -15.1 \\
\hline
\multicolumn{4}{c}{\textit{Food}} \\
banana & 379 & 702 & +85.2 \\
apple & 239 & 1,186 & +396.2 \\
sandwich & 177 & 157 & -11.3 \\
orange & 287 & 436 & +51.9 \\
broccoli & 316 & 695 & +119.9 \\
carrot & 371 & 2,039 & +449.6 \\
hot dog & 127 & 54 & -57.5 \\
pizza & 285 & 333 & +16.8 \\
donut & 338 & 746 & +120.7 \\
cake & 316 & 497 & +57.3 \\
\hline
\textbf{Total} & 7,080 & 10,735 & +51.6 \\
\hline
\end{tabular}
\end{table}

%% file: tables/gt_vs_vgt_table_cityscapes.tex
\begin{table}[htbp]
\centering
\caption{Class-wise comparison of annotation counts between the original and validated Cityscapes ground truth for the training and validation splits.}
\label{tab:gt_vs_vgt_table_cityscapes}
\begin{tabular}{l|ccc|ccc}
\hline
\multirow{2}{*}{Class} & \multicolumn{3}{c|}{Train (2,975 images)} & \multicolumn{3}{c}{Val (500 images)} \\
 & Original GT & Validated GT & $\Delta[\%]$ & Original GT & Validated GT & $\Delta[\%]$ \\
\hline
person      & 17,994 & 18,104 & +0.6  & 3,419 & 3,559 & +4.1  \\
rider       & 1,807  & 1,860  & +2.9  & 556  & 527  & -5.2  \\
car         & 27,155 & 28,526 & +5.0  & 4,667 & 4,074 & -12.7 \\
truck       & 489   & 539   & +10.2 & 93   & 84   & -9.7  \\
bus         & 385   & 380   & -1.3  & 98   & 97   & -1.0  \\
train       & 171   & 179   & +4.7  & 23   & 21   & -8.7  \\
motorcycle  & 739   & 705   & -4.6  & 149  & 149  & 0.0   \\
bicycle     & 3,729  & 5,180  & +38.9 & 1,175 & 1,872 & +59.4 \\
\hline
\textbf{Total} & 52,469 & 55,473 & +5.7 & 10,180 & 10,383 & +2.0 \\
\hline
\end{tabular}
\end{table}

%% file: tables/gt_vs_vgt_table_kitti.tex
\begin{table}[htbp]
\centering
\caption{Class-wise comparison of annotation counts between the original and validated KITTI ground truth for the training and validation splits.}
\label{tab:gt_vs_vgt_table_kitti}
\begin{tabular}{l|ccc|ccc}
\hline
\multirow{2}{*}{Class} & \multicolumn{3}{c|}{Train (5,984 images)} & \multicolumn{3}{c}{Val (1,497 images)} \\
 & Original GT & Validated GT & $\Delta[\%]$ & Original GT & Validated GT & $\Delta[\%]$ \\
\hline
car              & 22,923 & 36,627 & +59.8 & 5,819 & 10,357 & +78.0 \\
van              & 2,323  & 2,196  & -5.5  & 591  & 733   & +24.0 \\
truck            & 857    & 1,699  & +98.3 & 237  & 433   & +82.7 \\
pedestrian       & 3,591  & 5,912  & +64.6 & 896  & 1,469 & +64.0 \\
person sitting  & 182    & 801    & +340.1 & 40   & 229   & +472.5 \\
cyclist          & 1,297  & 1,859  & +43.3 & 330  & 453   & +37.3 \\
tram             & 416    & 457    & +9.9 & 95   & 123   & +29.5 \\
\hline
\textbf{Total}   & 31,589 & 49,551 & +56.9 & 8,008 & 13,797 & +72.3 \\
\hline
\end{tabular}
\end{table}

%% file: tables/false_negative_full.tex
\begin{table}[htbp]
\centering
\caption{False negative analysis for the original and validated ground truth.
The table reports the number of reviewed images and objects, the corresponding
false negative rates (FNRs), and the relative FNR reduction achieved by our
validated GT. For Cityscapes, we evaluate the complete validation set of
500 images, since only the validation set was fully re-annotated.}
\label{tab:fn_full}
\small
\resizebox{0.8\linewidth}{!}{
\begin{tabular}{l cc cc c}
\toprule
& \multicolumn{2}{c}{Review} 
& \multicolumn{2}{c}{FNR [\%]}
& \multirow{2}{*}{Relative Reduction [\%]} \\
\cmidrule(lr){2-3}
\cmidrule(lr){4-5}
Dataset
& Images
& Objects
& Original GT
& Validated GT (Ours)
& \\
\midrule
Pascal VOC  & 600 & 2,072  & 18.6 & 9.3  & 50.0 \\
COCO        & 600 & 4,666  & 13.1 & 6.1  & 53.2 \\
KITTI       & 600 & 5,908  & 41.5 & 11.6 & 72.0 \\
Cityscapes  & 500 & 11,594 & 11.1 & 3.8  & 65.8 \\
\bottomrule
\end{tabular}
}
\end{table}

%% file: imgs/collages-false-negative.tex
\begin{figure}[htbp]
\centering

\begin{subfigure}{0.24\linewidth}
    \centering
    \includegraphics[width=\linewidth]{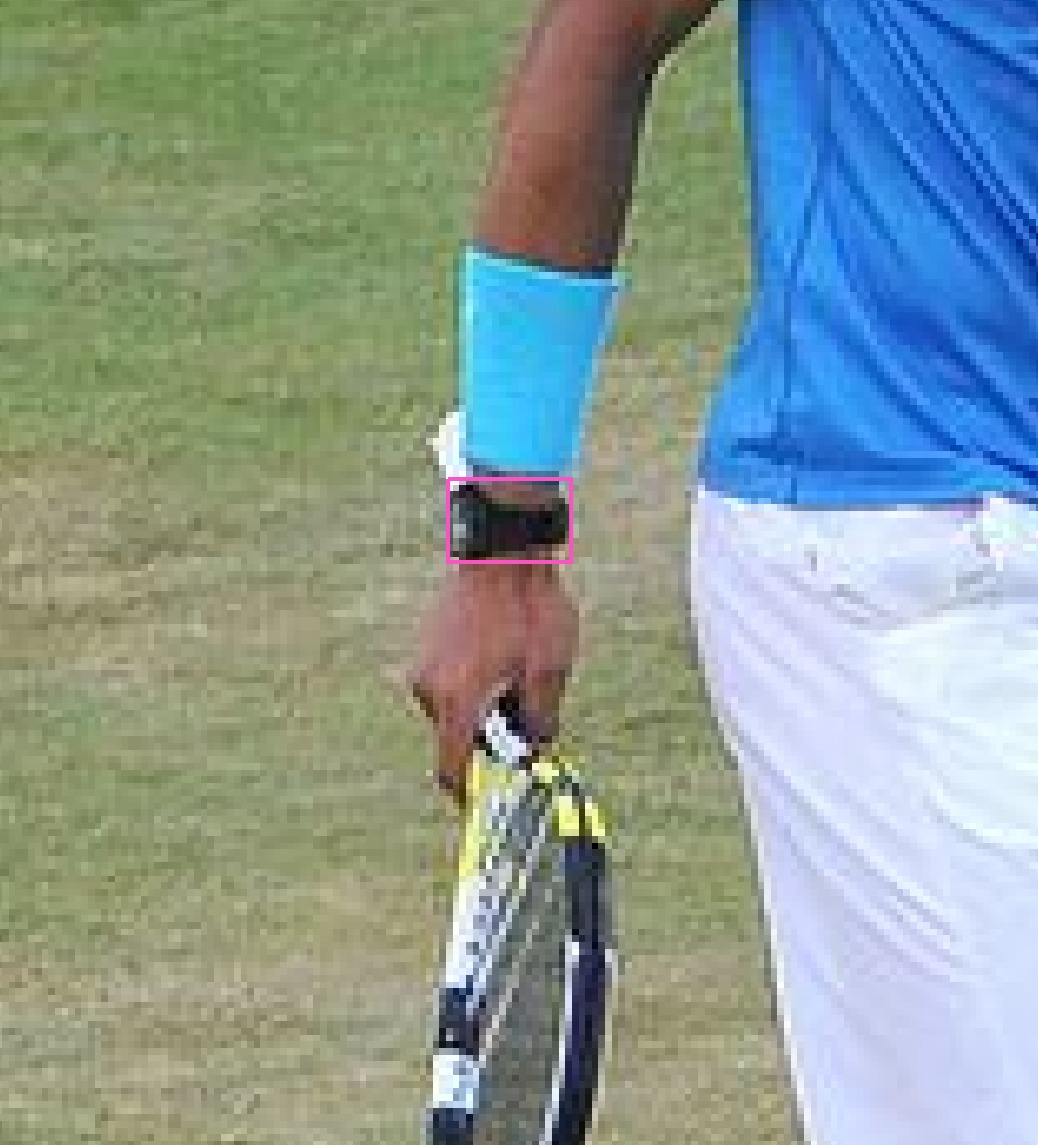}
     \caption{Clock}
\end{subfigure}
\hfill
\begin{subfigure}{0.24\linewidth}
    \centering
    \includegraphics[width=\linewidth]{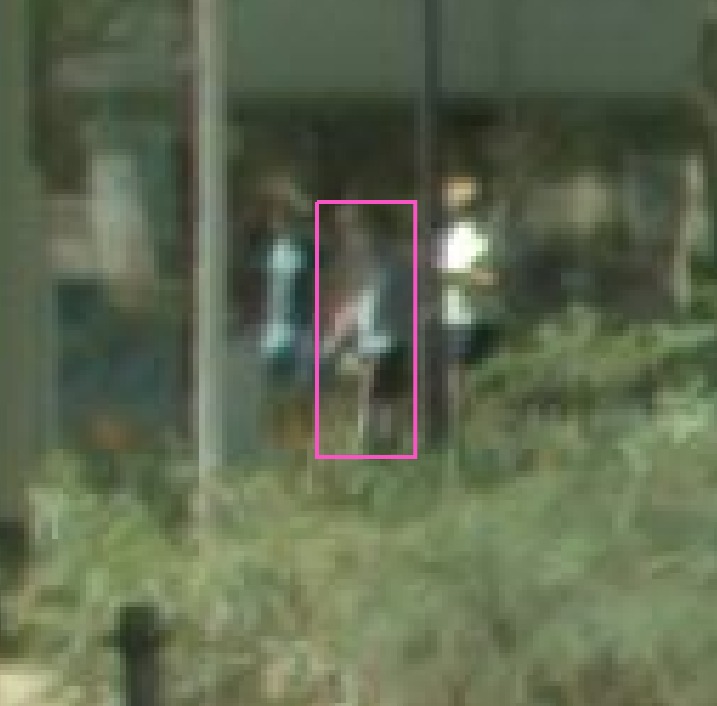}
     \caption{Person}
\end{subfigure}
\hfill
\begin{subfigure}{0.24\linewidth}
    \centering
    \includegraphics[width=\linewidth]{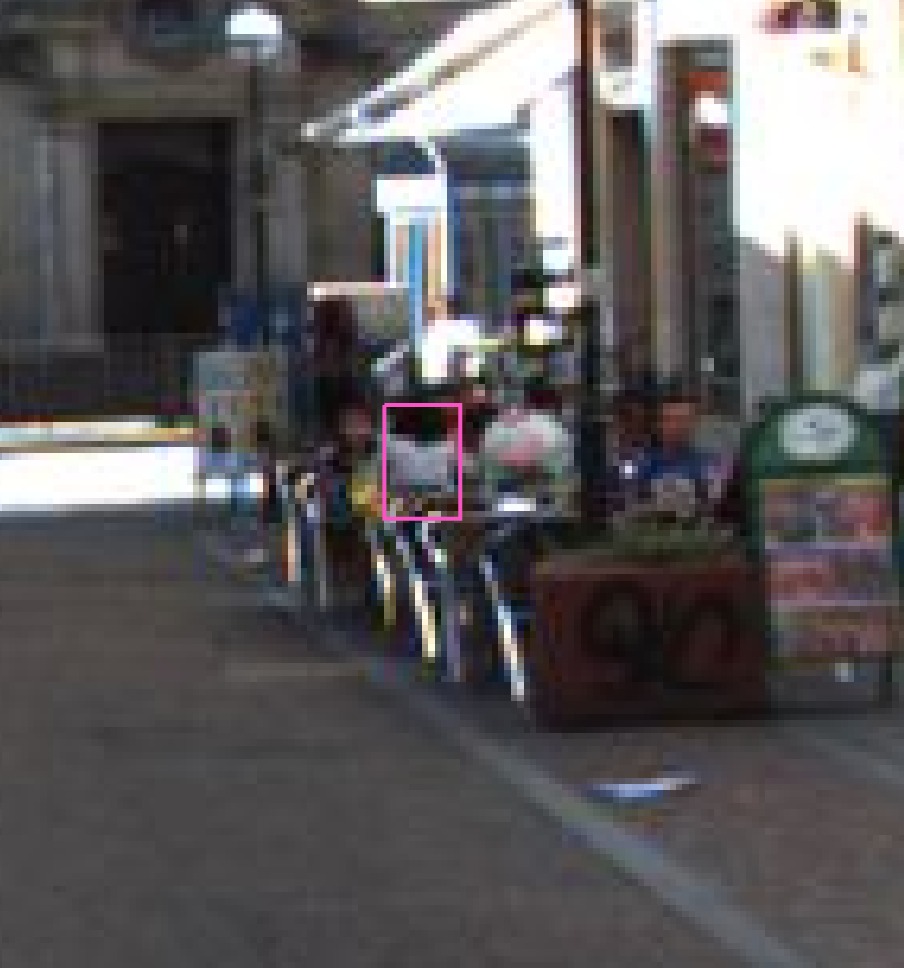}
     \caption{Person (Sitting)}
\end{subfigure}
\hfill
\begin{subfigure}{0.24\linewidth}
    \centering
    \includegraphics[width=\linewidth]{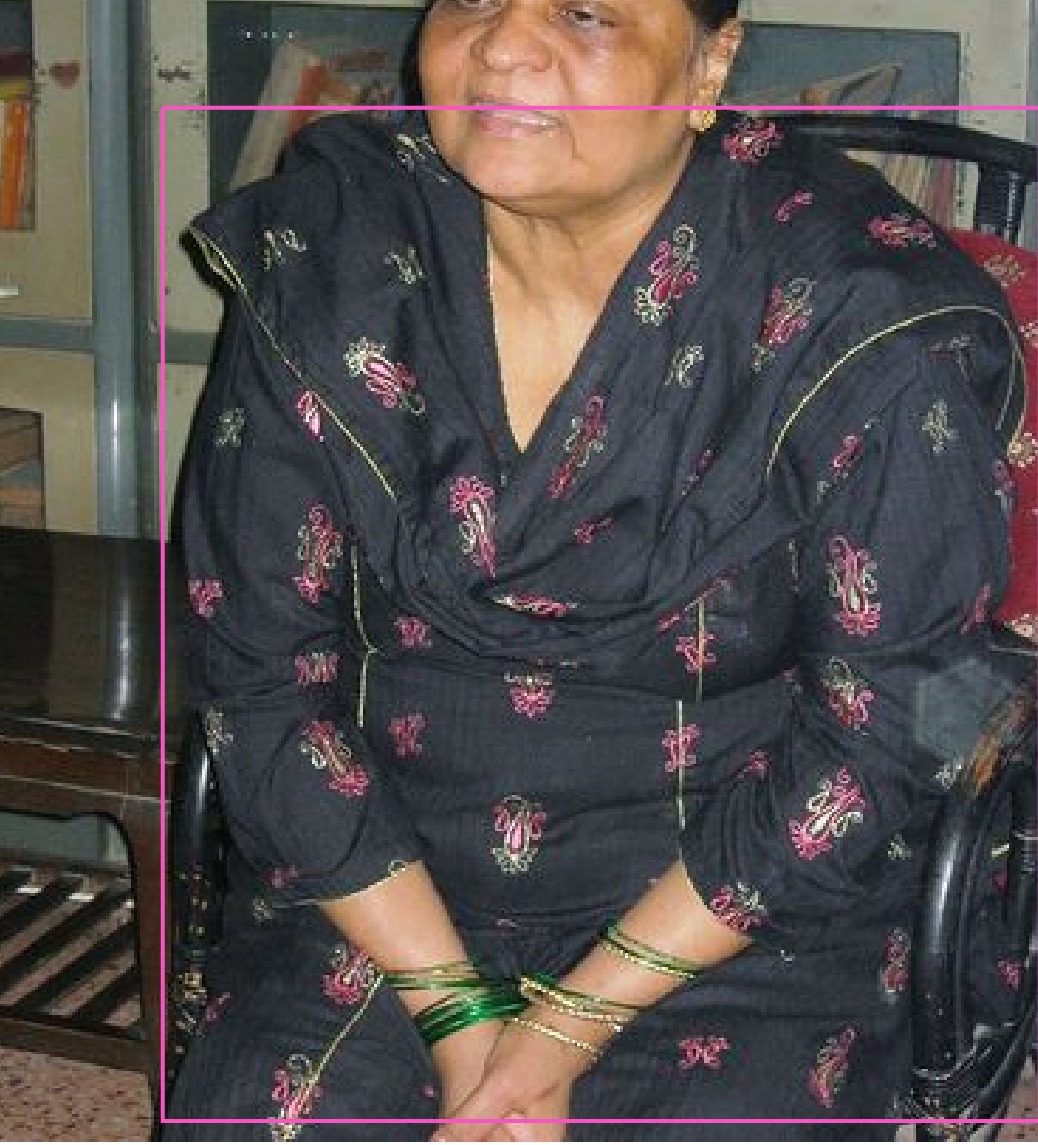}
     \caption{Chair}
\end{subfigure}

\vspace{0.3em}

\begin{subfigure}{0.24\linewidth}
    \centering
    \includegraphics[width=\linewidth]{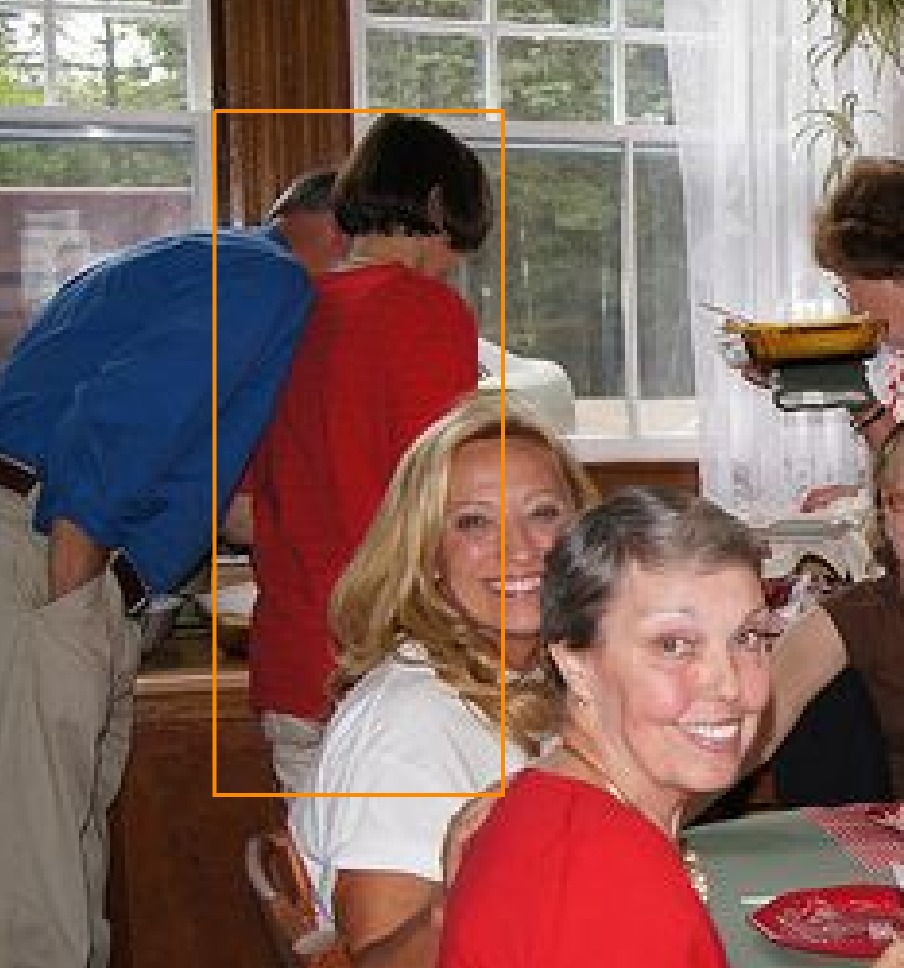}
     \caption{Person}
\end{subfigure}
\hfill
\begin{subfigure}{0.24\linewidth}
    \centering
    \includegraphics[width=\linewidth]{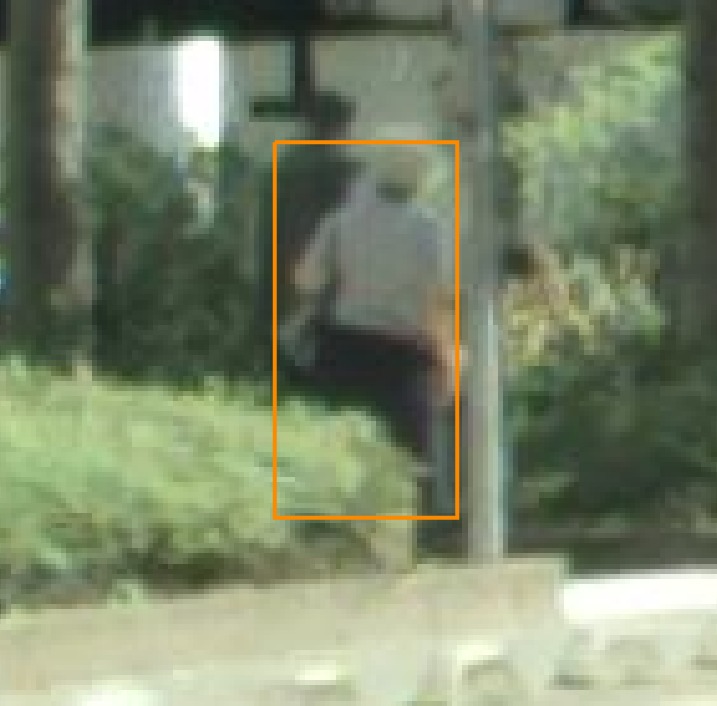}
    \caption{Person}
\end{subfigure}
\hfill
\begin{subfigure}{0.24\linewidth}
    \centering
    \includegraphics[width=\linewidth]{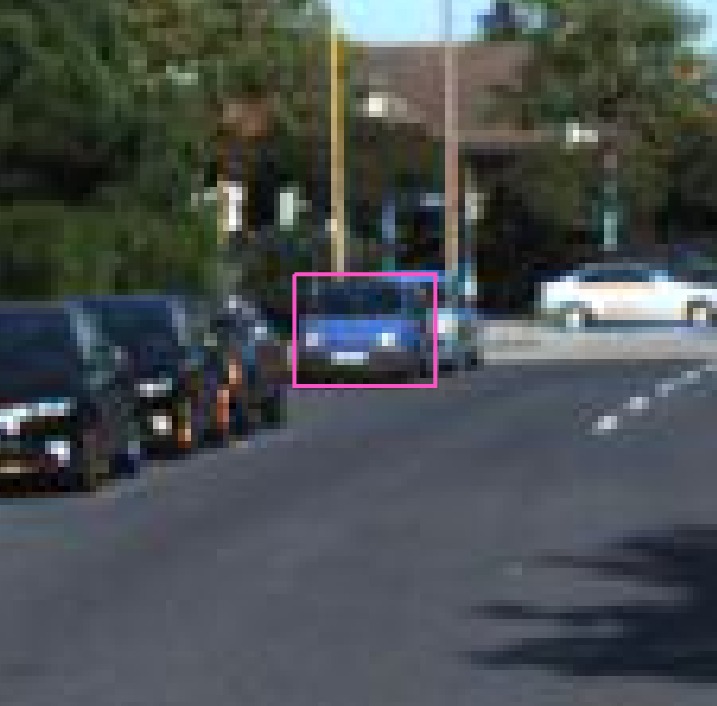}
    \caption{Car}
\end{subfigure}
\hfill
\begin{subfigure}{0.24\linewidth}
    \centering
    \includegraphics[width=\linewidth]{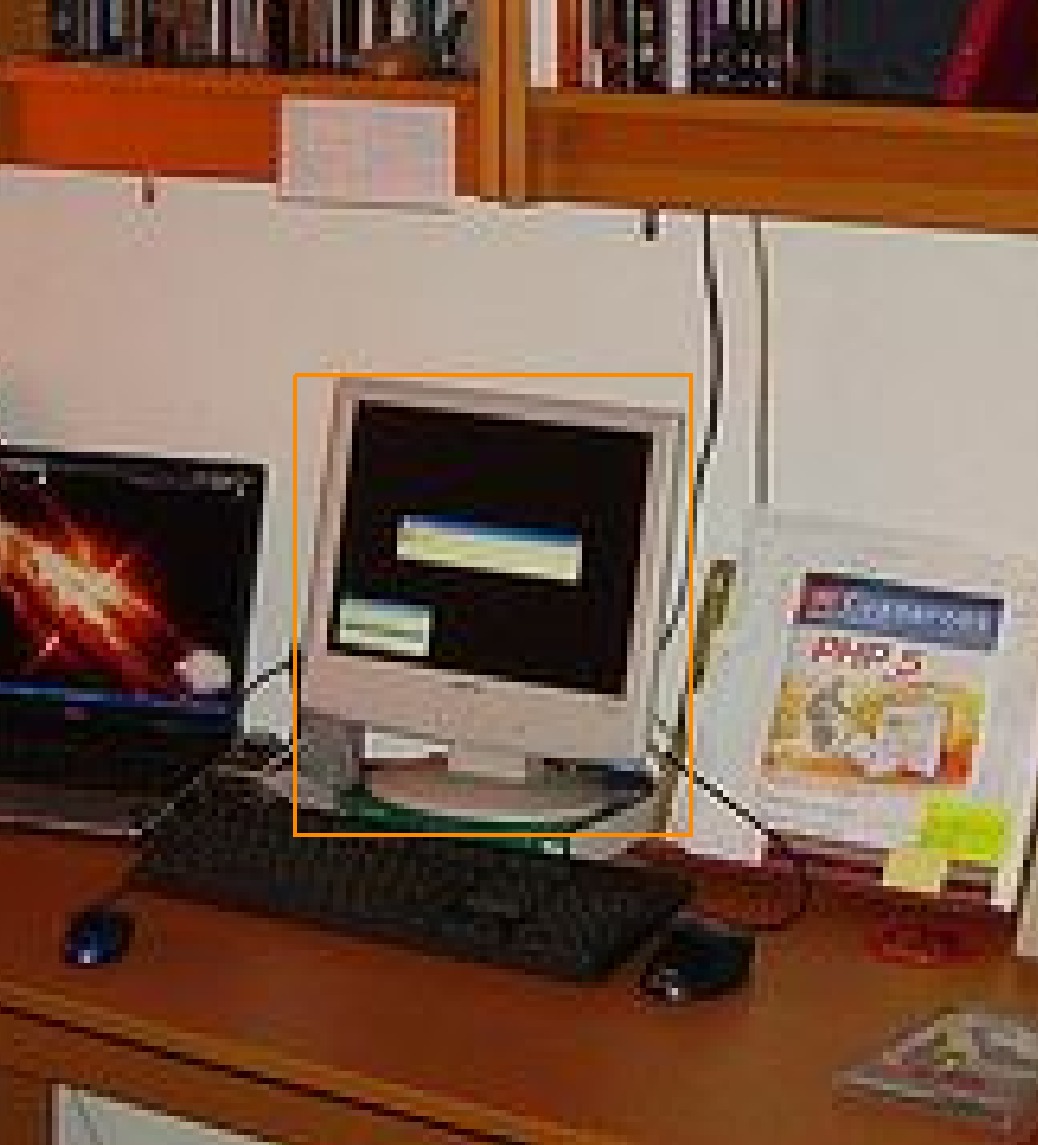}
    \caption{TV}
\end{subfigure}

\caption{Comparison of missing bounding boxes, Rows: Original GT, Validated GT (ours), Columns: COCO, Cityscapes, KITTI, Pascal VOC
For KITTI, we found no missing objects thus we are showing a second error of the original GT.
We see that all provided examples are clear errors which are missing from the validated or original ground truth, while for the original ground truth way more objects are missing in total.
}
\label{fig:missing_objects_examples}
\end{figure}

%% file: tables/further_analysis_vgt_table.tex
\begin{table}[htbp]
\centering
\small
\caption{Distribution of the number of annotations per bounding box and proportion of ambiguous instances where multiple classes share the highest probability.}
\label{tab:annotation_and_ambiguity}
\begin{tabular}{lccccc|cc}
\toprule
\textbf{Dataset} & \textbf{\#Boxes} & \textbf{11} [\%] & \textbf{22} [\%] & \textbf{33} [\%] & \textbf{44} [\%] & \textbf{\#Ambiguous} & \textbf{[\%]} \\
\midrule
Cityscapes & 69,478 & 92.26 & 7.33 & 0.16 & 0.25 & 42  & 0.06 \\
KITTI      & 65,711 & 50.12 & 47.03 & 1.09 & 1.75 & 30  & 0.05 \\
Pascal VOC & 36,134 & 71.04 & 27.69 & 0.48 & 0.79 & 30  & 0.08 \\
COCO       & 56,067 & 49.82 & 46.32 & 1.25 & 2.61 & 350 & 0.62 \\
\bottomrule
\end{tabular}
\end{table}

%% file: ambiguous_cases/plot_ambiguous_cases.tex
\begin{figure*}[htbp]
\centering

\begin{subfigure}[t]{\textwidth}
\centering
\includegraphics[width=0.19\linewidth]{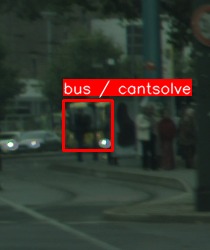}
\includegraphics[width=0.19\linewidth]{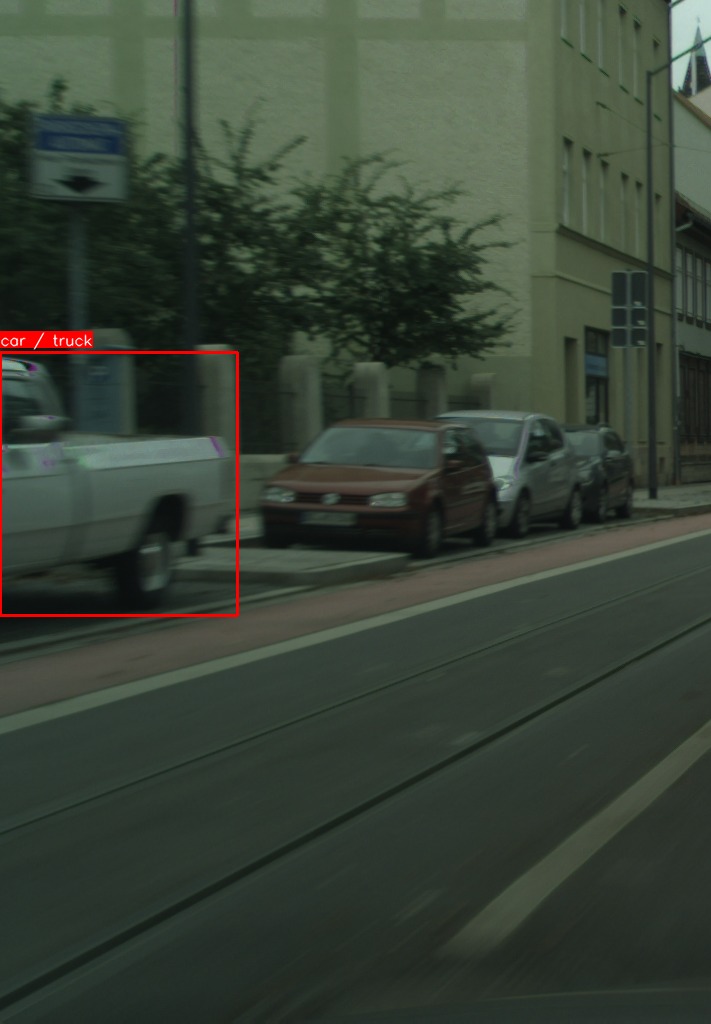}
\includegraphics[width=0.19\linewidth]{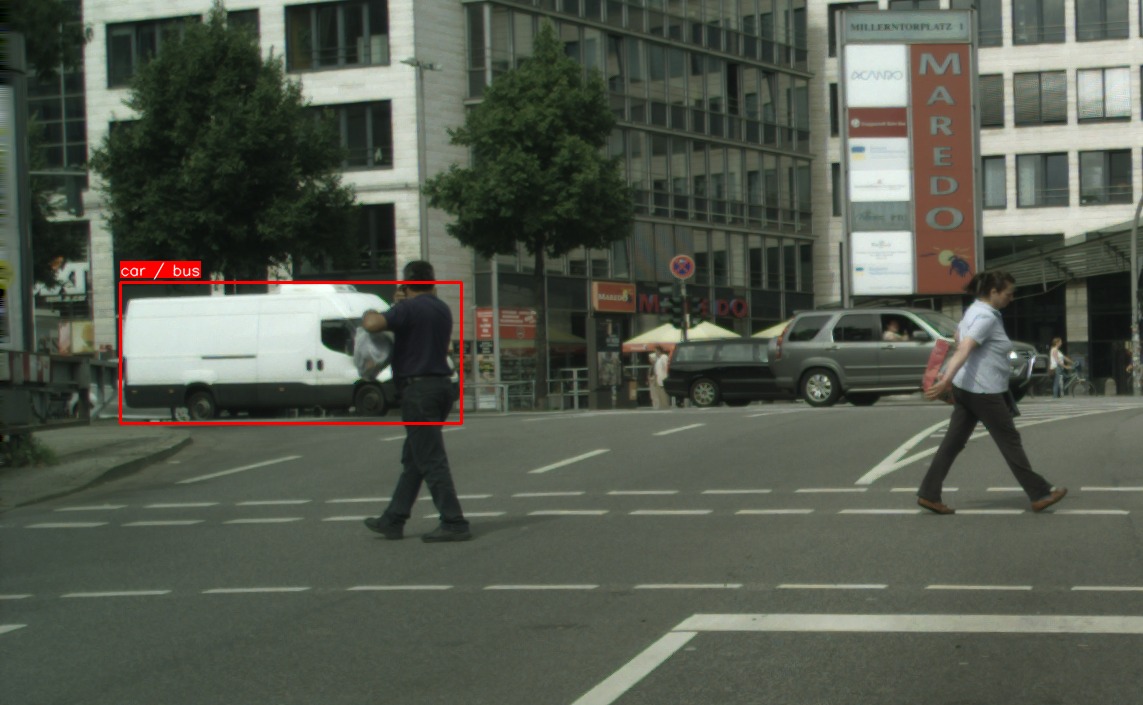}
\includegraphics[width=0.19\linewidth]{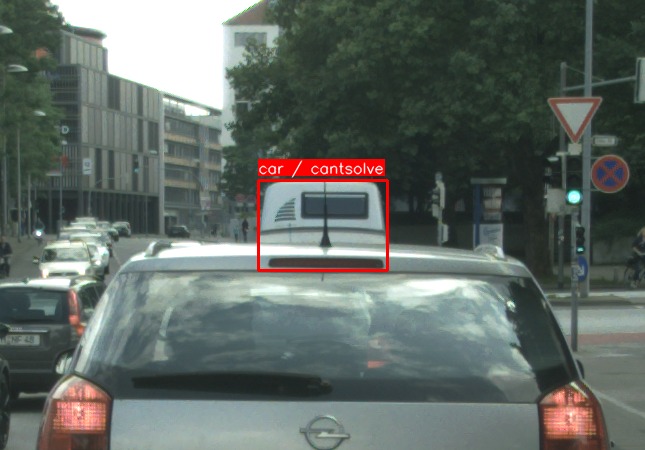}
\includegraphics[width=0.19\linewidth]{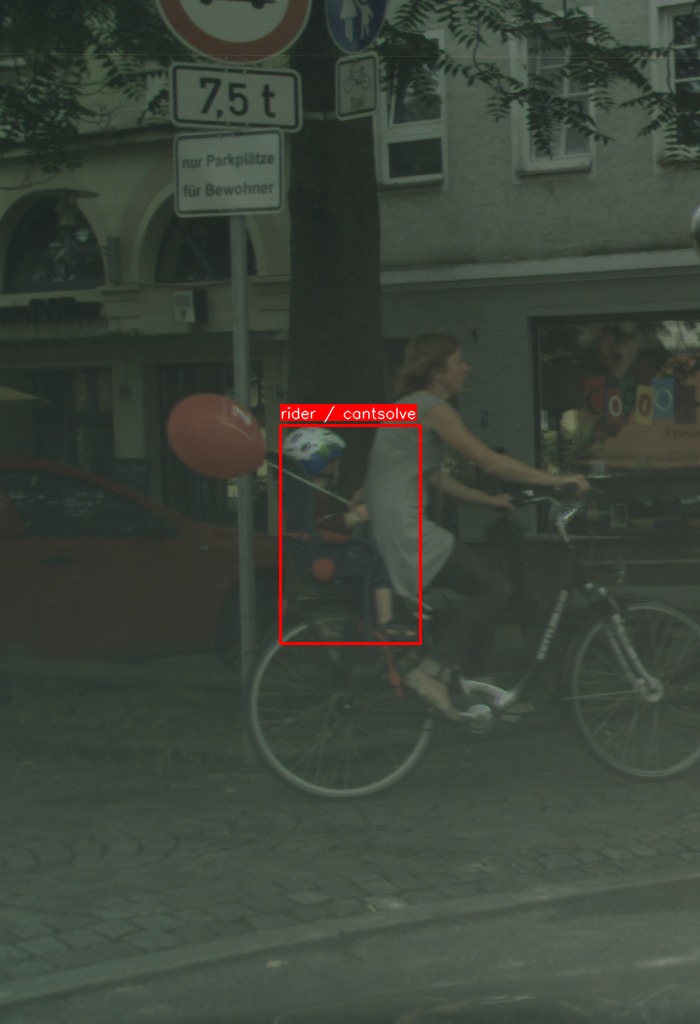}
\caption{Cityscapes}
\end{subfigure}

\vspace{0.4em}

\begin{subfigure}[t]{\textwidth}
\centering
\includegraphics[width=0.19\linewidth]{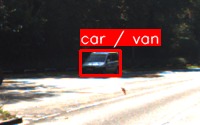}
\includegraphics[width=0.19\linewidth]{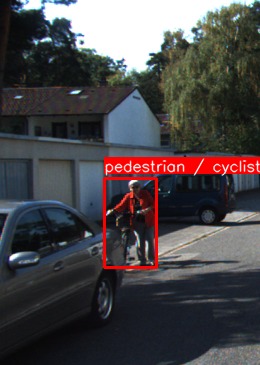}
\includegraphics[width=0.19\linewidth]{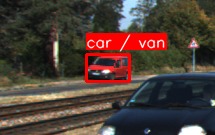}
\includegraphics[width=0.19\linewidth]{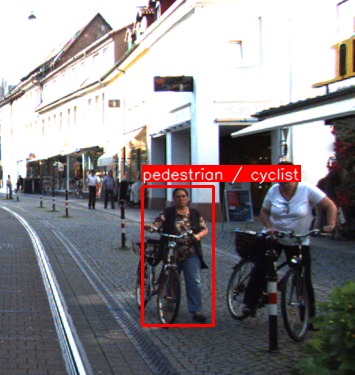}
\includegraphics[width=0.19\linewidth]{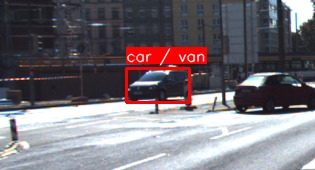}
\caption{KITTI}
\end{subfigure}

\vspace{0.4em}

\begin{subfigure}[t]{\textwidth}
\centering
\includegraphics[width=0.19\linewidth]{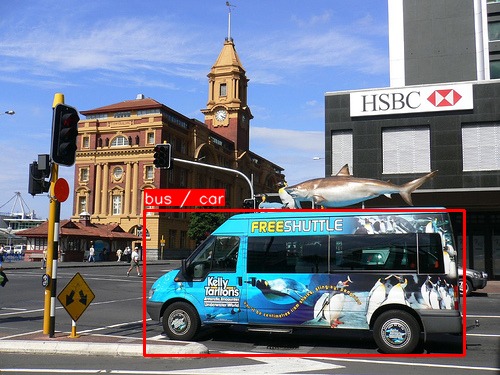}
\includegraphics[width=0.19\linewidth]{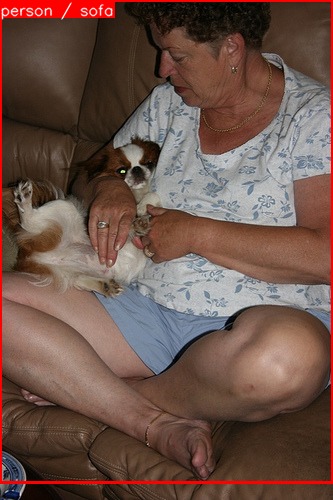}
\includegraphics[width=0.19\linewidth]{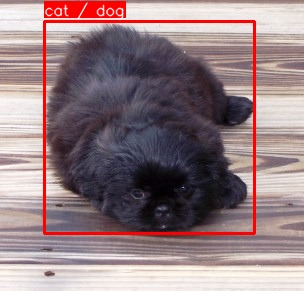}
\includegraphics[width=0.19\linewidth]{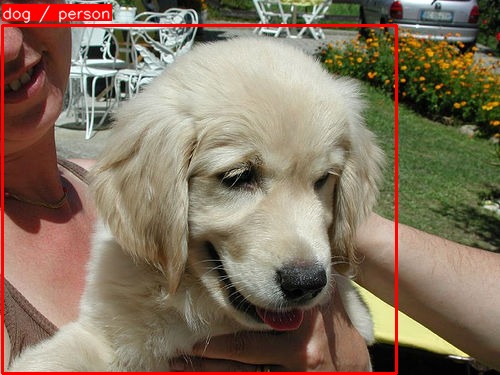}
\includegraphics[width=0.19\linewidth]{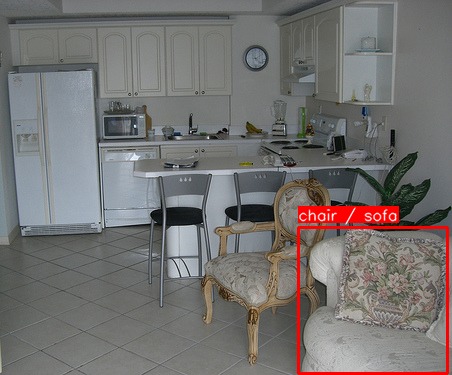}
\caption{Pascal VOC}
\end{subfigure}

\vspace{0.4em}

\begin{subfigure}[t]{\textwidth}
\centering
\includegraphics[width=0.19\linewidth]{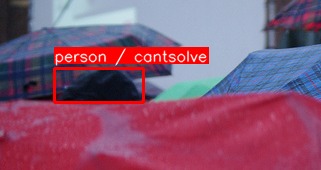}
\includegraphics[width=0.19\linewidth]{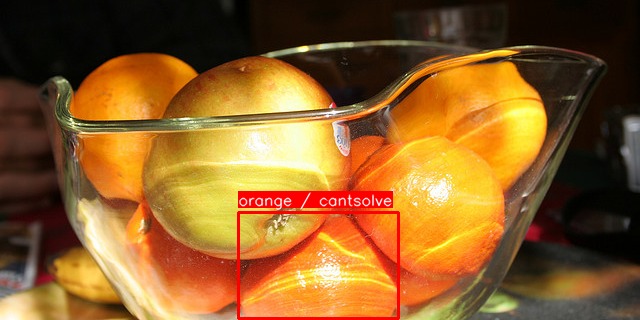}
\includegraphics[width=0.19\linewidth]{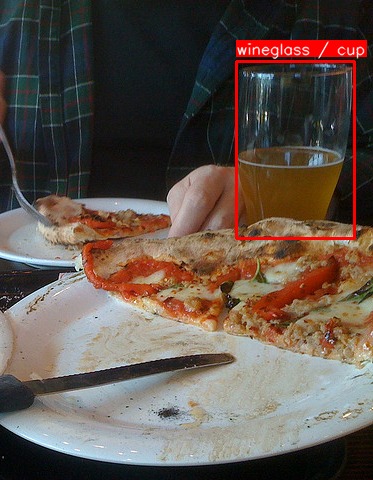}
\includegraphics[width=0.19\linewidth]{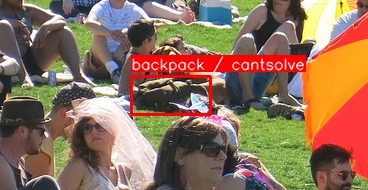}
\includegraphics[width=0.19\linewidth]{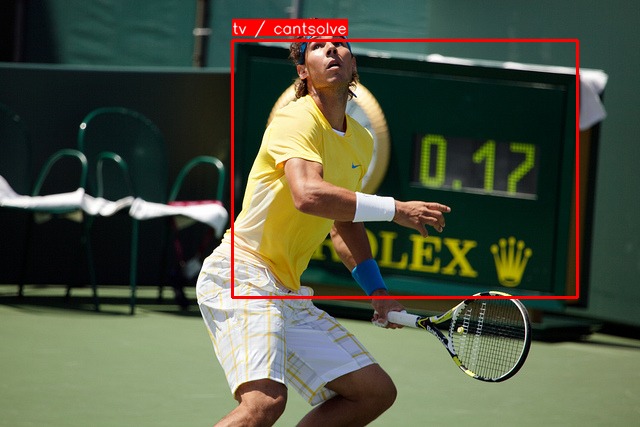}
\caption{COCO}
\end{subfigure}

\caption{Examples of ambiguous cases across all four datasets. Each bounding box corresponds to an instance where the soft label has no unique maximum, i.e., multiple classes share the highest probability. The labels shown above each box indicate the classes with equal highest probability.}
\label{fig:ambiguous_examples}
\end{figure*}

%% file: tables/Kitti_label_error_table_train.tex
\begin{table}[htbp]
\centering
\small
\caption{Number of identified label errors on the train split in the KITTI dataset depending on the probability threshold for soft label annotations and the minimal height of considered objects.}
\label{tab:train_label_errors_kitti}
\setlength{\tabcolsep}{6pt}
\begin{tabular}{lccc|ccc}
\toprule
Probability threshold & \multicolumn{3}{c|}{$p \geq 0.5$} & \multicolumn{3}{c}{$p \geq 0.8$} \\
\cmidrule(lr){2-4} \cmidrule(lr){5-7}
Object Height & Missing & Misaligned & Classification & Missing & Misaligned & Classification \\
\midrule
\multicolumn{7}{c}{\textit{Considering all objects}} \\
Arbitrary & 17,371 & 5,533 & 829 & 10,578 & 4,779 & 544 \\
$\geq 25$ pixels & 3,579 & 3,716 & 780 & 2,828 & 3,391 & 514 \\
$\geq 40$ pixels & 944 & 1,869 & 570 & 773 & 1,709 & 380 \\
\midrule
\multicolumn{7}{c}{\textit{Considering only objects outside of 'Don't Care' regions}} \\
Arbitrary & 4,933 & 5,161 & 826 & 2,577 & 4,470 & 543 \\
$\geq 25$ pixels & 1,441 & 3,509 & 777 & 1,012 & 3,214 & 513 \\
$\geq 40$ pixels & 425 & 1,808 & 567 & 310 & 1,661 & 379 \\
\bottomrule
\end{tabular}
\end{table}

%% file: tables/Kitti_label_error_table_val.tex
\begin{table}[htbp]
\centering
\small
\caption{Number of identified label errors on the val split in the KITTI dataset depending on the probability threshold for soft label annotations and the minimal height of considered objects.}
\label{tab:val_label_errors_kitti}
\setlength{\tabcolsep}{6pt}
\begin{tabular}{lccc|ccc}
\toprule
Probability threshold & \multicolumn{3}{c|}{$p \geq 0.5$} & \multicolumn{3}{c}{$p \geq 0.8$} \\
\cmidrule(lr){2-4} \cmidrule(lr){5-7}
Object Height & Missing & Misaligned & Classification & Missing & Misaligned & Classification \\
\midrule
\multicolumn{7}{c}{\textit{Considering all objects}} \\
Arbitrary & 5,339 & 1,634 & 197 & 2,527 & 1,319 & 141 \\
$\geq 25$ pixels & 951 & 1,054 & 185 & 674 & 946 & 133 \\
$\geq 40$ pixels & 235 & 511 & 138 & 188 & 465 & 99 \\
\midrule
\multicolumn{7}{c}{\textit{Considering only objects outside of 'Don't Care' regions}} \\
Arbitrary & 1,717 & 1,537 & 195 & 664 & 1,239 & 140 \\
$\geq 25$ pixels & 424 & 1,008 & 184 & 275 & 907 & 133 \\
$\geq 40$ pixels & 120 & 493 & 137 & 87 & 451 & 99 \\
\bottomrule
\end{tabular}
\end{table}

%% file: tables/Cityscapes_label_error_table_train.tex
\begin{table}[htbp]
\centering
\small
\caption{Number of identified label errors on the train split in the Cityscapes dataset depending on the probability threshold for soft label annotations and the minimal height of considered objects.}
\label{tab:train_label_errors_city}
\setlength{\tabcolsep}{6pt}
\begin{tabular}{lccc|ccc}
\toprule
Probability threshold & \multicolumn{3}{c|}{$p \geq 0.5$} & \multicolumn{3}{c}{$p \geq 0.8$} \\
\cmidrule(lr){2-4} \cmidrule(lr){5-7}
Object Height & Missing & Misaligned & Classification & Missing & Misaligned & Classification \\
\midrule
Arbitrary & 4,007 & 1,130 & 21 & 2,798 & 927 & 16 \\
$\geq 25$ pixels & 3,039 & 1,038 & 19 & 2,513 & 899 & 15 \\
$\geq 40$ pixels & 1,895 & 803 & 15 & 1,746 & 747 & 14 \\
\bottomrule
\end{tabular}
\end{table}

%% file: tables/Cityscapes_label_error_table_val.tex
\begin{table}[htbp]
\centering
\small
\caption{Number of identified label errors on the val split in the Cityscapes dataset depending on the probability threshold for soft label annotations and the minimal height of considered objects.}
\label{tab:val_label_errors_city}
\setlength{\tabcolsep}{6pt}
\begin{tabular}{lccc|ccc}
\toprule
Probability threshold & \multicolumn{3}{c|}{$p \geq 0.5$} & \multicolumn{3}{c}{$p \geq 0.8$} \\
\cmidrule(lr){2-4} \cmidrule(lr){5-7}
Object Height & Missing & Misaligned & Classification & Missing & Misaligned & Classification \\
\midrule
Arbitrary & 1,242 & 1,255 & 64 & 720 & 758 & 47 \\
$\geq 25$ pixels & 1,002 & 1,035 & 64 & 686 & 725 & 47 \\
$\geq 40$ pixels & 582 & 698 & 55 & 483 & 571 & 42 \\
\bottomrule
\end{tabular}
\end{table}

%% file: tables/COCO_label_error_table_val.tex
\begin{table}[htbp]
\centering
\small
\caption{Number of identified label errors on the val split in the COCO dataset depending on the probability threshold for soft label annotations and the minimal height of considered objects.}
\label{tab:val_label_errors_coco}
\setlength{\tabcolsep}{6pt}
\begin{tabular}{lccc|ccc}
\toprule
Probability threshold & \multicolumn{3}{c|}{$p \geq 0.5$} & \multicolumn{3}{c}{$p \geq 0.8$} \\
\cmidrule(lr){2-4} \cmidrule(lr){5-7}
Object Height & Missing & Misaligned & Classification & Missing & Misaligned & Classification \\
\midrule
Arbitrary & 16,219 & 7,736 & 254 & 11,855 & 6,152 & 146 \\
$\geq 25$ pixels & 7,224 & 5,570 & 226 & 5,949 & 4,781 & 128 \\
$\geq 40$ pixels & 3,182 & 3,983 & 172 & 2,685 & 3,497 & 98 \\
\bottomrule
\end{tabular}
\end{table}

%% file: tables/PascalVOC_label_error_table_train.tex
\begin{table}[htbp]
\centering
\small
\caption{Number of identified label errors on the train split in the Pascal VOC dataset depending on the probability threshold for soft label annotations and the minimal height of considered objects.}
\label{tab:train_label_errors_voc}
\setlength{\tabcolsep}{6pt}
\begin{tabular}{lccc|ccc}
\toprule
Probability threshold & \multicolumn{3}{c|}{$p \geq 0.5$} & \multicolumn{3}{c}{$p \geq 0.8$} \\
\cmidrule(lr){2-4} \cmidrule(lr){5-7}
Object Height & Missing & Misaligned & Classification & Missing & Misaligned & Classification \\
\midrule
Arbitrary & 1,835 & 1,120 & 29 & 957 & 856 & 11 \\
$\geq 25$ pixels & 1,004 & 932 & 29 & 629 & 752 & 11 \\
$\geq 40$ pixels & 597 & 780 & 27 & 383 & 644 & 10 \\
\bottomrule
\end{tabular}
\end{table}

%% file: tables/PascalVOC_label_error_table_val.tex
\begin{table}[htbp]
\centering
\small
\caption{Number of identified label errors on the val split in the Pascal VOC dataset depending on the probability threshold for soft label annotations and the minimal height of considered objects.}
\label{tab:val_label_errors_voc}
\setlength{\tabcolsep}{6pt}
\begin{tabular}{lccc|ccc}
\toprule
Probability threshold & \multicolumn{3}{c|}{$p \geq 0.5$} & \multicolumn{3}{c}{$p \geq 0.8$} \\
\cmidrule(lr){2-4} \cmidrule(lr){5-7}
Object Height & Missing & Misaligned & Classification & Missing & Misaligned & Classification \\
\midrule
Arbitrary & 1,819 & 1,090 & 23 & 846 & 852 & 17 \\
$\geq 25$ pixels & 1,030 & 919 & 23 & 574 & 756 & 17 \\
$\geq 40$ pixels & 636 & 774 & 21 & 378 & 646 & 15 \\
\bottomrule
\end{tabular}
\end{table}

%% file: labelerrors/original_only/plot_original_only.tex
\begin{figure*}[htbp]
\centering

\begin{subfigure}[t]{\textwidth}
\centering
\includegraphics[width=0.19\linewidth]{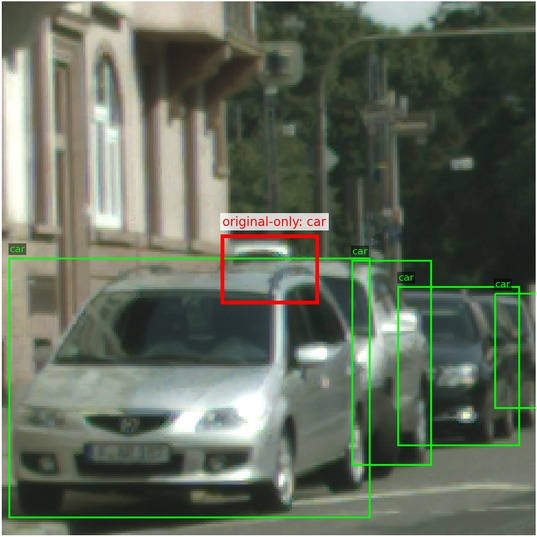}
\includegraphics[width=0.19\linewidth]{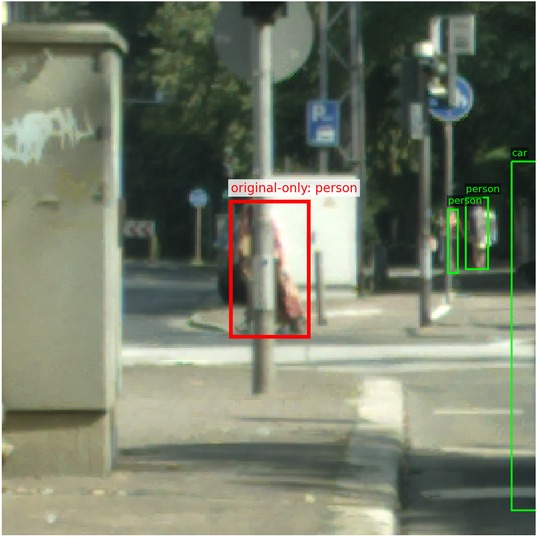}
\includegraphics[width=0.19\linewidth]{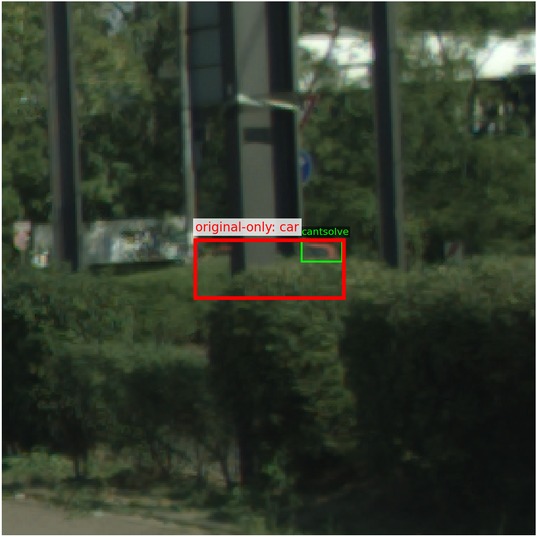}
\includegraphics[width=0.19\linewidth]{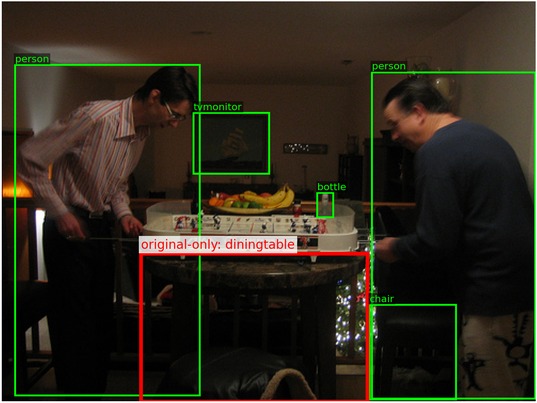}
\includegraphics[width=0.19\linewidth]{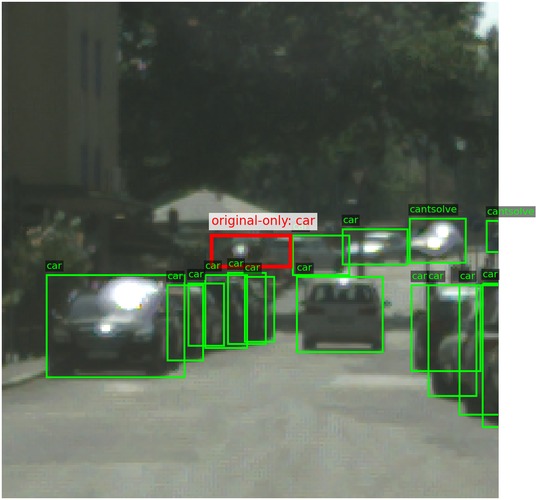}
\caption{Cityscapes}
\end{subfigure}

\vspace{0.4em}

\begin{subfigure}[t]{\textwidth}
\centering
\includegraphics[width=0.19\linewidth]{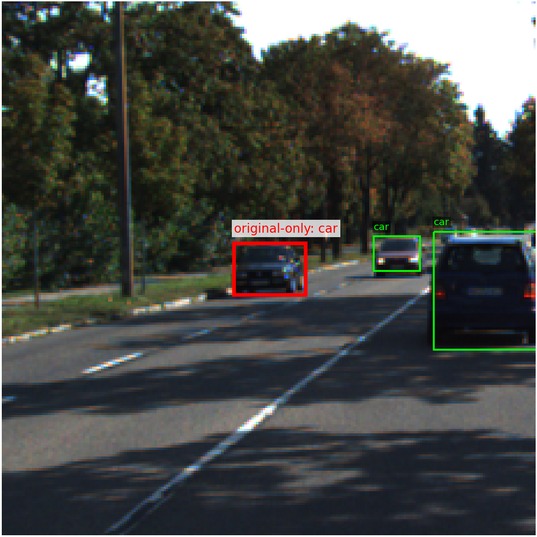}
\includegraphics[width=0.19\linewidth]{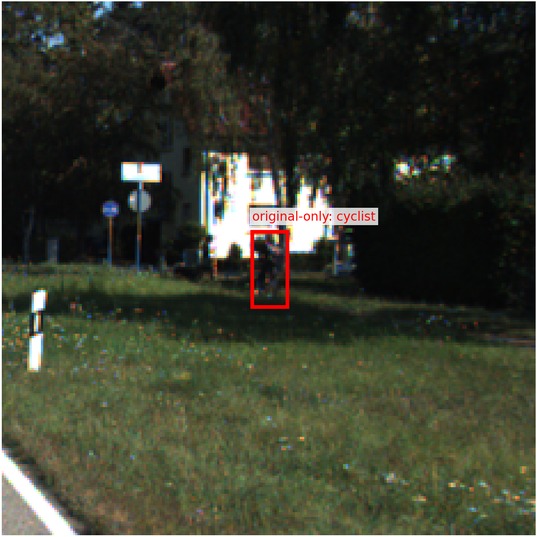}
\includegraphics[width=0.19\linewidth]{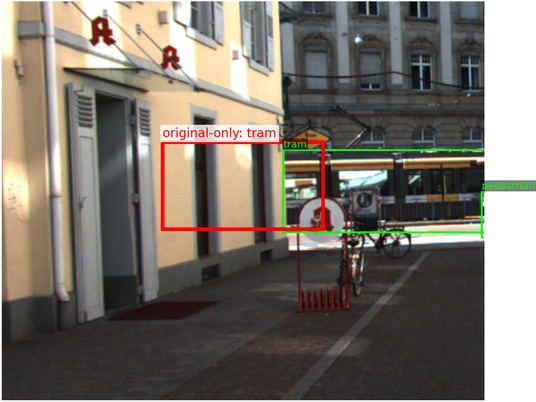}
\includegraphics[width=0.19\linewidth]{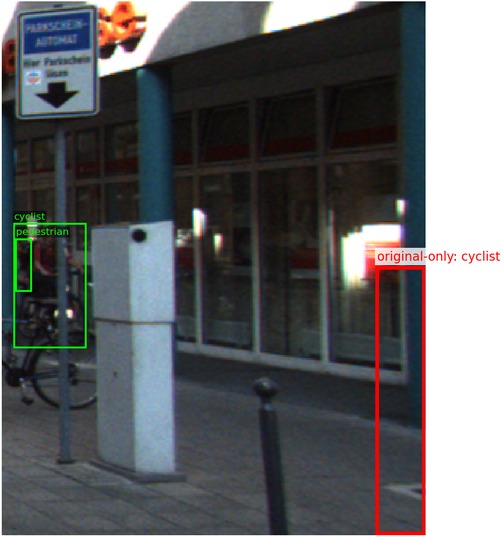}
\includegraphics[width=0.19\linewidth]{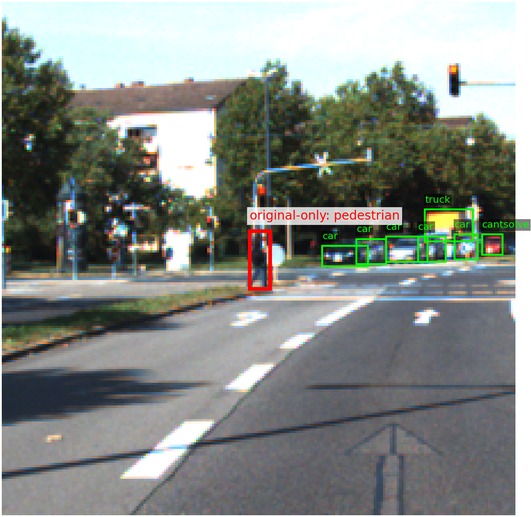}
\caption{KITTI}
\end{subfigure}

\vspace{0.4em}

\begin{subfigure}[t]{\textwidth}
\centering
\includegraphics[width=0.19\linewidth]{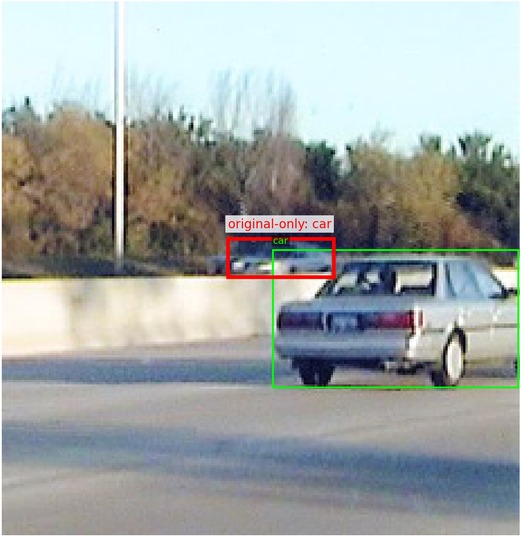}
\includegraphics[width=0.19\linewidth]{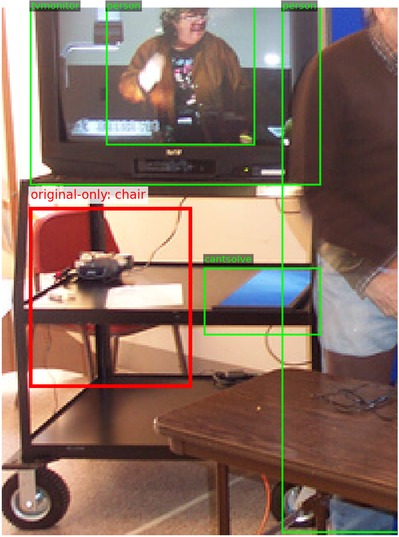}
\includegraphics[width=0.19\linewidth]{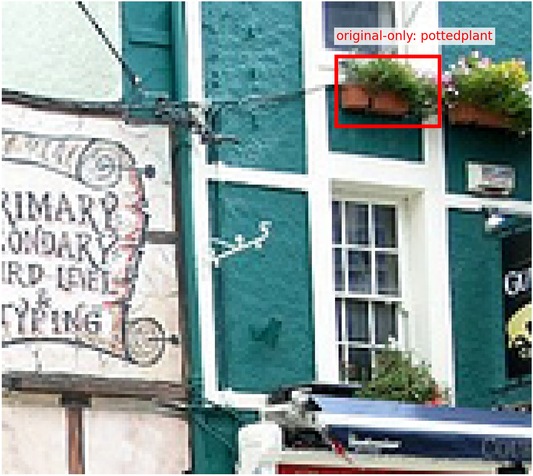}
\includegraphics[width=0.19\linewidth]{labelerrors/original_only/voc3.jpg}
\includegraphics[width=0.19\linewidth]{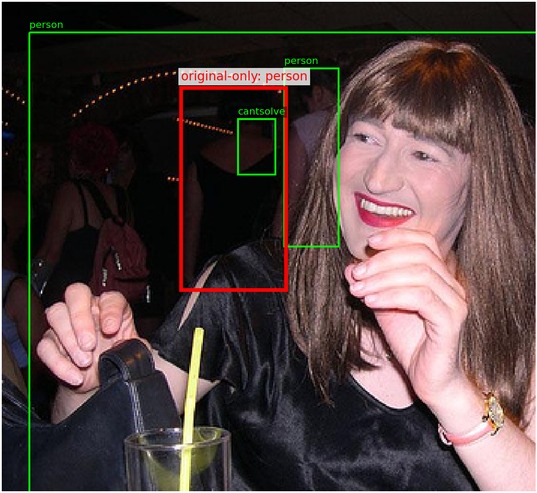}
\caption{Pascal VOC}
\end{subfigure}

\vspace{0.4em}

\begin{subfigure}[t]{\textwidth}
\centering
\includegraphics[width=0.19\linewidth]{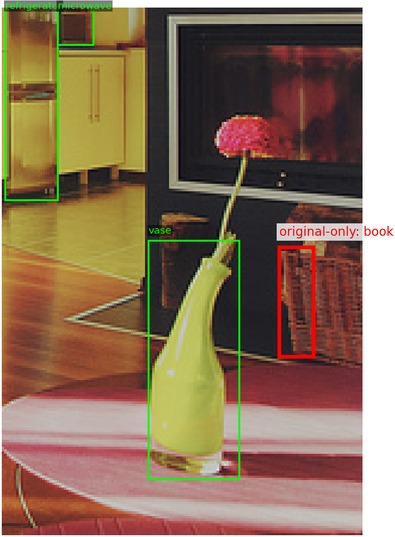}
\includegraphics[width=0.19\linewidth]{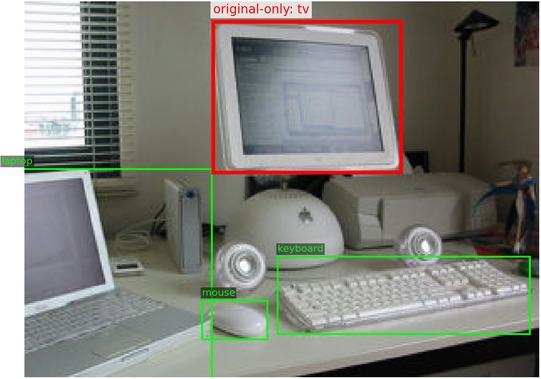}
\includegraphics[width=0.19\linewidth]{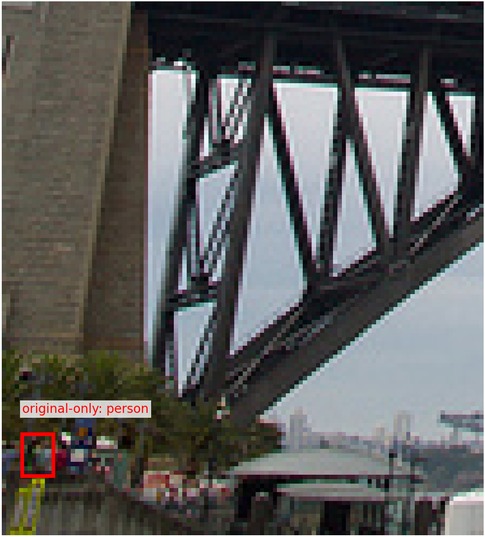}
\includegraphics[width=0.19\linewidth]{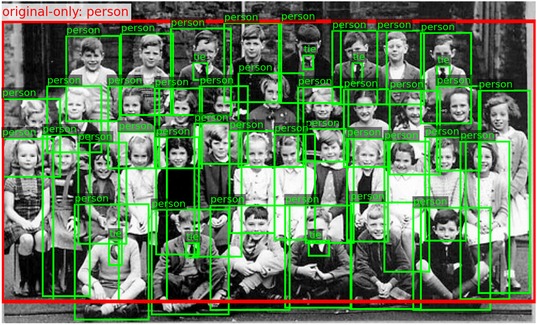}
\includegraphics[width=0.19\linewidth]{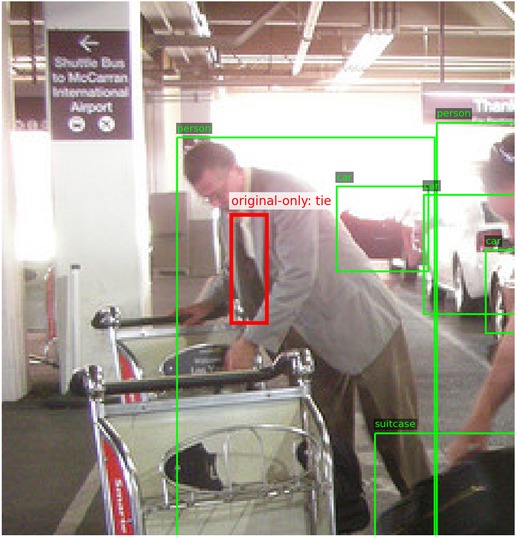}
\caption{COCO}
\end{subfigure}

\caption{
Examples of \textit{original-only} annotations across all datasets. 
Red bounding boxes correspond to annotations present only in the original ground truth, while green boxes show validated annotations. 
The examples illustrate that original-only cases arise from both missed objects during re-annotation (often small or occluded instances) and spurious or coarse annotations in the original ground truth, reflecting the inherent ambiguity of these cases.
}
\label{fig:originalonly_examples}
\end{figure*}

%% file: tables/original_only_table.tex
\begin{table}[htbp]
    \centering
    \caption{Manual review of original-only annotations across all datasets. For each original-only annotation, we manually review whether the bounding box corresponds to an object that should be included in the dataset. The acceptance rate denotes the fraction of accepted original-only annotations.}
    \label{tab:original_only_review}
    \small
    \begin{tabular}{cccccc}
        \toprule
        Dataset & Split & Original-Only & Accepted & Rejected & Accepted [\%] \\
        \midrule
        COCO
            & Val   & 3,057 & 2,037 & 1,020 & 66.63 \\
        \midrule
        Cityscapes
            & Val   & 454   & 242   & 212   & 53.30 \\
        \midrule
        \multirow{2}{*}{KITTI}
            & Train & 711   & 501   & 210   & 70.46 \\
            & Val   & 138   & 93    & 45    & 67.39 \\
        \midrule
        \multirow{2}{*}{Pascal VOC}
            & Train & 569   & 546   & 23    & 95.96 \\
            & Val   & 506   & 475   & 31    & 93.87 \\
        \bottomrule
    \end{tabular}
\end{table}

%% file: tables/reviewed_original_only_table.tex
\begin{table*}[htbp]
    \centering
    \caption{Comparison of the number of annotations in the original ground truth, the validated ground truth (VGT), and the VGT after        adding the reviewed original-only (OO) annotations. All counts in this table are based on hard label annotations. The relative increases of the VGT and VGT + original-only are computed with respect to the original GT.}
    \label{tab:vgt_original_only_hardlabels}
    \small
    \begin{tabular}{cccccccc}
        \toprule
        Dataset & Split
        & Original GT
        & VGT
        & $\Delta_{\mathrm{VGT}}$ [\%]
        & Original-Only Added
        & VGT + Original-Only
        & $\Delta_{\mathrm{VGT+OO}}$ [\%] \\
        \midrule

        Cityscapes
            & Val
            & 10,180
            & 10,383
            & +1.99
            & 209
            & 10,592
            & +4.05 \\

        \midrule

        \multirow{2}{*}{KITTI}
            & Train
            & 31,589
            & 49,551
            & +56.86
            & 376
            & 49,927
            & +58.05 \\
            & Val
            & 8,008
            & 13,797
            & +72.29
            & 72
            & 13,869
            & +73.19 \\

        \midrule

        \multirow{2}{*}{Pascal VOC}
            & Train
            & 15,774
            & 17,029
            & +7.96
            & 524
            & 17,553
            & +11.28 \\
            & Val
            & 15,787
            & 17,233
            & +9.16
            & 470
            & 17,703
            & +12.14 \\

        \midrule

        COCO
            & Val
            & 36,781
            & 51,487
            & +39.98
            & 1,553
            & 53,040
            & +44.20 \\

        \bottomrule
    \end{tabular}
\end{table*}

%% file: tables/fnr_original_only.tex
\begin{table}[ht]
\centering
\caption{Effect of incorporating reviewed original-only annotations on the false negative rate (FNR). Values are reported in \% with 95\% Wilson confidence intervals. The validated GT corresponds to the validated annotations before adding original-only annotations, while validated GT + original-only additionally includes the accepted original-only annotations.}
\label{tab:fnr_original_only}
\small
\resizebox{0.7\linewidth}{!}{
\begin{tabular}{c cc c}
\toprule
Dataset 
& Validated GT 
& Validated GT + Original-Only 
& Relative Reduction [\%] \\
\midrule
Pascal VOC  & $9.3 \pm 1.3$  & $7.2 \pm 1.1$  & 22.6 \\
COCO        & $6.1 \pm 0.7$  & $3.6 \pm 0.5$  & 41.0 \\
KITTI       & $11.6 \pm 0.8$ & $10.6 \pm 0.8$ & 8.6 \\
Cityscapes  & $3.8 \pm 0.3$  & $2.8 \pm 0.3$  & 26.3 \\
\bottomrule
\end{tabular}
}
\end{table}

%% file: tables/post_processing_results_full.tex
\begin{table}[ht]
\centering
\caption{Post-processing results per dataset. Here, $\bar{P}_\text{unbiased}$ denotes the mean probability of the proposed class for unbiased annotations, $\bar{P}_\text{biased}$ the mean probability of the proposed class for the biased annotations, and $\bar{P}_\text{post}$ the mean probability of the proposed class after post processing. CL = CleverLabel with parameters $\delta$ and avoid-overcorrection threshold $\alpha$; CL Lite uses only $\delta$ as a downweight factor. $\sqrt{\mathrm{JS}}$ denotes the square root of the Jensen--Shannon divergence. $\Delta\!\sqrt{\mathrm{JS}}$ is the average bootstrapped difference
$\sqrt{\mathrm{JS}}(\text{post},\text{unbiased})-\sqrt{\mathrm{JS}}(\text{biased},\text{unbiased})$; negative values indicate improvement.}
\label{tbl:post_process_results_full}
\small
\setlength{\tabcolsep}{3.5pt}
\resizebox{\linewidth}{!}{%
\begin{tabular}{l l c c c c c}
\toprule
Dataset
  & Method
  & $\bar{P}_\text{unbiased}$
  & $\bar{P}_\text{biased}$
  & $\bar{P}_\text{post}$
  & $\sqrt{\mathrm{JS}}$ (mean$\pm$std)
  & $\Delta\!\sqrt{\mathrm{JS}}$ (95\% CI) \\
\midrule
Cityscapes
  & CL $\delta\!=\!0.10$, $\alpha\!=\!0.5$
  & .827 & .888 & .882 & 0.1301 $\pm$ 0.1490 & $-0.00008$ ([+0.00035, $-0.00053$]) \\
Kitti
  & CL $\delta\!=\!0.00$
  & .886 & .868 & .871 & 0.1337 $\pm$ 0.1579 & $-0.00193$ ([$-0.00170$, $-0.00214$]) \\
Pascal VOC
  & CL Lite $\delta\!=\!0.50$
  & .845 & .915 & .877 & 0.0902 $\pm$ 0.1236 & $-0.00646$ ([$-0.00391$, $-0.00879$]) \\
\midrule
COCO/accessory
  & CL Lite $\delta\!=\!0.40$
  & .772 & .914 & .845 & 0.1626 $\pm$ 0.1636 & $-0.01468$ ([$-0.01132$, $-0.01824$]) \\
COCO/animal
  & CL $\delta\!=\!0.15$, $\alpha\!=\!0.4$
  & .784 & .829 & .819 & 0.0938 $\pm$ 0.1222 & $-0.00089$ ([$-0.00012$, $-0.00164$]) \\
COCO/electronic
  & CL $\delta\!=\!0.40$, $\alpha\!=\!0.8$
  & .868 & .946 & .937 & 0.1159 $\pm$ 0.1387 & $-0.00173$ ([$-0.00078$, $-0.00273$]) \\
COCO/food
  & CL Lite $\delta\!=\!0.50$
  & .872 & .946 & .918 & 0.1076 $\pm$ 0.1385 & $-0.00325$ ([$-0.00121$, $-0.00528$]) \\
COCO/furniture
  & CL $\delta\!=\!0.30$
  & .840 & .940 & .918 & 0.1401 $\pm$ 0.1460 & $-0.00281$ ([$-0.00147$, $-0.00419$]) \\
COCO/indoor
  & CL $\delta\!=\!0.00$, $\alpha\!=\!0.8$
  & .920 & .950 & .953 & 0.0890 $\pm$ 0.1018 & $-0.00099$ ([$-0.00066$, $-0.00133$]) \\
COCO/kitchen
  & CL Lite $\delta\!=\!0.30$
  & .734 & .902 & .806 & 0.1876 $\pm$ 0.1718 & $-0.01726$ ([$-0.01271$, $-0.02151$]) \\
COCO/outdoor
  & CL $\delta\!=\!0.20$, $\alpha\!=\!0.5$
  & .829 & .881 & .861 & 0.1164 $\pm$ 0.1223 & $-0.00293$ ([$-0.00159$, $-0.00429$]) \\
COCO/person
  & CL $\delta\!=\!0.00$, $\alpha\!=\!0.8$
  & .883 & .876 & .877 & 0.0572 $\pm$ 0.0842 & $-0.00043$ ([$-0.00021$, $-0.00066$]) \\
COCO/sports
  & CL Lite $\delta\!=\!0.50$
  & .823 & .928 & .890 & 0.1278 $\pm$ 0.1637 & $-0.00681$ ([$-0.00455$, $-0.00905$]) \\
COCO/vehicle
  & CL $\delta\!=\!0.05$, $\alpha\!=\!0.2$
  & .810 & .842 & .838 & 0.1111 $\pm$ 0.1373 & $+0.00011$ ([+0.00049, $-0.00029$]) \\
\bottomrule
\end{tabular}
}
\end{table}

%% file: tables/wrong_labels_full.tex
\begin{table}[htbp]
\centering
\caption{COCO. Bucket agreement between validated annotations and human expert judgments. Bucket accuracy is 62.8\,\% vs.\ 72.4\,\%, Error 1+ is 37.2\,\% vs.\ 27.6\,\%, and Error 2+ is 3.5\,\% vs.\ 0.0\,\%.}
\begin{subtable}[t]{0.48\linewidth}
\centering
\caption{Validated GT}
\resizebox{\linewidth}{!}{
\begin{tabular}{l ccc c}
\toprule
Avg. Reviewer $\backslash$ Prediction & no & maybe & clearly & Total \\
\midrule
no      & 24 & 4  & 0  & 28 \\
maybe   & 35 & 29 & 8  & 72 \\
clearly & 7  & 20 & 72 & 99 \\
\midrule
Total   & 66 & 53 & 80 & 199 \\
\bottomrule
\end{tabular}
}
\end{subtable}%
\hfill
\begin{subtable}[t]{0.48\linewidth}
\centering
\caption{Human Expert}
\resizebox{\linewidth}{!}{
\begin{tabular}{l ccc c}
\toprule
Reviewer $\backslash$ Avg. & no & maybe & clearly & Total \\
\midrule
no      & 71 & 56 & 0   & 127 \\
maybe   & 13 & 88 & 24  & 125 \\
clearly & 0  & 72 & 273 & 345 \\
\midrule
Total   & 84 & 216 & 297 & 597 \\
\bottomrule
\end{tabular}
}
\end{subtable}

\label{fig:bucket_coco}
\end{table}

\begin{table}[htbp]
\centering
\caption{Cityscapes. Bucket agreement between validated annotations and human expert judgments. Bucket accuracy is 68.0\,\% vs.\ 82.7\,\%, Error 1+ is 32.0\,\% vs.\ 17.3\,\%, and Error 2+ is 3.0\,\% vs.\ 0.0\,\%.}
\begin{subtable}[t]{0.48\linewidth}
\centering
\caption{Validated GT}
\resizebox{\linewidth}{!}{
\begin{tabular}{l ccc c}
\toprule
Avg. Reviewer $\backslash$ Prediction & no & maybe & clearly & Total \\
\midrule
no      & 6  & 4  & 0  & 10 \\
maybe   & 15 & 30 & 4  & 49 \\
clearly & 6  & 35 & 100 & 141 \\
\midrule
Total   & 27 & 69 & 104 & 200 \\
\bottomrule
\end{tabular}
}
\end{subtable}%
\hfill
\begin{subtable}[t]{0.48\linewidth}
\centering
\caption{Human Expert}
\resizebox{\linewidth}{!}{
\begin{tabular}{l ccc c}
\toprule
Reviewer $\backslash$ Avg. & no & maybe & clearly & Total \\
\midrule
no      & 25 & 23 & 0   & 48 \\
maybe   & 5  & 86 & 38  & 129 \\
clearly & 0  & 38 & 385 & 423 \\
\midrule
Total   & 30 & 147 & 423 & 600 \\
\bottomrule
\end{tabular}
}
\end{subtable}

\label{fig:bucket_cityscapes}
\end{table}

\begin{table}[htbp]
\centering
\caption{KITTI. Bucket agreement between validated annotations and human expert judgments. Bucket accuracy is 68.4\,\% vs.\ 74.3\,\%, Error 1+ is 31.6\,\% vs.\ 25.7\,\%, and Error 2+ is 2.6\,\% vs.\ 0.0\,\%.}
\begin{subtable}[t]{0.48\linewidth}
\centering
\caption{Validated GT}
\resizebox{\linewidth}{!}{
\begin{tabular}{l ccc c}
\toprule
Avg. Reviewer $\backslash$ Prediction & no & maybe & clearly & Total \\
\midrule
no      & 9  & 5  & 0  & 14 \\
maybe   & 18 & 51 & 3  & 72 \\
clearly & 5  & 31 & 74 & 110 \\
\midrule
Total   & 32 & 87 & 77 & 196 \\
\bottomrule
\end{tabular}
}
\end{subtable}%
\hfill
\begin{subtable}[t]{0.48\linewidth}
\centering
\caption{Human Expert}
\resizebox{\linewidth}{!}{
\begin{tabular}{l ccc c}
\toprule
Reviewer $\backslash$ Avg. & no & maybe & clearly & Total \\
\midrule
no      & 30 & 40 & 0   & 70 \\
maybe   & 12 & 106 & 29 & 147 \\
clearly & 0  & 70  & 301 & 371 \\
\midrule
Total   & 42 & 216 & 330 & 588 \\
\bottomrule
\end{tabular}
}
\end{subtable}

\label{fig:bucket_kitti}
\end{table}

\begin{table}[htbp]
\centering
\caption{Pascal VOC. Bucket agreement between validated annotations and human expert judgments. Bucket accuracy is 77.6\,\% vs.\ 82.7\,\%, Error 1+ is 22.4\,\% vs.\ 17.3\,\%, and Error 2+ is 2.0\,\% vs.\ 0.0\,\%.}
\begin{subtable}[t]{0.48\linewidth}
\centering
\caption{Validated GT}
\resizebox{\linewidth}{!}{
\begin{tabular}{l ccc c}
\toprule
Avg. Reviewer $\backslash$ Prediction & no & maybe & clearly & Total \\
\midrule
no      & 9  & 1  & 0  & 10 \\
maybe   & 14 & 34 & 5  & 53 \\
clearly & 4  & 20 & 109 & 133 \\
\midrule
Total   & 27 & 55 & 114 & 196 \\
\bottomrule
\end{tabular}
}
\end{subtable}%
\hfill
\begin{subtable}[t]{0.48\linewidth}
\centering
\caption{Human Expert}
\resizebox{\linewidth}{!}{
\begin{tabular}{l ccc c}
\toprule
Reviewer $\backslash$ Avg. & no & maybe & clearly & Total \\
\midrule
no      & 25 & 31 & 0   & 56 \\
maybe   & 5  & 79 & 17  & 101 \\
clearly & 0  & 49 & 382 & 431 \\
\midrule
Total   & 30 & 159 & 399 & 588 \\
\bottomrule
\end{tabular}
}
\end{subtable}

\label{fig:bucket_pascal}
\end{table}

\begin{table}[ht]
\centering
\caption{Original ground-truth labels: certainly wrong labels only. Counts are based on the ``definitely wrong'' annotations out of $n=200$ sampled labels per dataset.}
\label{tab:original_gt_certain_errors}
\small
\begin{tabular}{l cc}
\toprule
Dataset & $n$ & Certainly wrong \\
\midrule
COCO       & 200 & 6 \\
Cityscapes & 200 & 3 \\
KITTI      & 200 & 5 \\
PascalVOC  & 200 & 1 \\
\bottomrule
\end{tabular}
\end{table}

%% file: imgs/collage_wrong_class.tex
\begin{figure}[htbp]
\centering

\begin{subfigure}{0.24\linewidth}
    \centering
    \includegraphics[width=\linewidth]{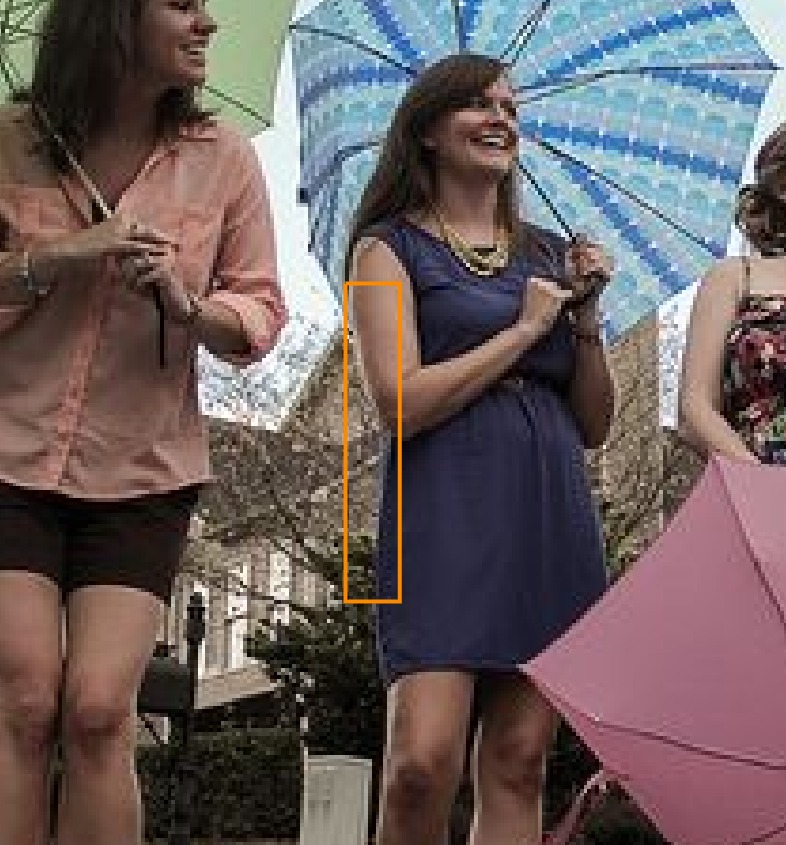}
     \caption{Human vs. \textbf{Can't Solve}}
\end{subfigure}
\hfill
\begin{subfigure}{0.24\linewidth}
    \centering
    \includegraphics[width=\linewidth]{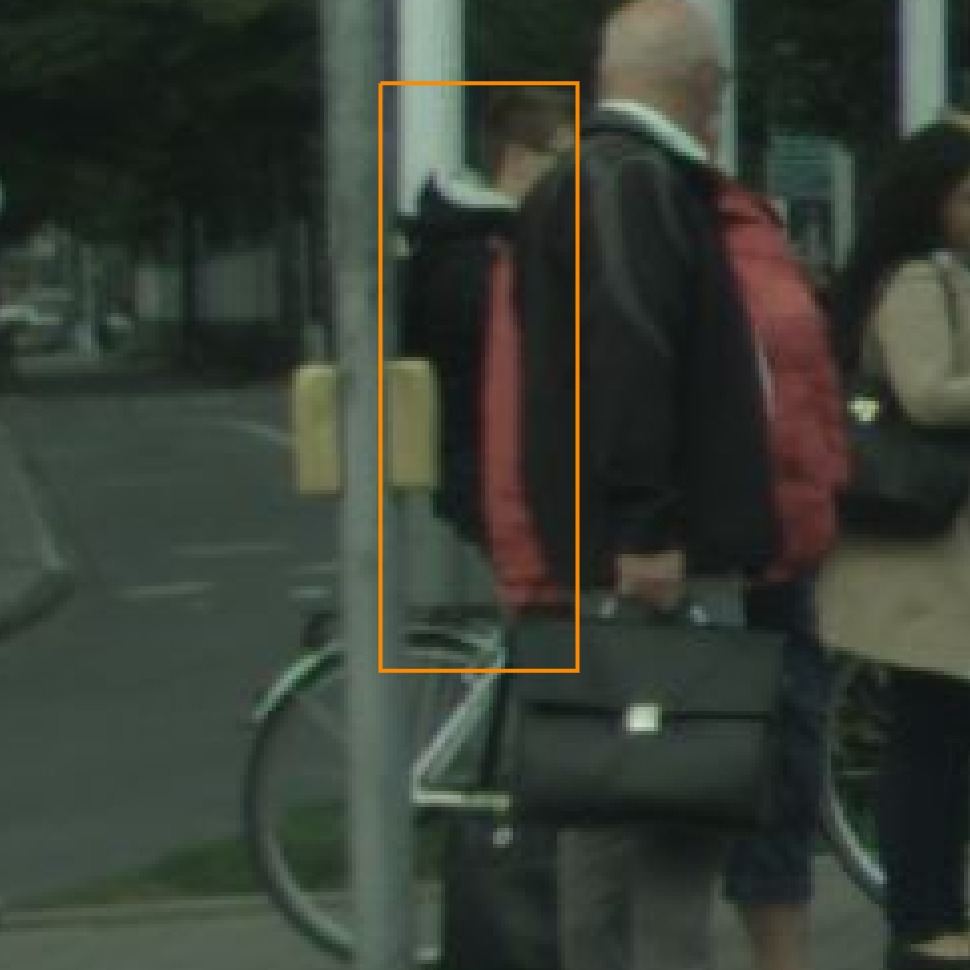}
     \caption{Person vs. \textbf{Rider}}
\end{subfigure}
\hfill
\begin{subfigure}{0.24\linewidth}
    \centering
    \includegraphics[width=\linewidth]{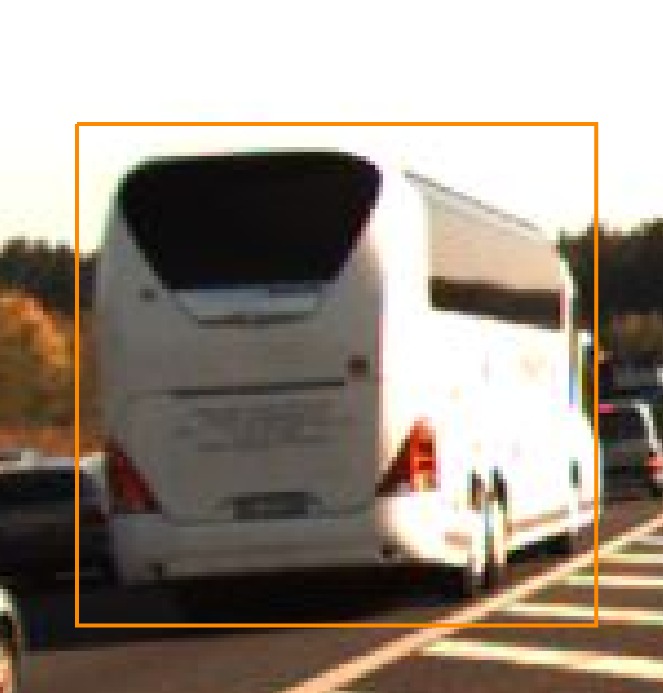}
     \caption{Misc vs. \textbf{Truck}}
\end{subfigure}
\hfill
\begin{subfigure}{0.24\linewidth}
    \centering
    \includegraphics[width=\linewidth]{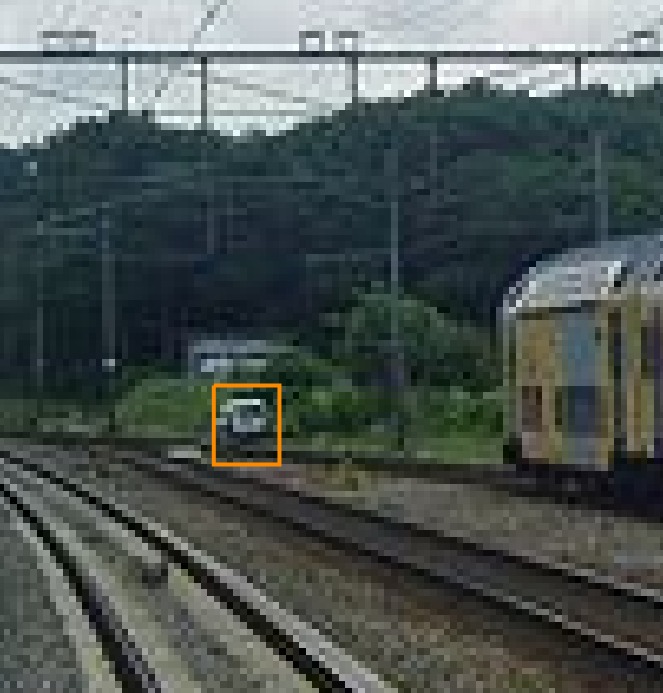}
     \caption{Car vs.  \textbf{Can't Solve}}
\end{subfigure}

\vspace{0.3em}

\begin{subfigure}{0.24\linewidth}
    \centering
    \includegraphics[width=\linewidth]{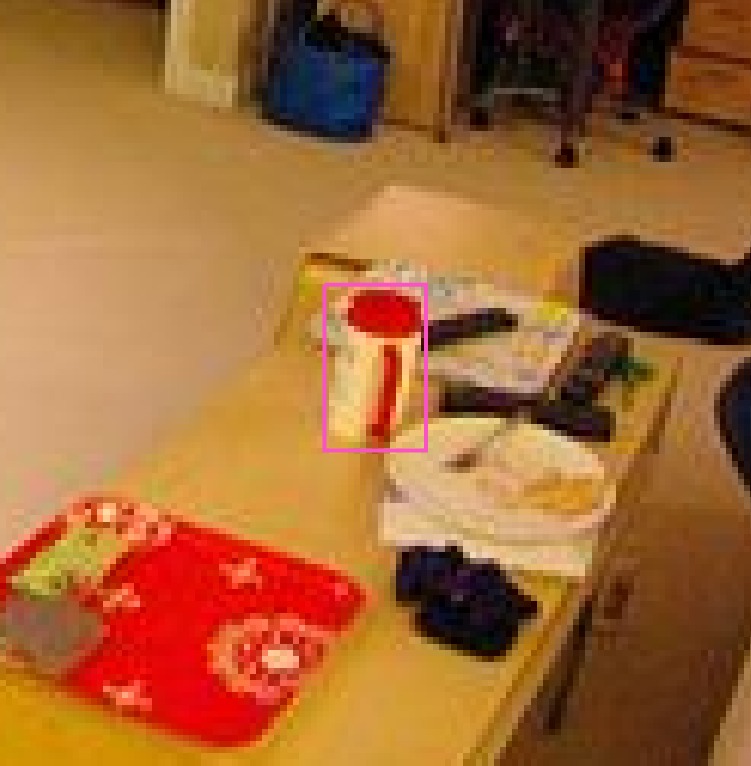}
     \caption{\textbf{cup} 0.19, bottle 0.07, can't solve 0.73}
\end{subfigure}
\hfill
\begin{subfigure}{0.24\linewidth}
    \centering
    \includegraphics[width=\linewidth]{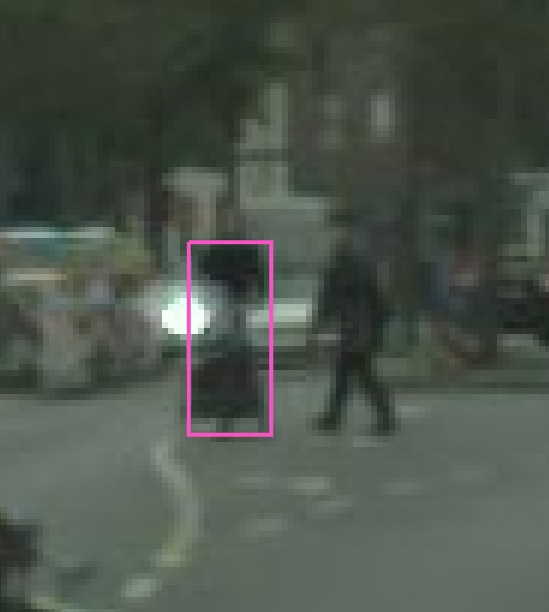}
    \caption{\textbf{rider} 0.05, person 0.76, can't solve 0.20}
\end{subfigure}
\hfill
\begin{subfigure}{0.24\linewidth}
    \centering
    \includegraphics[width=\linewidth]{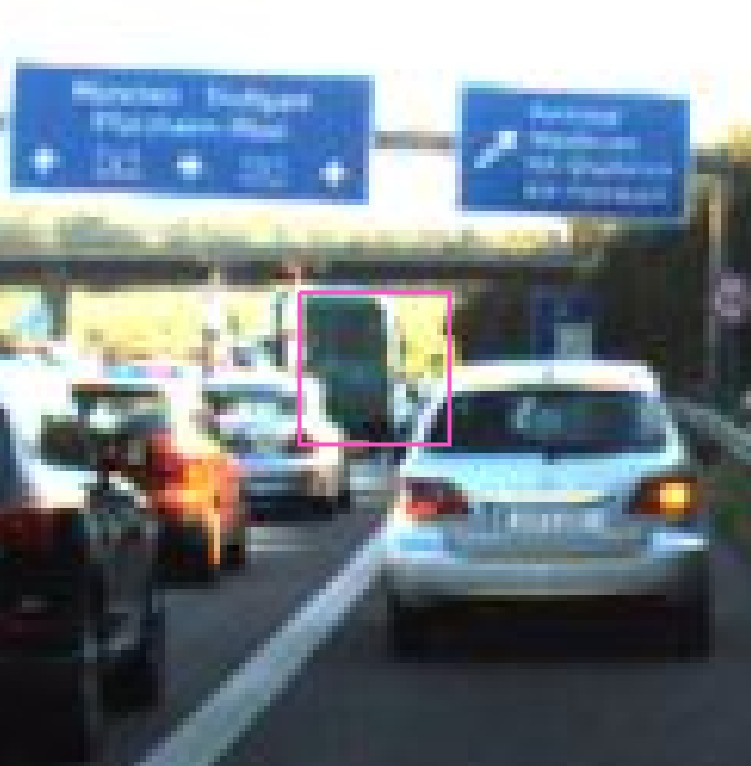}
    \caption{\textbf{truck} 0.08, van 0.69, can't solve 0.22}
\end{subfigure}
\hfill
\begin{subfigure}{0.24\linewidth}
    \centering
    \includegraphics[width=\linewidth]{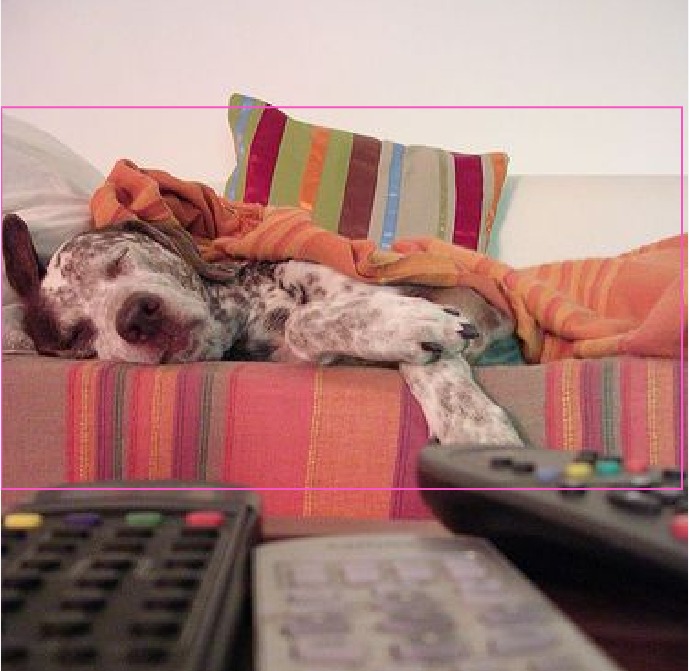}
    \caption{\textbf{dog} 0.08, sofa 0.69, can't solve 0.23}
\end{subfigure}

\caption{Comparison of wrong class errors, Rows: Original GT, Validated GT (ours), Columns: COCO, Cityscapes, KITTI, Pascal VOC
For the original GT the ground truth label is given versus the expert reviewed label in bold font.
For the validated GT the complete soft distribution is given and in bold the expert reviewed label.
We see that the original GT has several very clear errors, especially the human in COCO since the woman holding the umbrella already has a different bounding box (not shown) or truck in KITTI which includes buses.
For the validated GT, we always at least some signal for the correct class and for the example in Pascal VOC it is debatable if dog or sofa is actually correct given only one box while the other is not shown.
}
\label{fig:wrong_label_examples}
\end{figure}

%% file: tables/map_comparison_coco.tex
\begin{table}[tb]
\setlength{\tabcolsep}{3pt}
\centering
\caption{Performance of object detectors on original vs.\ refined annotations on COCO.}
 \label{tab:mAP_comparison_COCO}
\resizebox{\linewidth}{!}{
\begin{tabular}{ccccccccc|cccccccc|cccccccc}
\toprule
{Model} & 
\multicolumn{8}{c}{Original Annotations} & 
\multicolumn{8}{c}{Refined Annotations ($p>0.5$)} & 
\multicolumn{8}{c}{Refined Annotations ($p>0.8$)} \\ 
\cmidrule(lr){2-9} \cmidrule(lr){10-17} \cmidrule(lr){18-25}
& AP\textsubscript{50..95} & AP\textsubscript{50} & AP\textsubscript{75} & AP\textsubscript{S} & AP\textsubscript{M} & AP\textsubscript{L} & AR\textsubscript{10} & AR\textsubscript{100}
& AP\textsubscript{50..95} & AP\textsubscript{50} & AP\textsubscript{75} & AP\textsubscript{S} & AP\textsubscript{M} & AP\textsubscript{L} & AR\textsubscript{10} & AR\textsubscript{100}
& AP\textsubscript{50..95} & AP\textsubscript{50} & AP\textsubscript{75} & AP\textsubscript{S} & AP\textsubscript{M} & AP\textsubscript{L} & AR\textsubscript{10} & AR\textsubscript{100} \\ 
\midrule
Grounding DINO & \percent{0.5690} & \percent{0.7425} & \percent{0.6257} & \percent{0.4009} & \percent{0.6111} & \percent{0.7260} & \percent{0.7053} & \percent{0.7744} & \percent{0.5122} & \percent{0.7130} & \percent{0.5495} & \percent{0.2633} & \percent{0.5033} & \percent{0.7404} & \percent{0.6130} & \percent{0.6958} & \percent{0.5210} & \percent{0.7177} & \percent{0.5611} & \percent{0.2800} & \percent{0.5027} & \percent{0.7291} & \percent{0.6373} & \percent{0.7174} \\
YOLOv8-World v2 & \percent{0.4453} & \percent{0.6016} & \percent{0.4841} & \percent{0.2894} & \percent{0.4890} & \percent{0.5723} & \percent{0.5847} & \percent{0.6290} & \percent{0.4232} & \percent{0.5905} & \percent{0.4548} & \percent{0.1940} & \percent{0.4210} & \percent{0.6056} & \percent{0.5378} & \percent{0.5941} & \percent{0.4377} & \percent{0.6047} & \percent{0.4724} & \percent{0.2127} & \percent{0.4283} & \percent{0.6022} & \percent{0.5664} & \percent{0.6230} \\
OWL-ViT & \percent{0.2868} & \percent{0.4544} & \percent{0.3016} & \percent{0.1104} & \percent{0.2965} & \percent{0.4802} & \percent{0.4430} & \percent{0.4851} & \percent{0.2740} & \percent{0.4307} & \percent{0.2846} & \percent{0.0653} & \percent{0.2217} & \percent{0.4749} & \percent{0.4099} & \percent{0.4515} & \percent{0.2855} & \percent{0.4470} & \percent{0.2978} & \percent{0.0717} & \percent{0.2282} & \percent{0.4730} & \percent{0.4344} & \percent{0.4779} \\
RT-DETR-L & \percent{0.5239} & \percent{0.7064} & \percent{0.5677} & \percent{0.3364} & \percent{0.5699} & \percent{0.7054} & \percent{0.6359} & \percent{0.6963} & \percent{0.4614} & \percent{0.6558} & \percent{0.4864} & \percent{0.2094} & \percent{0.4442} & \percent{0.6856} & \percent{0.5503} & \percent{0.6211} & \percent{0.4720} & \percent{0.6642} & \percent{0.5002} & \percent{0.2281} & \percent{0.4437} & \percent{0.6720} & \percent{0.5764} & \percent{0.6475} \\
YOLOv8-L & \percent{0.5168} & \percent{0.6858} & \percent{0.5643} & \percent{0.3427} & \percent{0.5712} & \percent{0.6874} & \percent{0.6243} & \percent{0.6699} & \percent{0.4641} & \percent{0.6460} & \percent{0.4947} & \percent{0.2132} & \percent{0.4586} & \percent{0.6730} & \percent{0.5510} & \percent{0.6104} & \percent{0.4779} & \percent{0.6583} & \percent{0.5116} & \percent{0.2339} & \percent{0.4576} & \percent{0.6686} & \percent{0.5799} & \percent{0.6397} \\
Faster R-CNN (ResNet 50) & \percent{0.4709} & \percent{0.6817} & \percent{0.5147} & \percent{0.3163} & \percent{0.5090} & \percent{0.6042} & \percent{0.6063} & \percent{0.6497} & \percent{0.4168} & \percent{0.6357} & \percent{0.4398} & \percent{0.2038} & \percent{0.4190} & \percent{0.5994} & \percent{0.5253} & \percent{0.5780} & \percent{0.4312} & \percent{0.6502} & \percent{0.4576} & \percent{0.2183} & \percent{0.4243} & \percent{0.5982} & \percent{0.5484} & \percent{0.6000} \\
\bottomrule
\end{tabular}
}
\end{table}

%% file: tables/map_comparison_PascalVOC.tex
\begin{table}[tb]
\setlength{\tabcolsep}{3pt}
\centering
\caption{Performance of object detectors on original vs.\ refined annotations on Pascal VOC.}
 \label{tab:mAP_comparison_VOC}
\resizebox{\linewidth}{!}{
\begin{tabular}{ccccccccc|cccccccc|cccccccc}
\toprule
{Model} & 
\multicolumn{8}{c}{Original Annotations} & 
\multicolumn{8}{c}{Refined Annotations ($p>0.5$)} & 
\multicolumn{8}{c}{Refined Annotations ($p>0.8$)} \\ 
\cmidrule(lr){2-9} \cmidrule(lr){10-17} \cmidrule(lr){18-25}
& AP\textsubscript{50..95} & AP\textsubscript{50} & AP\textsubscript{75} & AP\textsubscript{S} & AP\textsubscript{M} & AP\textsubscript{L} & AR\textsubscript{10} & AR\textsubscript{100}
& AP\textsubscript{50..95} & AP\textsubscript{50} & AP\textsubscript{75} & AP\textsubscript{S} & AP\textsubscript{M} & AP\textsubscript{L} & AR\textsubscript{10} & AR\textsubscript{100}
& AP\textsubscript{50..95} & AP\textsubscript{50} & AP\textsubscript{75} & AP\textsubscript{S} & AP\textsubscript{M} & AP\textsubscript{L} & AR\textsubscript{10} & AR\textsubscript{100} \\ 
\midrule
Grounding DINO & \percent{0.8005} & \percent{0.9244} & \percent{0.8749} & \percent{0.3347} & \percent{0.6653} & \percent{0.8799} & \percent{0.8824} & \percent{0.9094} & \percent{0.7230} & \percent{0.8861} & \percent{0.7891} & \percent{0.3000} & \percent{0.5970} & \percent{0.8403} & \percent{0.8057} & \percent{0.8492} & \percent{0.7378} & \percent{0.8950} & \percent{0.8078} & \percent{0.3202} & \percent{0.6012} & \percent{0.8402} & \percent{0.8255} & \percent{0.8650} \\
YOLOv8-World v2 & \percent{0.6965} & \percent{0.8341} & \percent{0.7573} & \percent{0.2678} & \percent{0.5624} & \percent{0.7755} & \percent{0.8190} & \percent{0.8371} & \percent{0.6377} & \percent{0.8002} & \percent{0.6907} & \percent{0.2099} & \percent{0.5010} & \percent{0.7537} & \percent{0.7439} & \percent{0.7720} & \percent{0.6544} & \percent{0.8132} & \percent{0.7109} & \percent{0.2412} & \percent{0.5114} & \percent{0.7554} & \percent{0.7661} & \percent{0.7916} \\
OWL-ViT & \percent{0.5275} & \percent{0.7197} & \percent{0.5911} & \percent{0.1165} & \percent{0.3631} & \percent{0.6280} & \percent{0.7427} & \percent{0.7798} & \percent{0.4796} & \percent{0.6795} & \percent{0.5294} & \percent{0.0983} & \percent{0.3234} & \percent{0.6155} & \percent{0.6715} & \percent{0.7131} & \percent{0.4921} & \percent{0.6918} & \percent{0.5452} & \percent{0.1137} & \percent{0.3290} & \percent{0.6154} & \percent{0.6939} & \percent{0.7345} \\
RT-DETR-L & \percent{0.7028} & \percent{0.8598} & \percent{0.7694} & \percent{0.2931} & \percent{0.5536} & \percent{0.7759} & \percent{0.8234} & \percent{0.8776} & \percent{0.6261} & \percent{0.8078} & \percent{0.6824} & \percent{0.2186} & \percent{0.4828} & \percent{0.7447} & \percent{0.7363} & \percent{0.8030} & \percent{0.6418} & \percent{0.8211} & \percent{0.7024} & \percent{0.2415} & \percent{0.4898} & \percent{0.7439} & \percent{0.7603} & \percent{0.8257} \\
YOLOv8-L & \percent{0.7442} & \percent{0.8921} & \percent{0.8124} & \percent{0.3313} & \percent{0.5915} & \percent{0.8142} & \percent{0.8269} & \percent{0.8461} & \percent{0.6558} & \percent{0.8336} & \percent{0.7141} & \percent{0.2445} & \percent{0.5191} & \percent{0.7699} & \percent{0.7395} & \percent{0.7684} & \percent{0.6739} & \percent{0.8503} & \percent{0.7354} & \percent{0.2818} & \percent{0.5294} & \percent{0.7701} & \percent{0.7640} & \percent{0.7912} \\
Faster R-CNN (ResNet 50) & \percent{0.6125} & \percent{0.8438} & \percent{0.6918} & \percent{0.2552} & \percent{0.4960} & \percent{0.6740} & \percent{0.7287} & \percent{0.7451} & \percent{0.5352} & \percent{0.7606} & \percent{0.5937} & \percent{0.1702} & \percent{0.4228} & \percent{0.6436} & \percent{0.6498} & \percent{0.6739} & \percent{0.5557} & \percent{0.7865} & \percent{0.6180} & \percent{0.2106} & \percent{0.4393} & \percent{0.6475} & \percent{0.6747} & \percent{0.6971} \\
\bottomrule
\end{tabular}
}
\end{table}

%% file: tables/map_comparison_KITTI.tex
\begin{table}[htbp]
\setlength{\tabcolsep}{3pt}
\centering
\caption{Performance of object detectors on original vs.\ refined annotations on KITTI.}
 \label{tab:mAP_comparison_KITTI}
\resizebox{\linewidth}{!}{
\begin{tabular}{ccccccccc|cccccccc|cccccccc}
\toprule
{Model} & 
\multicolumn{8}{c}{Original Annotations} & 
\multicolumn{8}{c}{Refined Annotations ($p>0.5$)} & 
\multicolumn{8}{c}{Refined Annotations ($p>0.8$)} \\ 
\cmidrule(lr){2-9} \cmidrule(lr){10-17} \cmidrule(lr){18-25}
& AP\textsubscript{50..95} & AP\textsubscript{50} & AP\textsubscript{75} & AP\textsubscript{S} & AP\textsubscript{M} & AP\textsubscript{L} & AR\textsubscript{10} & AR\textsubscript{100}
& AP\textsubscript{50..95} & AP\textsubscript{50} & AP\textsubscript{75} & AP\textsubscript{S} & AP\textsubscript{M} & AP\textsubscript{L} & AR\textsubscript{10} & AR\textsubscript{100}
& AP\textsubscript{50..95} & AP\textsubscript{50} & AP\textsubscript{75} & AP\textsubscript{S} & AP\textsubscript{M} & AP\textsubscript{L} & AR\textsubscript{10} & AR\textsubscript{100} \\ 
\midrule
Grounding DINO & \percent{0.2918} & \percent{0.5133} & \percent{0.2951} & \percent{0.2141} & \percent{0.3069} & \percent{0.3620} & \percent{0.5067} & \percent{0.5626} & \percent{0.3449} & \percent{0.5188} & \percent{0.3834} & \percent{0.1684} & \percent{0.4342} & \percent{0.6740} & \percent{0.5417} & \percent{0.6133} & \percent{0.3942} & \percent{0.5726} & \percent{0.4481} & \percent{0.2052} & \percent{0.4352} & \percent{0.6748} & \percent{0.6168} & \percent{0.6824} \\
YOLOv8-World v2 & \percent{0.1986} & \percent{0.3361} & \percent{0.2101} & \percent{0.1883} & \percent{0.1872} & \percent{0.2469} & \percent{0.3171} & \percent{0.3354} & \percent{0.2127} & \percent{0.3015} & \percent{0.2266} & \percent{0.0771} & \percent{0.2705} & \percent{0.4516} & \percent{0.3165} & \percent{0.3487} & \percent{0.2444} & \percent{0.3429} & \percent{0.2622} & \percent{0.1054} & \percent{0.2716} & \percent{0.4436} & \percent{0.3768} & \percent{0.4136} \\
OWL-ViT & \percent{0.0928} & \percent{0.1996} & \percent{0.0744} & \percent{0.0134} & \percent{0.0682} & \percent{0.1918} & \percent{0.2226} & \percent{0.2439} & \percent{0.0837} & \percent{0.1668} & \percent{0.0695} & \percent{0.0141} & \percent{0.0850} & \percent{0.2734} & \percent{0.1870} & \percent{0.2027} & \percent{0.0955} & \percent{0.1910} & \percent{0.0801} & \percent{0.0180} & \percent{0.0851} & \percent{0.2707} & \percent{0.2196} & \percent{0.2387} \\
RT-DETR-L & \percent{0.7846} & \percent{0.9621} & \percent{0.8782} & \percent{0.7616} & \percent{0.7956} & \percent{0.8022} & \percent{0.8216} & \percent{0.8548} & \percent{0.2341} & \percent{0.4314} & \percent{0.2294} & \percent{0.0947} & \percent{0.3414} & \percent{0.4457} & \percent{0.3107} & \percent{0.3471} & \percent{0.2733} & \percent{0.5018} & \percent{0.2682} & \percent{0.1316} & \percent{0.3419} & \percent{0.4438} & \percent{0.3691} & \percent{0.4116} \\
YOLOv8-L & \percent{0.7468} & \percent{0.9351} & \percent{0.8421} & \percent{0.6748} & \percent{0.7671} & \percent{0.7754} & \percent{0.7834} & \percent{0.7923} & \percent{0.2287} & \percent{0.4264} & \percent{0.2177} & \percent{0.0913} & \percent{0.3349} & \percent{0.4201} & \percent{0.2959} & \percent{0.3013} & \percent{0.2666} & \percent{0.4962} & \percent{0.2545} & \percent{0.1293} & \percent{0.3344} & \percent{0.4188} & \percent{0.3538} & \percent{0.3602} \\
Faster R-CNN (ResNet 50) & \percent{0.6529} & \percent{0.8997} & \percent{0.7352} & \percent{0.6960} & \percent{0.6618} & \percent{0.6725} & \percent{0.7191} & \percent{0.7370} & \percent{0.2292} & \percent{0.4562} & \percent{0.2071} & \percent{0.1033} & \percent{0.3346} & \percent{0.4007} & \percent{0.3459} & \percent{0.3693} & \percent{0.2672} & \percent{0.5282} & \percent{0.2439} & \percent{0.1385} & \percent{0.3358} & \percent{0.3997} & \percent{0.4040} & \percent{0.4284} \\
\bottomrule
\end{tabular}
}
\end{table}

%% file: tables/map_comparison_Cityscapes.tex
\begin{table}[htbp]
\setlength{\tabcolsep}{3pt}
\centering
\caption{Performance of object detectors on original vs.\ refined annotations on Cityscapes.}
 \label{tab:mAP_comparison_Cityscapes}
\resizebox{\linewidth}{!}{
\begin{tabular}{ccccccccc|cccccccc|cccccccc}
\toprule
{Model} & 
\multicolumn{8}{c}{Original Annotations} & 
\multicolumn{8}{c}{Refined Annotations ($p>0.5$)} & 
\multicolumn{8}{c}{Refined Annotations ($p>0.8$)} \\ 
\cmidrule(lr){2-9} \cmidrule(lr){10-17} \cmidrule(lr){18-25}
& AP\textsubscript{50..95} & AP\textsubscript{50} & AP\textsubscript{75} & AP\textsubscript{S} & AP\textsubscript{M} & AP\textsubscript{L} & AR\textsubscript{10} & AR\textsubscript{100}
& AP\textsubscript{50..95} & AP\textsubscript{50} & AP\textsubscript{75} & AP\textsubscript{S} & AP\textsubscript{M} & AP\textsubscript{L} & AR\textsubscript{10} & AR\textsubscript{100}
& AP\textsubscript{50..95} & AP\textsubscript{50} & AP\textsubscript{75} & AP\textsubscript{S} & AP\textsubscript{M} & AP\textsubscript{L} & AR\textsubscript{10} & AR\textsubscript{100} \\ 
\midrule
Grounding DINO & \percent{0.3419} & \percent{0.5331} & \percent{0.3346} & \percent{0.0789} & \percent{0.2367} & \percent{0.5231} & \percent{0.5164} & \percent{0.5882} & \percent{0.4020} & \percent{0.5773} & \percent{0.4186} & \percent{0.1208} & \percent{0.2736} & \percent{0.5782} & \percent{0.5790} & \percent{0.6652} & \percent{0.4382} & \percent{0.6163} & \percent{0.4601} & \percent{0.1442} & \percent{0.2880} & \percent{0.5795} & \percent{0.6312} & \percent{0.7106} \\
YOLOv8-World v2 & \percent{0.2411} & \percent{0.3628} & \percent{0.2436} & \percent{0.0215} & \percent{0.1439} & \percent{0.4133} & \percent{0.3055} & \percent{0.3350} & \percent{0.2662} & \percent{0.3875} & \percent{0.2744} & \percent{0.0762} & \percent{0.1607} & \percent{0.4341} & \percent{0.3373} & \percent{0.3669} & \percent{0.3035} & \percent{0.4396} & \percent{0.3142} & \percent{0.0432} & \percent{0.1737} & \percent{0.4442} & \percent{0.3870} & \percent{0.4199} \\
OWL-ViT & \percent{0.1160} & \percent{0.2168} & \percent{0.1113} & \percent{0.0029} & \percent{0.0487} & \percent{0.2212} & \percent{0.1735} & \percent{0.1873} & \percent{0.1220} & \percent{0.2262} & \percent{0.1182} & \percent{0.0023} & \percent{0.0526} & \percent{0.2256} & \percent{0.1848} & \percent{0.1982} & \percent{0.1378} & \percent{0.2552} & \percent{0.1350} & \percent{0.0055} & \percent{0.0597} & \percent{0.2301} & \percent{0.2122} & \percent{0.2270} \\
RT-DETR-L & \percent{0.3822} & \percent{0.5979} & \percent{0.3848} & \percent{0.0574} & \percent{0.2754} & \percent{0.5705} & \percent{0.4230} & \percent{0.4768} & \percent{0.3906} & \percent{0.6055} & \percent{0.4093} & \percent{0.0911} & \percent{0.2731} & \percent{0.5770} & \percent{0.4486} & \percent{0.5013} & \percent{0.4248} & \percent{0.6421} & \percent{0.4476} & \percent{0.1177} & \percent{0.2883} & \percent{0.5780} & \percent{0.5007} & \percent{0.5542} \\
YOLOv8-L & \percent{0.3814} & \percent{0.5867} & \percent{0.3744} & \percent{0.0392} & \percent{0.2598} & \percent{0.6167} & \percent{0.4369} & \percent{0.4740} & \percent{0.4060} & \percent{0.6047} & \percent{0.4252} & \percent{0.0385} & \percent{0.2626} & \percent{0.6332} & \percent{0.4552} & \percent{0.4897} & \percent{0.4435} & \percent{0.6530} & \percent{0.4683} & \percent{0.0827} & \percent{0.2776} & \percent{0.6319} & \percent{0.5151} & \percent{0.5522} \\
Faster R-CNN (ResNet 50) & \percent{0.3591} & \percent{0.6151} & \percent{0.3413} & \percent{0.0873} & \percent{0.2724} & \percent{0.5355} & \percent{0.4489} & \percent{0.4927} & \percent{0.3792} & \percent{0.6239} & \percent{0.3882} & \percent{0.1021} & \percent{0.2819} & \percent{0.5532} & \percent{0.4661} & \percent{0.5054} & \percent{0.4102} & \percent{0.6599} & \percent{0.4280} & \percent{0.1312} & \percent{0.2969} & \percent{0.5529} & \percent{0.5136} & \percent{0.5508} \\
\bottomrule
\end{tabular}
}
\end{table}

%% file: tables/ECE_before_and_after_calibration.tex
\begin{table}[htbp]
\centering
\caption{ECE before and after calibration using Platt Scaling (PS) and Isotonic Regression (IR).}
\label{tab:calibration_results}
\begin{tabular}{llccc}
\toprule
Dataset & Model & ECE (Raw) & ECE (PS) & ECE (IR) \\
\midrule
\multirow{6}{*}{COCO}  & Grounding DINO~\cite{liu2023grounding} & \num{3.0e-02} & \num{4.4e-03} & \num{1.8e-04} \\
 & YOLOv8-World v2~\cite{cheng2024yoloworld} & \num{6.8e-02} & \num{2.1e-02} & \num{1.1e-03} \\
 & OWL-ViT~\cite{minderer2022owlvit} & \num{1.4e-02} & \num{2.5e-03} & \num{1.4e-04} \\
 & RT-DETR-L~\cite{lv2023detrs} & \num{1.0e-02} & \num{4.2e-03} & \num{1.7e-04} \\
 & YOLOv8-L~\cite{yolov8_ultralytics} & \num{9.8e-02} & \num{8.0e-03} & \num{1.7e-03} \\
 & Faster R-CNN (ResNet 50)~\cite{fasterRCNN} & \num{1.5e-02} & \num{3.9e-03} & \num{1.7e-04} \\
\midrule
\multirow{6}{*}{Cityscapes}  & Grounding DINO & \num{6.3e-02} & \num{9.6e-03} & \num{6.5e-04} \\
 & YOLOv8-World v2 & \num{5.5e-02} & \num{3.4e-02} & \num{2.9e-03} \\
 & OWL-ViT & \num{1.5e-02} & \num{3.1e-03} & \num{4.9e-04} \\
 & RT-DETR-L & \num{5.8e-02} & \num{3.3e-03} & \num{6.4e-04} \\
 & YOLOv8-L & \num{1.9e-02} & \num{9.9e-03} & \num{6.7e-04} \\
 & Faster R-CNN (ResNet 50) & \num{8.5e-02} & \num{2.4e-05} & \num{1.2e-05} \\
\midrule
\multirow{6}{*}{KITTI}  & Grounding DINO & \num{5.3e-02} & \num{7.2e-04} & \num{1.8e-04} \\
 & YOLOv8-World v2 & \num{7.2e-03} & \num{6.4e-03} & \num{6.1e-04} \\
 & OWL-ViT & \num{1.0e-02} & \num{1.1e-03} & \num{1.7e-04} \\
 & RT-DETR-L & \num{6.3e-03} & \num{3.1e-04} & \num{8.4e-05} \\
 & YOLOv8-L & \num{3.2e-02} & \num{9.4e-03} & \num{3.6e-03} \\
 & Faster R-CNN (ResNet 50) & \num{2.3e-02} & \num{7.3e-05} & \num{2.7e-06} \\
\midrule
\multirow{6}{*}{Pascal VOC}  & Grounding DINO & \num{2.6e-02} & \num{7.0e-04} & \num{3.7e-05} \\
 & YOLOv8-World v2 & \num{9.3e-03} & \num{3.1e-03} & \num{3.4e-04} \\
 & OWL-ViT & \num{8.3e-03} & \num{1.2e-03} & \num{5.7e-05} \\
 & RT-DETR-L & \num{3.8e-02} & \num{5.2e-04} & \num{5.3e-05} \\
 & YOLOv8-L & \num{7.3e-03} & \num{3.2e-03} & \num{4.4e-04} \\
 & Faster R-CNN (ResNet 50) & \num{1.6e-02} & \num{1.2e-05} & \num{6.1e-07} \\
\bottomrule
\end{tabular}
\end{table}

%% file: tables/soft_label_evaluation_coco.tex
\begin{table}[htbp]
\centering
\caption{Soft label-based evaluation of object detector performance on COCO.}
\label{tab:soft_label_based_benchmark_coco}
\resizebox{\linewidth}{!}{
\begin{tabular}{ccccccccccc}
\toprule
{Model} 
& AUSRC\textsubscript{50..95}\,$\uparrow$ & AUSRC\textsubscript{50} $\uparrow$ & AUSRC\textsubscript{75} $\uparrow$ & AUSRC\textsubscript{S} $\uparrow$ & AUSRC\textsubscript{M} $\uparrow$ & AUSRC\textsubscript{L} $\uparrow$ & TVD\,$\downarrow$ & TVD\textsubscript{TP} $\downarrow$  & $\sqrt{\mathrm{JS}}\,\downarrow$ & $\sqrt{\mathrm{JS}}$\textsubscript{TP} $\downarrow$ \\
\midrule
Grounding DINO & \percent{0.5508} & \percent{0.7905} & \percent{0.5801} & \percent{0.3842} & \percent{0.6033} & \percent{0.7875} & \percent{0.1198} & \percent{0.3823} & \percent{0.1636} & \percent{0.3495} \\
YOLOv8-World v2 & \percent{0.4808} & \percent{0.7006} & \percent{0.5015} & \percent{0.3060} & \percent{0.5602} & \percent{0.7032} & \percent{0.1641} & \percent{0.3758} & \percent{0.2230} & \percent{0.3570} \\
OWL-ViT & \percent{0.3583} & \percent{0.5865} & \percent{0.3543} & \percent{0.1402} & \percent{0.4264} & \percent{0.6781} & \percent{0.1386} & \percent{0.7524} & \percent{0.1845} & \percent{0.6305} \\
RT-DETR-L & \percent{0.4910} & \percent{0.7325} & \percent{0.5084} & \percent{0.3226} & \percent{0.5360} & \percent{0.7454} & \percent{0.1406} & \percent{0.4688} & \percent{0.1934} & \percent{0.4397} \\
YOLOv8-L & \percent{0.4664} & \percent{0.6906} & \percent{0.4815} & \percent{0.3032} & \percent{0.5342} & \percent{0.6779} & \percent{0.1756} & \percent{0.3312} & \percent{0.2374} & \percent{0.3170} \\
Faster R-CNN (ResNet 50) & \percent{0.4395} & \percent{0.6734} & \percent{0.4508} & \percent{0.2965} & \percent{0.5073} & \percent{0.6129} & \percent{0.1815} & \percent{0.2468} & \percent{0.2190} & \percent{0.2552} \\
\bottomrule
\end{tabular}
}
\end{table}

%% file: tables/soft_label_evaluation_VOC.tex
\begin{table}[htbp]
\centering
\caption{Soft label-based evaluation of object detector performance on Pascal VOC.}
\label{tab:soft_label_based_benchmark_VOC}
\resizebox{\linewidth}{!}{
\begin{tabular}{ccccccccccc}
\toprule
{Model} 
& AUSRC\textsubscript{50..95}\,$\uparrow$ & AUSRC\textsubscript{50} $\uparrow$ & AUSRC\textsubscript{75} $\uparrow$ & AUSRC\textsubscript{S} $\uparrow$ & AUSRC\textsubscript{M} $\uparrow$ & AUSRC\textsubscript{L} $\uparrow$ & TVD\,$\downarrow$ & TVD\textsubscript{TP} $\downarrow$  & $\sqrt{\mathrm{JS}}\,\downarrow$ & $\sqrt{\mathrm{JS}}$\textsubscript{TP} $\downarrow$ \\
\midrule
Grounding DINO & \percent{0.8173} & \percent{0.9744} & \percent{0.8768} & \percent{0.5822} & \percent{0.7603} & \percent{0.9022} & \percent{0.0169} & \percent{0.2417} & \percent{0.0217} & \percent{0.2484} \\
YOLOv8-World v2 & \percent{0.7520} & \percent{0.9360} & \percent{0.8126} & \percent{0.5131} & \percent{0.6950} & \percent{0.8247} & \percent{0.0442} & \percent{0.3481} & \percent{0.0695} & \percent{0.3429} \\
OWL-ViT & \percent{0.6929} & \percent{0.9224} & \percent{0.7460} & \percent{0.3311} & \percent{0.5997} & \percent{0.8370} & \percent{0.0086} & \percent{0.5354} & \percent{0.0192} & \percent{0.4802} \\
RT-DETR-L & \percent{0.7879} & \percent{0.9556} & \percent{0.8372} & \percent{0.4936} & \percent{0.7127} & \percent{0.9051} & \percent{0.0057} & \percent{0.3601} & \percent{0.0146} & \percent{0.3649} \\
YOLOv8-L & \percent{0.7225} & \percent{0.9056} & \percent{0.7827} & \percent{0.4588} & \percent{0.6654} & \percent{0.8209} & \percent{0.0592} & \percent{0.2475} & \percent{0.0844} & \percent{0.2600} \\
Faster R-CNN (ResNet 50) & \percent{0.3997} & \percent{0.5105} & \percent{0.4454} & \percent{0.3561} & \percent{0.3714} & \percent{0.3148} & \percent{0.4896} & \percent{0.7115} & \percent{0.4359} & \percent{0.5969} \\
\bottomrule
\end{tabular}
}
\end{table}

%% file: tables/soft_label_evaluation_KITTI.tex
\begin{table}[htbp]
\centering
\caption{Soft label-based evaluation of object detector performance on KITTI.}
\label{tab:soft_label_based_benchmark_KITTI}
\resizebox{\linewidth}{!}{
\begin{tabular}{ccccccccccc}
\toprule
{Model} 
&  AUSRC\textsubscript{50..95}\,$\uparrow$ & AUSRC\textsubscript{50} $\uparrow$ & AUSRC\textsubscript{75} $\uparrow$ & AUSRC\textsubscript{S} $\uparrow$ & AUSRC\textsubscript{M} $\uparrow$ & AUSRC\textsubscript{L} $\uparrow$ & TVD\,$\downarrow$ & TVD\textsubscript{TP} $\downarrow$  & $\sqrt{\mathrm{JS}}\,\downarrow$ & $\sqrt{\mathrm{JS}}$\textsubscript{TP} $\downarrow$ \\
\midrule
Grounding DINO & \percent{0.6054} & \percent{0.8769} & \percent{0.6335} & \percent{0.4450} & \percent{0.7328} & \percent{0.8821} & \percent{0.0423} & \percent{0.4025} & \percent{0.0681} & \percent{0.3836} \\
YOLOv8-World v2 & \percent{0.4970} & \percent{0.7072} & \percent{0.5178} & \percent{0.2844} & \percent{0.6874} & \percent{0.7911} & \percent{0.0739} & \percent{0.4961} & \percent{0.1089} & \percent{0.4504} \\
OWL-ViT & \percent{0.2911} & \percent{0.5617} & \percent{0.2624} & \percent{0.0984} & \percent{0.4047} & \percent{0.7532} & \percent{0.0232} & \percent{0.6799} & \percent{0.0469} & \percent{0.6000} \\
RT-DETR-L & \percent{0.4051} & \percent{0.6192} & \percent{0.4508} & \percent{0.1966} & \percent{0.5811} & \percent{0.7586} & \percent{0.0110} & \percent{0.2838} & \percent{0.0149} & \percent{0.2758} \\
YOLOv8-L & \percent{0.2854} & \percent{0.4793} & \percent{0.2964} & \percent{0.1295} & \percent{0.4257} & \percent{0.5163} & \percent{0.2788} & \percent{0.0649} & \percent{0.2638} & \percent{0.0862} \\
Faster R-CNN (ResNet 50) & \percent{0.3276} & \percent{0.5036} & \percent{0.3579} & \percent{0.1750} & \percent{0.4582} & \percent{0.5881} & \percent{0.2310} & \percent{0.3462} & \percent{0.2377} & \percent{0.3267} \\
\bottomrule
\end{tabular}
}
\end{table}

%% file: tables/soft_label_evaluation_Cityscapes.tex
\begin{table}[htbp]
\centering
\caption{Soft label-based evaluation of object detector performance on Cityscapes.}
\label{tab:soft_label_based_benchmark_Cityscapes}
\resizebox{\linewidth}{!}{
\begin{tabular}{ccccccccccc}
\toprule
{Model} 
& AUSRC\textsubscript{50..95}\,$\uparrow$ & AUSRC\textsubscript{50} $\uparrow$ & AUSRC\textsubscript{75} $\uparrow$ & AUSRC\textsubscript{S} $\uparrow$ & AUSRC\textsubscript{M} $\uparrow$ & AUSRC\textsubscript{L} $\uparrow$ & TVD\,$\downarrow$ & TVD\textsubscript{TP} $\downarrow$  & $\sqrt{\mathrm{JS}}\,\downarrow$ & $\sqrt{\mathrm{JS}}$\textsubscript{TP} $\downarrow$\\
\midrule
Grounding DINO & \percent{0.5486} & \percent{0.7939} & \percent{0.5713} & \percent{0.3162} & \percent{0.5440} & \percent{0.7273} & \percent{0.1237} & \percent{0.3581} & \percent{0.1529} & \percent{0.3623} \\
YOLOv8-World v2 & \percent{0.4106} & \percent{0.6252} & \percent{0.4221} & \percent{0.0852} & \percent{0.3998} & \percent{0.7100} & \percent{0.1190} & \percent{0.3823} & \percent{0.1660} & \percent{0.3717} \\
OWL-ViT & \percent{0.2419} & \percent{0.4883} & \percent{0.2116} & \percent{0.0195} & \percent{0.1775} & \percent{0.5574} & \percent{0.0241} & \percent{0.5469} & \percent{0.0483} & \percent{0.4973} \\
RT-DETR-L & \percent{0.5297} & \percent{0.8434} & \percent{0.5297} & \percent{0.2574} & \percent{0.5331} & \percent{0.7545} & \percent{0.0549} & \percent{0.3308} & \percent{0.0857} & \percent{0.3125} \\
YOLOv8-L & \percent{0.4552} & \percent{0.7243} & \percent{0.4666} & \percent{0.1341} & \percent{0.4490} & \percent{0.7383} & \percent{0.0822} & \percent{0.1976} & \percent{0.1257} & \percent{0.2157} \\
Faster R-CNN (ResNet 50) & \percent{0.1273} & \percent{0.1987} & \percent{0.1338} & \percent{0.0641} & \percent{0.1184} & \percent{0.2035} & \percent{0.7302} & \percent{0.5867} & \percent{0.6026} & \percent{0.5085} \\
\bottomrule
\end{tabular}
}
\end{table}

%% file: tables/soft_label_training_results.tex
\begin{table}[tbhp]
\setlength{\tabcolsep}{2.5pt}
\centering
\caption{Performance and calibration of models trained on hard vs.\ soft labels ($\ast$). Values in \%.}
\label{tab:training_comparison}
\resizebox{\linewidth}{!}{
\begin{tabular}{lccccc|ccccc|ccccc|ccccc}
\toprule
 & 
\multicolumn{5}{c}{COCO} & 
\multicolumn{5}{c}{Pascal VOC} & 
\multicolumn{5}{c}{KITTI} & 
\multicolumn{5}{c}{Cityscapes} \\ 
\cmidrule(lr){2-6} \cmidrule(lr){7-11} \cmidrule(lr){12-16} \cmidrule(lr){17-21}
{Model} & AP\,$\uparrow$ & AS\,$\uparrow$ & ECE\,$\downarrow$ & TVD\,$\downarrow$ & $\sqrt{\mathrm{JS}}$\,$\downarrow$
& AP\,$\uparrow$ & AS\,$\uparrow$ & ECE\,$\downarrow$ & TVD\,$\downarrow$ & $\sqrt{\mathrm{JS}}$\,$\downarrow$
& AP\,$\uparrow$ & AS\,$\uparrow$ & ECE\,$\downarrow$ & TVD\,$\downarrow$ & $\sqrt{\mathrm{JS}}$\,$\downarrow$
& AP\,$\uparrow$ & AS\,$\uparrow$ & ECE\,$\downarrow$ & TVD\,$\downarrow$ & $\sqrt{\mathrm{JS}}$\,$\downarrow$
\\
\midrule
RT-DETR-L & \percent{0.4868} & \percent{0.5270} & \percent{0.0302} & \percent{0.1020} & \percent{0.1567} & \percent{0.6323} & \percent{0.8006} & \percent{0.0348} & \percent{0.0058} & \percent{0.0139} & \percent{0.5618} & \percent{0.7216} & \percent{0.0222} & \percent{0.0131} & \percent{0.0307} & \percent{0.4052} & \percent{0.5313} & \percent{0.0558} & \percent{0.0519} & \percent{0.0826} \\
RT-DETR-L$^\ast$ & \percent{0.4903} & \percent{0.5291} & \percent{0.0245} & \percent{0.0996} & \percent{0.1536} & \percent{0.6380} & \percent{0.8006} & \percent{0.0330} & \percent{0.0055} & \percent{0.0139} & \percent{0.5681} & \percent{0.7196} & \percent{0.0172} & \percent{0.0126} & \percent{0.0303} & \percent{0.4027} & \percent{0.5271} & \percent{0.0335} & \percent{0.0609} & \percent{0.0950} \\
\addlinespace[0.25em]
YOLOv8-L & \percent{0.5024} & \percent{0.5204} & \percent{0.0666} & \percent{0.1415} & \percent{0.2053} & \percent{0.6665} & \percent{0.7556} & \percent{0.0135} & \percent{0.0385} & \percent{0.0593} & \percent{0.4998} & \percent{0.5650} & \percent{0.0467} & \percent{0.0992} & \percent{0.1371} & \percent{0.4001} & \percent{0.4510} & \percent{0.0154} & \percent{0.0859} & \percent{0.1264} \\
YOLOv8-L$^\ast$ & \percent{0.5016} & \percent{0.5185} & \percent{0.0728} & \percent{0.1488} & \percent{0.2124} & \percent{0.6678} & \percent{0.7534} & \percent{0.0124} & \percent{0.0408} & \percent{0.0620} & \percent{0.4986} & \percent{0.5626} & \percent{0.0438} & \percent{0.0991} & \percent{0.1369} & \percent{0.3988} & \percent{0.4458} & \percent{0.0176} & \percent{0.0881} & \percent{0.1305} \\
\bottomrule
\end{tabular}
}
\end{table}

%% file: imgs/visual_examples/overconfident_gt.tex
\begin{figure*}[!t]
    \centering
    \begin{subfigure}[b]{0.19\linewidth}
        \includegraphics[width=\linewidth]{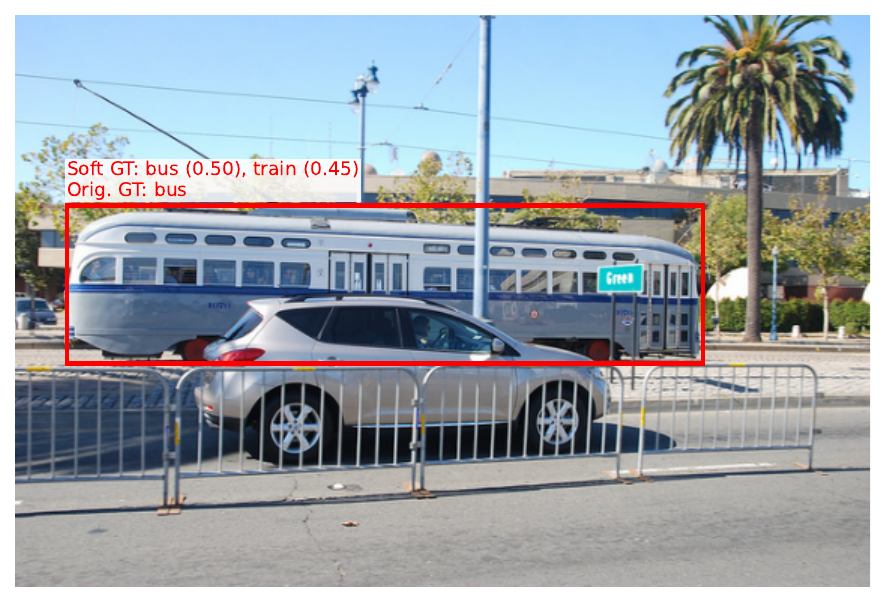}
        \caption{\tiny COCO - 000000086220}
        \label{fig:overconfident-gt-0}
    \end{subfigure}
    \hfill
    \begin{subfigure}[b]{0.19\linewidth}
        \includegraphics[width=\linewidth]{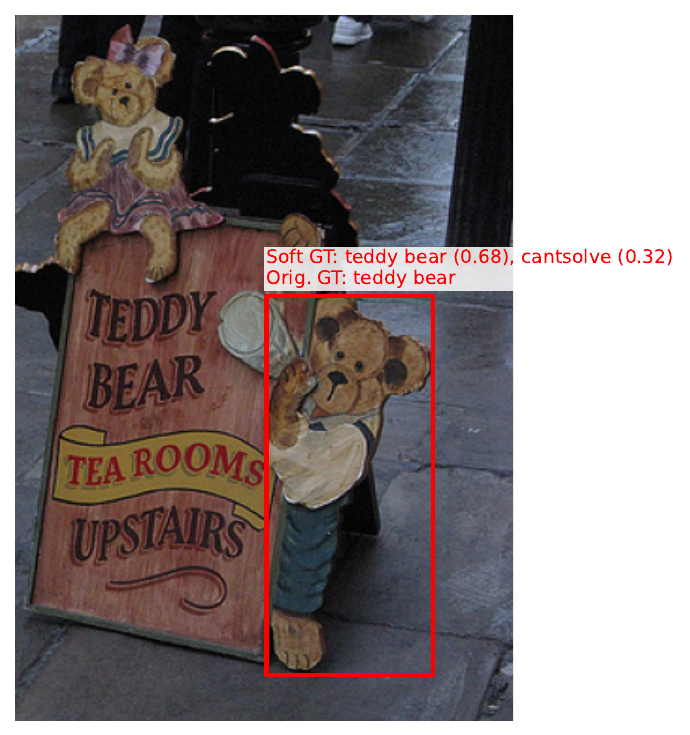}
        \caption{\tiny COCO - 000000162035}
        \label{fig:overconfident-gt-1}
    \end{subfigure}
    \hfill
    \begin{subfigure}[b]{0.19\linewidth}
        \includegraphics[width=\linewidth]{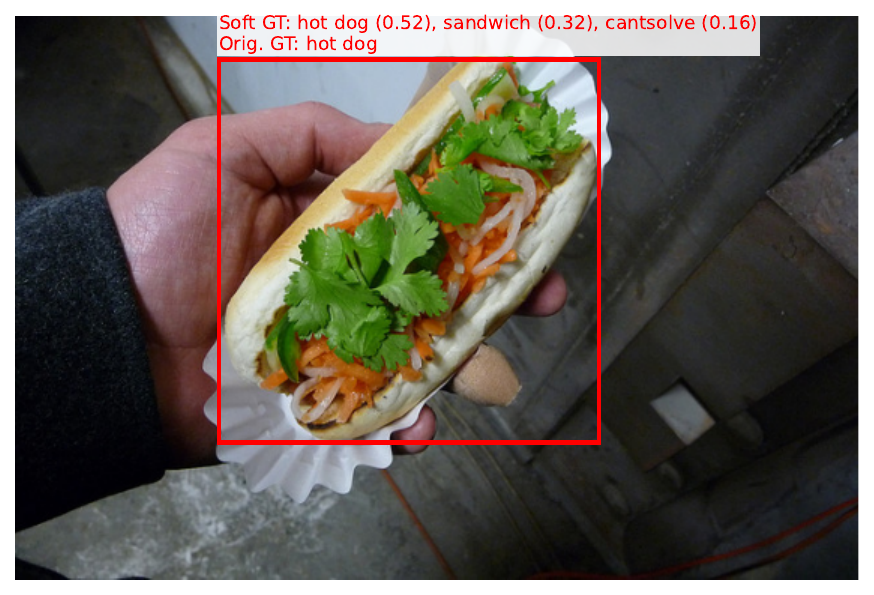}
        \caption{\tiny COCO - 000000118367}
        \label{fig:overconfident-gt-2}
    \end{subfigure}
    \hfill
    \begin{subfigure}[b]{0.19\linewidth}
        \includegraphics[width=\linewidth]{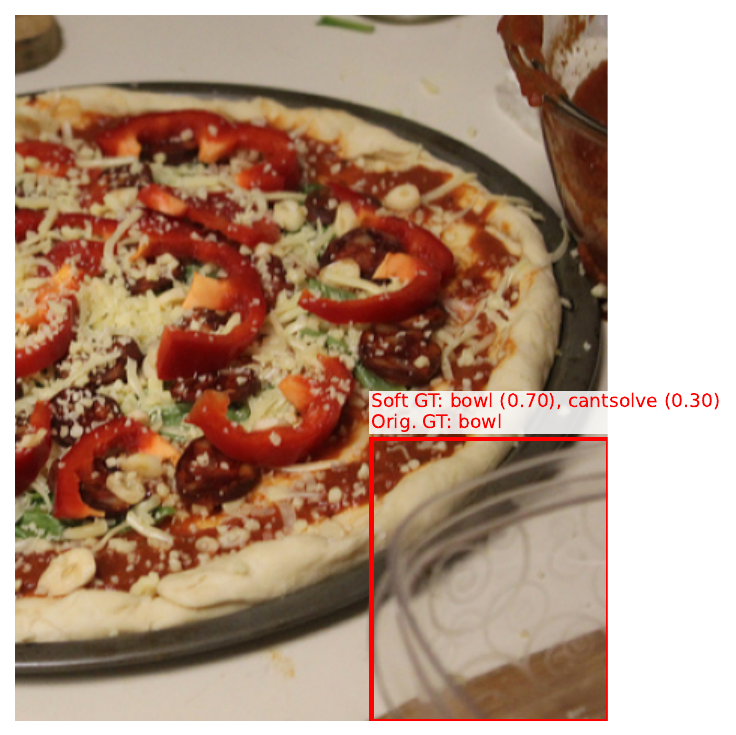}
        \caption{\tiny COCO - 000000360393}
        \label{fig:overconfident-gt-3}
    \end{subfigure}
    \hfill
    \begin{subfigure}[b]{0.19\linewidth}
        \includegraphics[width=\linewidth]{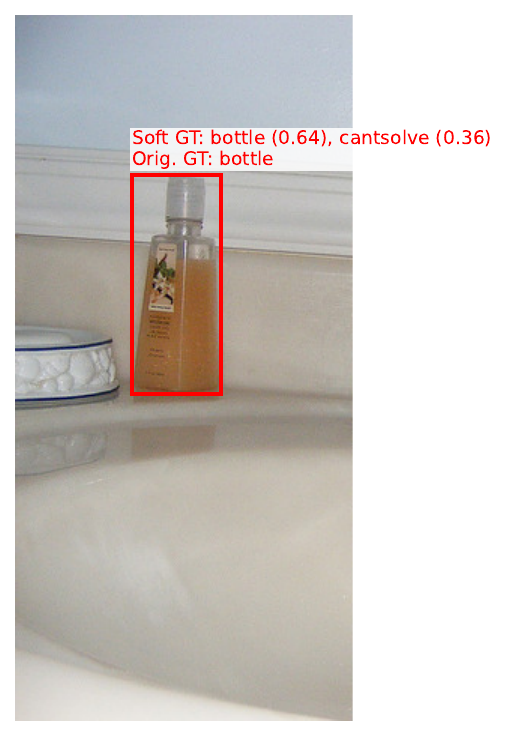}
        \caption{\tiny COCO - 000000047121}
        \label{fig:overconfident-gt-4}
    \end{subfigure}
    \vspace{0.5em}
    \begin{subfigure}[b]{0.19\linewidth}
        \includegraphics[width=\linewidth]{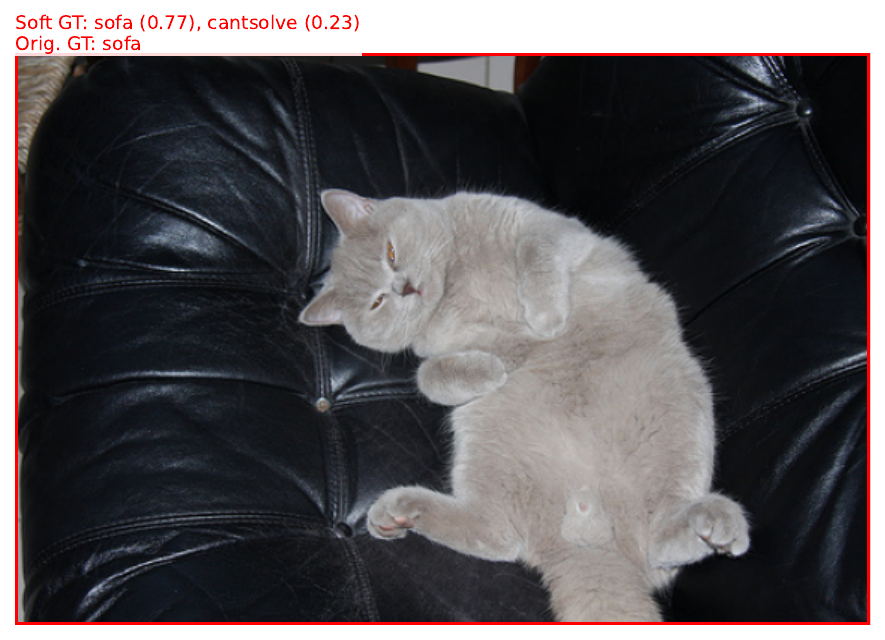}
        \caption{\tiny Pascal VOC - 2009\_004291}
        \label{fig:overconfident-gt-5}
    \end{subfigure}
    \hfill
    \begin{subfigure}[b]{0.19\linewidth}
        \includegraphics[width=\linewidth]{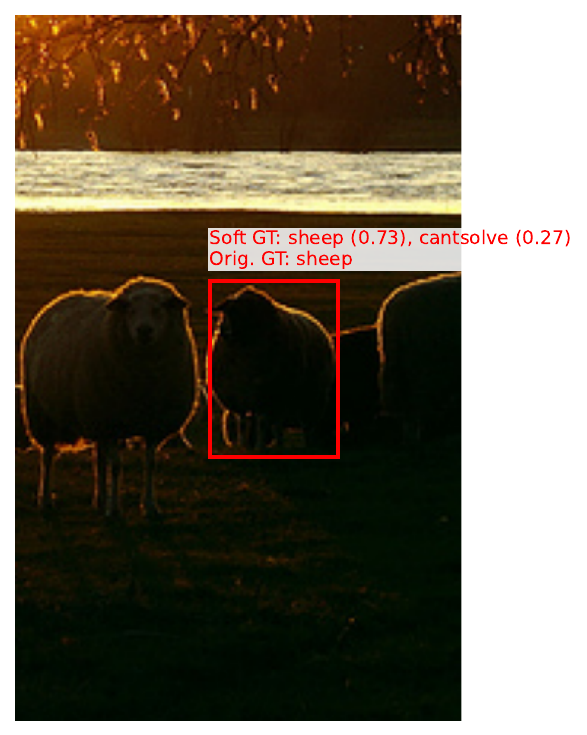}
        \caption{\tiny Pascal VOC - 2010\_001257}
        \label{fig:overconfident-gt-6}
    \end{subfigure}
    \hfill
    \begin{subfigure}[b]{0.19\linewidth}
        \includegraphics[width=\linewidth]{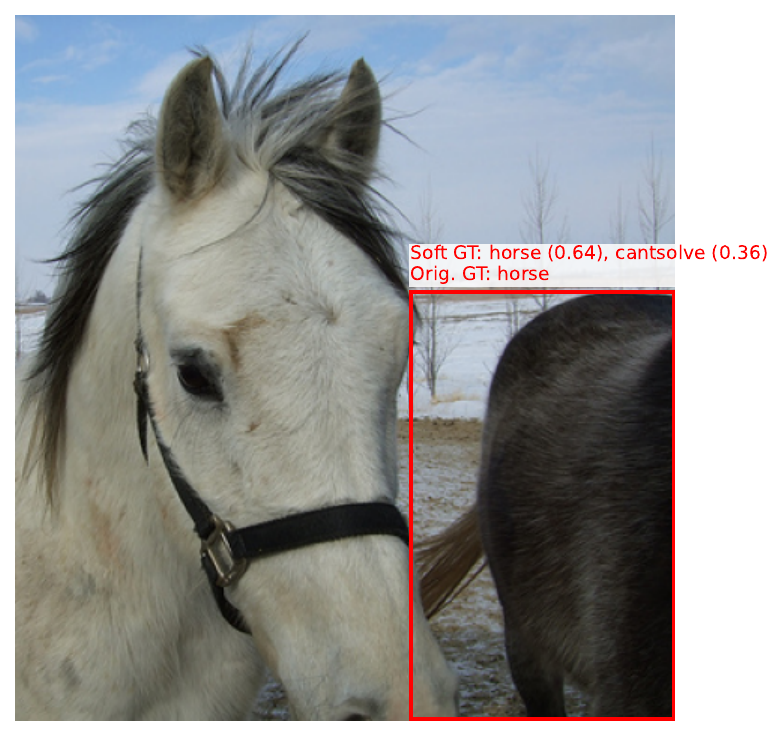}
        \caption{\tiny Pascal VOC - 2009\_004497}
        \label{fig:overconfident-gt-7}
    \end{subfigure}
    \hfill
    \begin{subfigure}[b]{0.19\linewidth}
        \includegraphics[width=\linewidth]{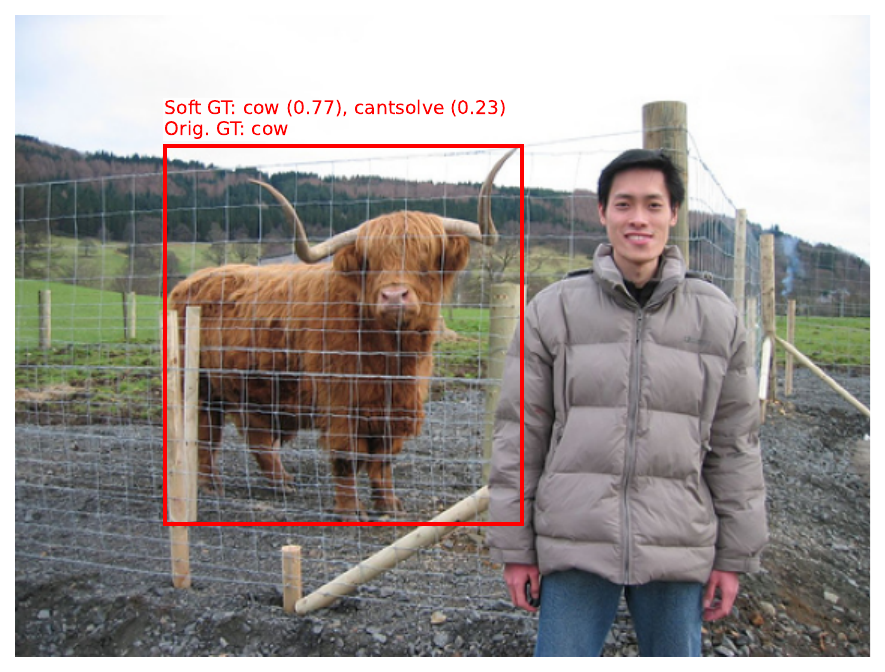}
        \caption{\tiny Pascal VOC - 2008\_007586}
        \label{fig:overconfident-gt-8}
    \end{subfigure}
    \hfill
    \begin{subfigure}[b]{0.19\linewidth}
        \includegraphics[width=\linewidth]{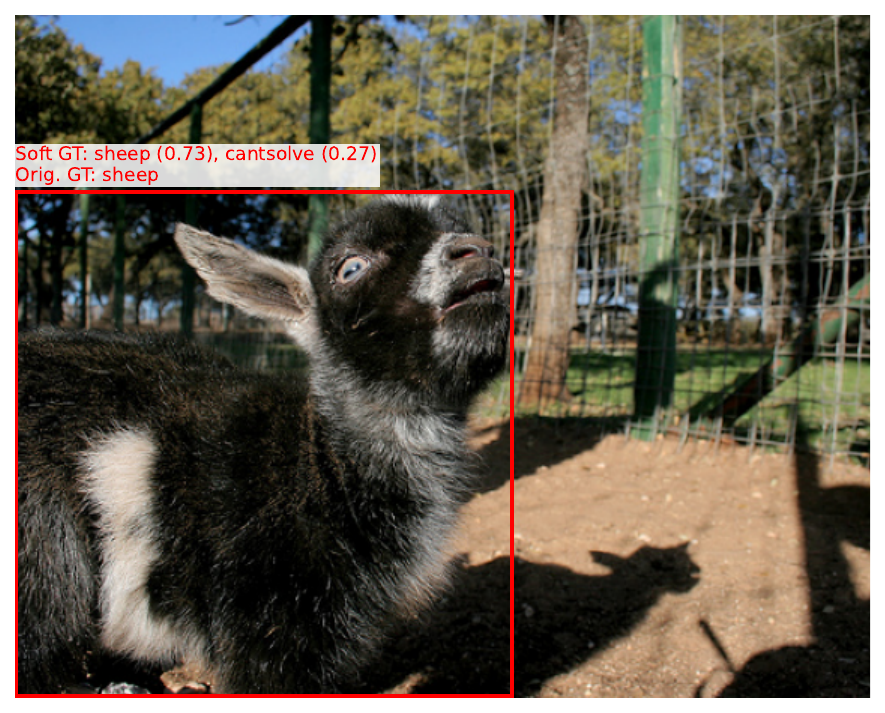}
        \caption{\tiny Pascal VOC - 2008\_006477}
        \label{fig:overconfident-gt-9}
    \end{subfigure}
    \vspace{0.5em}
    \begin{subfigure}[b]{0.19\linewidth}
        \includegraphics[width=\linewidth]{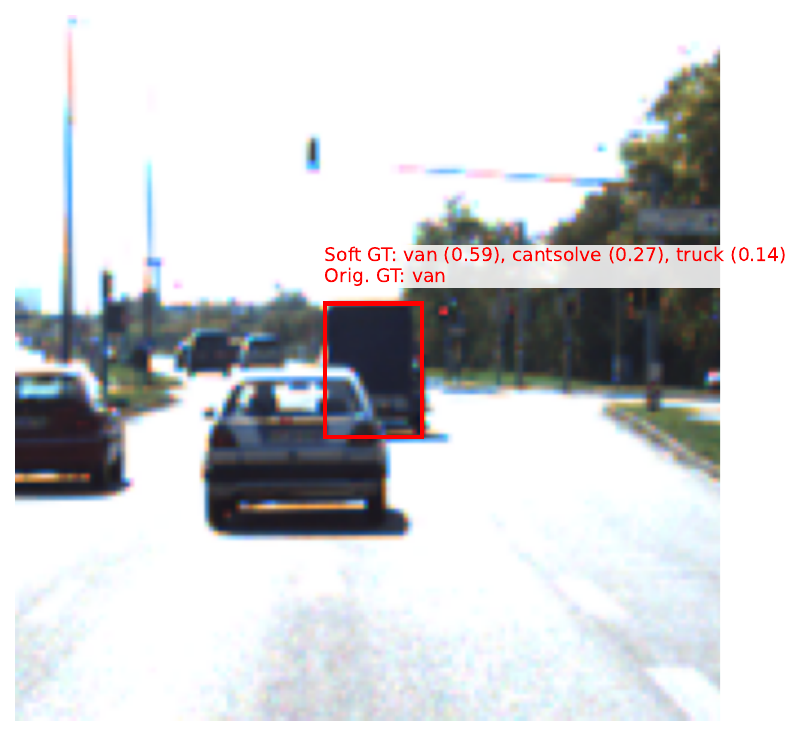}
        \caption{\tiny KITTI - 002924}
        \label{fig:overconfident-gt-10}
    \end{subfigure}
    \hfill
    \begin{subfigure}[b]{0.19\linewidth}
        \includegraphics[width=\linewidth]{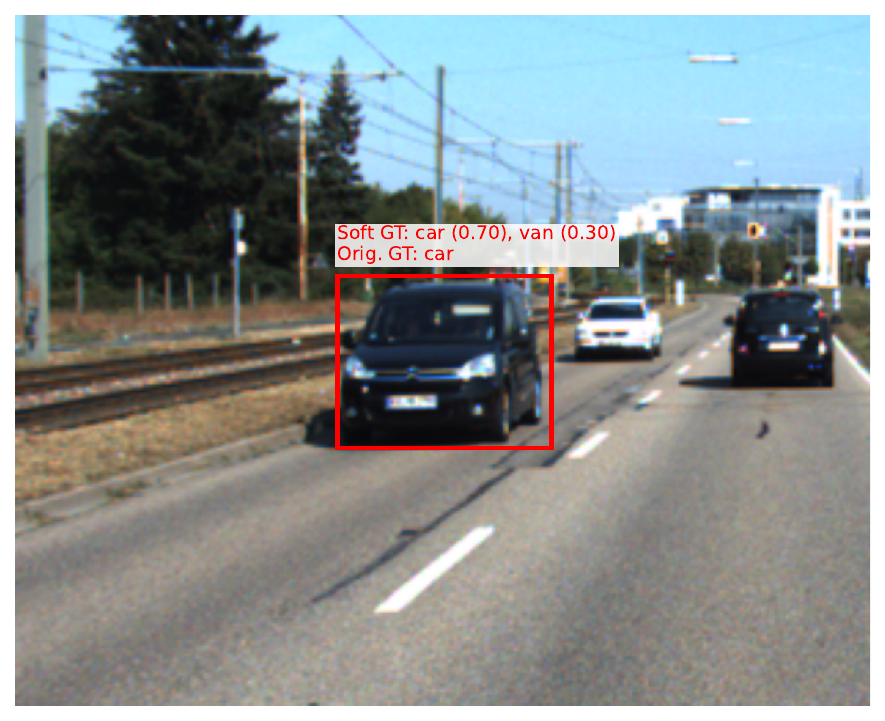}
        \caption{\tiny KITTI - 003278}
        \label{fig:overconfident-gt-11}
    \end{subfigure}
    \hfill
    \begin{subfigure}[b]{0.19\linewidth}
        \includegraphics[width=\linewidth]{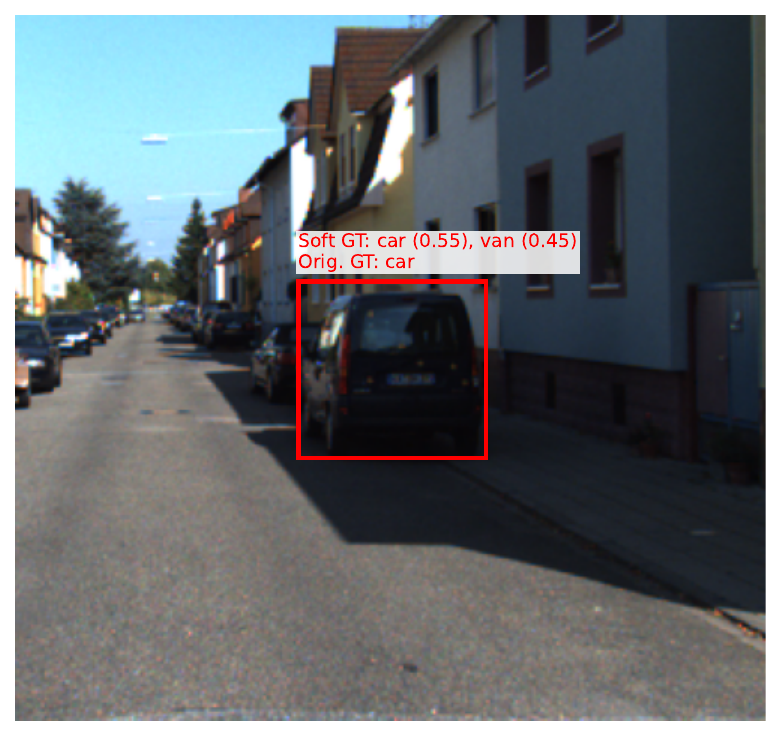}
        \caption{\tiny KITTI - 000403}
        \label{fig:overconfident-gt-12}
    \end{subfigure}
    \hfill
    \begin{subfigure}[b]{0.19\linewidth}
        \includegraphics[width=\linewidth]{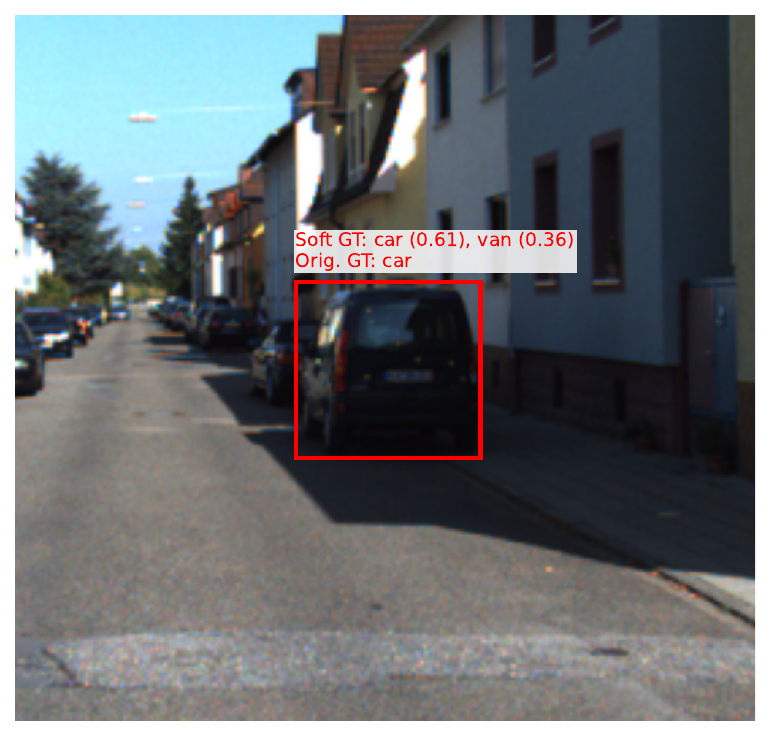}
        \caption{\tiny KITTI - 004598}
        \label{fig:overconfident-gt-13}
    \end{subfigure}
    \hfill
    \begin{subfigure}[b]{0.19\linewidth}
        \includegraphics[width=\linewidth]{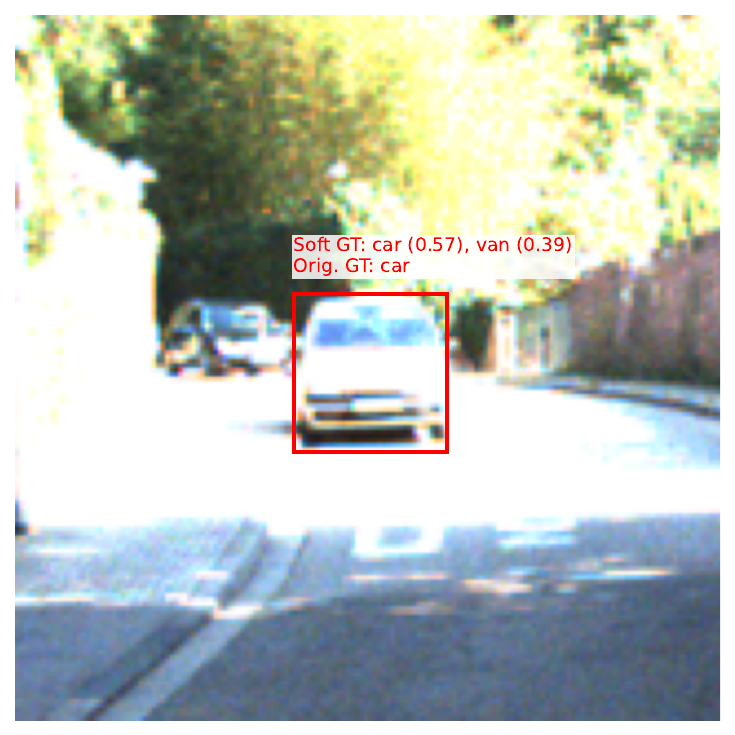}
        \caption{\tiny KITTI - 002941}
        \label{fig:overconfident-gt-14}
    \end{subfigure}
    \vspace{0.5em}
    \begin{subfigure}[b]{0.19\linewidth}
        \includegraphics[width=\linewidth]{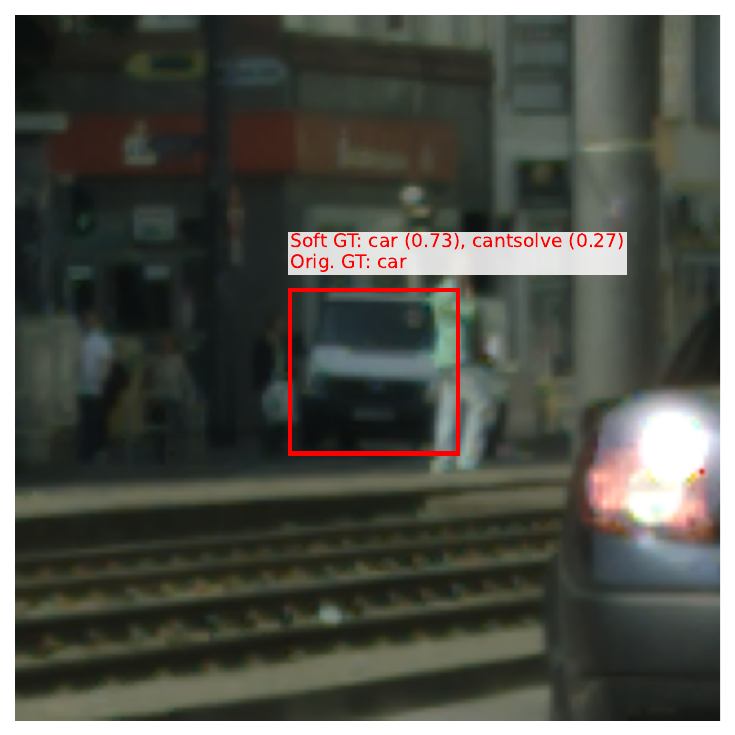}
        \caption{\tiny Cityscapes - frankfurt\_000001\_038645}
        \label{fig:overconfident-gt-15}
    \end{subfigure}
    \hfill
    \begin{subfigure}[b]{0.19\linewidth}
        \includegraphics[width=\linewidth]{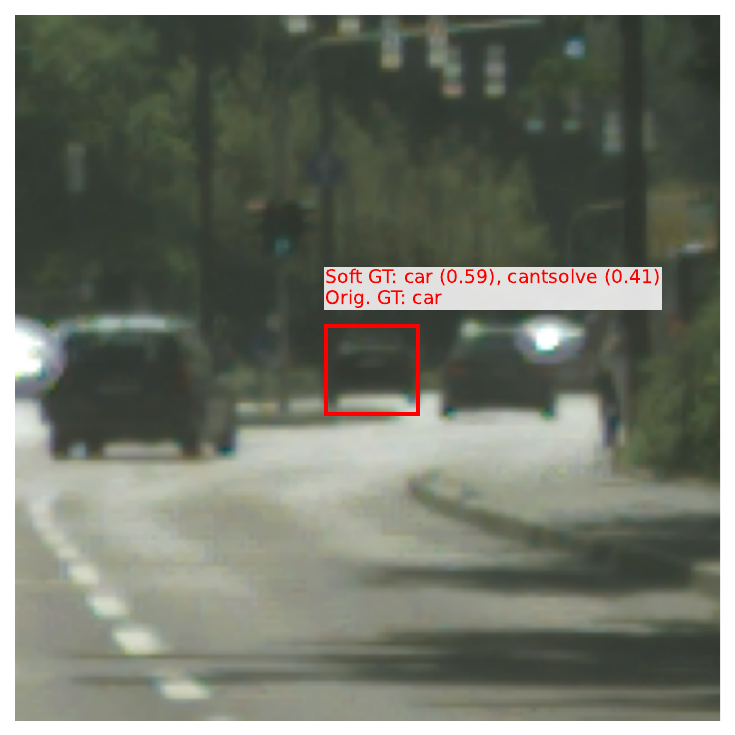}
        \caption{\tiny Cityscapes - frankfurt\_000001\_049078}
        \label{fig:overconfident-gt-16}
    \end{subfigure}
    \hfill
    \begin{subfigure}[b]{0.19\linewidth}
        \includegraphics[width=\linewidth]{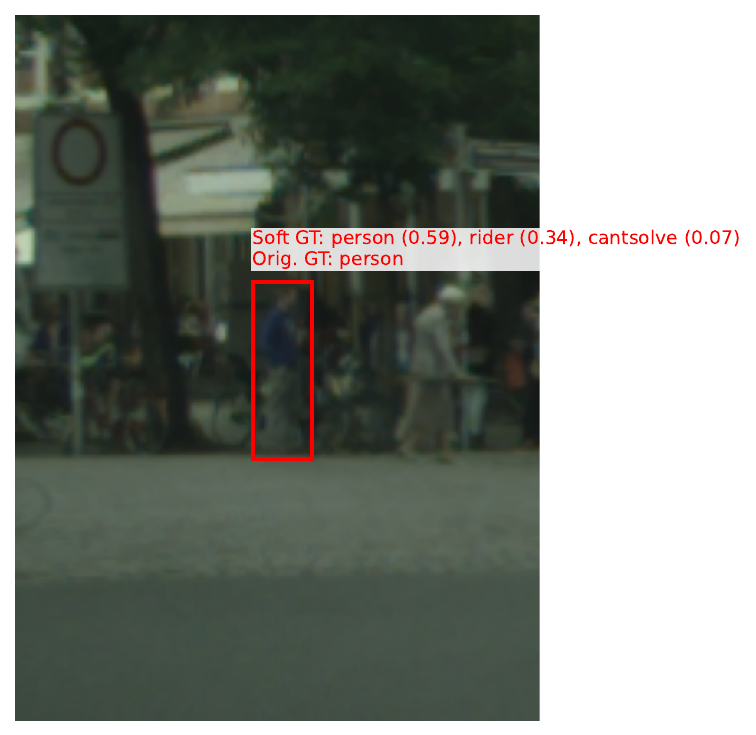}
        \caption{\tiny Cityscapes - munster\_000059\_000019}
        \label{fig:overconfident-gt-17}
    \end{subfigure}
    \hfill
    \begin{subfigure}[b]{0.19\linewidth}
        \includegraphics[width=\linewidth]{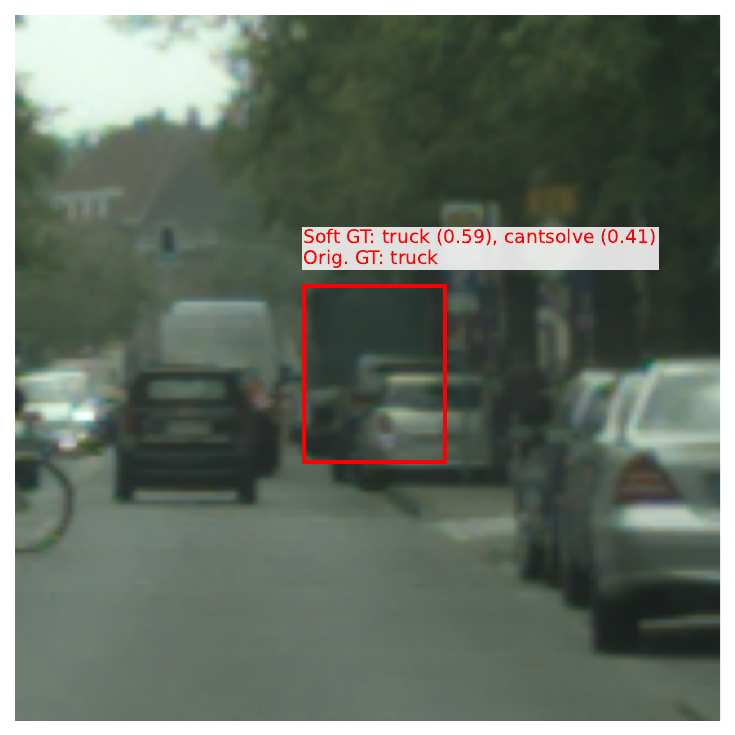}
        \caption{\tiny Cityscapes - munster\_000011\_000019}
        \label{fig:overconfident-gt-18}
    \end{subfigure}
    \hfill
    \begin{subfigure}[b]{0.19\linewidth}
        \includegraphics[width=\linewidth]{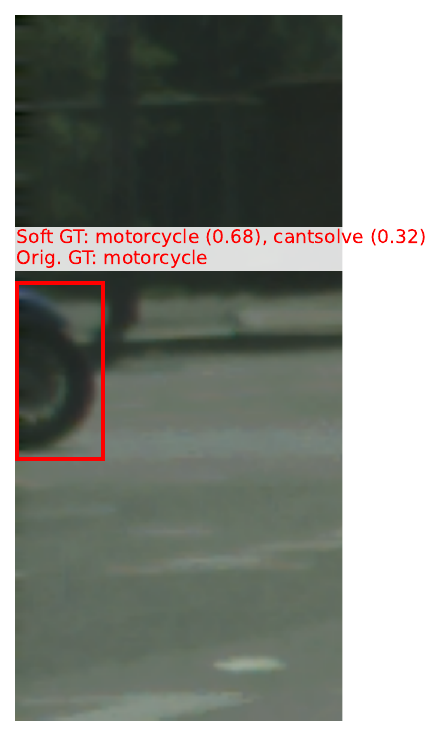}
        \caption{\tiny Cityscapes - frankfurt\_000001\_060906}
        \label{fig:overconfident-gt-19}
    \end{subfigure}
    \caption{Overconfident annotations in the original ground truth. In each case the original hard label was assigned with implicit confidence 1.0, but our crowd-sourced soft-labels indicate ambiguity for these objects.}
    \label{fig:visual-overconfident-gt}
\end{figure*}

%% file: imgs/visual_examples/model_confidence_discrepancy.tex
\begin{figure*}[t]
    \centering
    \begin{subfigure}[b]{0.19\linewidth}
        \includegraphics[width=\linewidth]{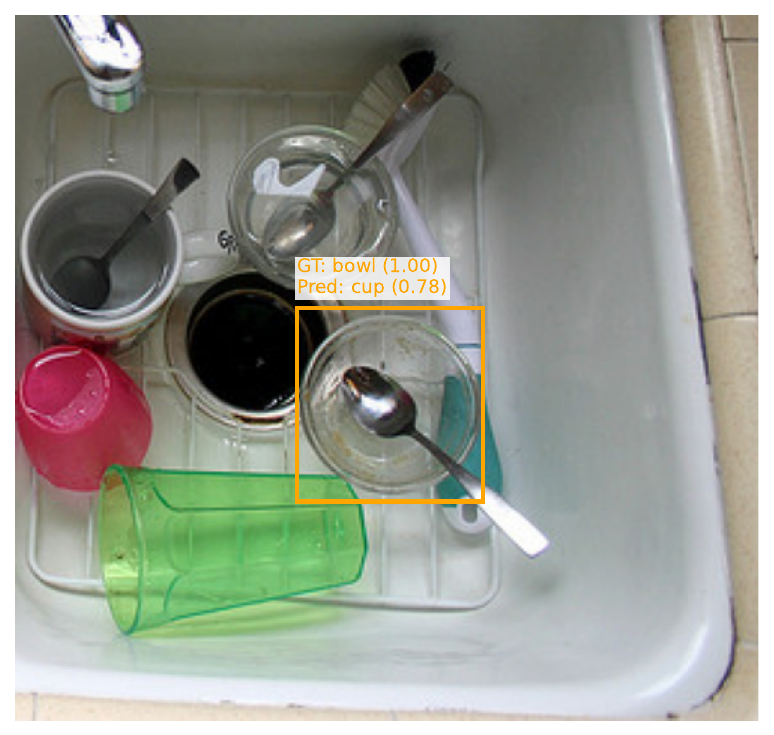}
        \caption{\tiny COCO - 000000239627}
        \label{fig:model-gt-discrepancy-0}
    \end{subfigure}
    \hfill
    \begin{subfigure}[b]{0.19\linewidth}
        \includegraphics[width=\linewidth]{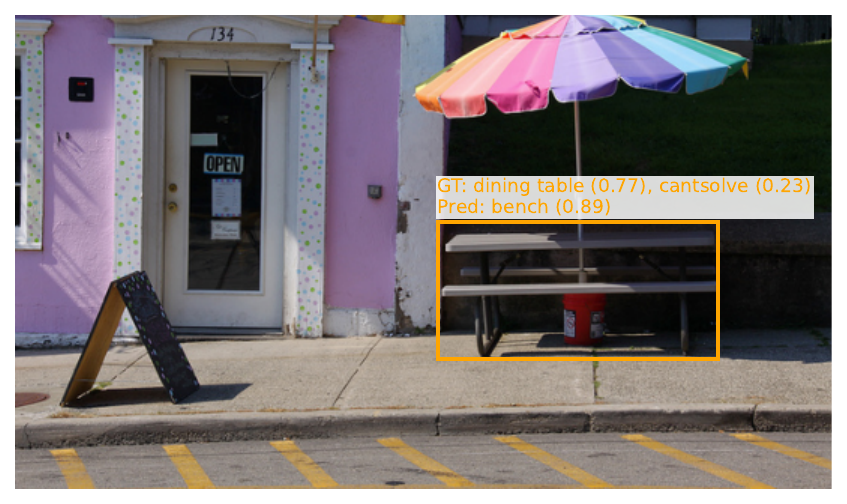}
        \caption{\tiny COCO - 000000203580}
        \label{fig:model-gt-discrepancy-1}
    \end{subfigure}
    \hfill
    \begin{subfigure}[b]{0.19\linewidth}
        \includegraphics[width=\linewidth]{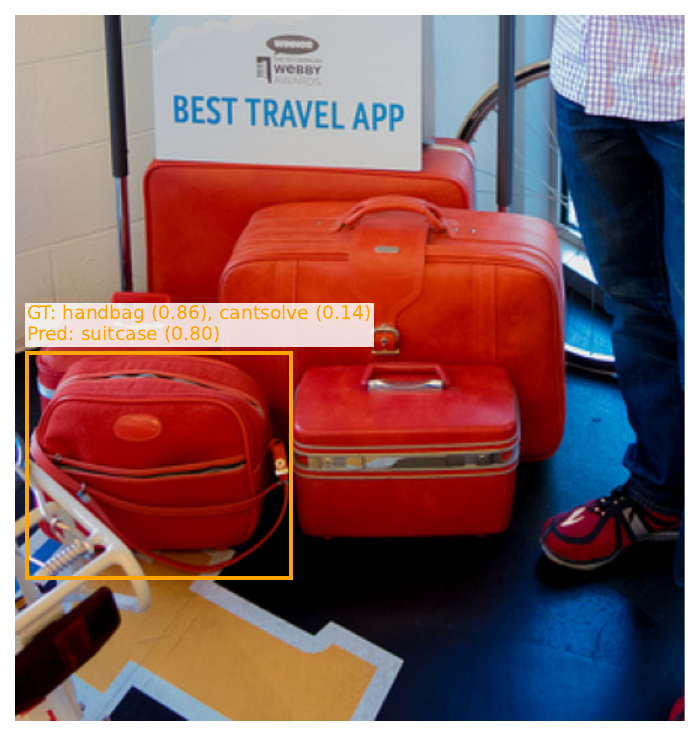}
        \caption{\tiny COCO - 000000341469}
        \label{fig:model-gt-discrepancy-2}
    \end{subfigure}
    \hfill
    \begin{subfigure}[b]{0.19\linewidth}
        \includegraphics[width=\linewidth]{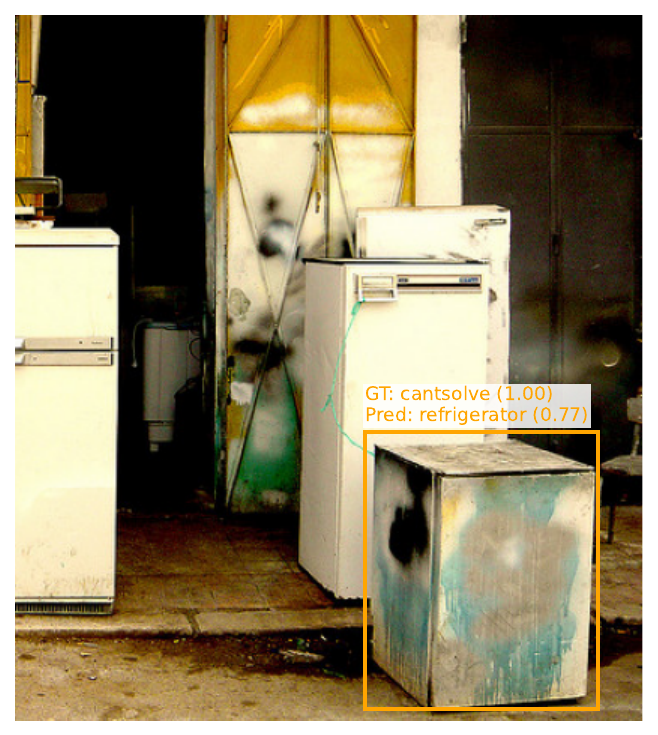}
        \caption{\tiny COCO - 000000462629}
        \label{fig:model-gt-discrepancy-3}
    \end{subfigure}
    \hfill
    \begin{subfigure}[b]{0.19\linewidth}
        \includegraphics[width=\linewidth]{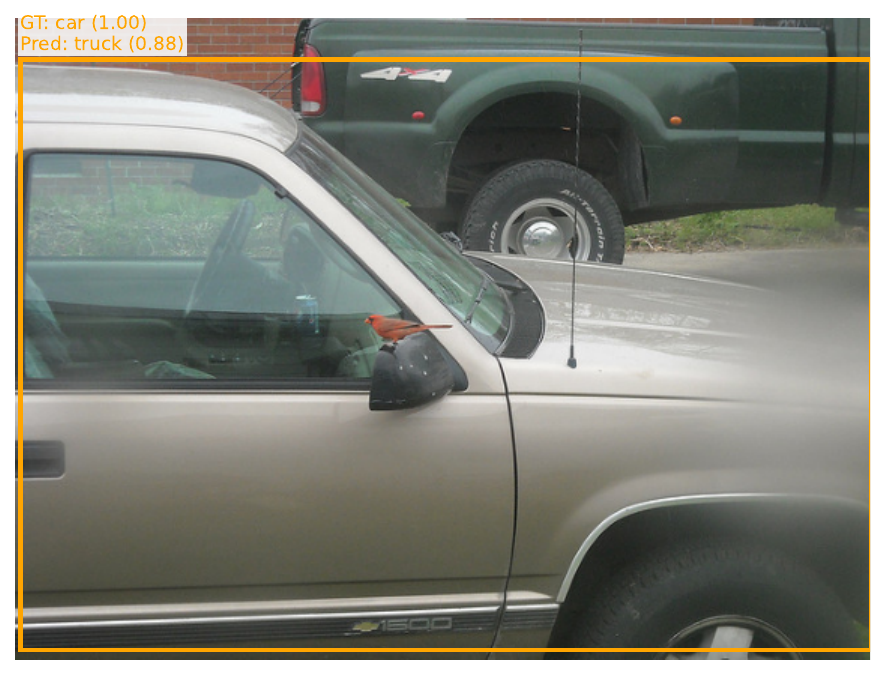}
        \caption{\tiny COCO - 000000523782}
        \label{fig:model-gt-discrepancy-4}
    \end{subfigure}
    \vspace{0.5em}
    \begin{subfigure}[b]{0.19\linewidth}
        \includegraphics[width=\linewidth]{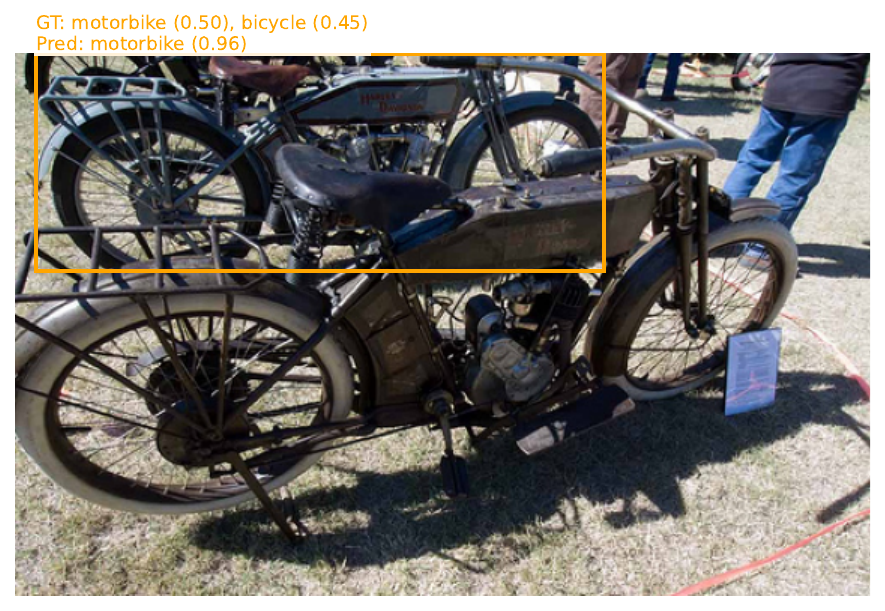}
        \caption{\tiny Pascal VOC - 2009\_000676}
        \label{fig:model-gt-discrepancy-5}
    \end{subfigure}
    \hfill
    \begin{subfigure}[b]{0.19\linewidth}
        \includegraphics[width=\linewidth]{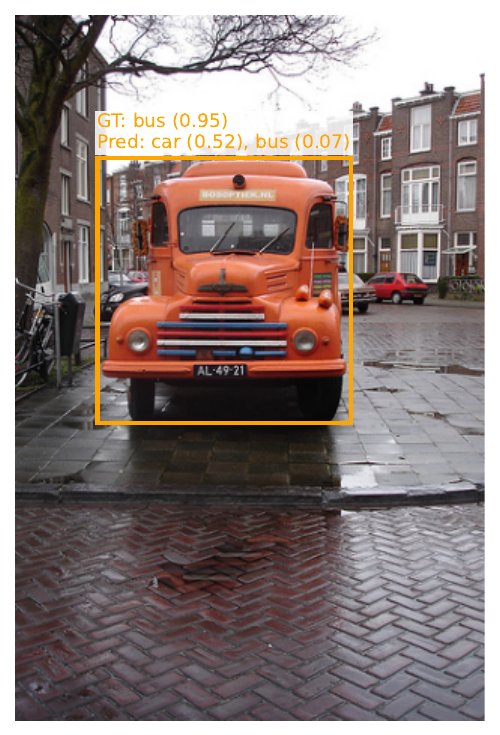}
        \caption{\tiny Pascal VOC - 2010\_000590}
        \label{fig:model-gt-discrepancy-6}
    \end{subfigure}
    \hfill
    \begin{subfigure}[b]{0.19\linewidth}
        \includegraphics[width=\linewidth]{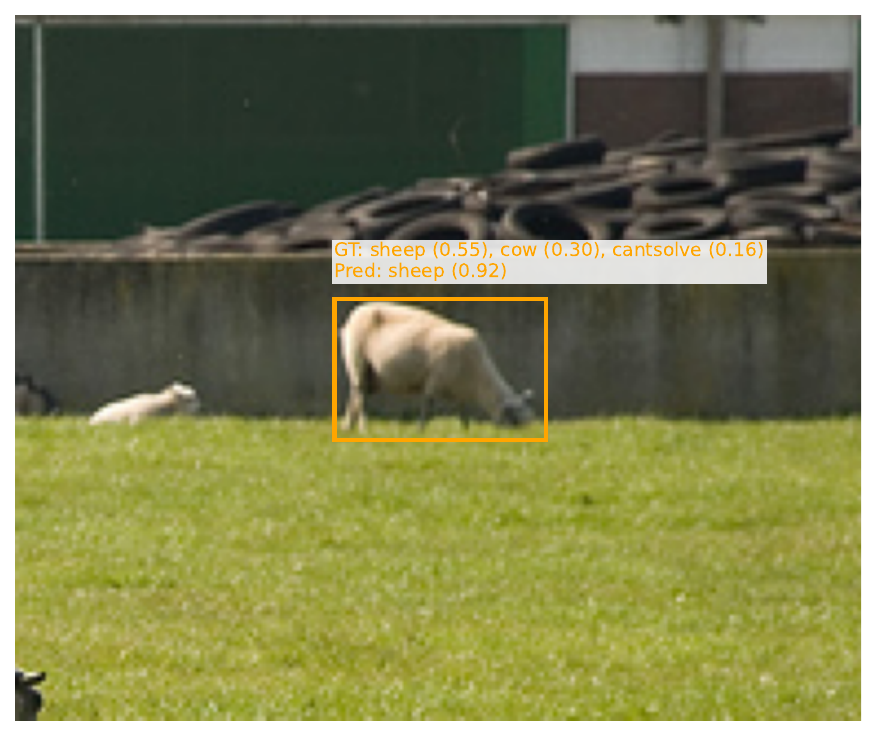}
        \caption{\tiny Pascal VOC - 2010\_000309}
        \label{fig:model-gt-discrepancy-7}
    \end{subfigure}
    \hfill
    \begin{subfigure}[b]{0.19\linewidth}
        \includegraphics[width=\linewidth]{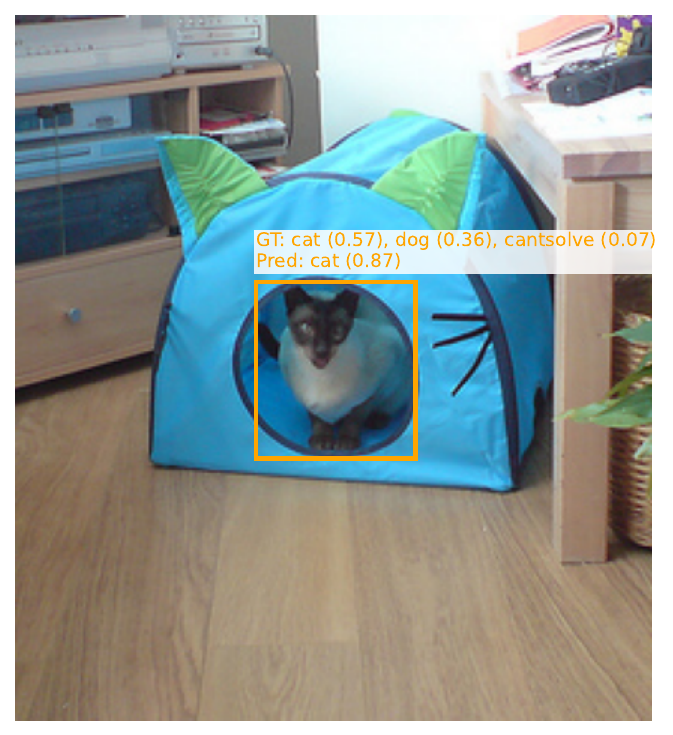}
        \caption{\tiny Pascal VOC - 2009\_000488}
        \label{fig:model-gt-discrepancy-8}
    \end{subfigure}
    \hfill
    \begin{subfigure}[b]{0.19\linewidth}
        \includegraphics[width=\linewidth]{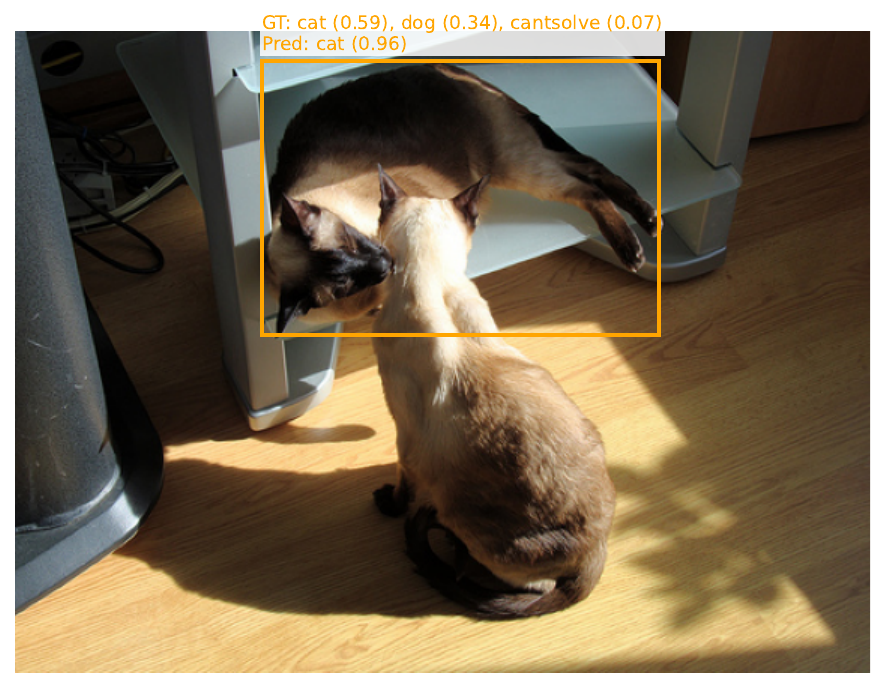}
        \caption{\tiny Pascal VOC - 2009\_002836}
        \label{fig:model-gt-discrepancy-9}
    \end{subfigure}
    \vspace{0.5em}
    \begin{subfigure}[b]{0.19\linewidth}
        \includegraphics[width=\linewidth]{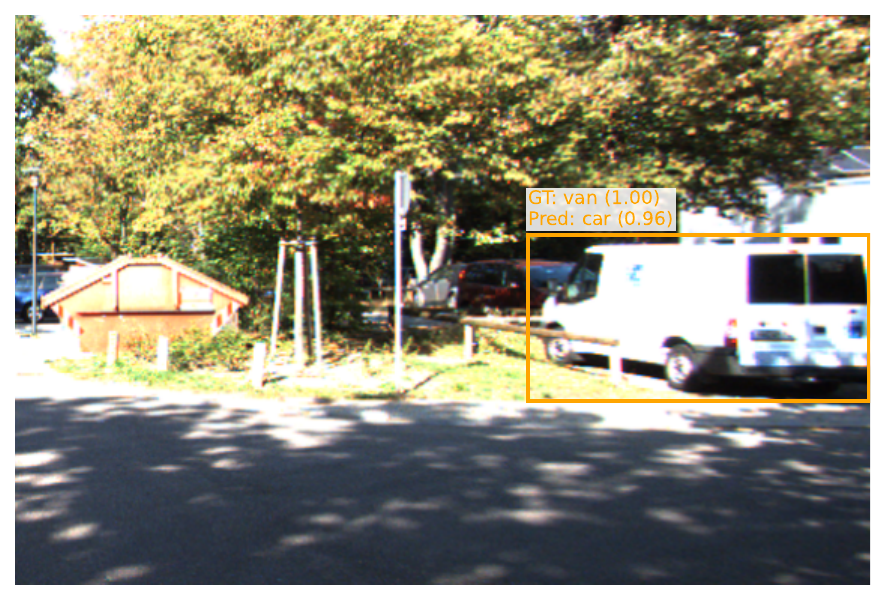}
        \caption{\tiny KITTI - 006622}
        \label{fig:model-gt-discrepancy-10}
    \end{subfigure}
    \hfill
    \begin{subfigure}[b]{0.19\linewidth}
        \includegraphics[width=\linewidth]{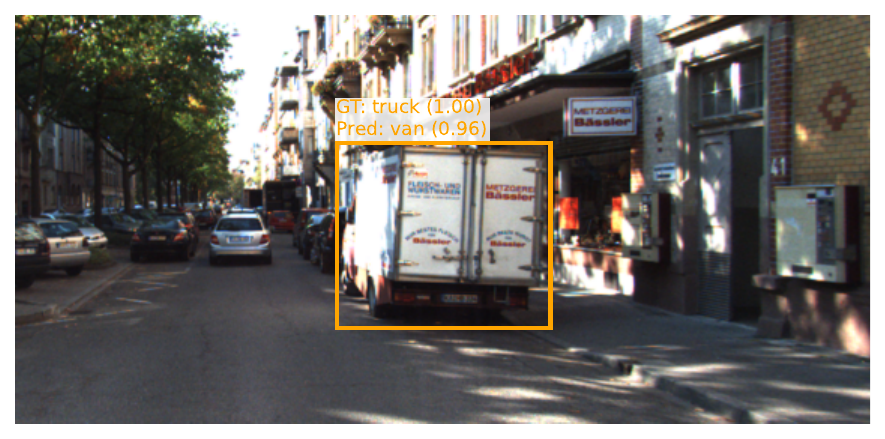}
        \caption{\tiny KITTI - 001580}
        \label{fig:model-gt-discrepancy-11}
    \end{subfigure}
    \hfill
    \begin{subfigure}[b]{0.19\linewidth}
        \includegraphics[width=\linewidth]{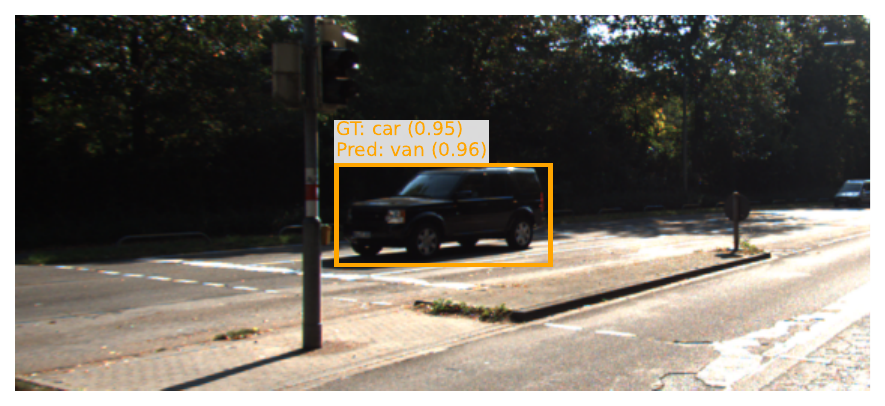}
        \caption{\tiny KITTI - 001185}
        \label{fig:model-gt-discrepancy-12}
    \end{subfigure}
    \hfill
    \begin{subfigure}[b]{0.19\linewidth}
        \includegraphics[width=\linewidth]{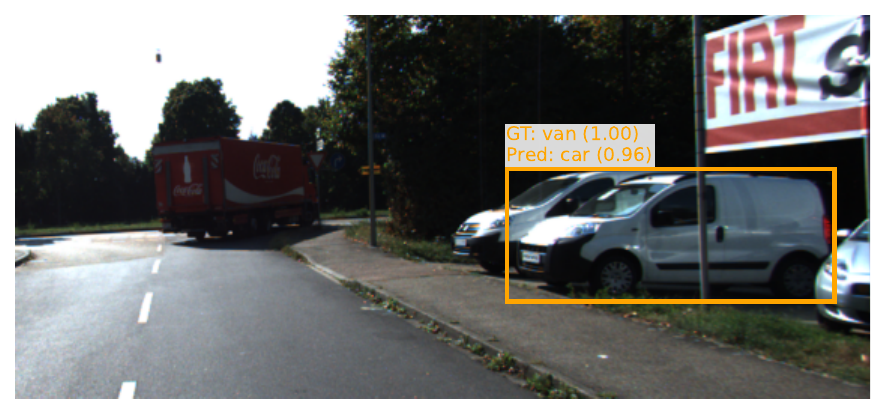}
        \caption{\tiny KITTI - 003320}
        \label{fig:model-gt-discrepancy-13}
    \end{subfigure}
    \hfill
    \begin{subfigure}[b]{0.19\linewidth}
        \includegraphics[width=\linewidth]{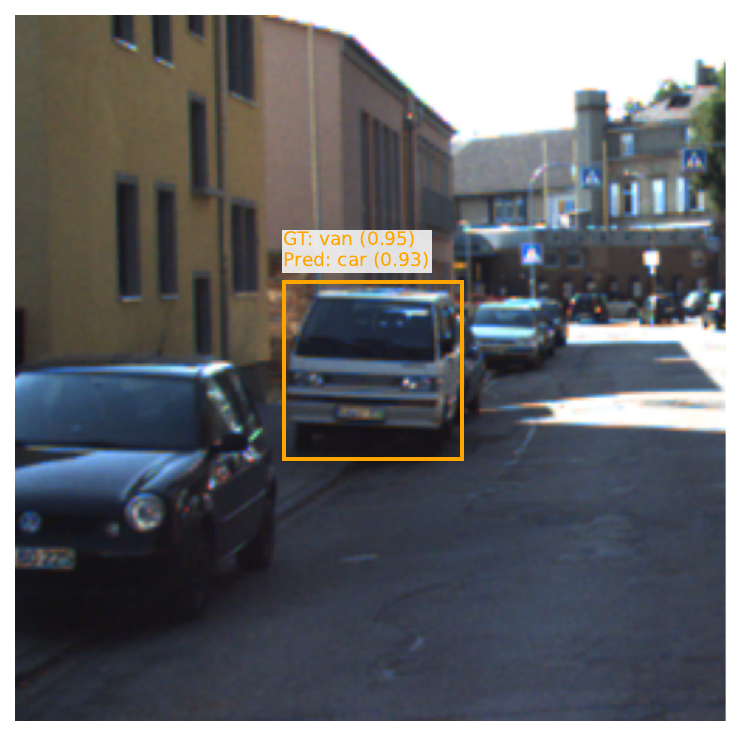}
        \caption{\tiny KITTI - 001458}
        \label{fig:model-gt-discrepancy-14}
    \end{subfigure}
    \vspace{0.5em}
    \begin{subfigure}[b]{0.19\linewidth}
        \includegraphics[width=\linewidth]{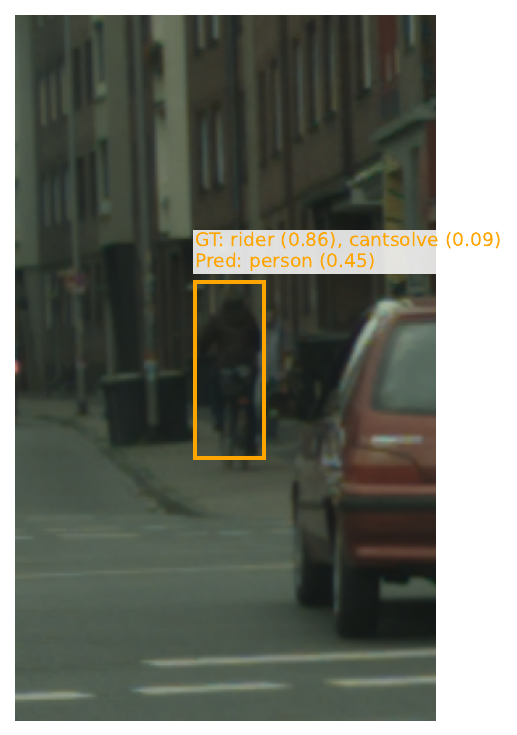}
        \caption{\tiny Cityscapes - munster\_000078\_000019}
        \label{fig:model-gt-discrepancy-15}
    \end{subfigure}
    \hfill
    \begin{subfigure}[b]{0.19\linewidth}
        \includegraphics[width=\linewidth]{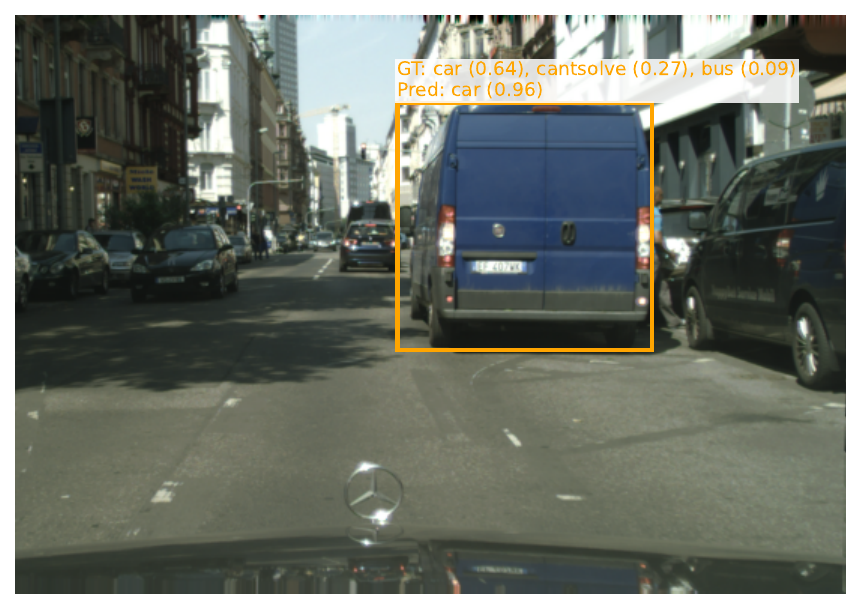}
        \caption{\tiny Cityscapes - frankfurt\_000001\_031416}
        \label{fig:model-gt-discrepancy-16}
    \end{subfigure}
    \hfill
    \begin{subfigure}[b]{0.19\linewidth}
        \includegraphics[width=\linewidth]{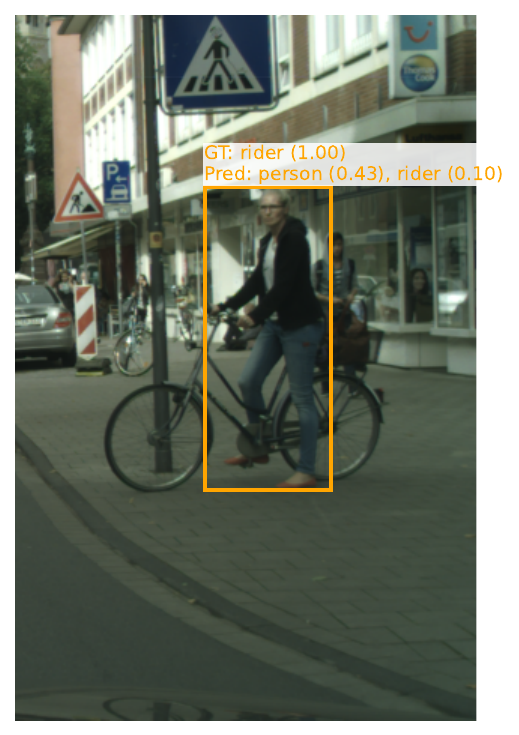}
        \caption{\tiny Cityscapes - munster\_000040\_000019}
        \label{fig:model-gt-discrepancy-17}
    \end{subfigure}
    \hfill
    \begin{subfigure}[b]{0.19\linewidth}
        \includegraphics[width=\linewidth]{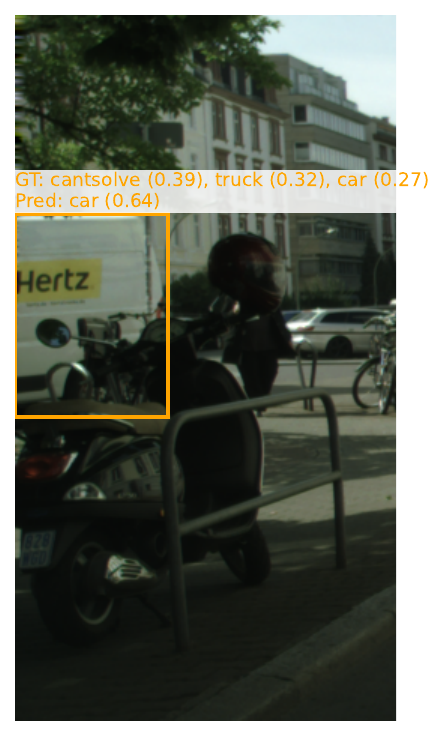}
        \caption{\tiny Cityscapes - frankfurt\_000001\_064130}
        \label{fig:model-gt-discrepancy-18}
    \end{subfigure}
    \hfill
    \begin{subfigure}[b]{0.19\linewidth}
        \includegraphics[width=\linewidth]{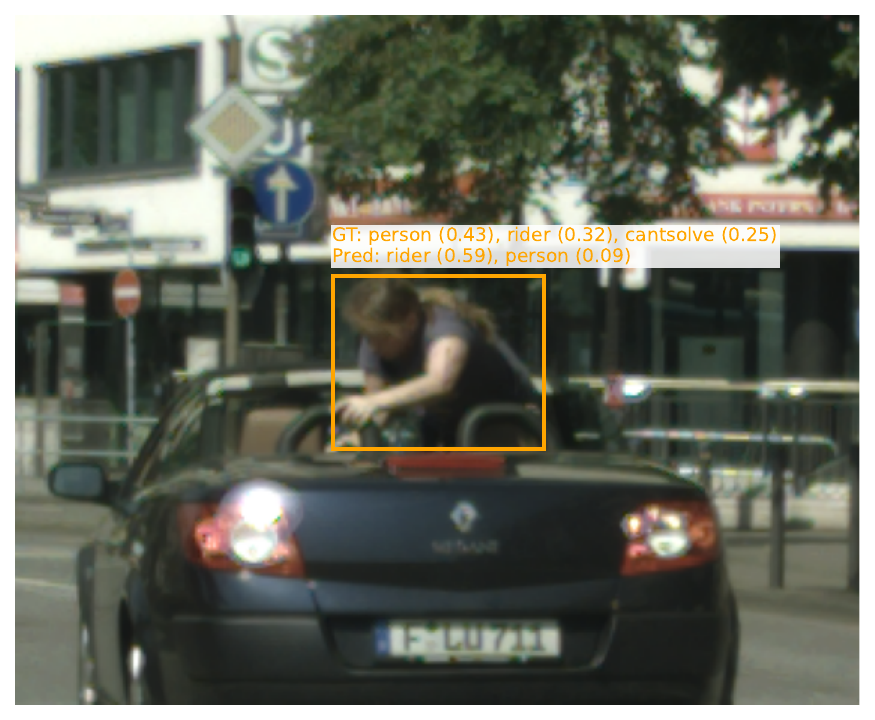}
        \caption{\tiny Cityscapes - frankfurt\_000001\_038418}
        \label{fig:model-gt-discrepancy-19}
    \end{subfigure}
    \caption{Model predictions vs.\ soft labels on objects where both disagree. We show illustrative cases with a discrepancy between the YOLOv8 predicted probabilities and the corresponding soft label for each dataset. We display all class probabilities larger than 0.05 in the images instead of the full distribution for readability.}
    \label{fig:visual-model-gt-discrepancy}
\end{figure*}

%% file: tables/preds_perf_gt_table.tex
\begin{table}[htbp]
\centering
\small
\setlength{\tabcolsep}{4pt}

\caption{Detection performance comparison of Grounding DINO and Cascade R-CNN with respect to the original ground truth on the evaluated datasets using COCO evaluation metrics.}
\label{tab:gd_vs_cascade}

\begin{subtable}{0.48\textwidth}
\centering
\caption{Grounding DINO}
\label{tab:gd_detection_original}
\scalebox{0.85}{
\begin{tabular}{lcccccc}
\hline
Dataset & AP & AP\textsubscript{50} & AP\textsubscript{75} & AP\textsubscript{S} & AP\textsubscript{M} & AP\textsubscript{L} \\
\hline
Pascal VOC  & 74.86 & 92.31 & 81.58 & 30.64 & 60.50 & 85.50 \\
COCO       & 46.93 & 64.33 & 50.90 & 23.26 & 46.27 & 62.08 \\
Cityscapes & 33.09 & 52.51 & 32.57 & 7.69 & 22.96 & 50.75 \\
KITTI      & 25.22 & 45.52 & 25.09 & 17.51 & 26.59 & 32.79 \\
\hline
\end{tabular}}
\end{subtable}
\hfill
\begin{subtable}{0.48\textwidth}
\centering
\caption{Cascade R-CNN}
\label{tab:cascade_detection_original}
\scalebox{0.85}{
\begin{tabular}{lcccccc}
\hline
Dataset & AP & AP\textsubscript{50} & AP\textsubscript{75} & AP\textsubscript{S} & AP\textsubscript{M} & AP\textsubscript{L} \\
\hline
Pascal VOC  & 23.30 & 40.99 & 23.24 & 9.35 & 14.81 & 28.19 \\
COCO       & 20.28 & 33.77 & 21.04 & 8.99 & 19.59 & 27.56 \\
Cityscapes & 30.30 & 51.45 & 30.79 & 7.22 & 25.94 & 44.69 \\
KITTI      & 54.32 & 78.42 & 59.72 & 60.32 & 58.58 & 49.91 \\
\hline
\end{tabular}}
\end{subtable}

\end{table}

%% file: tables/cascade_rcnn_training_parameters.tex
\begin{table}[htbp]
\centering
\caption{Training parameters of the Cascade R-CNN models used to generate predictions for the loss inspection label error detection method. We report the batch size, input image resolution, number of training iterations, and learning rate for each dataset. }
\small
\begin{tabular}{lcccc}
\toprule
\textbf{Dataset} & \textbf{Batch Size} & \textbf{Image Resolution} & \textbf{\# Iterations} & \textbf{Learning Rate} \\
\midrule
Cityscapes & 4  & $1333 \times 800$ & 120,000 & 0.01 \\
COCO       & 12 & $1000 \times 600$ & 250,000 & 0.02 \\
KITTI      & 6  & $1000 \times 600$ & 70,000  & 0.01 \\
Pascal VOC & 6  & $1000 \times 600$ & 70,000  & 0.02 \\
\bottomrule
\end{tabular}

\label{tab:cascade_rcnn_training}
\end{table}

%% file: tables/label_error_benchmark_table_appendix.tex
\begin{table*}[htbp]
\centering
\caption{
Label error detection performance across multiple datasets and representative label error detection methods on the stricter label error ground truth (including only label errors with a minimum height of 40 pixels and soft label probability $\geq 0.8$).
Evaluation is based on predictions from Grounding DINO.
For all score-based methods, Average Precision, Precision, Recall, and F1 are reported at the method-specific threshold that maximizes F1-Score on the corresponding precision-recall curve.
The naive baseline has no ranking score and is therefore evaluated as a single operating point; AP is not defined for this baseline.
Evaluation is performed using an IoU threshold of 0.1 for all methods.
}
\begin{minipage}{0.49\textwidth}
\centering
\small
\begin{tabular}{lcccc}
\toprule
\textbf{Cityscapes} & \textbf{AP} & \textbf{Prec} & \textbf{Rec} & \textbf{F1}\\
\midrule
Naive Baseline   & - & 0.64 & 100.00 & 1.28 \\
Score Baseline   & 17.16  & 20.64  & 47.35  & 28.75   \\
Loss Inspection  & 8.83  & 10.63 & 35.95  & 16.41 \\
ObjectLab        & 13.45 & 17.25 & 41.70 & 24.41 \\
MetaDetect       & 16.24  & 21.20  & 34.58  & 26.28   \\
\bottomrule
\end{tabular}
\end{minipage}
\hfill
\begin{minipage}{0.49\textwidth}
\centering
\small
\begin{tabular}{lcccc}
\toprule
\textbf{COCO} & \textbf{AP} & \textbf{Prec} & \textbf{Rec} & \textbf{F1}\\
\midrule
Naive Baseline   & - & 0.56 & 99.94 & 1.10 \\
Score Baseline   & 26.02 & 27.28  & 44.51  & 33.83   \\
Loss Inspection  & 7.64  & 10.64 & 28.33  & 15.47 \\
ObjectLab        & 11.02 & 22.95 & 24.67 & 23.78  \\
MetaDetect       & 26.08  & 32.40  & 37.01  & 34.55    \\
\bottomrule
\end{tabular}
\end{minipage}

\vspace{0.5cm}

\begin{minipage}{0.49\textwidth}
\centering
\small
\begin{tabular}{lcccc}
\toprule
\textbf{KITTI} & \textbf{AP} & \textbf{Prec} & \textbf{Rec} & \textbf{F1}\\
\midrule
Naive Baseline   & - & 0.19 & 100.00  & 0.37 \\
Score Baseline   & 17.82 & 20.41 & 48.80 & 28.78  \\
Loss Inspection  & 1.53  & 4.80 & 14.49  & 7.21 \\
ObjectLab        & 14.30 & 23.42 & 27.66  & 25.37 \\
MetaDetect       & 28.46 & 35.59 & 47.61 & 40.73   \\
\bottomrule
\end{tabular}
\end{minipage}
\hfill
\begin{minipage}{0.49\textwidth}
\centering
\small
\begin{tabular}{lcccc}
\toprule
\textbf{Pascal VOC} & \textbf{AP} & \textbf{Prec} & \textbf{Rec} & \textbf{F1}\\
\midrule
Naive Baseline   & - & 0.14 & 100.00 & 0.28 \\
Score Baseline   & 21.69 & 25.76 & 36.48 & 30.20  \\
Loss Inspection  & 3.16  & 5.66 & 20.12  & 8.84 \\
ObjectLab        & 11.00  & 21.87 & 23.00 & 22.42 \\
MetaDetect       & 18.95  & 25.57  & 39.08  & 30.91   \\
\bottomrule
\end{tabular}
\end{minipage}

\label{tab:label_error_benchmark_appendix}
\end{table*}

%% file: tables/cascade_label_error_benchmark_less_restrictive.tex
\begin{table*}[htbp]
\centering
\scalebox{1.0}{
\begin{minipage}{\textwidth}

\caption{
Label error detection performance across multiple datasets and representative methods on the less restrictive label error ground truth (all label errors with soft label probability $\geq 0.5$).
Evaluation is based on predictions from Cascade R-CNN.
For all score-based methods, results are reported at the threshold that maximizes the F1-score on the corresponding precision-recall curve.
The naive baseline has no ranking score and is therefore evaluated as a single operating point (no AP).
Evaluation is performed using an IoU threshold of 0.1.
}

\label{tab:cascade_label_error_benchmark_less_Restrictive}

\begin{minipage}{0.49\textwidth}
\centering
\small
\begin{tabular}{lcccc}
\toprule
\textbf{Cityscapes} & \textbf{AP} & \textbf{Prec} & \textbf{Rec} & \textbf{F1}\\
\midrule
Naive Baseline & - & 6.48 & 79.23 & 11.98\\
Score Baseline & 23.68 & 27.66 & 40.34 & 32.82 \\
Loss Inspection & 15.53 & 18.89 & 41.47 & 25.96 \\
ObjectLab & 23.04 & 25.57 & 41.90 & 31.76 \\
MetaDetect & 19.23 & 25.52 & 28.97 & 27.14 \\
\bottomrule
\end{tabular}
\end{minipage}
\hfill
\begin{minipage}{0.49\textwidth}
\centering
\small
\begin{tabular}{lcccc}
\toprule
\textbf{COCO} & \textbf{AP} & \textbf{Prec} & \textbf{Rec} & \textbf{F1}\\
\midrule
Naive Baseline & - & 3.47 & 56.72 & 6.55 \\
Score Baseline & 13.54 & 21.77 & 28.89 & 24.83 \\
Loss Inspection & 10.42 & 13.85 & 31.12 & 19.17 \\
ObjectLab & 15.64 & 26.70 & 29.30 & 27.94  \\
MetaDetect & 12.46 & 20.86 & 27.51 & 23.73 \\
\bottomrule
\end{tabular}
\end{minipage}

\vspace{0.5cm}

\begin{minipage}{0.49\textwidth}
\centering
\small
\begin{tabular}{lcccc}
\toprule
\textbf{KITTI} & \textbf{AP} & \textbf{Prec} & \textbf{Rec} & \textbf{F1}\\
\midrule
Naive Baseline & - & 8.91 & 62.13 & 15.58\\
Score Baseline & 20.28 & 28.15 & 30.49 & 29.27 \\
Loss Inspection & 6.72 & 21.28 & 24.81 & 22.91 \\
ObjectLab & 14.67 & 20.87 & 29.64 & 24.49 \\
MetaDetect & 17.04 & 15.77 & 43.14 & 23.10 \\
\bottomrule
\end{tabular}
\end{minipage}
\hfill
\begin{minipage}{0.49\textwidth}
\centering
\small
\begin{tabular}{lcccc}
\toprule
\textbf{Pascal VOC} & \textbf{AP} & \textbf{Prec} & \textbf{Rec} & \textbf{F1}\\
\midrule
Naive Baseline & - & 0.56 & 78.10 & 1.11\\
Score Baseline & 3.36 & 6.43 & 19.30 & 9.65  \\
Loss Inspection & 3.27 & 7.19 & 14.29 & 9.56 \\
ObjectLab & 2.79 & 7.75 & 13.03 & 9.72  \\
MetaDetect & 3.07 & 7.15 & 12.48 & 9.09 \\
\bottomrule
\end{tabular}
\end{minipage}

\end{minipage}
}
\end{table*}

%% file: tables/cascade_label_error_benchmark_stricter.tex
\begin{table*}[htbp]
\centering
\caption{
Label error detection performance across multiple datasets and representative label error detection methods on the stricter label error ground truth (including only label errors with a minimum height of 40 pixels and soft label probability $\geq 0.8$).
Evaluation is based on predictions from Cascade R-CNN.
For all score-based methods, Average Precision, Precision, Recall, and F1 are reported at the method-specific threshold that maximizes F1-Score on the corresponding precision-recall curve.
The naive baseline has no ranking score and is therefore evaluated as a single operating point; AP is not defined for this baseline.
Evaluation is performed using an IoU threshold of 0.1 for all methods.
}
\begin{minipage}{0.49\textwidth}
\centering
\small
\begin{tabular}{lcccc}
\toprule
\textbf{Cityscapes} & \textbf{AP} & \textbf{Prec} & \textbf{Rec} & \textbf{F1}\\
\midrule
Naive Baseline   & - & 3.04 & 86.86 & 5.88\\
Score Baseline   & 16.77 & 22.30 & 30.57 & 25.79 \\
Loss Inspection  & 8.83  & 10.63 & 35.95 & 16.41 \\
ObjectLab        & 15.53 & 19.89 & 34.12 & 25.13 \\
MetaDetect       & 14.83 & 24.55 & 19.71 & 21.86 \\
\bottomrule
\end{tabular}
\end{minipage}
\hfill
\begin{minipage}{0.49\textwidth}
\centering
\small
\begin{tabular}{lcccc}
\toprule
\textbf{COCO} & \textbf{AP} & \textbf{Prec} & \textbf{Rec} & \textbf{F1}\\
\midrule
Naive Baseline   & - & 1.38 & 86.64 & 2.71 \\
Score Baseline   & 14.72 & 18.28 & 33.15 & 23.57 \\
Loss Inspection  & 7.64  & 10.64 & 28.33  & 15.47 \\
ObjectLab        & 14.49 & 20.21 & 36.51 & 26.02  \\
MetaDetect       & 14.84 & 20.36 & 29.89 & 24.22  \\
\bottomrule
\end{tabular}
\end{minipage}

\vspace{0.5cm}

\begin{minipage}{0.49\textwidth}
\centering
\small
\begin{tabular}{lcccc}
\toprule
\textbf{KITTI} & \textbf{AP} & \textbf{Prec} & \textbf{Rec} & \textbf{F1}\\
\midrule
Naive Baseline   & - & 1.43 & 95.08 & 2.82\\
Score Baseline   & 11.39 & 16.29 & 26.73 & 20.24 \\
Loss Inspection  & 1.53  & 4.80 & 14.49  & 7.21 \\
ObjectLab        & 9.30 & 13.42 & 27.26 & 17.98\\
MetaDetect       & 11.64 & 16.63 & 26.73 & 20.50   \\
\bottomrule
\end{tabular}
\end{minipage}
\hfill
\begin{minipage}{0.49\textwidth}
\centering
\small
\begin{tabular}{lcccc}
\toprule
\textbf{Pascal VOC} & \textbf{AP} & \textbf{Prec} & \textbf{Rec} & \textbf{F1}\\
\midrule
Naive Baseline   & - & 0.25 & 96.63 & 0.49 \\
Score Baseline   & 3.67 & 6.78 & 13.28 & 8.98  \\
Loss Inspection  & 3.16  & 5.66 & 20.12  & 8.84 \\
ObjectLab        & 2.98 & 7.07 & 17.52 & 10.07 \\
MetaDetect       & 3.76 & 6.72 & 16.94 & 9.62   \\
\bottomrule
\end{tabular}
\end{minipage}

\label{tab:cascade_label_error_benchmark_stricter}
\end{table*}

%% file: labelerrors/missing/plot_missing_label_errors.tex
\begin{figure*}[htbp]
\centering

\begin{subfigure}[t]{\textwidth}
\centering

\includegraphics[width=0.19\linewidth]{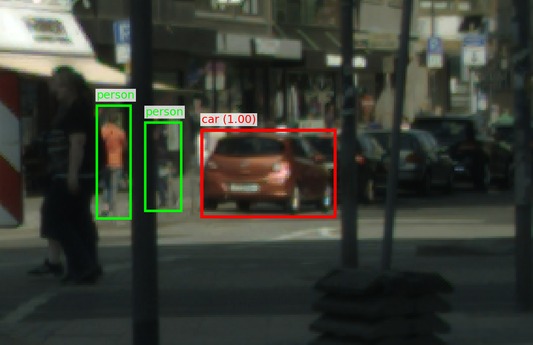}
\scalebox{0.8}{\includegraphics[width=0.19\linewidth, trim={0 4cm 0 4cm}, clip]{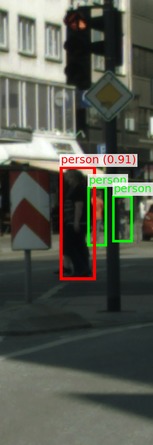}}
\includegraphics[width=0.19\linewidth, trim={0 4cm 0 4cm}, clip]{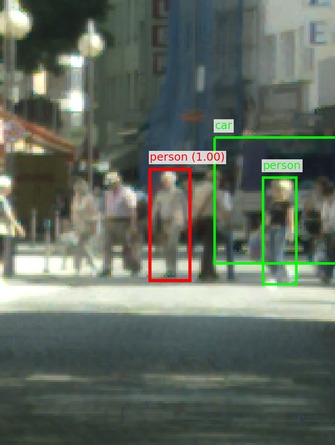}
\scalebox{0.8}{\includegraphics[width=0.19\linewidth]{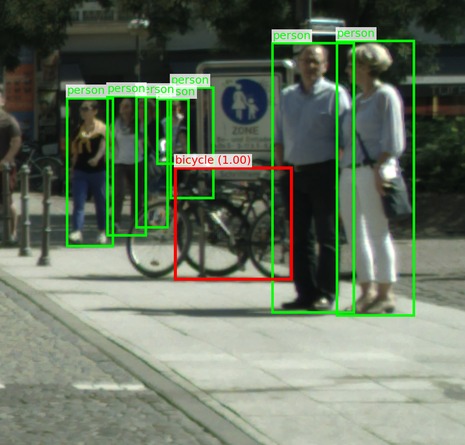}}
\scalebox{0.8}{\includegraphics[width=0.19\linewidth]{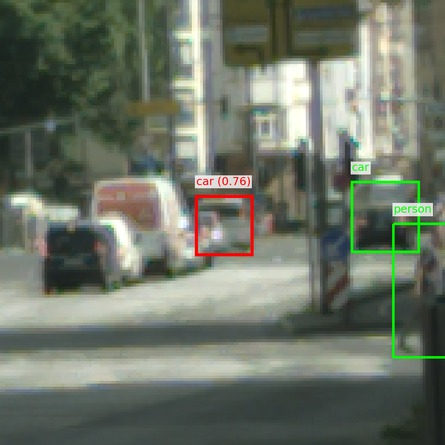}}

\vspace{0.3em}

\includegraphics[width=0.19\linewidth, trim={0 2cm 0 2cm}, clip]{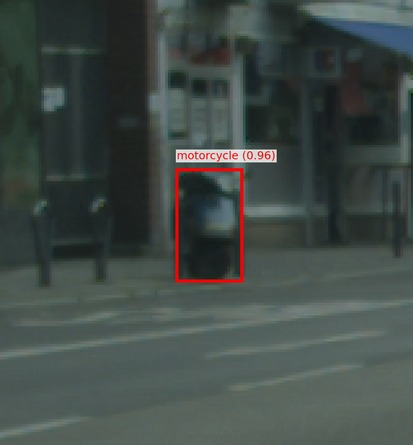}
\includegraphics[width=0.19\linewidth, trim={0 2cm 0 2cm}, clip]{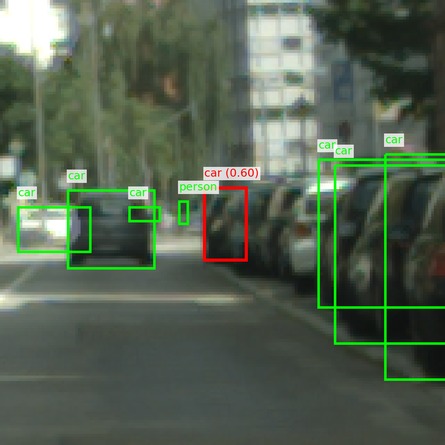}
\includegraphics[width=0.19\linewidth, trim={0 2cm 0 2cm}, clip]{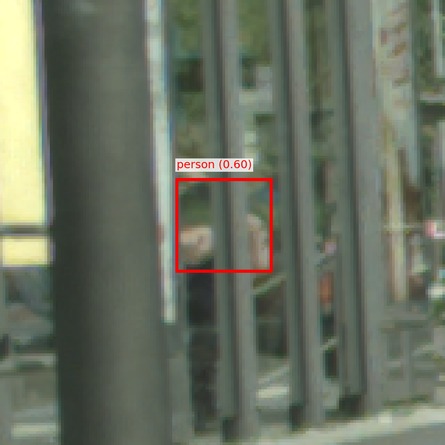}
\includegraphics[width=0.19\linewidth, trim={0 2cm 0 2cm}, clip]{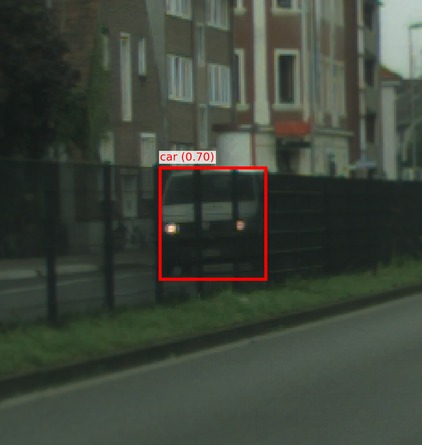}
\includegraphics[width=0.19\linewidth]{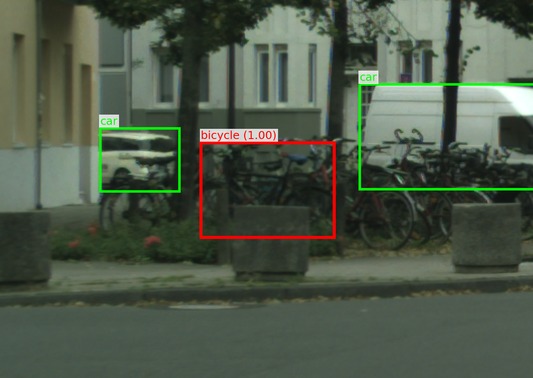}

\caption{Cityscapes}
\end{subfigure}

\vspace{0.6em}

\begin{subfigure}[t]{\textwidth}
\centering

\includegraphics[width=0.19\linewidth, trim={0 3cm 0 3cm}, clip]{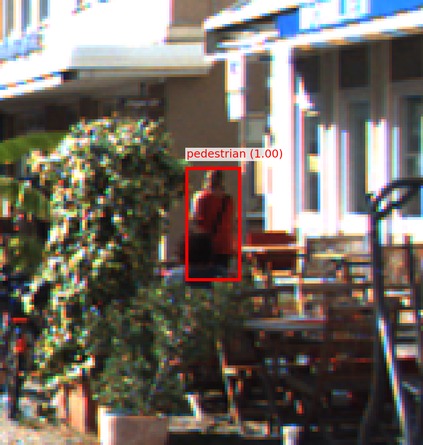}
\includegraphics[width=0.19\linewidth, trim={0 3cm 0 3cm}, clip]{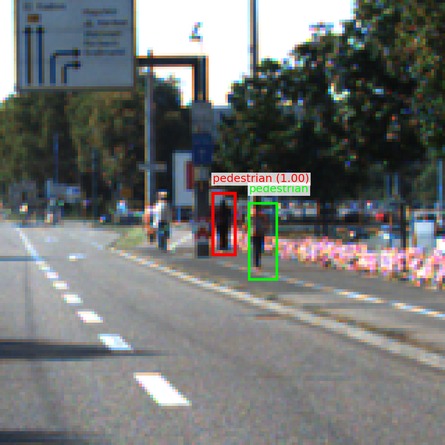}
\includegraphics[width=0.19\linewidth, trim={0 3cm 0 3cm}, clip]{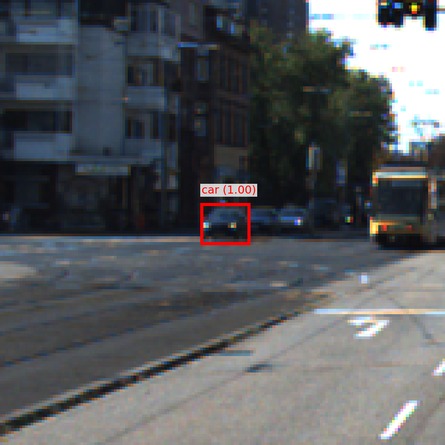}
\includegraphics[width=0.19\linewidth, trim={0 3cm 0 3cm}, clip]{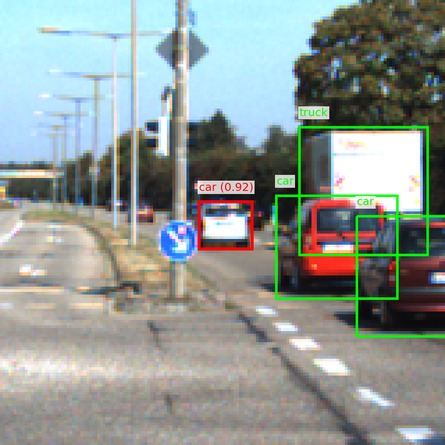}
\includegraphics[width=0.19\linewidth, trim={0 3cm 0 3cm}, clip]{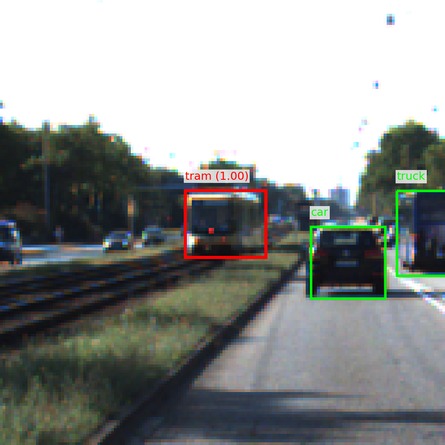}

\vspace{0.3em}

\includegraphics[width=0.19\linewidth, trim={0 3cm 0 3cm}, clip]{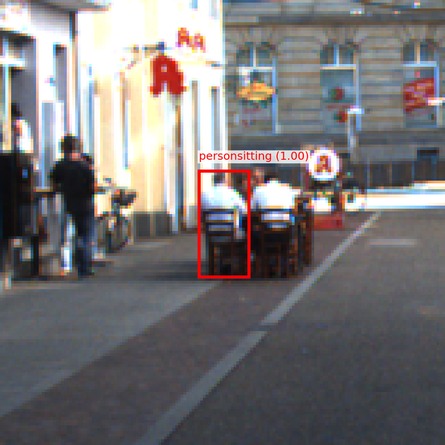}
\includegraphics[width=0.19\linewidth, trim={0 3cm 0 3cm}, clip]{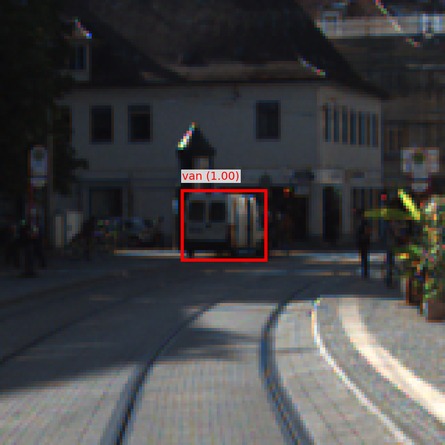}
\includegraphics[width=0.19\linewidth, trim={0 3cm 0 3cm}, clip]{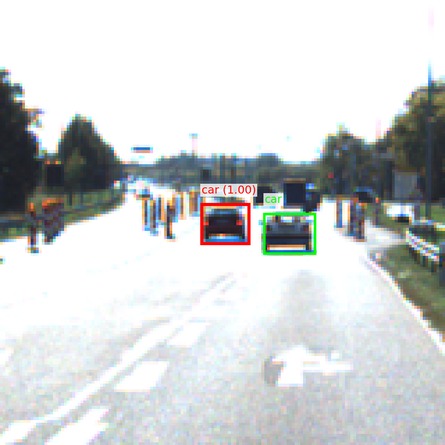}
\includegraphics[width=0.19\linewidth, trim={0 3cm 0 3cm}, clip]{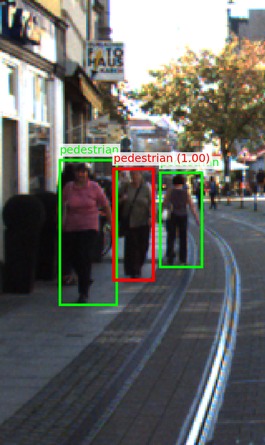}
\includegraphics[width=0.19\linewidth, trim={0 3cm 0 3cm}, clip]{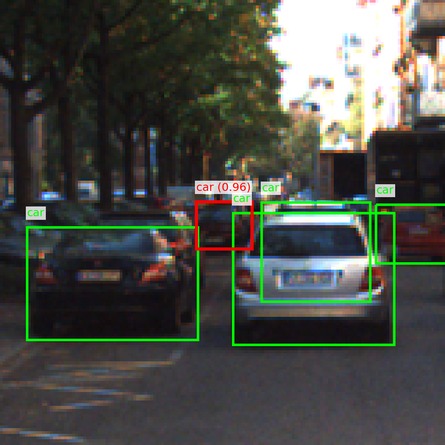}

\caption{KITTI}
\end{subfigure}

\vspace{0.6em}

\begin{subfigure}[t]{\textwidth}
\centering

\includegraphics[width=0.19\linewidth]{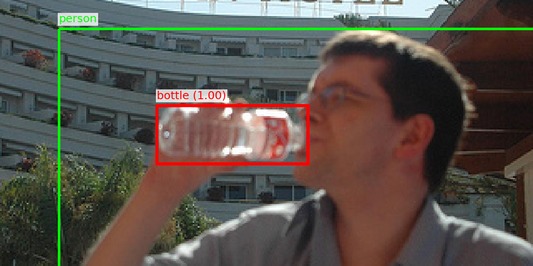}
\includegraphics[width=0.19\linewidth, trim={0 2cm 0 2cm}, clip]{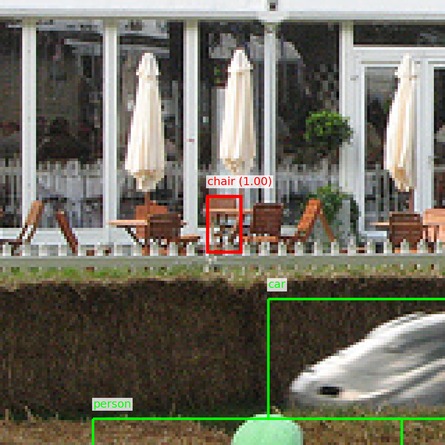}
\includegraphics[width=0.19\linewidth]{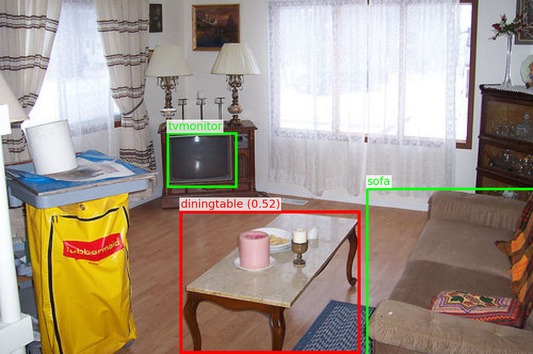}
\includegraphics[width=0.19\linewidth, trim={0 2cm 0 2cm}, clip]{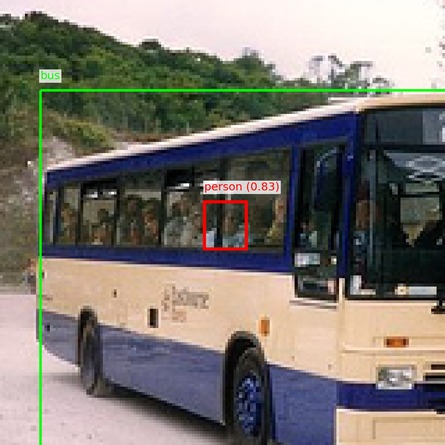}
\includegraphics[width=0.19\linewidth, trim={0 3cm 0 3cm}, clip]{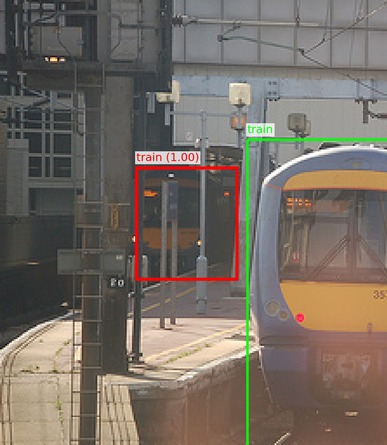}

\vspace{0.3em}

\includegraphics[width=0.19\linewidth, trim={0 3cm 0 3cm}, clip]{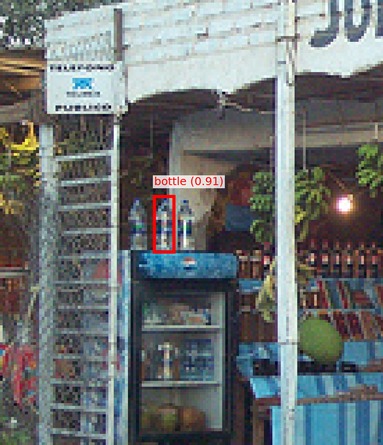}
\includegraphics[width=0.19\linewidth, trim={0 3cm 0 3cm}, clip]{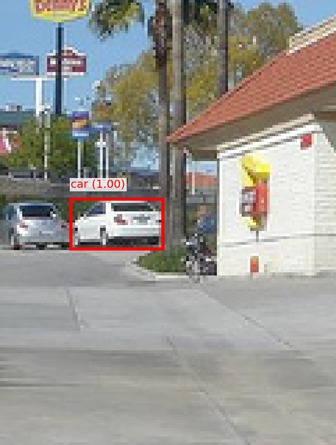}
\includegraphics[width=0.19\linewidth, trim={0 3cm 0 3cm}, clip]{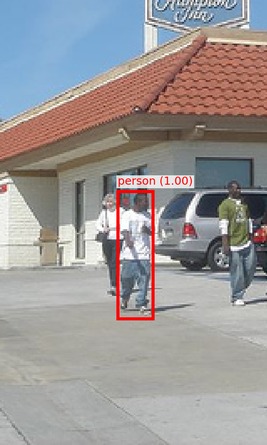}
\includegraphics[width=0.19\linewidth]{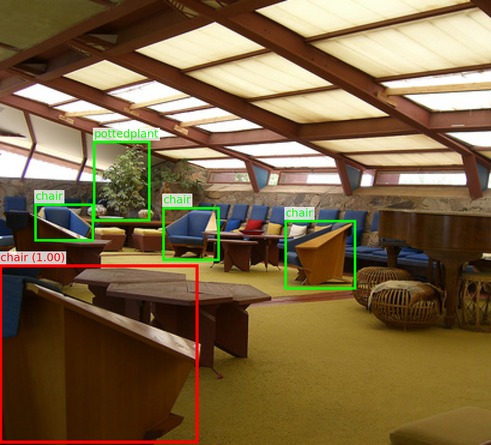}
\includegraphics[width=0.19\linewidth]{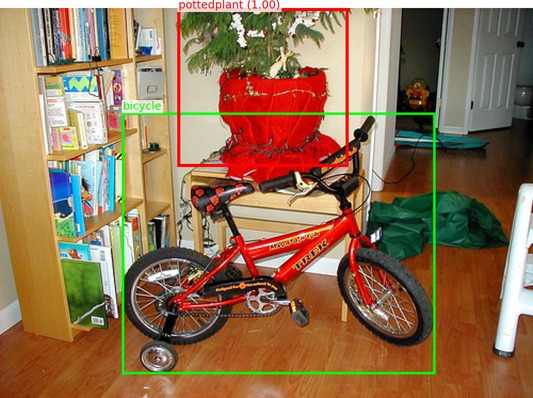}

\caption{Pascal VOC}
\end{subfigure}

\vspace{0.6em}

\begin{subfigure}[t]{\textwidth}
\centering

\includegraphics[width=0.19\linewidth, trim={0 0 0 4cm}, clip]{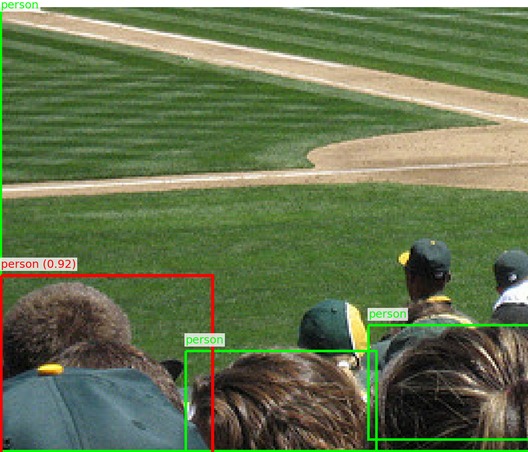}
\includegraphics[width=0.19\linewidth, trim={0 4cm 0 4cm}, clip]{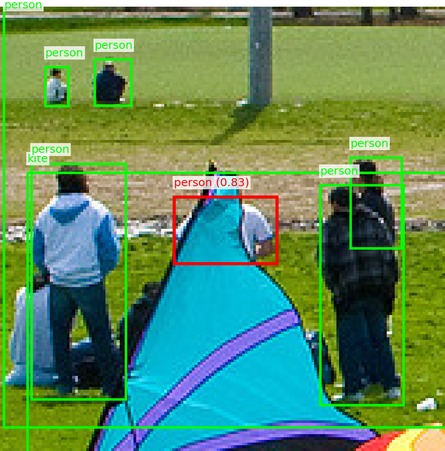}
\includegraphics[width=0.19\linewidth, trim={0 4cm 0 4cm}, clip]{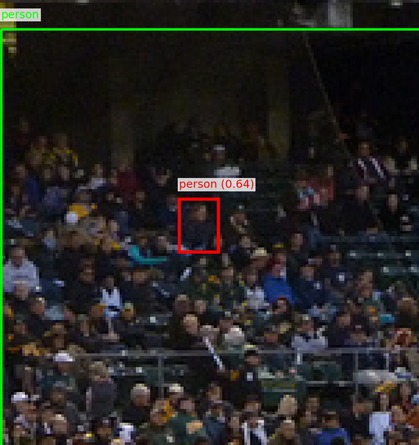}
\includegraphics[width=0.19\linewidth, trim={0 4cm 0 4cm}, clip]{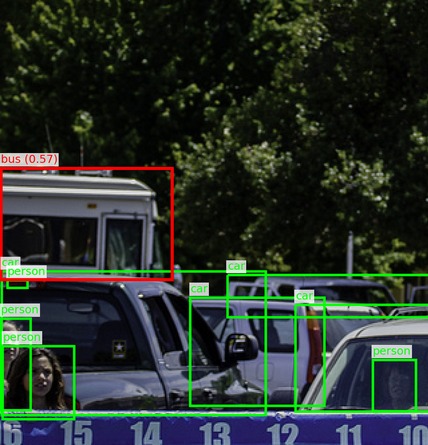}
\includegraphics[width=0.19\linewidth, trim={0 4cm 0 4cm}, clip]{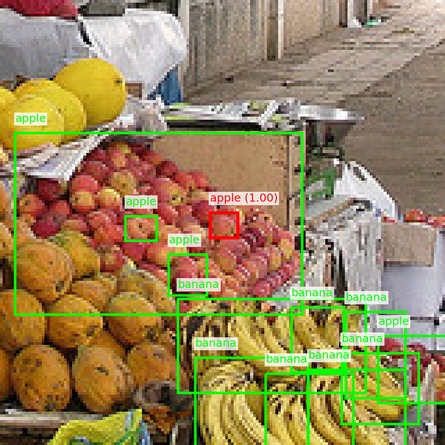}

\vspace{0.3em}

\scalebox{0.8}{\includegraphics[width=0.19\linewidth, trim={0 4cm 0 0}, clip]{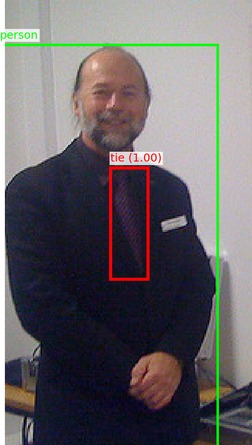}}
\scalebox{0.8}{\includegraphics[width=0.19\linewidth, trim={0 2cm 0 2cm}, clip]{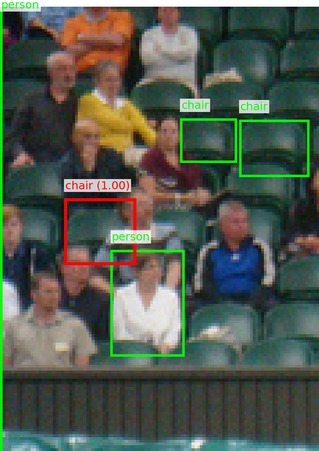}}
\includegraphics[width=0.19\linewidth, trim={0 4cm 0 4cm}, clip]{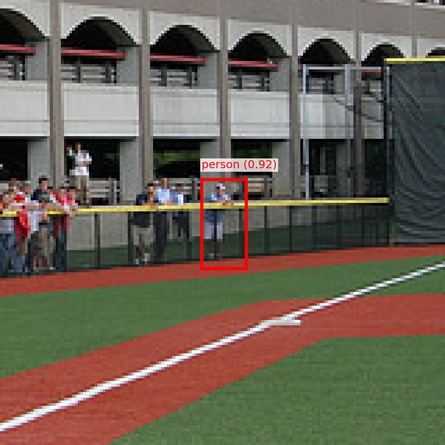}
\scalebox{0.8}{\includegraphics[width=0.19\linewidth, trim={0 4cm 0 0}, clip]{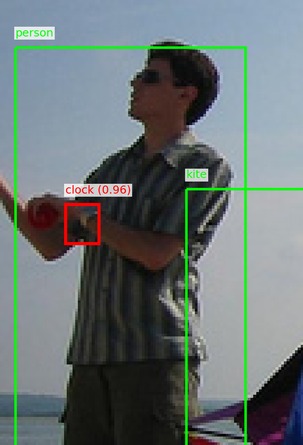}}
\includegraphics[width=0.19\linewidth, trim={0 4cm 0 4cm}, clip]{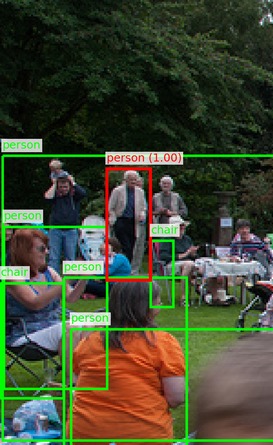}

\caption{COCO}
\end{subfigure}

\caption{Examples of \textit{missing} label errors across all four datasets. The green bounding boxes denote the original ground truth annotations, while the red bounding box highlights the label error with corresponding soft label probability $p$.}
\label{fig:missing_label_errors}
\end{figure*}

%% file: labelerrors/misaligned/plot_misaligned_label_errors.tex
\begin{figure*}[htbp]
\centering

\begin{subfigure}[t]{\textwidth}
\centering

\includegraphics[width=0.18\linewidth,trim={0 3cm 0 2cm}, clip]{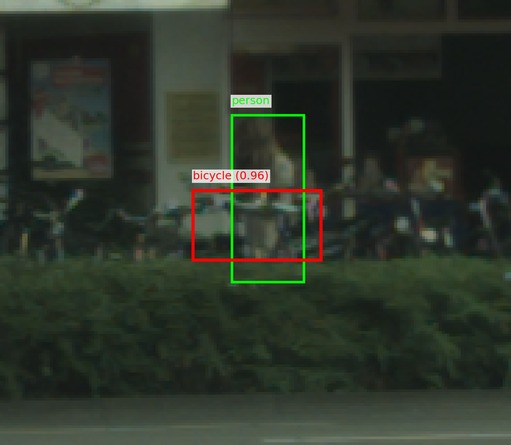}
\includegraphics[width=0.18\linewidth, trim={0 4cm 0 4cm}, clip]{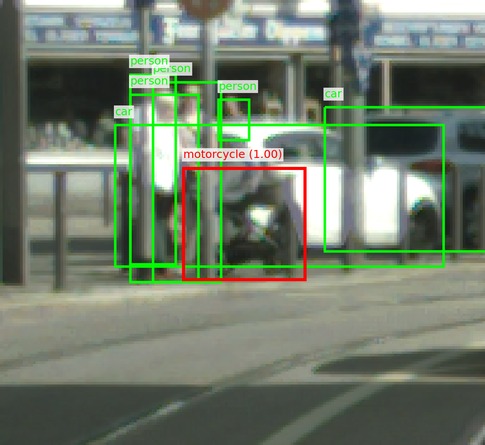}
\includegraphics[width=0.18\linewidth, trim={0 4cm 0 4cm}, clip]{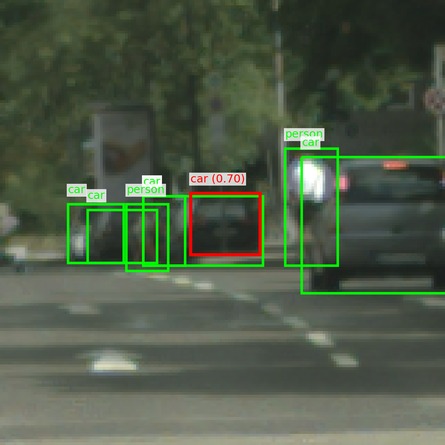}
\includegraphics[width=0.18\linewidth, trim={0 4cm 0 4cm}, clip]{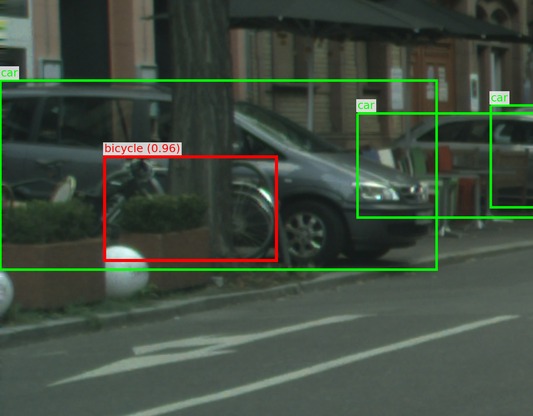}
\includegraphics[width=0.18\linewidth, trim={0 4cm 0 4cm}, clip]{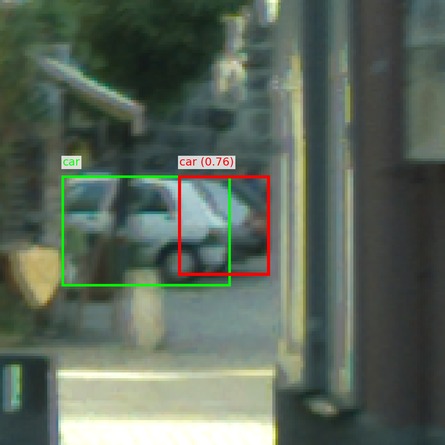}

\vspace{0.3em}

\includegraphics[width=0.18\linewidth,trim={0 4cm 0 4cm}, clip]{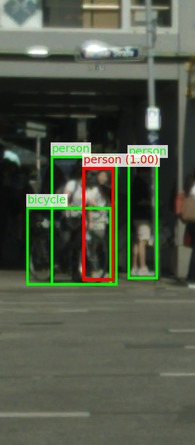}
\includegraphics[width=0.18\linewidth,trim={0 2.5cm 0 2.5cm}, clip]{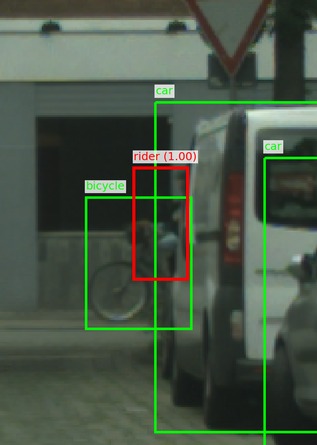}
\scalebox{0.8}{\includegraphics[width=0.18\linewidth,trim={0 4cm 0 3cm}, clip]{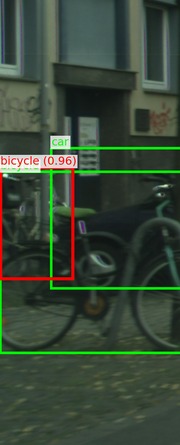}}
\includegraphics[width=0.18\linewidth,trim={0 3cm 0 3cm}, clip]{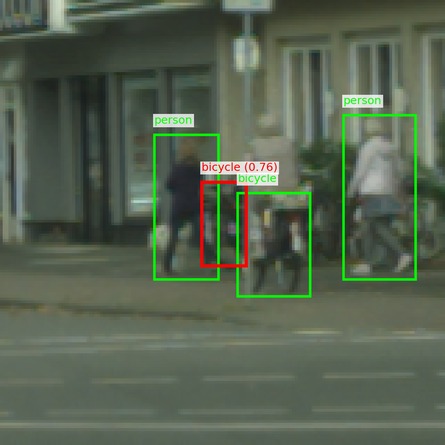}
\includegraphics[width=0.18\linewidth,trim={0 4cm 0 2cm}, clip]{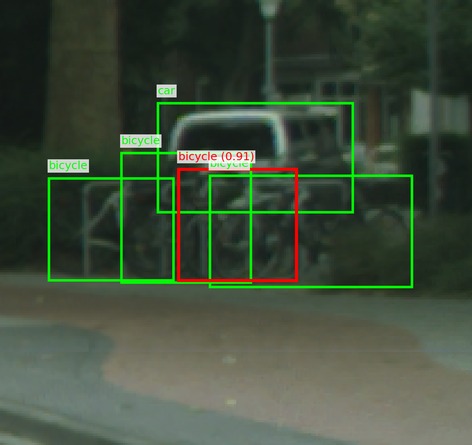}

\caption{Cityscapes}
\end{subfigure}

\vspace{0.6em}

\begin{subfigure}[t]{\textwidth}
\centering

\includegraphics[width=0.18\linewidth,trim={0 3cm 0 3cm}, clip]{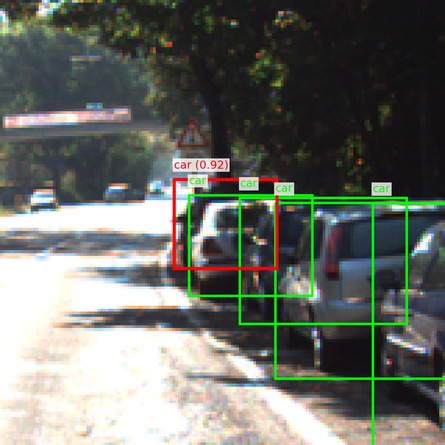}
\includegraphics[width=0.18\linewidth,trim={0 3cm 0 3cm}, clip]{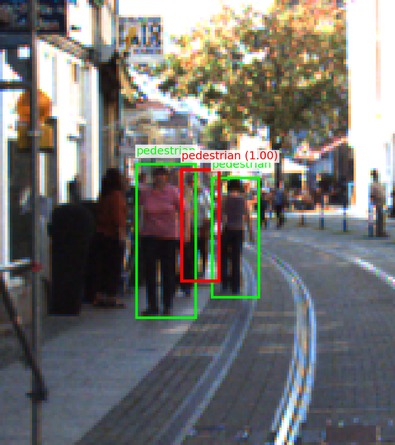}
\includegraphics[width=0.18\linewidth,trim={0 3cm 0 3cm}, clip]{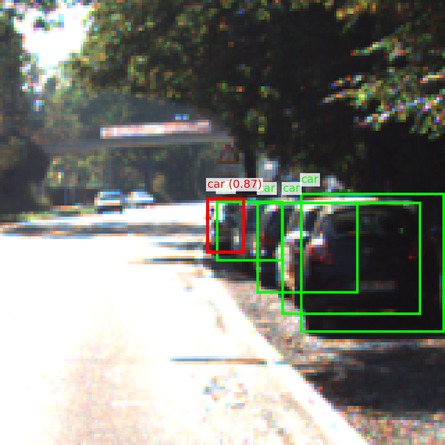}
\includegraphics[width=0.18\linewidth,trim={0 2cm 0 2cm}, clip]{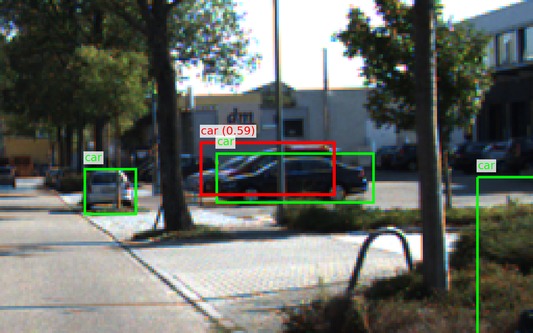}
\includegraphics[width=0.18\linewidth]{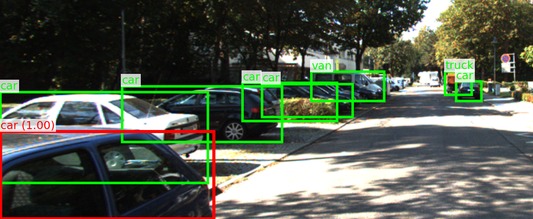}

\vspace{0.3em}

\includegraphics[width=0.18\linewidth,trim={0 3cm 0 3cm}, clip]{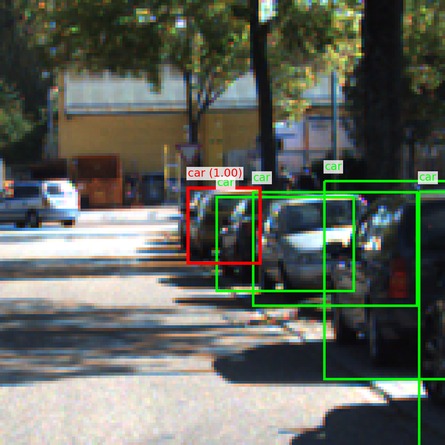}
\includegraphics[width=0.18\linewidth,trim={0 3cm 0 3cm}, clip]{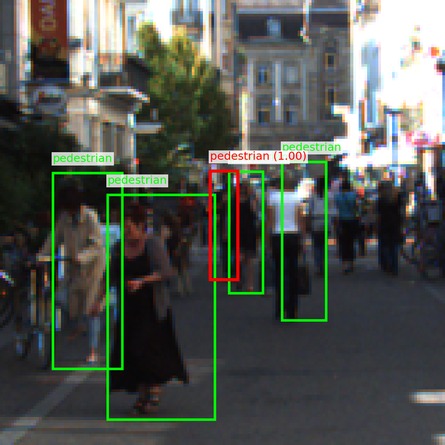}
\includegraphics[width=0.18\linewidth,trim={0 3cm 0 3cm}, clip]{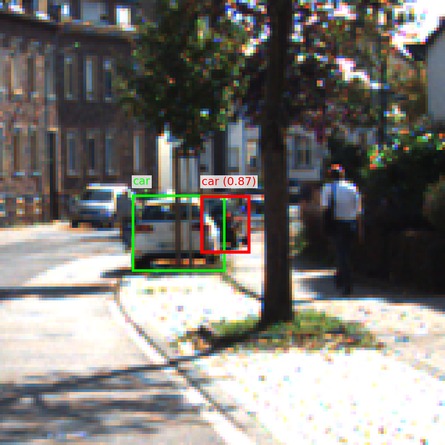}
\includegraphics[width=0.18\linewidth,trim={0 3cm 0 3cm}, clip]{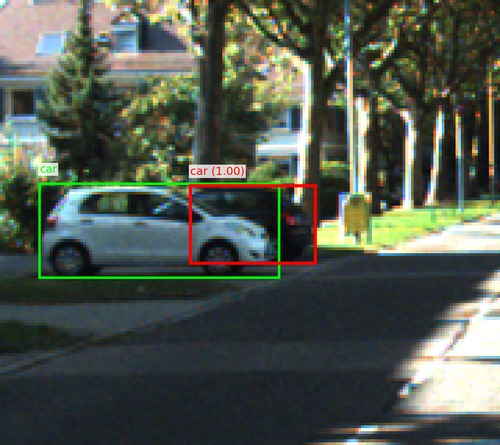}
\includegraphics[width=0.18\linewidth,trim={0 3cm 0 3cm}, clip]{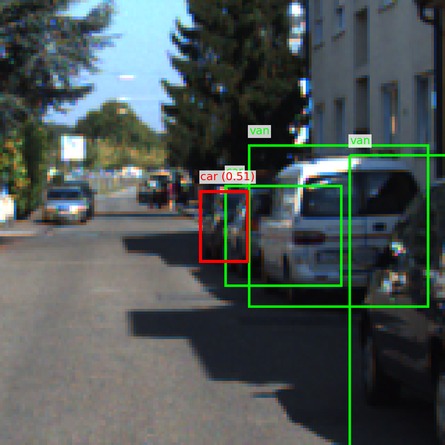}

\caption{KITTI}
\end{subfigure}

\vspace{0.6em}

\begin{subfigure}[t]{\textwidth}
\centering

\includegraphics[width=0.18\linewidth,trim={0 3cm 0 2cm}, clip]{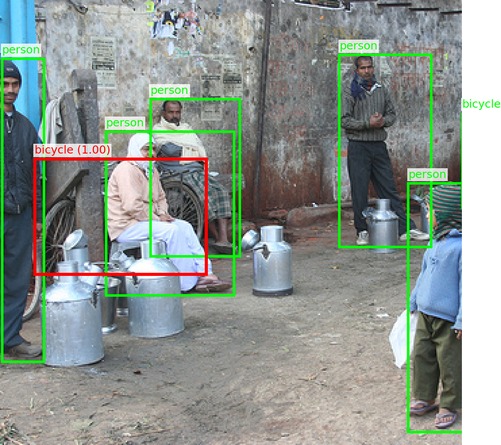}
\includegraphics[width=0.18\linewidth,trim={0 1cm 0 1cm}, clip]{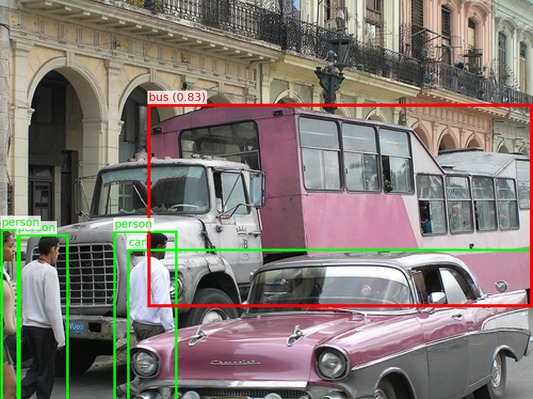}
\includegraphics[width=0.18\linewidth,trim={0 0 0 2cm}, clip]{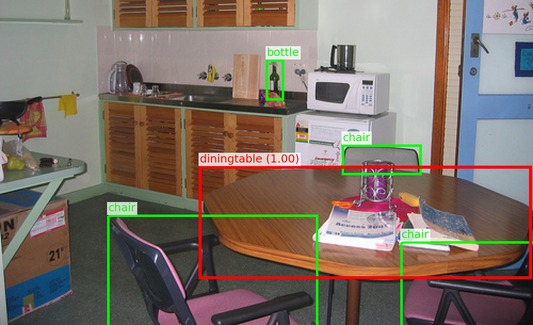}
\includegraphics[width=0.18\linewidth]{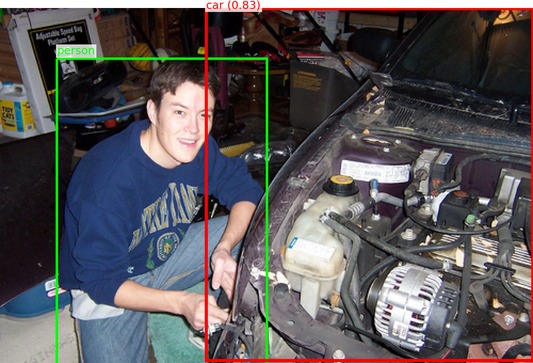}
\includegraphics[width=0.18\linewidth,trim={0 0 0 3cm}, clip]{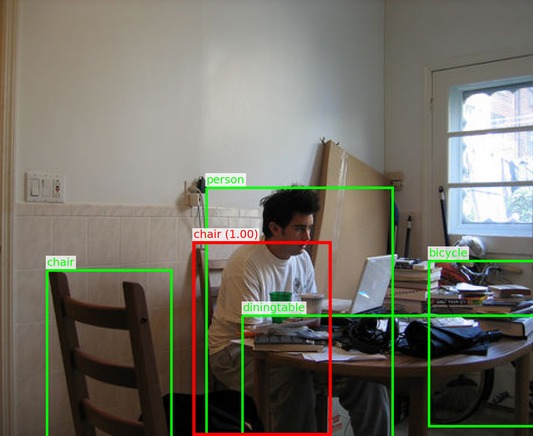}

\vspace{0.3em}

\includegraphics[width=0.18\linewidth,trim={0 3cm 0 3cm}, clip]{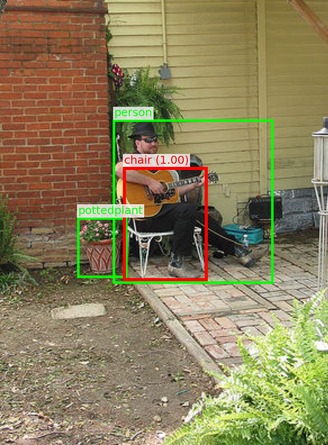}
\includegraphics[width=0.18\linewidth,trim={0 3cm 0 0}, clip]{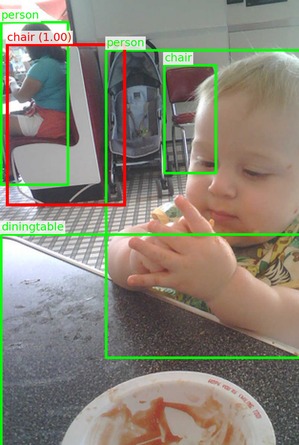}
\scalebox{0.8}{\includegraphics[width=0.18\linewidth]{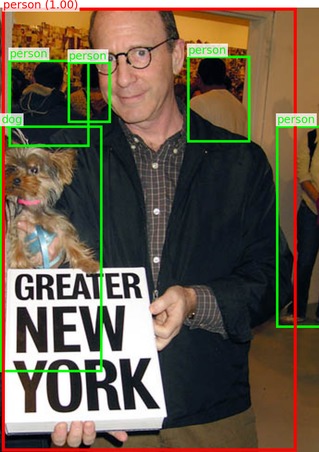}}
\includegraphics[width=0.18\linewidth,trim={0 2cm 0 2cm}, clip]{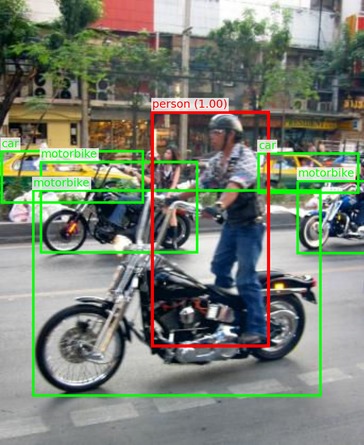}
\includegraphics[width=0.18\linewidth]{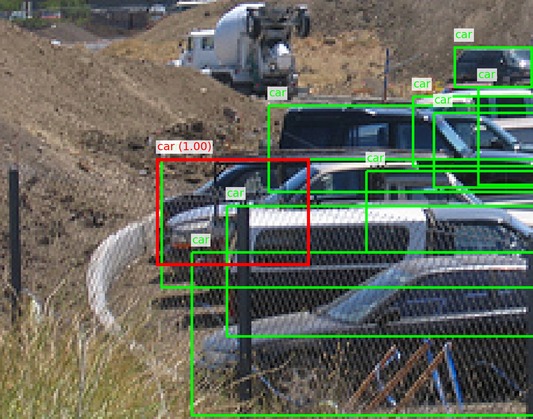}

\caption{Pascal VOC}
\end{subfigure}

\vspace{0.6em}

\begin{subfigure}[t]{\textwidth}
\centering

\includegraphics[width=0.18\linewidth]{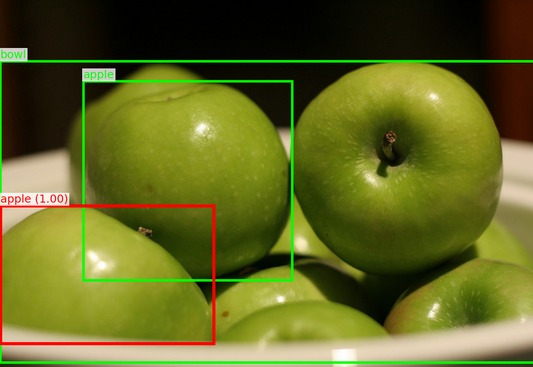}
\includegraphics[width=0.18\linewidth,trim={0 3cm 0 0}, clip]{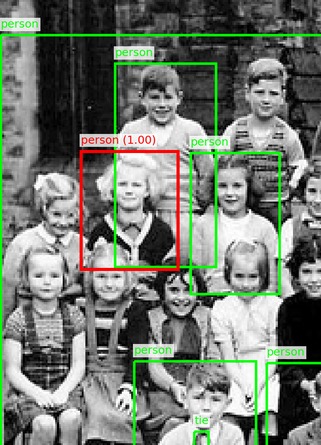}
\includegraphics[width=0.18\linewidth]{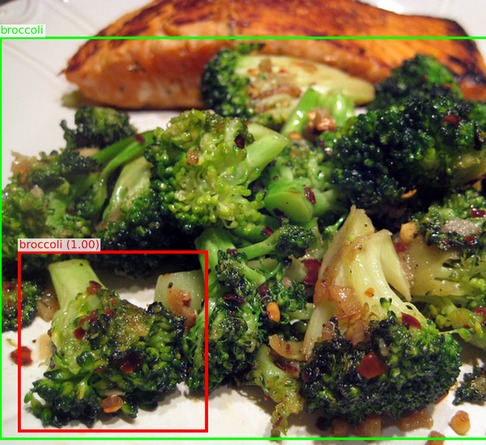}
\includegraphics[width=0.18\linewidth,trim={0 2cm 0 2cm}, clip]{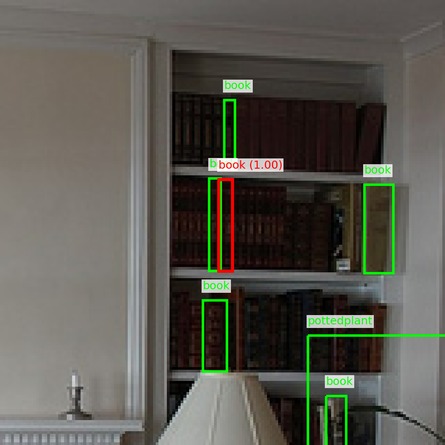}
\includegraphics[width=0.18\linewidth,trim={0 4cm 0 4cm}, clip]{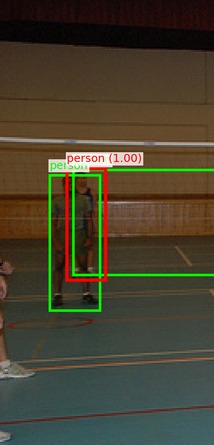}

\vspace{0.3em}

\includegraphics[width=0.18\linewidth]{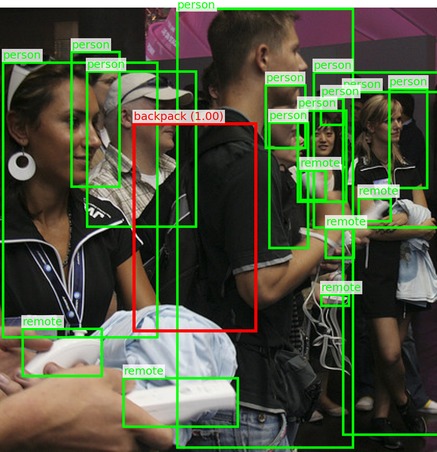}
\includegraphics[width=0.18\linewidth]{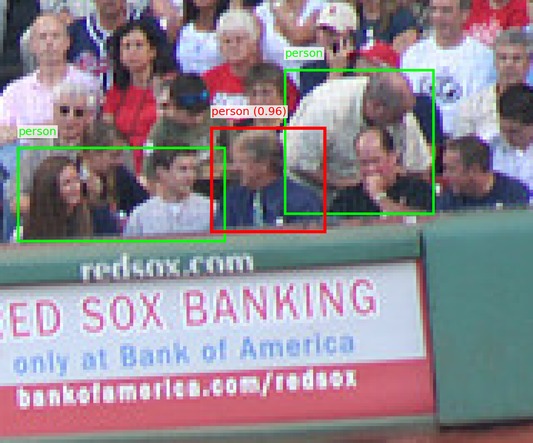}
\includegraphics[width=0.18\linewidth]{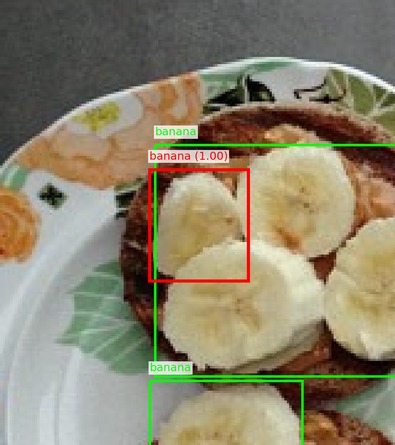}
\includegraphics[width=0.18\linewidth]{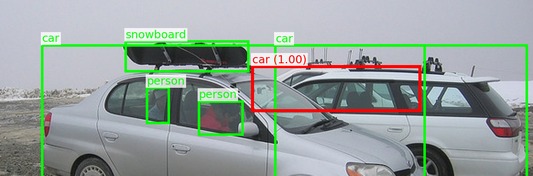}
\includegraphics[width=0.18\linewidth,trim={0 3cm 0 3cm}, clip]{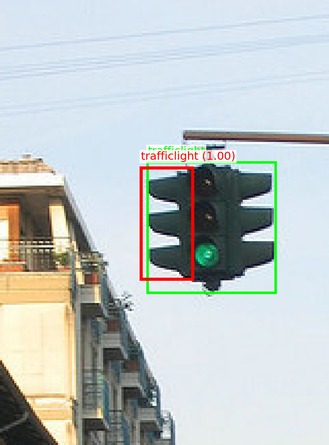}

\caption{COCO}
\end{subfigure}

\caption{Examples of \textit{misaligned} label errors across all four datasets. The green bounding boxes denote the original ground truth annotations, while the red bounding box highlights the label error with corresponding soft label probability $p$.}
\label{fig:misaligned_label_errors}
\end{figure*}

%% file: labelerrors/classification/plot_classification_label_errors.tex
\begin{figure*}[htbp]
\centering

\begin{subfigure}[t]{\textwidth}
\centering

\includegraphics[width=0.19\linewidth, trim={0 4cm 0 4cm}, clip]{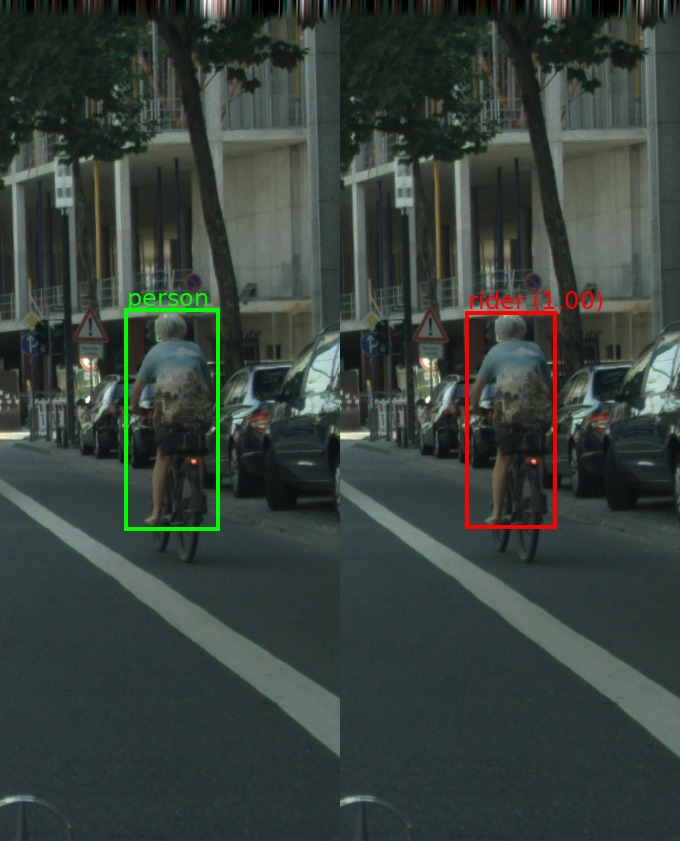}
\includegraphics[width=0.19\linewidth]{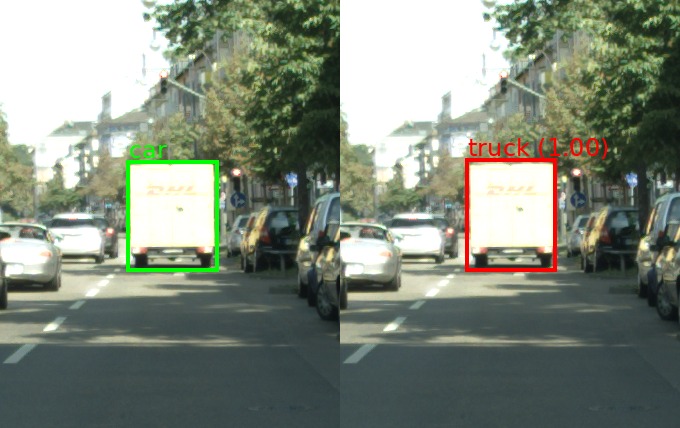}
\includegraphics[width=0.19\linewidth]{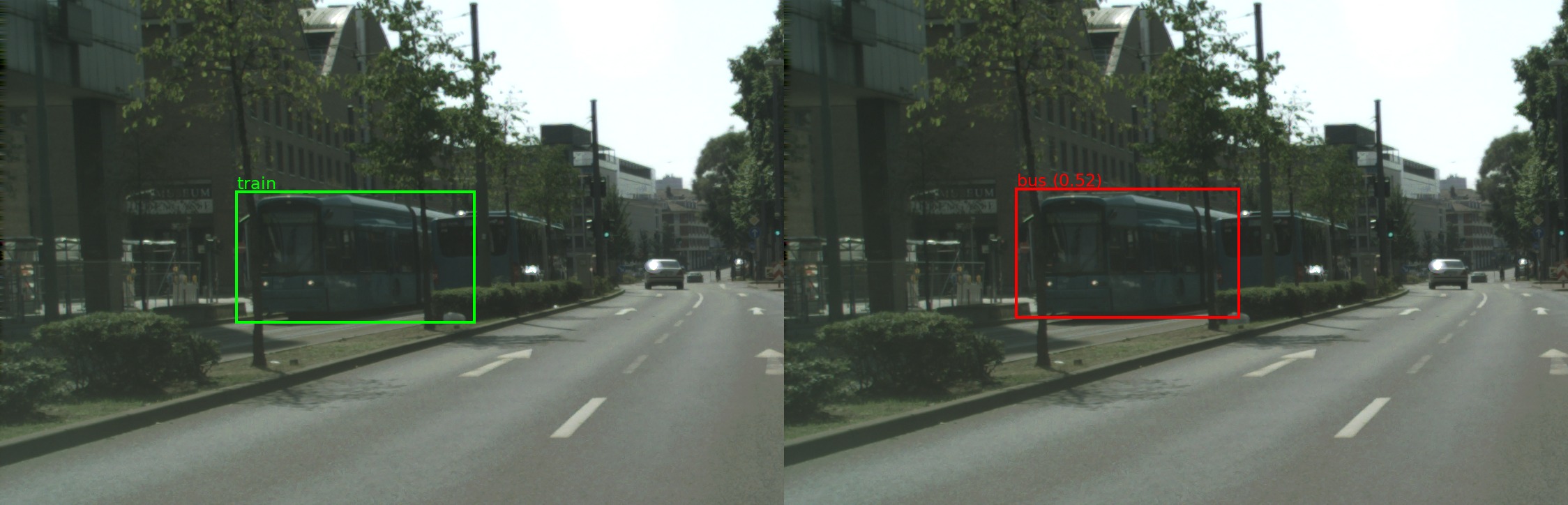}
\includegraphics[width=0.19\linewidth, trim={0 4cm 0 4cm}, clip]{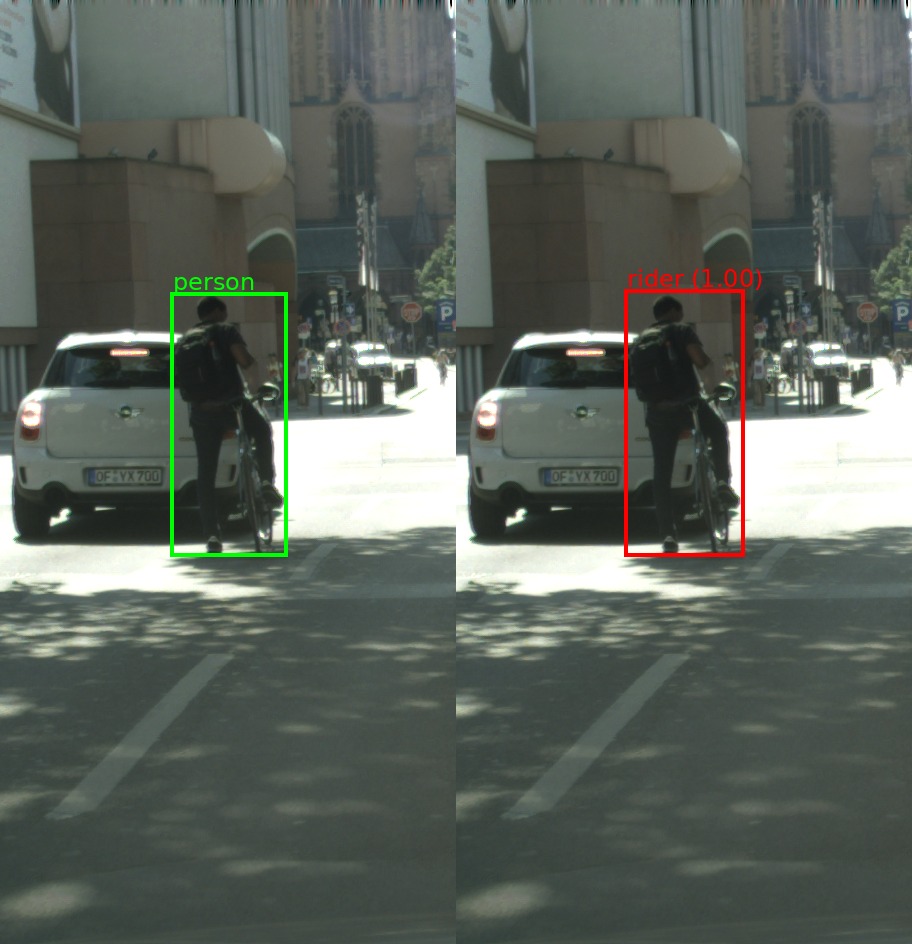}
\includegraphics[width=0.19\linewidth]{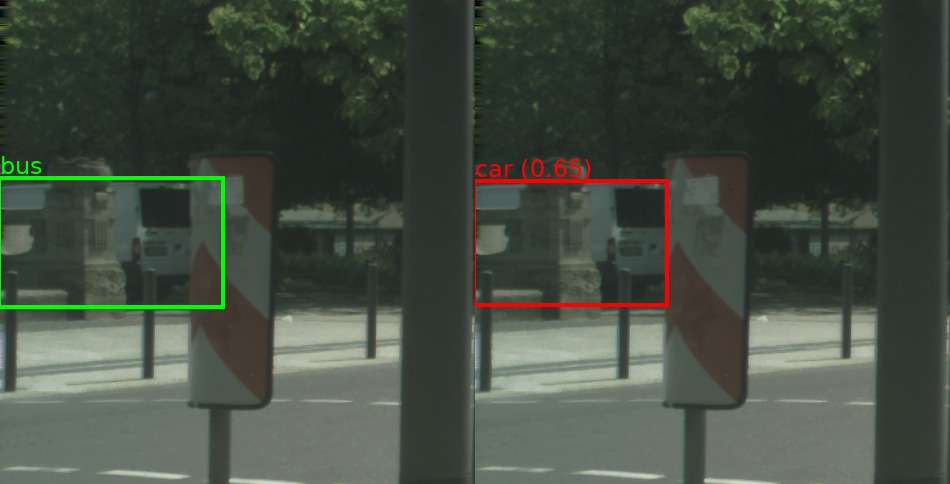}

\vspace{0.3em}

\includegraphics[width=0.19\linewidth]{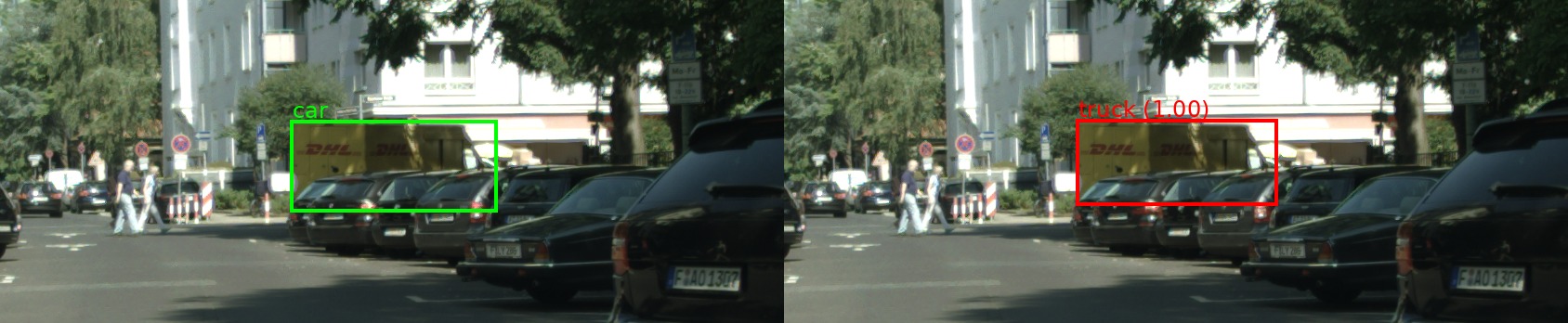}
\includegraphics[width=0.19\linewidth]{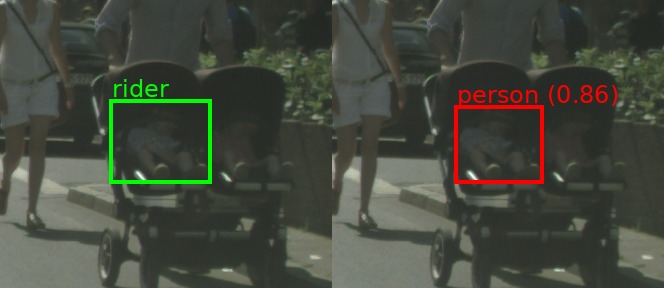}
\includegraphics[width=0.19\linewidth]{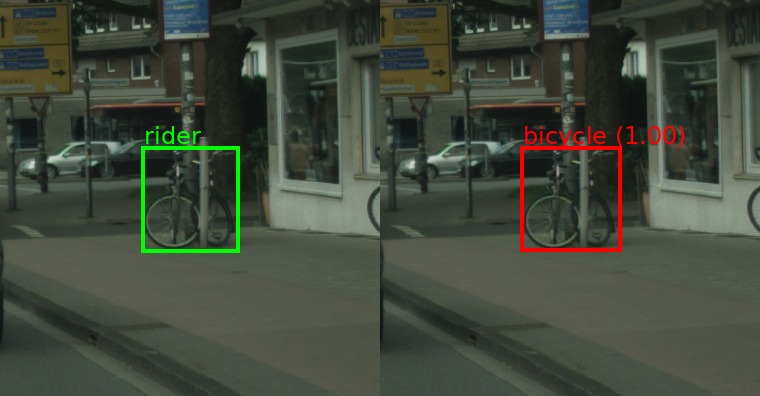}
\includegraphics[width=0.19\linewidth, trim={0 4cm 0 4cm}, clip]{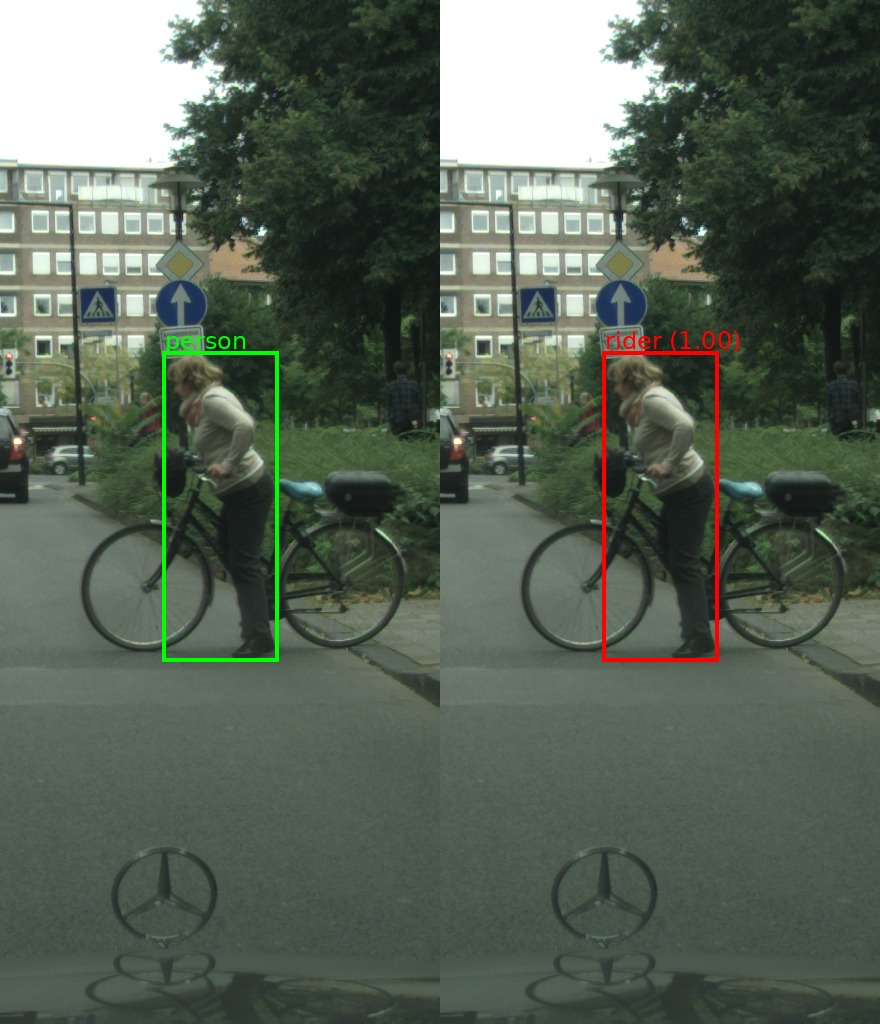}
\includegraphics[width=0.19\linewidth]{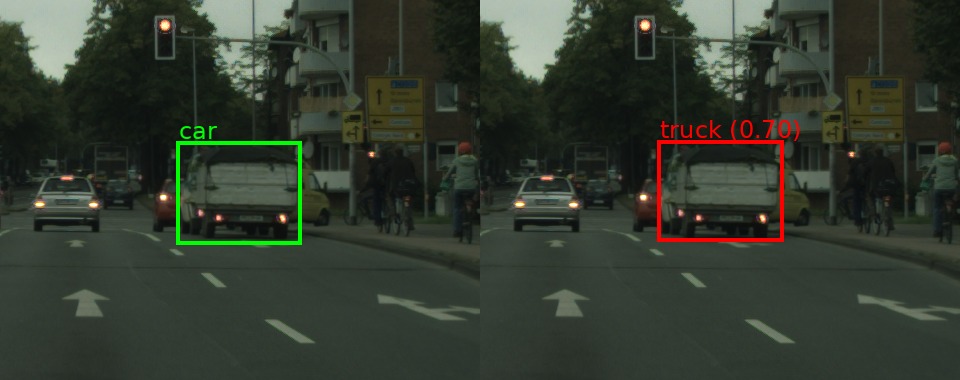}

\caption{Cityscapes}
\end{subfigure}

\vspace{0.6em}

\begin{subfigure}[t]{\textwidth}
\centering

\includegraphics[width=0.19\linewidth]{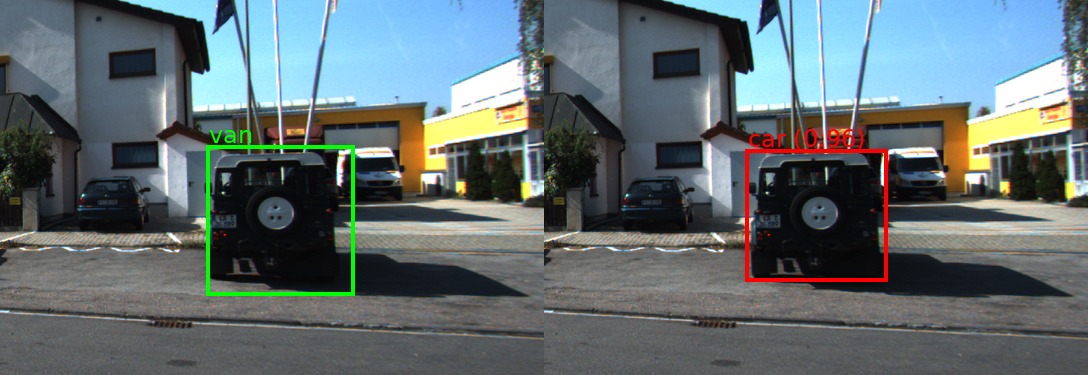}
\includegraphics[width=0.19\linewidth]{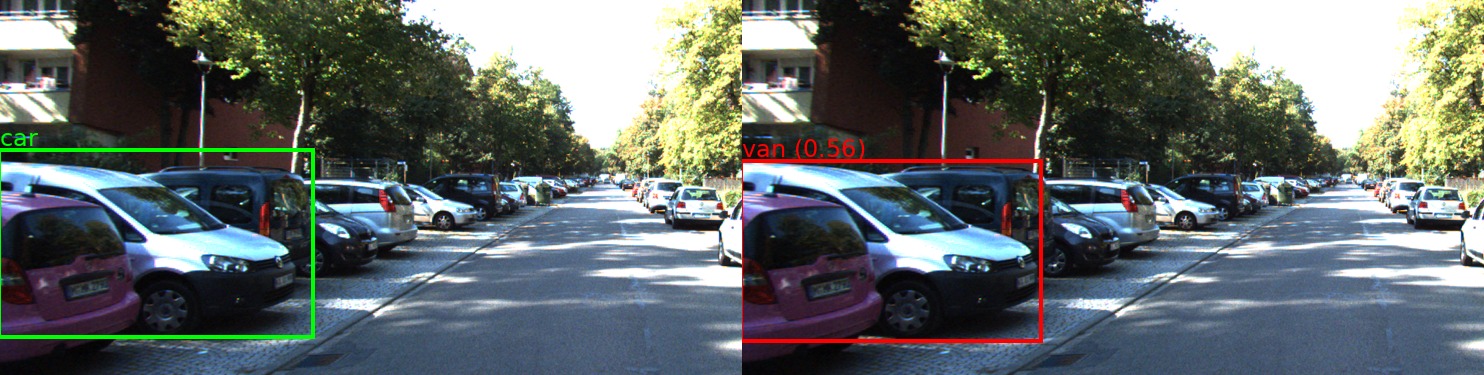}
\includegraphics[width=0.19\linewidth]{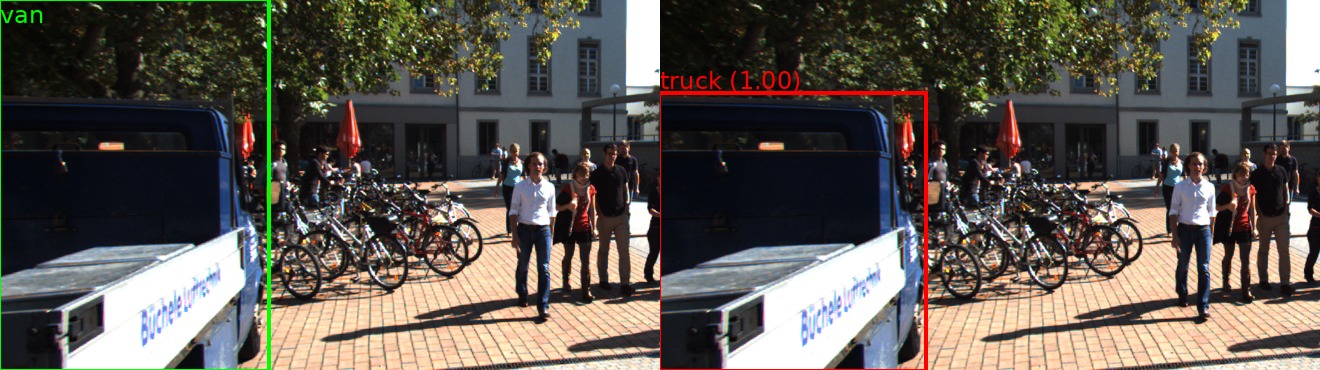}
\includegraphics[width=0.19\linewidth]{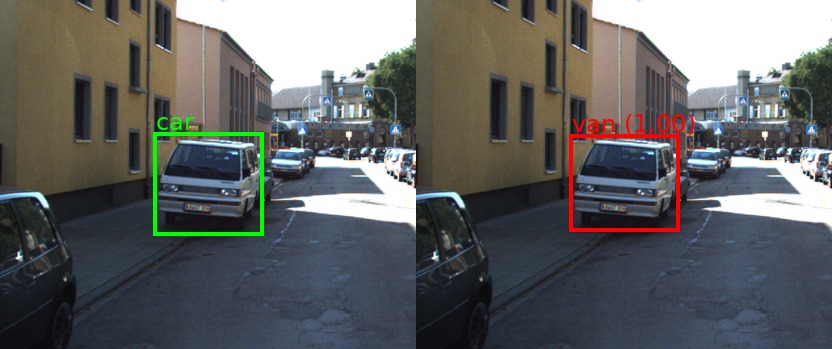}
\includegraphics[width=0.19\linewidth]{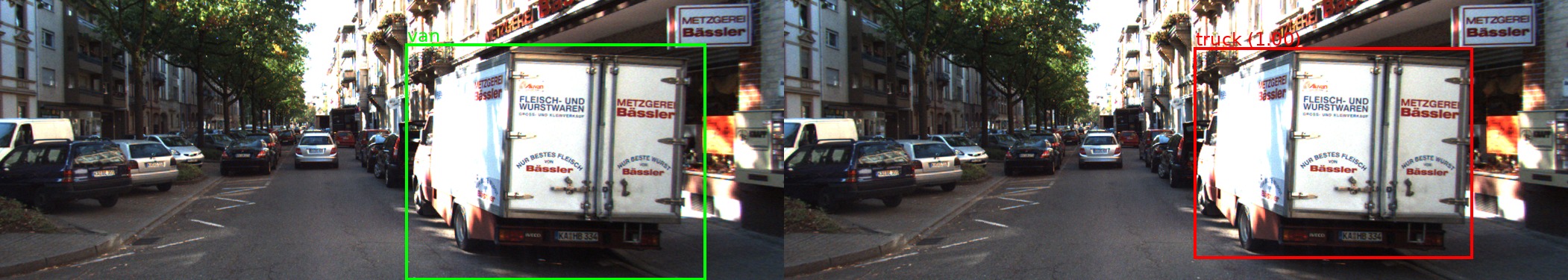}

\vspace{0.3em}

\includegraphics[width=0.19\linewidth]{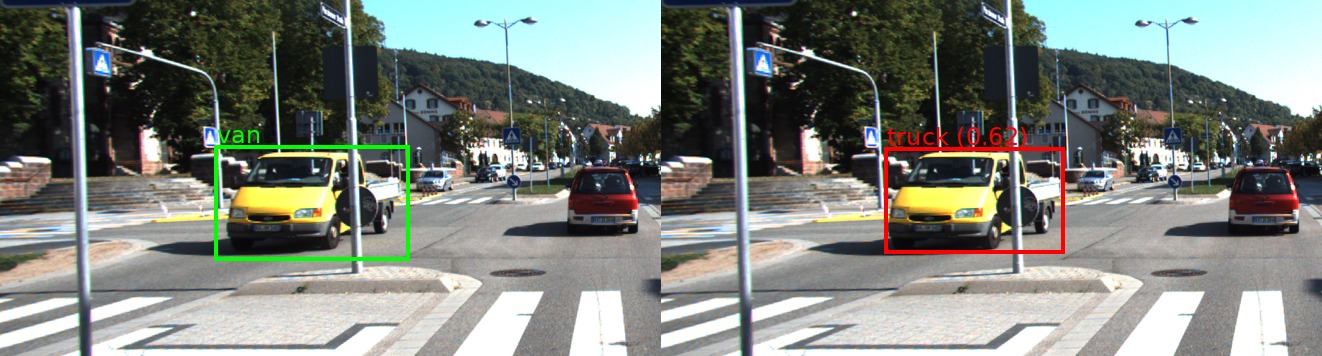}
\includegraphics[width=0.19\linewidth]{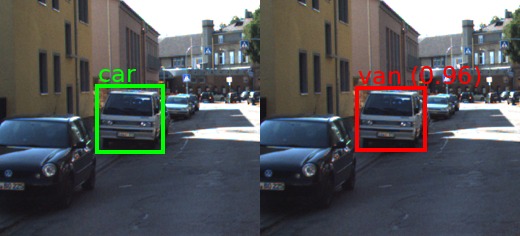}
\includegraphics[width=0.19\linewidth]{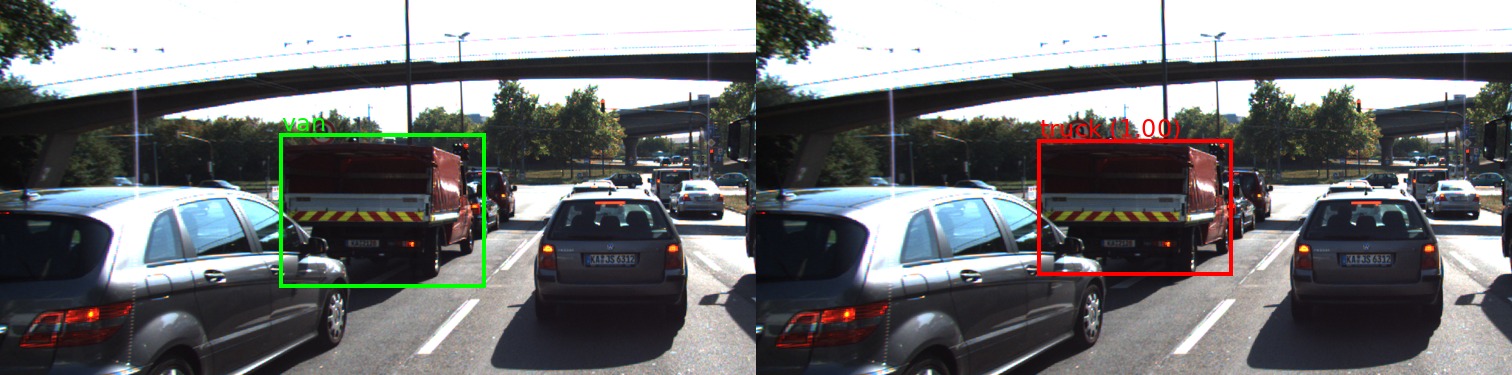}
\includegraphics[width=0.19\linewidth]{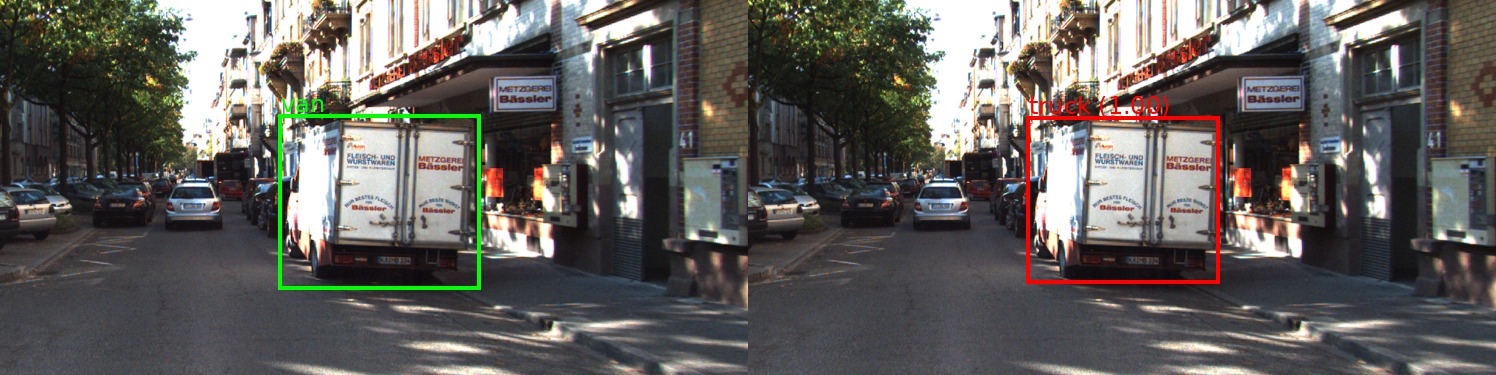}
\includegraphics[width=0.19\linewidth]{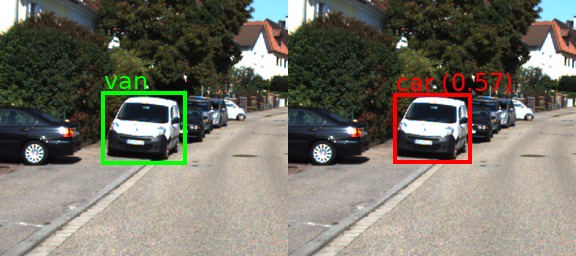}

\caption{KITTI}
\end{subfigure}

\vspace{0.6em}

\begin{subfigure}[t]{\textwidth}
\centering

\includegraphics[width=0.19\linewidth]{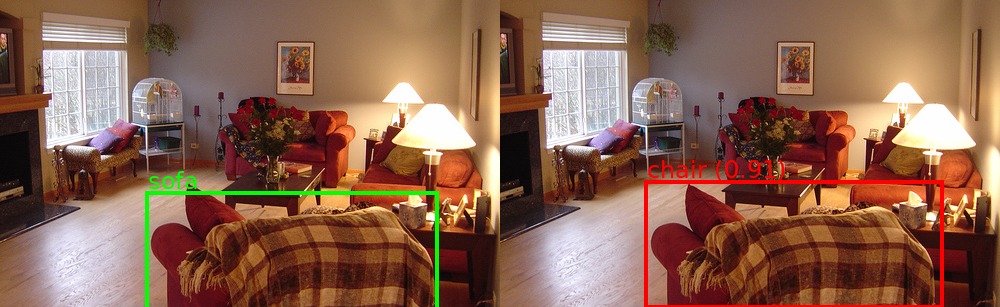}
\includegraphics[width=0.19\linewidth]{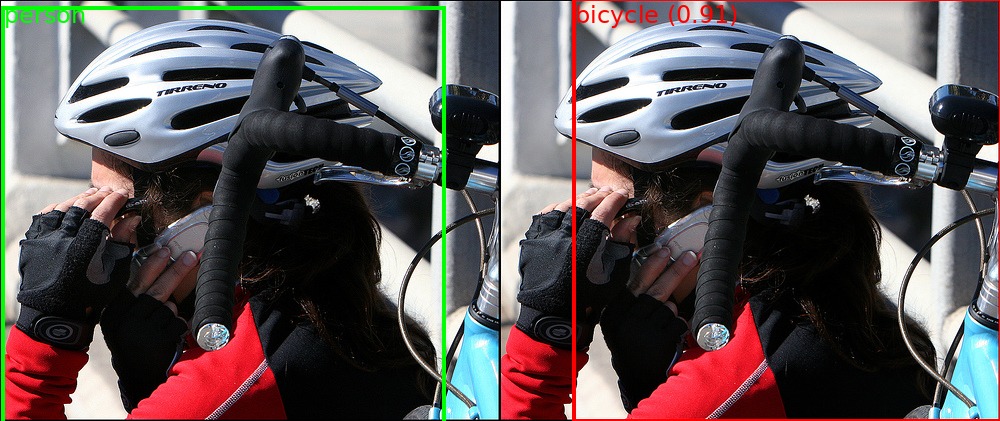}
\includegraphics[width=0.19\linewidth]{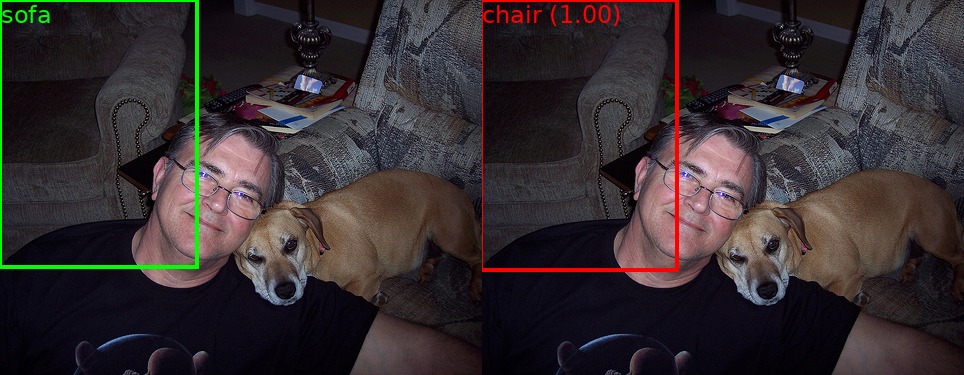}
\includegraphics[width=0.19\linewidth]{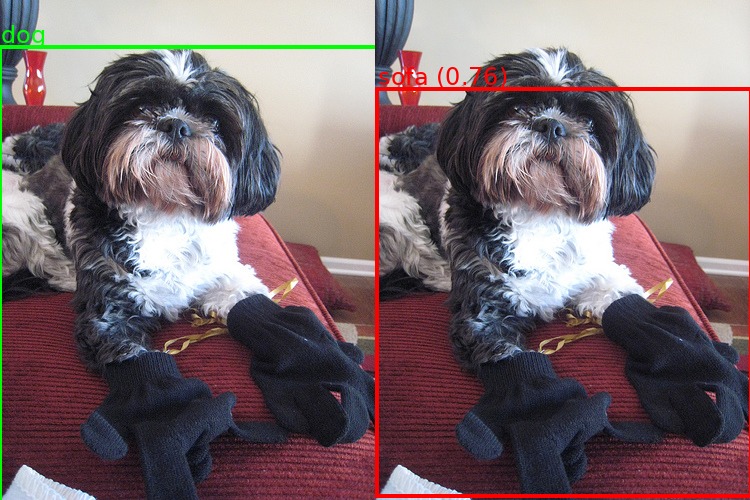}
\includegraphics[width=0.19\linewidth]{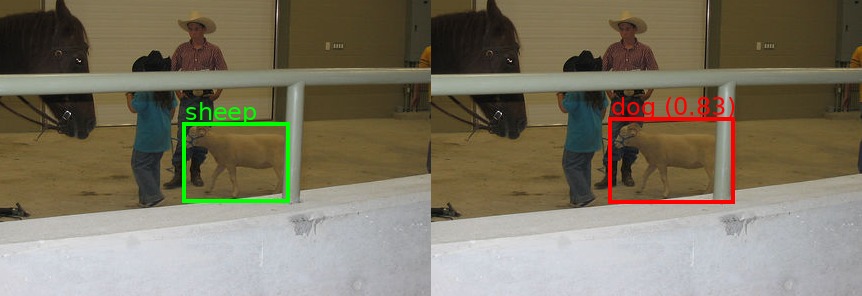}

\vspace{0.3em}

\includegraphics[width=0.19\linewidth]{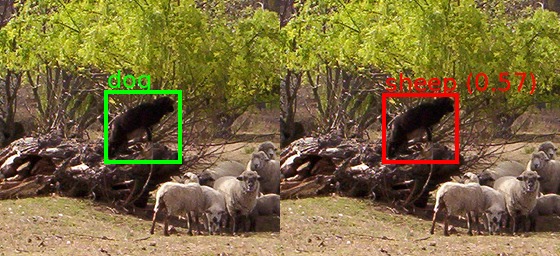}
\includegraphics[width=0.19\linewidth]{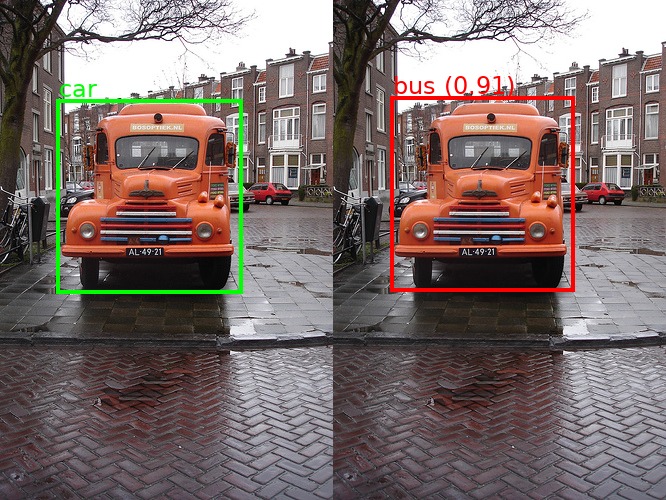}
\includegraphics[width=0.19\linewidth]{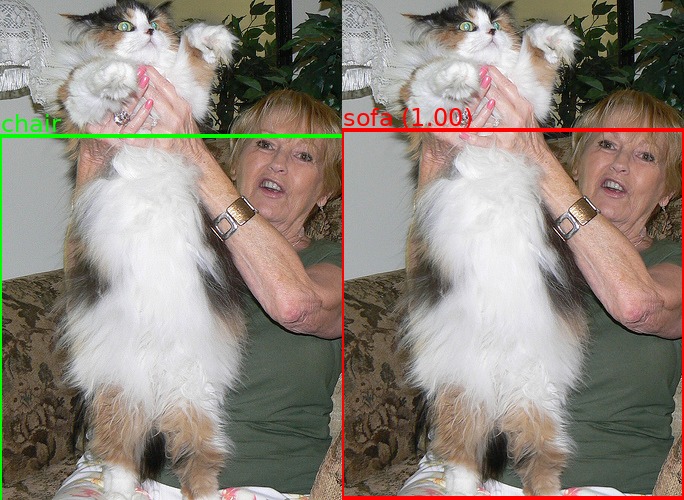}
\includegraphics[width=0.19\linewidth]{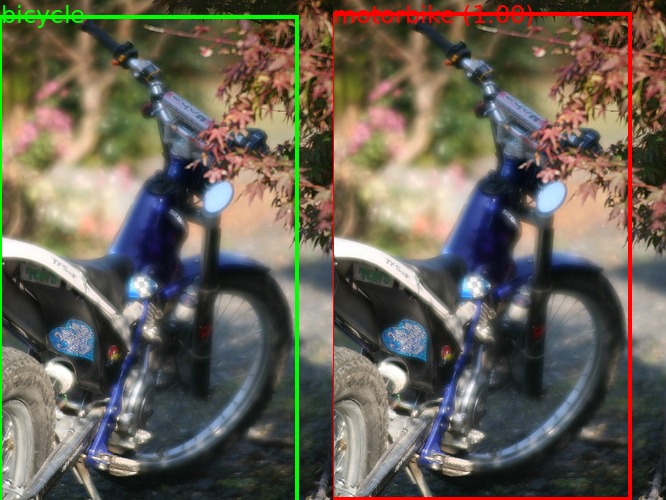}
\includegraphics[width=0.19\linewidth]{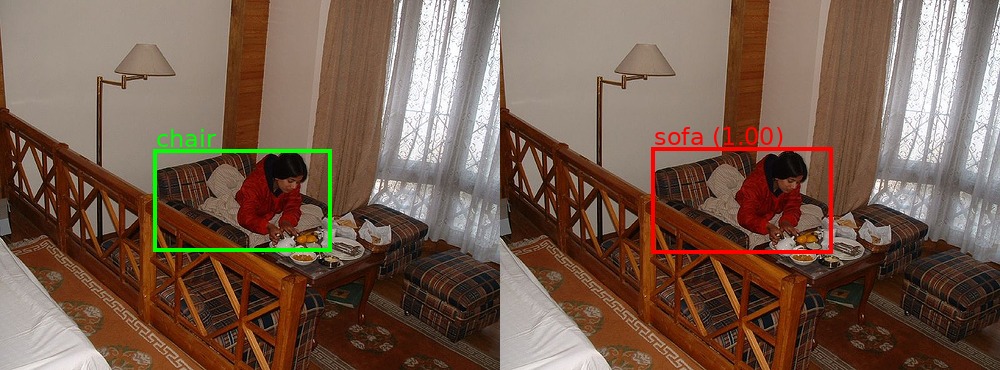}

\caption{Pascal VOC}
\end{subfigure}

\vspace{0.6em}

\begin{subfigure}[t]{\textwidth}
\centering
\includegraphics[width=0.19\linewidth]{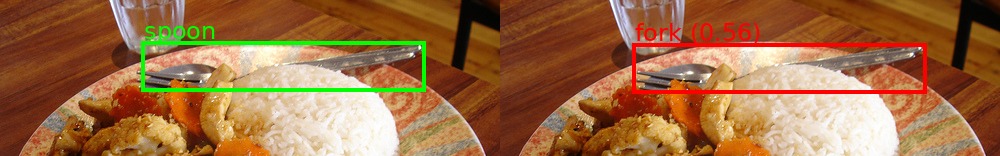}
\includegraphics[width=0.19\linewidth]{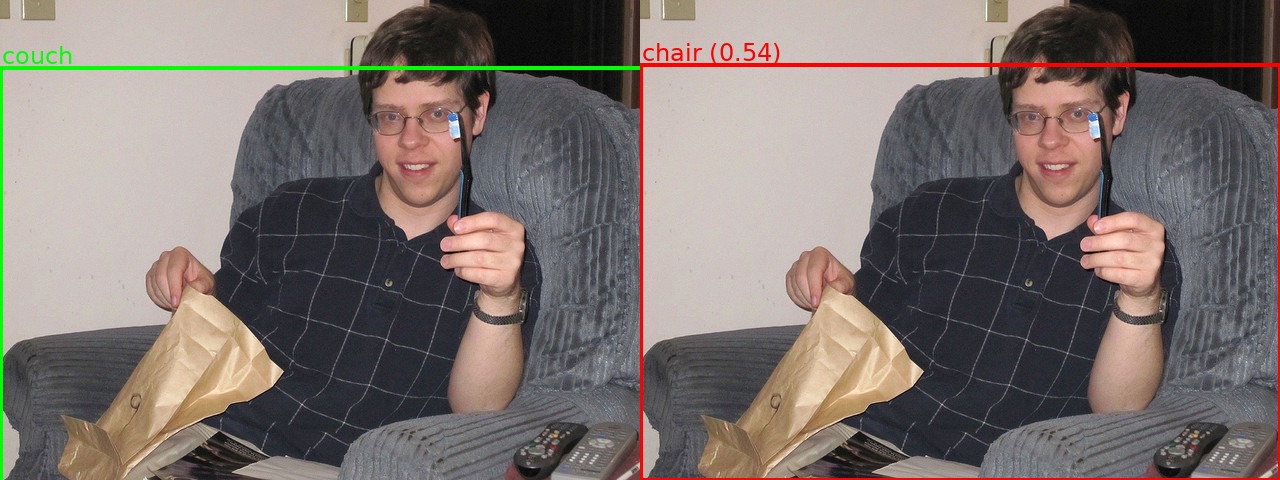}
\includegraphics[width=0.19\linewidth]{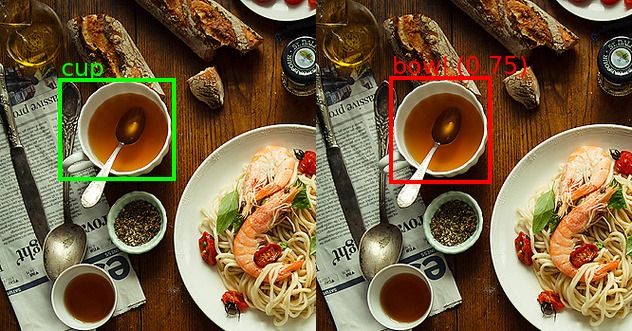}
\includegraphics[width=0.19\linewidth]{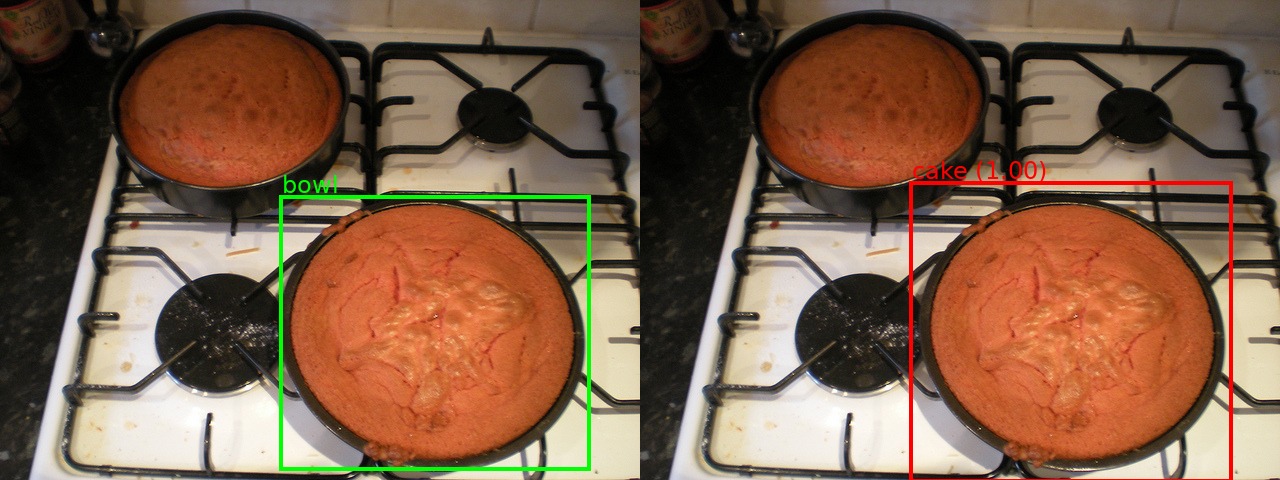}
\includegraphics[width=0.19\linewidth]{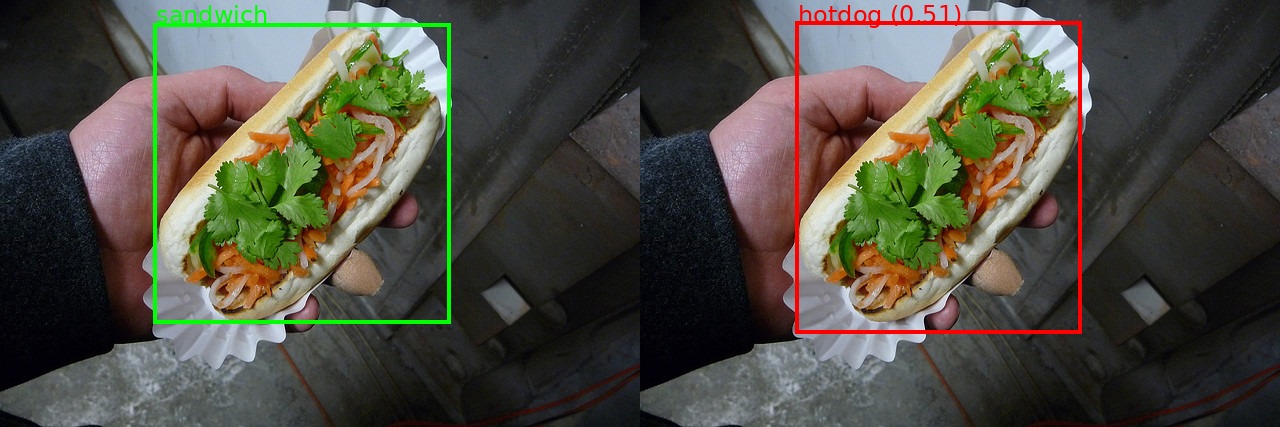}

\vspace{0.3em}

\includegraphics[width=0.19\linewidth]{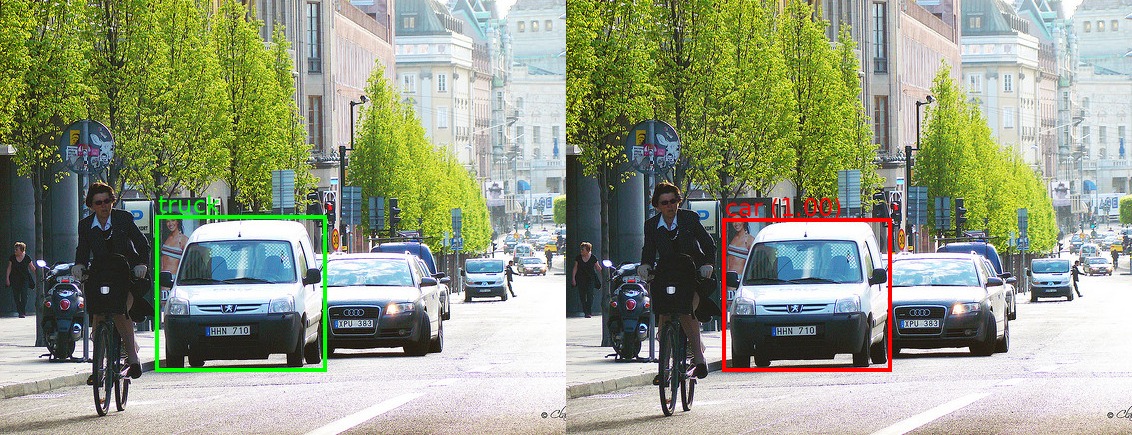}
\includegraphics[width=0.19\linewidth]{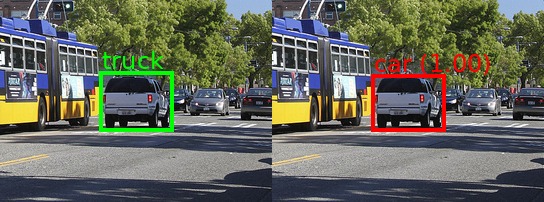}
\includegraphics[width=0.19\linewidth]{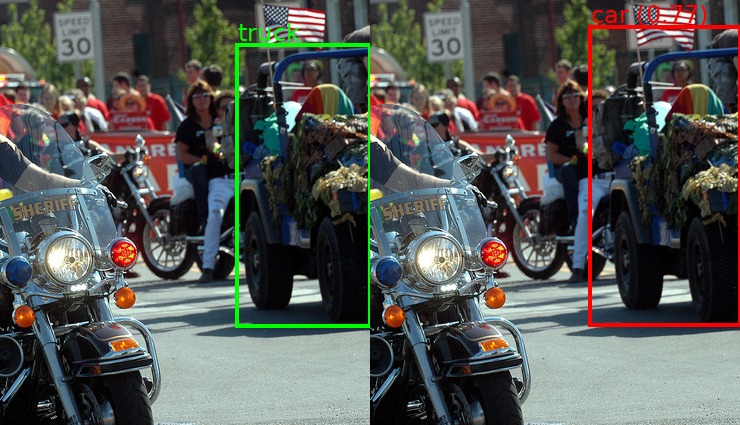}
\includegraphics[width=0.19\linewidth]{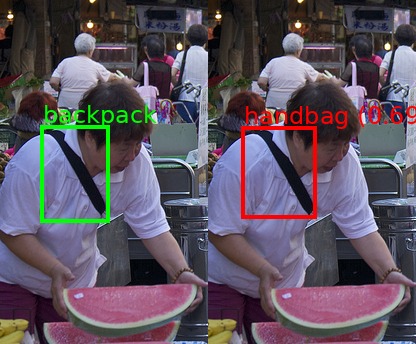}
\includegraphics[width=0.19\linewidth]{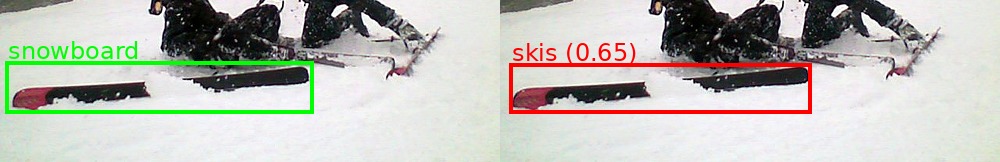}

\caption{COCO}
\end{subfigure}

\caption{Examples of classification label errors across all four datasets. The left (green) bounding box shows the original ground truth annotation with its corresponding label, while the right bounding box and label correspond to the validated ground truth with soft label probability $p$.}
\label{fig:classification_label_errors}
\end{figure*}